\documentclass[MAL,chapter,biber]{nowfnt} 

\usepackage[utf8]{inputenc}

\usepackage{amsmath}
\usepackage{amssymb}
\usepackage{amsfonts}
\usepackage{graphicx}
\usepackage{tikz}
\usepackage{algorithm}
\usepackage{algorithmic}
\usepackage{hhline}
\usepackage{booktabs}
\usepackage{array}
\usepackage{multirow}
\usepackage{url}
\usepackage{enumitem}
\usepackage[utf8]{inputenc}
\usepackage{xcolor}
\usepackage{comment}

\title{Dynamical Variational Autoencoders:  A Comprehensive Review}

\subtitle{Dynamical Variational Autoencoders:  A Comprehensive Review}

\maintitleauthorlist{
Laurent Girin \\
Univ.~Grenoble Alpes, CNRS, Grenoble-INP, GIPSA-lab \\
laurent.girin@grenoble-inp.fr
\and
Simon Leglaive \\
CentraleSup\'elec, IETR \\
simon.leglaive@centralesupelec.fr
\and
Xiaoyu Bie \\
Inria, Univ.~Grenoble Alpes, CNRS, LJK \\
xiaoyu.bie@inria.fr
\and
Julien Diard \\
Univ.~Grenoble Alpes, CNRS, LPNC \\
julien.diard@univ-grenoble-alpes.fr
\and
Thomas Hueber \\
Univ.~Grenoble Alpes, CNRS, Grenoble-INP, GIPSA-lab \\
thomas.hueber@grenoble-inp.fr
\and
Xavier Alameda-Pineda \\
Inria, Univ.~Grenoble Alpes, CNRS, LJK \\
xavier.alameda-pineda@inria.fr
}

\issuesetup
{%
 copyrightowner={now Publishers Inc.},
 volume        = 15,
 issue         = 1-2,
 pubyear       = 2021,
 isbn          = xxx-x-xxxxx-xxx-x,
 eisbn         = xxx-x-xxxxx-xxx-x,
 doi           = 10.1561/2200000089,
 firstpage     = 1, 
 lastpage      = 175
 }

\usepackage{mwe}

\usetikzlibrary{fit,positioning,bayesnet}
\setlist[itemize]{label=$\bullet$}

\definecolor{grey30}{RGB}{84,84,84}
\definecolor{ored}{RGB}{255,36,0}
\definecolor{metal}{RGB}{35,107,142}
\definecolor{dsblue}{RGB}{0,104,139}
\definecolor{darkgreen}{rgb}{0 0.6 0}

\newcommand{\vect}[1]{\mathbf{#1}}

\newcommand{\inp}{\textrm{\footnotesize in}}

\newcommand{\hid}{\textrm{\footnotesize hid}}
\newcommand{\rec}{\textrm{\footnotesize rec}}
\newcommand{\out}{\textrm{\footnotesize out}}
\newcommand{\enc}{\textrm{\footnotesize enc}}

\definecolor{dgreen}{RGB}{0,128,0}
\definecolor{dblue}{RGB}{0,0,128}
\definecolor{dred}{RGB}{128,0,0}
\definecolor{dpurple}{RGB}{128,0,128}

\newcommand*{\myx}[1][]{{\color{dgreen}\vect{x}_{#1}}}
\newcommand*{\myxh}[1][]{{\color{dgreen}\hat{\vect{x}}_{#1}}}
\newcommand*{\myxn}[1][]{{\color{dgreen}\vect{x}_{#1}^{(n)}}}
\newcommand*{\myxN}[1][]{{\color{dgreen}\vect{x}_{#1}^{(1:N)}}}
\newcommand*{\myX}[1][]{{\color{dgreen}\vect{X}_{#1}}}

\newcommand*{\myz}[1][]{{\color{dred}\vect{z}_{#1}}}
\newcommand*{\myzr}[1][]{{\color{dred}\vect{z}_{#1}^{(r)}}}
\newcommand*{\myzn}[1][]{{\color{dred}\vect{z}_{#1}^{(n)}}}
\newcommand*{\myzN}[1][]{{\color{dred}\vect{z}_{#1}^{(1:N)}}}
\newcommand*{\myZ}[1][]{{\color{dred}\vect{Z}_{#1}}}

\newcommand*{\myh}[1][]{{\color{dblue}\vect{h}_{#1}}}
\newcommand*{\myhfw}[1][]{{\color{dblue}\overrightarrow{\vect{h}}_{#1}}}
\newcommand*{\myhbw}[1][]{{\color{dblue}\overleftarrow{\vect{h}}_{#1}}}
\newcommand*{\myhn}[1][]{{\color{dblue}\vect{h}_{#1}^{(n)}}}

\newcommand*{\myk}[1][]{{\color{dblue}\vect{k}_{#1}}}

\newcommand*{\myg}[1][]{{\color{dblue}\vect{g}_{#1}}}
\newcommand*{\mygfw}[1][]{{\color{dblue}\overrightarrow{\vect{g}}_{#1}}}
\newcommand*{\mygbw}[1][]{{\color{dblue}\overleftarrow{\vect{g}}_{#1}}}
\newcommand*{\mygfwz}[1][]{{\color{dblue}\overrightarrow{\vect{g}}_{#1}^{\mathbf{z}}}}
\newcommand*{\mygbwz}[1][]{{\color{dblue}\overleftarrow{\vect{g}}_{#1}^{\mathbf{z}}}}
\newcommand*{\mygfwx}[1][]{{\color{dblue}\overrightarrow{\vect{g}}_{#1}^{\mathbf{x}}}}
\newcommand*{\mygbwx}[1][]{{\color{dblue}\overleftarrow{\vect{g}}_{#1}^{\mathbf{x}}}}
\newcommand*{\mygfwv}[1][]{{\color{dblue}\overrightarrow{\vect{g}}_{#1}^{\mathbf{v}}}}
\newcommand*{\mygbwv}[1][]{{\color{dblue}\overleftarrow{\vect{g}}_{#1}^{\mathbf{v}}}}

\newcommand*{\mygv}[1][]{{\color{dblue}\vect{g}_{#1}^{\mathbf{v}}}}
\newcommand*{\mygz}[1][]{{\color{dblue}\vect{g}_{#1}^{\mathbf{z}}}}

\newcommand*{\mya}[1][]{{\color{dblue}\vect{a}_{#1}}}

\newcommand*{\myu}[1][]{{\color{dpurple}\vect{u}_{#1}}}

\newcommand*{\myv}[1][]{{\color{dred}\vect{v}_{#1}}}

\newcommand*{\myw}[1][]{{\color{dred}\vect{w}_{#1}}}
\newcommand*{\myvn}[1][]{{\color{dred}\vect{v}_{#1}^{(n)}}}

\newcommand*{\myvN}[1][]{{\color{dred}\vect{v}_{#1}^{(1:N)}}}

\tikzset{  net/.style={draw,trapezium,trapezium angle=65,shape border rotate=270} }
\tikzset{  rnn/.style={draw,rectangle} }

\definecolor{goldenpoppy}{rgb}{0.99, 0.76, 0.0}
\definecolor{silver}{rgb}{0.75, 0.75, 0.75}

\author[1]{Girin, Laurent}
\author[2]{Leglaive, Simon}
\author[3]{Bie, Xiaoyu}
\author[4]{Diard, Julien}
\author[1]{Hueber, Thomas}
\author[3]{Alameda-Pineda, Xavier}

\affil[1]{Univ.~Grenoble Alpes, CNRS, Grenoble-INP, GIPSA-lab; laurent.girin@grenoble-inp.fr, thomas.hueber@grenoble-inp.fr}
\affil[2]{CentraleSup\'elec, IETR; simon.leglaive@centralesupelec.fr}
\affil[3]{Inria, Univ.~Grenoble Alpes, CNRS, LJK; xiaoyu.bie@inria.fr, xavier.alameda-pineda@inria.fr}
\affil[4]{Univ.~Grenoble Alpes, CNRS, LPNC; julien.diard@univ-grenoble-alpes.fr}

\articledatabox{\nowfntstandardcitation}

\begin{document}

\thispagestyle{empty}

\begin{center}
\vspace*{\fill}
The version of record is available at:\\ 
\url{http://dx.doi.org/10.1561/2200000089}
\vspace*{\fill}
\end{center}

\newpage

\makeabstracttitle

\begin{abstract}
Variational autoencoders (VAEs) are powerful deep generative models widely used to represent high-dimensional complex data through a low-dimensional latent space learned in an unsupervised manner. In the original VAE model, the input data vectors are processed independently. Recently, a series of papers have presented different extensions of the VAE to process sequential data, which model not only the latent space but also the temporal dependencies within a sequence of data vectors and corresponding latent vectors, relying on recurrent neural networks or state-space models. In this paper, we perform a literature review of these models. We introduce and discuss a general class of models, called dynamical variational autoencoders (DVAEs), which encompasses a large subset of these temporal VAE extensions. Then, we present in detail seven recently proposed DVAE models, with an aim to homogenize the notations and presentation lines, as well as to relate these models with existing classical temporal models. We have reimplemented those seven DVAE models and present the results of an experimental benchmark conducted on the speech analysis-resynthesis task (the PyTorch code is made publicly available). The paper concludes with a discussion on important issues concerning the DVAE class of models and future research guidelines.

\end{abstract}

\tableofcontents

\chapter{Introduction}
\label{sec:intro}
Deep Generative Models (DGMs) constitute a large family of probabilistic models that are currently of high interest in the machine learning and signal processing. They result from the combination of conventional (i.e., non-deep) generative probabilistic models and Deep Neural Networks (DNNs). 
For both conventional models and DGMs, different nonconflicting taxonomies can be established due to the domain richness and percolation across the different approaches. Nevertheless, these models can be grossly classified into the following two categories. 
Using the terminology of \citet{diggle1984monte}, the first category corresponds to \textit{prescribed} models for which the probability density function (pdf) of the generative model is defined explicitly, generally through a parametric form. The second category corresponds to \textit{implicit} models that can generate data ``directly,'' without using an explicit formulation and manipulation of a pdf model. Generative adversarial networks (GANs) are a popular example of this second category \citep{Goodfellow2014,DLBook,goodfellow2016nips}.

In the present review, we focus on the first category, in which a parametric pdf model is used. A suitable feature of generative models based on an explicit formulation of the pdf is that they can be easily plugged into a more general Bayesian framework, not only for generating data but also for modeling the data structure (without actually generating them) in various applications (e.g., data denoising or data transformation). 
In any case, the pdf model must be as close as possible to the true pdf of the data, which is generally unknown. To achieve this aim, the model must be trained from data, and model parameters are generally estimated by following the maximum likelihood methodology \citep{DLBook, BishopBook, koller2009probabilistic}. These principles are valid for both conventional generative models and DGMs; however, in the case of DGMs, the pdf parameters are generally the output of DNNs, which makes model training potentially difficult. 

\section{Deep Dynamical Bayesian Networks}
\label{sec:DDBN}

\begin{figure}
    \centering
\resizebox{\textwidth}{!}{

\begin{tikzpicture}

\def\xbn{2}
\def\xdeepbn{6}
\def\xvae{11}

\def\yig{-3}
\def\ybn{-6}
\def\ydbn{-9}
\def\yrnn{-12}

\tikzset{  model/.style={draw, rectangle, rounded corners=3pt, align=center, minimum width=2cm, minimum height=1cm} }
\tikzset{  section/.style={draw, circle, align=center, fill=black!25} }

\node[model] (gpm) at (0, 0) {generative \\ probabilistic \\ model};
\node[model] (ig) at (-2, \yig) {implicit \\ generative};
\node[model] (pg) at (\xbn, \yig) {prescribed \\ generative};
\node[model] (bn) at (\xbn, \ybn) {BN};
\node[model] (deepbn) at (\xdeepbn, \ybn) {Deep BN};
\node[model] (vae) at (\xvae, \ybn) {VAE};
\node[model] (dbn) at (\xbn, \ydbn) {DBN};
\node[model] (ddbn) at (\xdeepbn, \ydbn) {DDBN};
\node[model] (dvae) at (\xvae, \ydbn) {DVAE};
\node[rectangle, minimum width=9cm, minimum height=8cm] (dvaecercle) at (\xvae, \ydbn-.5) {~};
\node[model] (rnn) at (\xbn-1.5, \yrnn) {RNN};
\node[model] (ssm) at (\xbn+1.5, \yrnn) {SSM};
\node[model] (dkf) at (dvaecercle.-163) {DKF};
\node[model] (kvae) at (dvaecercle.-150) {KVAE};
\node[model] (storn) at (dvaecercle.-135) {STORN};
\node[model] (vrnn) at(dvaecercle.-110) {VRNN};
\node[model] (srnn) at (dvaecercle.-70) {SRNN};
\node[model] (rvae) at (dvaecercle.-45) {RVAE};
\node[model] (dsae) at (dvaecercle.-30) {DSAE};
\node[model] (others) at (dvaecercle.-17) {other \\ models};

\node[section] (sectvae) at (vae.east) {2};
\node[section] (sectrnn) at (rnn.east) {3};
\node[section] (sectssm) at (ssm.east) {3};
\node[section] (sectdvae) at (dvae.east) {4};
\node[section] (sectdkf) at (dkf.east) {5};
\node[section] (sectkvae) at (kvae.east) {6};
\node[section] (sectstorn) at (storn.east) {7};
\node[section] (sectvrnn) at (vrnn.east) {8};
\node[section] (sectsrnn) at (srnn.east) {9};
\node[section] (sectrvae) at (rvae.east) {10};
\node[section] (sectdsae) at (dsae.east) {11};
\node[section] (sectothers) at (others.east) {12};

\draw[->] (gpm) -- (ig) node[midway, left]{pdf-free\;};
\draw[->] (gpm) -- (pg) node[midway, right]{\;pdf-explicit};
\draw[->] (pg) -- (bn) node[midway, right]{directed acyclic graph};

\draw[->] (bn) -- (deepbn) node[midway, above]{deep};
\draw[->] (deepbn) -- (vae) node[midway, above]{variational};

\draw[->] (bn) -- (dbn) node[midway, right]{dynamical};
\draw[->] (deepbn) -- (ddbn) node[midway, right]{dynamical};
\draw[->] (vae) -- (dvae) node[midway, right]{dynamical};

\draw[->] (dbn) -- (ddbn) node[midway, above]{deep};
\draw[->] (ddbn) -- (dvae) node[midway, above]{variational};

\draw[->] (dbn) -- (rnn) node[pos=0.4, left]{deterministic\;};
\draw[->] (dbn) -- (rnn) node[pos=0.6, left]{dynamics\;};

\draw[->] (dbn) -- (ssm) node[pos=0.4, right]{\;stochastic};
\draw[->] (dbn) -- (ssm) node[pos=0.6, right]{\;dynamics};

\edge{dvae} {dkf, kvae, storn, vrnn, srnn, rvae, dsae, others};

\end{tikzpicture}
}
\caption{A graphical taxonomy of generative probabilistic models. Only the branch corresponding to the models covered in this review is detailed. Nodes represent classes of models, and arrow labels specify some of the relationships between the classes of models. Please refer to the text for details and acronym definitions. The numbered gray circles indicate the section number in which the corresponding class of models is detailed.}
    \label{fig:Taxonomy}
\end{figure}

In the present review, we focus on an important subfamily of DGMs, namely the deep dynamical Bayesian networks (DDBNs), which are built on the following models:
\begin{itemize}[label=\textbullet]
\item Bayesian networks (BNs) are a popular class of probabilistic models for which i) the dependencies among all involved random variables are explicitly represented by conditional pdfs (i.e.~BNs are prescribed models), and ii) these dependencies can be schematically represented using a directed acyclic graph \citep{BishopBook, koller2009probabilistic}. The structure of these dependencies often reflects (or originates from) an underlying hierarchical generative process. 
 \item Dynamical Bayesian networks are BNs that include temporal dependencies and are widely used to model dynamical systems and/or data sequences. Dynamical BNs are BNs ``repeated over time''; that is, they exhibit a repeating dependency structure (a time-slice at discrete time $t$) and some dependencies across these time-slices (the dynamical models).
Recurrent neural networks (RNNs) and state-space models (SSMs) can be considered special cases of dynamical BNs. In fact, a temporal dependency in a dynamical BN is often implemented either as a deterministic recursive process, as in RNNs, or as a first-order Markovian process, as in usual SSMs.
\item Deep Bayesian networks combine BNs with DNNs. DNNs are used to generate the parameters of the modeled distributions. This enables them to be high-dimensional and highly multi-modal while having a reasonable number of parameters. In short, deep BNs have can appropriately combine the ``explainability'' of Bayesian models with the modeling power of DNNs.
\end{itemize}
DDBNs are thus a combination of all these aspects, as illustrated in Figure~\ref{fig:Taxonomy}. They can be equally seen as dynamical versions of deep BNs (i.e., deep BNs including temporal dependencies) or deep versions of dynamical BNs (i.e., dynamical BNs mixed with DNNs). As an extension of dynamical BNs, DDBNs are expected to be powerful tools for modeling dynamical systems and/or data sequences. 
However, as mentioned above, the combination of probabilistic modeling with DNNs in deep BNs can result in a complex and costly model training. This is an even more serious issue for DDBNs, in which the repeating structure due to temporal modeling adds a level of complexity. 

\section{Variational inference and VAEs}

Recently, the application of the variational inference methodology \citetext{\citealp{jordan1999introduction}; \citealp[Chapter~10]{BishopBook}; \citealp{vsmidl2006variational}; \citealp[Chapter~21]{murphy2012machine}} to a fundamental deep BN architecture --a low-dimensional to a high-dimensional generative feed-forward DNN-- has led to efficient inference and training of the resulting model, called a variational autoencoder (VAE) \citep{Kingma2014}. A similar approach was proposed the same year \citep{rezende2014stochastic}.\footnote{\citet{Kingma2014} and \citet{rezende2014stochastic} were both pre-published in 2013 as ArXiv papers. Connections also exist with \citet{mnih2014neural}.} 
The VAE is directly connected to the concepts of a \emph{latent} variable and \emph{unsupervised representation learning}: the observed random variable representing the data of interest is assumed to be generated from an unobserved latent variable through a probabilistic process. Often, this latent variable is of lower dimension than the observed data (which can be high-dimensional) and is assumed to ``encode'' the observed data in a compact manner so that new data can be generated from new values of the latent variable. Moreover, one wishes to extract a latent representation that is \emph{disentangled} (i.e., different latent coefficients encode different properties or different factors of variation in the data). When successful, this provides good interpretability and control of the data generation/transformation process.

The automatic discovery of a latent space structure is part of the model training process. The inference process, which is defined in the present context as the estimation of latent variables from the observed data, also plays a major role. As presented in detail later, in a deep BN, the exact posterior distribution (i.e., the posterior distribution of the latent variable given the observed variable corresponding to the generative model) is generally not tractable. It is thus replaced with a parametric approximate posterior distribution (i.e., an inference model) that is implemented with a DNN. As the observed data likelihood function is also not tractable, the model parameters are estimated by chaining the inference model (also known as the \emph{encoder} in the VAE framework) and the generative model (the \emph{decoder}) and maximizing a lower bound of the log-likelihood function, called the variational lower bound (VLB), over a training dataset.\footnote{The idea of using an artificial neural network to approximate an inference model and chaining the encoder and decoder dates back to the early studies of \citet{hinton1995wake} and \citet{dayan1995helmholtz}. However, the algorithms presented in these papers for model training are different from the one used to optimize the VAE.} Hereinafter, we refer to this general variational inference and training methodology as \emph{the VAE methodology}. 

In summary, the VAE methodology enables deep unsupervised representation learning while providing efficient inference and parameter estimation in a Bayesian framework. As a result, the seminal papers by \citet{Kingma2014} and \citet{rezende2014stochastic} have had and continue to have a strong impact on the machine learning community. VAEs have been applied to many signal processing problems, such as the generation and transformation of images and speech signals (we provide a few references in Chapter~\ref{sec:VAEs}). 

\section{Dynamical VAEs}
\label{sec:intro-DVAEs}

As a deep BN, the original VAE proposed by \citet{Kingma2014} did not include temporal modeling. This means that each data vector was processed independently of the other data vectors (and the corresponding latent vector was also processed independently of the other latent vectors). This is clearly suboptimal for the modeling of correlated (temporal) vector sequences.

In the years following the publication of \citet{Kingma2014} and \citet{rezende2014stochastic}, the VAE methodology was extended and successfully applied to several more complex deep BNs. In particular, it was applied to deep BNs with a temporal model (i.e., DDBNs) dedicated to the modeling of sequential data exhibiting temporal correlation. In the present review, we are particularly interested in the models presented in the following papers: \citep{bayer2014learning,
krishnan2015deep, chung2015recurrent, gu2015neural, fraccaro2016sequential, krishnan2017structured, fraccaro2017disentangled, goyal2017z, hsu2017unsupervised, yingzhen2018disentangled, leglaive2020recurrent}. In addition to including temporal dependencies, the unsupervised representation learning essence of the VAE is preserved and cherished in these studies. These DDBNs combine the observed and latent variables and aim at modeling not only data dynamics but also discovering the latent factors governing them. 

To achieve this aim, these models are trained using the VAE methodology (i.e.,~design of an inference model and maximization of the corresponding VLB). We can thus encompass these models under the common class and terminology of \textit{variational DDBNs} (i.e., DDBNs immersed in the VAE framework). In the following of the paper, as well as the title, we prefer to refer to them as \emph{dynamical VAEs} (DVAEs) (i.e., VAEs including a temporal model for modeling sequential data). This is simply because we assume that the term ``VAE'' is currently more popular than the term ``DBN,'' and ``dynamical VAEs'' gives a more speaking-first evocation of these models, compared to ``variational DDBNs.''
This convergence of DDBNs and VAEs into DVAEs is illustrated in Figure~\ref{fig:Taxonomy}.

In practice, these different DVAE models vary in how they define the dependencies between the observed and latent variables, how they define and parameterize the corresponding generative pdfs, and how they define and parameterize the inference model. They also differ in how they combine the variables with RNNs to model temporal dependencies, at both generation and inference. 
In contrast, they are all characterized by the following common set of features. 

First, as stated above, they are all trained using the VAE methodology, possibly with a few adaptations and refinements. In this paper, we do not review models based on GANs and, more generally, on adversarial training. Examples of extensions of ``static'' GANs to sequence modeling and generation can be found in the literature \citep{mathieu2016disentangling, villegas2017decomposing, denton2017unsupervised, tulyakov2018mocogan, lee2018acoustic}. This approach is particularly popular for separating content and motion in videos.

Second, even if the observed random vectors can be continuous or discrete, as in the original VAE formulation, they all feature \emph{continuous latent random variables}. In the present review, we do not consider the case of discrete latent random variables. The latter can be incorporated in DVAE models, in the line with the case of, for example, conditional VAEs \citep{sohn2015learning,zhao2017learning}. Temporal models with binary observed and latent random variables have been proposed \citep{boulanger2012modeling, gan2015deep}. These models are based on restricted Boltzman machines (RBMs) or sigmoid belief networks (SBNs) combined with RNNs. A detailed analysis of such models is beyond the scope of the present review. 

Third, all DVAE models we consider feature \emph{a discrete-time sequence of (continuous or discrete) observed random vectors associated with a corresponding discrete-time sequence of (continuous) latent random vectors}. In other words, these models function in a \emph{sequence-to-sequence mode for both encoding and decoding}.
Thus, we do not focus on VAE-based models specifically designed for text and dialogue generation \citep{bowman2015generating, miao2016neural, serban2016piecewise, serban2017hierarchical, yang2017improved, semeniuta2017hybrid, hu2017toward, zhao2018unsupervised, jang2019recurrent} or (2D) image modeling  \citep{gulrajani2016pixelvae, chen2017variational, lucas2018auxiliary,shang2018channel}. These models generally have a many-to-one encoder and a one-to-many decoder; that is, a long sequence of data (e.g., words or pixels) is encoded into a single latent vector, which is in turn decoded into a whole data sequence (see also \citet{roberts2018hierarchical} and \citet{pereira2018unsupervised} for examples on music score modeling and anomaly detection in energy time series, respectively). Even if those models can include a hierarchical structure at encoding and/or at decoding, they do not consider a temporal sequence of latent vectors.

All these latent-variable deep temporal models, the ones we detail and unify in the DVAE class, and the ones we do not detail, remain strongly connected, with a similar overall encoding-decoding architecture and possibly a similar inference and training VAE methodology. Therefore, we must keep in mind that some of the propositions made in the literature for one type of model can be adapted and be beneficial to the other.

\section{Aim, contributions, and outline of the paper}
\label{sec:intro-aim-contributions}

This paper aims to provide a comprehensive overview of DVAE models. The contributions of this paper are detailed as follows.

\vspace{0.25cm}
\noindent\textbf{We provide a formal definition of the general class of DVAEs}. We describe its main properties and characteristics and how this class is related to previous classical models, such as VAEs, RNNs, and SSMs. We discuss the structure of dependencies between the observed and latent random variables in DVAE pdfs, as well as how these dependencies are implemented with neural networks. We discuss the design of inference models considering the general methodology used to identify the actual dependencies of the latent variables at inference time. We also discuss the VLB computation for training DVAEs. All these points are presented in Chapter~\ref{sec:DVAE}. To the best of our knowledge, this is the first time this class of models has been presented in such a general and unified manner. 

\vspace{0.25cm}
\noindent\textbf{We provide a detailed and complete technical description of seven DVAE models selected from the literature}. In Chapter~\ref{sec:DKF}, we start with the deep Kalman filter (DKF) \citep{krishnan2015deep,krishnan2017structured}, which is a basic combination of an SSM with DNNs. Then, we examine the Kalman variational autoencoder (KVAE) \citep{fraccaro2017disentangled} in Chapter~\ref{sec:KVAE}, the stochastic recurrent neural network (STORN) \citep{bayer2014learning} in Chapter~\ref{sec:STORN}, the variational recurrent neural network (VRNN) \citep{chung2015recurrent, goyal2017z} in Chapter~\ref{sec:VRNN}, another type of stochastic recurrent neural network (SRNN) \citep{fraccaro2016sequential} in Chapter~\ref{sec:SRNN}, the recurrent variational autoencoder (RVAE) \citep{leglaive2020recurrent} in Chapter~\ref{sec:RVAE}, and finally the disentangled sequential autoencoder (DSAE) \citep{yingzhen2018disentangled} in Chapter~\ref{sec:DSAE}. 

We have spent effort on consistency of presentation. For all seven models that we detail, we first present the generative equations in time-step form and then for an entire data sequence. Then, we present the structure of the exact posterior distribution of the latent variables given the observed data and present the inference model as proposed in the original papers. Finally, we present the corresponding VLB. In the original papers, some parts of this complete picture are often overviewed or even missing (not always the same parts), independently of the authors' goodwill, because of lack of space. 

We discuss the links, similarities, and differences of the selected DVAE models. We comment on the choices of the authors of the reviewed papers regarding the inference model, its relation to the exact posterior distribution, and implementation issues. In the present review, we discuss only high-level implementation issues related to the general structure of the neural network that implements a given DVAE at generation or inference (e.g., the type of RNN). We do not discuss practical implementation issues (e.g., the number of layers), which are too low-level in the present technical review context. 

We have also spent some effort making the notations homogeneous across all models. This is valid for both the review of the seven detailed models and the other sections of the paper, including the general presentation of the DVAE class of models in Chapter~\ref{sec:DVAE}. For some models, we have changed the time indexation notation, and in some instances, the names of some variables compared to the original papers. We have taken great care to do that consistently in the generative model, the inference part, and the VLB so that these notation changes do not affect the essence and functioning of the model. Together with consistency of presentation, this enables us to better put in evidence the commonalities and differences across models and make their comparison easier. Notation remarks are specified in independent dedicated paragraphs throughout the paper to facilitate  connections with the original papers. 

\vspace{0.25cm}
\noindent In complement to the detailed review of the seven selected models, \textbf{we provide a more rapid overview of other DVAE models presented in the recent literature} in Chapter~\ref{sec:other-models}. 

\vspace{0.25cm}
\noindent\textbf{We relate the recent developments in DVAEs to the history and technical background of the classical models DVAEs are built on, namely VAEs, RNNs, and SSMs}. Although there are already many papers on VAEs, including tutorials, we present them in Chapter~\ref{sec:VAEs} because all subsequent DVAE models rely on the VAE methodology. Then, as the introduction of temporal models in the VAE framework is closely linked to RNNs and SSMs, we briefly present these two classes of models in Chapter~\ref{sec:RNNs-SSMs}. The unified notation that we use will help readers from different communities (e.g., machine learning, signal processing, and control theory) who are not familiar with the relations among VAEs, RNNs, and SSMs to discover them comfortably.  

\vspace{0.25cm}
\noindent\textbf{We provide a quantitative benchmark of the selected DVAE models in an analysis-resynthesis task, as well as qualitative examples of data generation}. We have reimplemented the seven DVAE models detailed in this review and evaluated them on two different datasets (speech signals and 3D human motion data). This benchmark is presented in Chapter~\ref{sec:experiments}. 

The performance comparison of the different models from the literature review is a difficult task for many reasons. First, all models are not evaluated on the same data. Then, a newly proposed model generally performs better than some previously proposed model(s), at least on some aspect(s), but this can depend on model tuning, task, data, and experimental setup. Moreover, the comparison performed with a subset of previous models is incomplete in essence. In short, an extended benchmark of DVAE models is not yet available in the literature. Conducting an extended benchmark is a huge endeavor, as there are many possible configurations for the models and many tasks for evaluating them. In particular, it is not yet clear how to evaluate the degree of ``disentanglement'' of the extracted latent space. 
The presented experiments are a first step in that direction. We plan to exploit and compare the models more extensively and on more complex tasks in future studies. For example, we compared three models in the recently proposed DVAE-based unsupervised speech enhancement method \citep{bie2021unsupervised}.

\textbf{The code reimplementing the seven DVAE models and used on the benchmark task is made available to the community.} A link to the open-source code and the best-trained models can be found at \url{https://team.inria.fr/robotlearn/dvae/}. We have also taken care, in the code, to follow the unified presentation and notation used in the paper, making it, hopefully, a useful and pedagogical resource.

\vspace{0.25cm}
\noindent\textbf{We provide a discussion to put the DVAE class of models into perspective}. We summarize the outcome of this review and discuss the future challenges and possible improvements of VAEs and DVAEs. This is presented in Chapter~\ref{sec:discussion}.

\vspace{0.25cm}
\noindent
In summary, we believe that comparison of models across papers is a difficult task in essence, regardless of the efforts spent by the authors of the original papers, because of the use of different notations, presentation lines, missing information, etc. We hope that the present review paper and accompanying code will enable the readers to access the technical substance of the different DVAE models, their connections with classical models, their cross-connections, and their unification in the DVAE class more rapidly and ``comfortably'' than by analyzing and comparing the original papers by themselves. 

\vspace{0.25cm}
We wish the reader to enjoy this DVAE tour.

\chapter{Variational Autoencoders}
\label{sec:VAEs}
In this section, we present the VAE and the associated methodology for model training and approximate posterior distribution estimation (i.e., inference) with variational methods \citep{Kingma2014,rezende2014stochastic}. An extended tutorial on VAEs can be found in \citeauthor{kingma2019introduction}'s \citeyearpar{kingma2019introduction} paper.

\section{Principle}
\label{sec:VAE-principle}

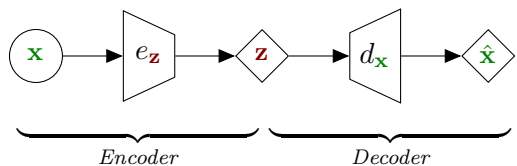
\begin{figure}
\centering
\begin{tikzpicture}[->]
\node[align=center] at (1.35,-1.2) {$\underbrace{\mbox{\hspace{3.2cm}}}_{\textit{Encoder}}$};
\node[align=center] at (4.7,-1.2) {$\underbrace{\mbox{\hspace{3.2cm}}}_{\textit{Decoder}}$};
\node[latent] (x) at (0,0) {$\myx$};
\node[net, rotate=0, fill=white, minimum width=12mm] (fenc) at (1.5,0) {$e_{\myz}$};
\node[det] (z) at (3,0) {$\myz$};
\node[net, rotate=180, fill=white, minimum width=12mm] (fdec) at (4.5,0) {\rotatebox{180}{$d_{\myx}$}};
\node[det] (xh) at (6,0) {$\myxh$};
\edge{x} {fenc};
\edge{z} {fdec};
\edge{fenc}{z};
\edge{fdec} {xh};
\end{tikzpicture}
\caption{Schematic representation of an AE. The left trapezoid represents a high-to-low-dimensional encoder DNN (denoted $e_{\myz}$), and the right trapezoid represents a low-to-high-dimensional decoder DNN (denoted $d_{\myx}$). Calculation of the latent variable $\myz$ and output $\myxh$ from input $\myx$ is deterministic. In line with probabilistic graphical models, deterministic variables are represented within diamonds.}
\label{fig:AE}
\end{figure}

For clarity of presentation, let us start with an autoencoder (AE). As illustrated in Figure~\ref{fig:AE}, an AE is a DNN that is trained to replicate an input vector $\myx \in \mathbb{R}^F$ at the output \citep{Hinton2006, vincent2010stacked}. At training time, the target output is thus set equal to $\myx$, and at test time, the output $\myxh$ is an estimated value of  $\myx$ (i.e., we have $\myxh \approx \myx$). An AE usually has a diabolo shape. The left part of the AE, the \emph{encoder}, provides a low-dimensional latent representation $\myz \in \mathbb{R}^L$ of the data vector $\myx$, with $L \ll F$, at the so-called bottleneck layer. The right part of the AE, the \emph{decoder}, tries to reconstruct $\myx$ from $\myz$. 
So far, everything is deterministic: At test time, each time the AE is fed with a specific input vector $\myx[0]$, it will provide the same corresponding output $\myxh[0]$.

The VAE was initially proposed by \citet{Kingma2014} and \citet{rezende2014stochastic}. It can be seen as a probabilistic version of an AE, where the output of the decoder is not directly a value of $\myx$ but the parameters of a probability distribution of $\myx$. As shown below, the same probabilistic formulation applies to the encoding of $\myz$. The resulting probabilistic model can be used to generate new data from new values of $\myz$. It can also be used to transform existing data within an encoding-modification-decoding scheme. For instance, the seminal papers on VAEs and many subsequent ones have considered image generation and transformation. Examples of speech/music signals transformation based on a VAE can be found in the literature \citep{blaauw2016modeling, hsu2017learning, esling2018bridging, roche2019, bitton2020neural}. Finally, it can be employed as a prior distribution of $\myx$ in more complex Bayesian models for, for example, speech enhancement  \citep{bando2017statistical, leglaive2018variance, pariente2019statistically, leglaive2019semi} or source separation \citep{kameoka2018semi}.

For clarity of presentation, at this point, it is convenient to separate the presentation of the VAE decoder (i.e., the generative model) and that of the VAE encoder (i.e., the inference model).

\section{VAE generative model}

In the following, $\mathcal{N}\big(\cdot; \boldsymbol{\mu}, \boldsymbol{\Sigma}\big)$ denotes a multivariate Gaussian distribution with mean vector $\boldsymbol{\mu}$ and covariance matrix $\boldsymbol{\Sigma}$, $\text{diag}\{\cdot\}$ is the operator that forms a diagonal matrix from a vector by putting the vector entries on the diagonal, $\mathbf{0}_L$ is the zero-vector of size $L$, and $\mathbf{I}_L$ is the identity matrix of size $L$. $p_{\theta_{\myx}}(\myx)$ is a generic notation for a parametric pdf of the random variable $\myx$, where $\theta_{\myx}$ is the set of parameters. It is equivalent to $p(\myx; \theta_{\myx})$.

\begin{figure}
\centering
\begin{tikzpicture}[->]
\node[align=center] at (2.35,-1) {$\underbrace{\mbox{\hspace{5.2cm}}}_{\textit{Encoder}}$};
\node[align=center] at (7.65,-1) {$\underbrace{\mbox{\hspace{5.2cm}}}_{\textit{Decoder}}$};
\node[obs] (x) at (0,0) {$\myx$};
\node[net, rotate=0, fill=white, minimum width=12mm] (fenc) at (1.5,0) {$e_{\myz}$};
\node[rnn, fill=white] (paramtilde) at (3, 0) {\parbox[]{1cm}{$\boldsymbol{\mu}_{\phi}(\myx)$ \\ $\boldsymbol{\sigma}_{\phi}(\myx)$}};
\node[latent] (z) at (5,0) {$\myz$};
\node[net, rotate=180, fill=white,minimum width=12mm] (fdec) at (6.5,0) {\rotatebox{180}{$d_{\myx}$}};
\node[rnn, fill=white] (param) at (8, 0) {\parbox[]{1cm}{$\boldsymbol{\mu}_{\theta}(\myz)$ \\ $\boldsymbol{\sigma}_{\theta}(\myz)$}};
\node[obs] (xh) at (10,0) {$\myx$};
\edge{x} {fenc};
\edge{z} {fdec};
\edge{fenc}{paramtilde};
\path(paramtilde) edge[dashed] (z) ;
\edge{fdec} {param};
\path(param) edge[dashed] (xh);
\end{tikzpicture}
\caption{Schematic representation of the VAE: Encoder (left) and decoder (right). Dashed lines represent a sampling process. In line with probabilistic graphical models, latent variables are represented within empty circles and observed variables are represented within shaded circles. When the encoder and decoder are cascaded, using the same variable name $\myx$ at both the input and output is an abuse of notation, but this is done to be more consistent with the separate encoder and decoder equations.}
\label{fig:VAE}
\end{figure}
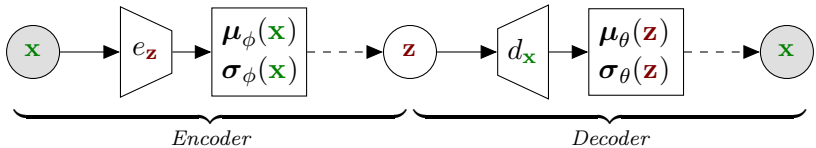

Formally, the VAE decoder is defined by 
\begin{align}
\label{eq:joint-pdf-1}
p_{\theta}(\myx,\myz) &= p_{\theta_{\myx}}(\myx|\myz)p_{\theta_{\myz}}(\myz),
\end{align}
with
\begin{align}
p_{\theta_{\myz}}(\myz) &=  \mathcal{N}(\myz;\mathbf{0}_L, \mathbf{I}_L), \label{eq:VAE-decoder-b}
\end{align}
and $p_{\theta_{\myx}}(\myx|\myz)$ is a parametric conditional distribution, the parameters of which are a nonlinear function of $\myz$ modeled by a DNN. This DNN is called the \emph{decoder network}, or the \emph{generation network}, and is parametrized by a set of weights and biases denoted $\theta_{\myx}$. In the standard VAE, the set of parameters $\theta_{\myz}$ is empty, but we write it explicitly to be coherent with the rest of the paper, and we have here $\theta=\theta_{\myx}\cup\theta_{\myz}=\theta_{\myx}$. The decoder network is illustrated in Figure~\ref{fig:VAE} (right). 

The VAE model and associated variational methodology was introduced by \citet{Kingma2014} in the general framework of parametric distributions, independently of the practical choice of the pdf $p_{\theta_{\myx}}(\myx|\myz)$ (and to a lesser extend of $p_{\theta_{\myz}}(\myz)$). The observed variable $\myx$ can be a continuous or discrete random variable with any arbitrary conditional distribution. The Gaussian case was then presented by \citet{Kingma2014} as a major example. Of course, other pdfs (rather than Gaussian) can be used depending on the nature of the data vector $\myx$. For example, Gamma distributions better fit the natural statistics of speech/audio power spectra \citep{girin2019notes}. For simplicity of presentation and consistency across models, in the present review, $p_{\theta_{\myx}}(\myx|\myz)$ is assumed to be a Gaussian distribution with diagonal covariance matrix for all models; that is,
\begin{align}
p_{\theta_{\myx}}(\myx|\myz) &= \mathcal{N}\big(\myx;\boldsymbol{\mu}_{\theta_{\myx}}(\myz),\text{diag}\{\boldsymbol{\sigma}_{\theta_{\myx}}^2(\myz)\}\big) \label{eq:VAE-decoder-a} \\ 
&= \prod_{f=1}^{F} p_{\theta_{\myx}}({\color{dgreen}x_{f}}|\myz) = \prod_{f=1}^{F}\mathcal{N}\big({\color{dgreen}x_{f}};\mu_{\theta_{\myx},f}(\myz),\sigma_{\theta_{\myx},f}^2(\myz)\big),
\end{align}
where the subscript $f$ denotes the $f$-th entry of a vector,  and $\boldsymbol{\mu}_{\theta_{\myx}}: \mathbb{R}^L \mapsto \mathbb{R}^F$ and $\boldsymbol{\sigma}_{\theta_{\myx}}: \mathbb{R}^L \mapsto \mathbb{R}_+^F$ are nonlinear functions of $\myz$ modeled by the decoder DNN. Although from a mathematical perspective, we could choose to work with full covariance matrices, assuming diagonal covariance matrices is preferable for computational reasons, since the number of free parameters of a covariance matrix grows quadratically with the variable dimension. This is problematic not only because we need to learn the neural network that computes all these parameters but also because covariance matrices often need to be inverted. The use of full covariance matrices also requires choosing an appropriate representation, for instance, based on the Cholesky decomposition. Please refer to Section~2.5.1 of~\citet{kingma2019introduction} for an extended discussion on this topic.

For maintaining consistency with the presentation of the other models in the next sections, we gather into $d_{\myx}$ the functions implemented by the decoder DNN; that is,
\begin{align}
[\boldsymbol{\mu}_{\theta_{\myx}}(\myz), \boldsymbol{\sigma}_{\theta_{\myx}}(\myz)] = d_{\myx}(\myz). \label{eq:VAE-decoder-a-bis} 
\end{align}

A VAE decoder can be considered a generalization of the probabilistic principal component analysis (PPCA)~\citep{tipping1999probabilistic} with a nonlinear (instead of linear) relationship between $\myz$ and the parameters $\theta_{\myx}$. It can also be considered the generalization of a generative mixture models, with a continuous conditional latent variable instead of a discrete one \citep{kingma2019introduction}. 
Indeed, the marginal distribution of $\myx$, $p_{\theta}(\myx)$, is given by
\begin{align}
p_{\theta}(\myx) &=  \int p_{\theta_{\myx}}(\myx|\myz)p_{\theta_{\myz}}(\myz)d\myz. \label{eq:marginal}
\end{align}
As any conditional distribution $p_{\theta_{\myx}}(\myx|\myz)$ can provide a mode, $p_{\theta}(\myx)$ can be highly multimodal (in addition to being potentially high-dimensional). Unlike PPCA, in VAEs, the posterior distribution cannot be written analytically and has to be approximated, as discussed in the next section.

\section{Learning with variational inference}
\label{sec:learning_with_VI}

Training the generative model defined in \eqref{eq:joint-pdf-1}--\eqref{eq:VAE-decoder-a} amounts to estimating the parameters $\theta$ so as to minimize the Kullback-Leibler (KL) divergence between the \emph{true data distribution} $p^\star(\myx)$ and the \emph{model distribution} (i.e., the marginal likelihood) $p_{\theta}(\myx)$:
\begin{align}
&\underset{\theta}{\min}\,\, \Big\{ D_{\text{KL}}\big( p^\star(\myx) \parallel p_\theta(\myx) \big) = \mathbb{E}_{p^\star(\myx)}\big[ \log p^\star(\myx) - \log p_\theta(\myx) \big] \Big\} \nonumber \\ 
\Leftrightarrow\,\, & \underset{\theta}{\max}\,\, \mathbb{E}_{p^\star(\myx)}\big[ \log p_\theta(\myx) \big]. 
\label{KL_min_training}
\end{align}
This equivalence relation shows that this definition of model training actually corresponds to the maximum (marginal) likelihood parameter estimation. In practice, the true data distribution $p^\star(\myx)$ is unknown, but we assume the availability of a training dataset $\myX = \{\myx[n] \in \mathbb{R}^F \}_{n=1}^{N}$, where the training examples $\myx[n]$ are independent and identically distributed (i.i.d.) according to $p^\star(\myx)$. Following the principle of empirical risk minimization, where the risk is here defined as the negative log-marginal likelihood, the intractable expectation in \eqref{KL_min_training} is replaced by a Monte Carlo estimate:
\begin{equation}
\underset{\theta}{\max}\,\, \frac{1}{N} \sum_{n=1}^{N} \log p_\theta(\myx[n]), \qquad \myx[n] \overset{i.i.d.}{\sim}  p^\star(\myx).
\label{emp_likelihood_training}
\end{equation}
The estimated model parameters can then be used, for example, to generate new data from \eqref{eq:joint-pdf-1}. 

For many generative models with latent variables, directly solving this optimization problem is difficult, if not impossible when the marginal likelihood is analytically intractable, because of the integral in \eqref{eq:marginal}, which cannot be computed in closed form. For the VAE, this intractability arises from the nonlinear relationship between the latent and observed variables, the latter being generated from the former through a DNN, which makes $p_{\theta_{\myx}}(\myx|\myz)$ in \eqref{eq:marginal} a nonlinear function of $\myz$. One standard approach then involves leveraging the latent variable nature of the model in to maximize a lower bound of the intractable log-marginal likelihood \citep{neal1998view}, which precisely depends on the \emph{posterior distribution} of the latent variables or its approximation. This strategy leads to the expectation-maximization (EM) algorithm \citep{dempster1977maximum} and its variants when the posterior distribution is intractable, such as Monte Carlo EM \citep{wei1990monte} and variational EM \citep{jordan1999introduction} algorithms.

As the name suggests, a VAE builds upon variational inference techniques, the general principles of which will now be briefly reviewed. Let $\mathcal{F}$ denote a \emph{variational family} defined as a set of pdfs over the latent variables $\myz$. For any variational distribution of pdf $q(\myz) \in \mathcal{F}$, the following decomposition of the log-marginal likelihood holds \citep{neal1998view}:
\begin{equation}
\log p_\theta(\myx) = \mathcal{L}(\theta, q(\myz); \myx) + D_{\text{KL}}\big( q(\myz) \parallel p_\theta(\myz | \myx) \big),
\label{log_likelihood_decomposition}
\end{equation}
where $\mathcal{L}(\theta, q(\myz); \myx)$ is referred to in the literature as the evidence lower bound (ELBO), the negative variational free energy, or the VLB, and is defined as
\begin{equation}
\mathcal{L}(\theta, q(\myz); \myx) = \mathbb{E}_{q(\myz)} \big[ \log p_\theta(\myx,\myz) - \log q(\myz) \big] \le \log p_\theta(\myx).
\label{varFreeEnergy}
\end{equation}
The inequality in \eqref{varFreeEnergy} is obtained from \eqref{log_likelihood_decomposition} based on the fact that $D_{\text{KL}}(\cdot \parallel \cdot) \ge 0$. Equality holds (i.e., the VLB is tight to the log-marginal likelihood) if and only if the variational distribution $q(\myz)$ is equal to the exact posterior distribution $p_\theta(\myz | \myx)$.

The EM algorithm \citep{dempster1977maximum} is an iterative algorithm that consists in alternatively maximizing the VLB with respect to $q(\myz) \in \mathcal{F}$ in the E-step and with respect to $\theta$ in the M-step \citep{neal1998view}. From \eqref{log_likelihood_decomposition}, we see that the E-step involves finding the variational distribution $q(\myz)$ in the variational family $\mathcal{F}$ that best approximates the true posterior $p_\theta(\myz | \myx)$ according to the KL divergence measure of fit:
\begin{equation}
q^\star(\myz) = \underset{q \in \mathcal{F}}{\arg\max}\,\, \mathcal{L}(\theta, q(\myz); \myx) = \underset{q \in \mathcal{F}}{\arg\min}\,\, D_{\text{KL}}\big( q(\myz) \parallel p_\theta(\myz | \myx) \big).
\label{E-Step}
\end{equation}
In the exact EM algorithm, the variational family $\mathcal{F}$ is unconstrained, so the solution to the E-step is given by the exact posterior distribution: $q^\star(\myz) = p_\theta(\myz|\myx)$. As this optimal variational distribution over $\myz$ is actually conditioned on $\myx$, we will now use the notation $q(\myz | \myx)$ instead of $q(\myz)$. The difficulty arises when the posterior distribution $p_\theta(\myz|\myx)$ is intractable, which prevents us from solving the E-step analytically. Variational inference then consists in \emph{constraining the variational family $\mathcal{F}$} and resorting to optimization methods for solving the E-step \citep{jordan1999introduction}. 

Seminal works on variational inference relied on the so-called mean-field approximation, which constrains the variational family $\mathcal{F}$ to be a set of completely factorized distributions (i.e., multivariate distributions over $\myz$ that are written as a product of univariate marginal distributions over the entries of $\myz$). All marginal posterior dependencies between different entries of $\myz$ are ignored here, while the ``structured'' mean-field approximation \citep{saul1996exploiting} partially restores some of them. Solving the E-step under the mean-field approximation leads to a set of closed-form coupled solutions for each univariate distribution in the factorization. This approach is also referred to as coordinate-ascent variational inference in the literature \citep{BishopBook, blei2017variational}. However, closed-form updates are usually only available for conjugate-exponential models \citep{winn2005variational} when the distribution of each scalar latent variable, conditionally on its parents, belongs to the exponential family and is conjugate with respect to the distribution of these parent variables. Moreover, this coordinate-ascent approach does not scale well for high-dimensional and large-scale inference problems \citep{hoffman2013stochastic}.

An alternative to the mean-field approximation is then to define the variational family as a set of distributions with a certain parametric form $q_\lambda(\myz | \myx)$, where the parameters $\lambda$ govern the shape of the distribution. For example, we can define the Gaussian variational family where the parameters $\lambda$ correspond to the mean vector and covariance matrix:
\begin{equation}
\mathcal{F} = \big\{ q_\lambda(\myz | \myx) = \mathcal{N}(\myz; \boldsymbol{\mu}, \boldsymbol{\Sigma}), \, \lambda = \{\boldsymbol{\mu}, \boldsymbol{\Sigma}\} \big\}.
\end{equation}
As shown below, the optimal parameters $\lambda$ that maximize the VLB depend on $\myx$ and $\theta$.
This approach is called fixed-form or structured variational inference \citep{honkela2010approximate, salimans2013fixed}. The VLB in \eqref{varFreeEnergy} then becomes a function of both the generative model parameters $\theta$ and \emph{variational parameters} $\lambda$: 
\begin{equation}
\mathcal{L}(\theta, \lambda; \myx) = \mathbb{E}_{q_\lambda(\myz | \myx)} \big[ \log p_\theta(\myx,\myz) - \log q_\lambda(\myz | \myx) \big].
\label{varFreeEnergyParam}
\end{equation}
The E-step in \eqref{E-Step} consequently reduces to a parametric optimization problem:
\begin{equation}
\lambda^\star = \underset{\lambda}{\arg\max}\,\, \mathcal{L}(\theta, \lambda; \myx) = \underset{\lambda}{\arg\min}\,\, D_{\text{KL}}\big( q_\lambda(\myz | \myx) \parallel p_\theta(\myz | \myx) \big).
\end{equation}
As the objective function depends on the observed data vector $\myx$ and the generative model parameters $\theta$, so does the solution $\lambda^\star$. In fact, the optimal variational distribution depends on $\myx$ through the parameters $\lambda^\star$. The M-step remains unchanged in fixed-form variational inference; that is, it consists in updating the generative model parameters by maximizing $\mathcal{L}(\theta, \lambda; \myx)$ w.r.t.~$\theta$, using the current estimate of the variational parameters. If the expectation in \eqref{varFreeEnergyParam} and its gradient w.r.t.~$\lambda$ can be computed analytically, the optimization problem of the E-step can be solved using gradient-based optimization methods.

In general, given a dataset of i.i.d. data vectors $\myX = \{\myx[1],...,\myx[N]\}$, one needs to find the parameters $\Lambda = \{\lambda_1,...,\lambda_N\}$ of the variational distributions $q_{\lambda_n}(\myz[n] | \myx[n])$,  $n=1,...,N$. Taking the same example as before, with $q_{\lambda_n}(\myz[n] | \myx[n]) = \mathcal{N}(\myz[n]; \boldsymbol{\mu}_n, \boldsymbol{\Sigma}_n)$, we have here $\lambda_n = \{\boldsymbol{\mu}_n, \boldsymbol{\Sigma}_n\}$.  This problem is solved by maximizing the following total VLB, which is the sum (or equivalently, the mean) of the local VLB defined in \eqref{varFreeEnergyParam} over each vector in the training dataset:
\begin{equation}
\mathcal{L}(\theta, \Lambda; \myX) = \sum_{n=1}^{N} \mathcal{L}(\theta, \lambda_n; \myx[n]).
\label{VLB_sum_dataset}
\end{equation}
To scale to large amounts of data, \emph{stochastic variational inference} \citep{hoffman2013stochastic} relies on gradient-based stochastic optimization \citep{robbins1951stochastic, Bottou2004} for maximizing the total VLB in \eqref{VLB_sum_dataset} w.r.t. the generative model parameters $\theta$. The gradient of the total VLB, $\mathcal{L}(\theta, \Lambda; \myX)$, is the sum of the gradients of the local VLBs, $\mathcal{L}(\theta, \lambda_n; \myx[n])$, defined for each sample $\myx[n]$ in the dataset. For large datasets, computing this sum to perform a single update of $\theta$ with a step of gradient ascent can be inefficient. Therefore, stochastic variational inference exploits a noisy stochastic estimate of the gradient, computed from a single example $\myx[n]$ or from a mini-batch of examples in the dataset. This is the same principle as that used in stochastic and mini-batch gradient descent \citep{Bottou2004}, such that stochastic variational inference inherits from the same convergence properties \citep{robbins1951stochastic}.

However, the estimation of the complete set of variational parameters can remain expensive for large datasets. Thus, \emph{amortized variational inference} makes a stronger assumption for defining the variational family, by introducing an \emph{inference model} $f_\phi$ such that
\begin{equation}
\lambda_n = f_\phi(\myx[n]),
\end{equation}
where $\phi$ is a set of parameters that is shared among all variational distributions $q_{\lambda_n}(\myz[n] | \myx[n])$. This inference model is used to map the observation $\myx[n]$ to the local variational parameter $\lambda_n$. The variational family $\mathcal{F}$ then corresponds to the set of variational distributions parametrized by $\phi$, which are denoted by $q_\phi(\myz[n] | \myx[n])$. For instance, for $q_\phi(\myz[n] | \myx[n]) = \mathcal{N}(\myz[n]; \boldsymbol{\mu}_n, \boldsymbol{\Sigma}_n)$, we have $\lambda_n = [\boldsymbol{\mu}_n, \boldsymbol{\Sigma}_n] = f_\phi(\myx[n])$. This amortization principle corresponds to a stronger assumption for the variational family compared to nonamortized fixed-form and mean-field approximations. Therefore, the KL divergence between the exact posterior and its approximation is likely to be larger in the amortized case than in the previous cases. The total VLB for the complete training dataset then becomes a function of $\phi$:
\begin{equation}
\mathcal{L}(\theta, \phi; \myX) = \sum_{n=1}^{N} \mathbb{E}_{q_\phi(\myz[n] | \myx[n])} \big[ \log p_\theta(\myx[n],\myz[n]) - \log q_\phi(\myz[n] | \myx[n]) \big].
\label{eq:VLB-a}
\end{equation}
This means that the optimization of the set of local variational parameters $\Lambda = \{\lambda_1,...,\lambda_N\}$ is replaced by the optimization of the shared set of inference model parameters $\phi$. Hereinafter, we will use the term inference model to directly denote the variational distribution $q_\phi(\myz[n] | \myx[n])$.

\section{VAE inference model}

VAEs belong to the family of amortized variational inference techniques, where the VLB in \eqref{eq:VLB-a} is optimized using stochastic gradient-based optimization techniques. The VAE generative model $p_\theta(\myx,\myz)$ has already been defined in \eqref{eq:joint-pdf-1}--\eqref{eq:VAE-decoder-a}. It involves a decoder neural network through $p_{\theta}(\myx | \myz)$. To fully specify the VLB, which is required to learn the generative model parameters $\theta$, it is also necessary to define the inference model $q_\phi(\myz | \myx)$, which approximates the intractable exact posterior $p_{\theta}(\myz | \myx)$.

Similar to the generative model, the inference model for $q_\phi(\myz | \myx)$ is defined by an \emph{encoder neural network}. A common choice for the approximate posterior distribution $q_{\phi}(\myz | \myx)$ is to use a Gaussian distribution:
\begin{align}
q_{\phi}(\myz | \myx) &= \mathcal{N}\big(\myz; \boldsymbol{\mu}_{\phi}(\myx), \text{diag}\{\boldsymbol{\sigma}_{\phi}^2(\myx)\} \big) \label{eq:VAE-encoder-a} \\
&= \prod_{l=1}^{L} q_{\phi}({\color{dred}z_l} | \myx) = \prod_{l=1}^{L} \mathcal{N}\big({\color{dred}z_l}; \mu_{\phi, l}(\myx), \sigma_{\phi, l}^2(\myx) \big),
\end{align}
where index $l \in \{1,...,L\}$ is used to denote the $l$-th entry of the corresponding vectors, and $\boldsymbol{\mu}_{\phi}: \mathbb{R}^{F} \mapsto \mathbb{R}^L$ and  $\boldsymbol{\sigma}_{\phi}: \mathbb{R}^{F} \mapsto \mathbb{R}_+^L$ are nonlinear functions of $\myx$, modeled by a DNN called the encoder or recognition network, which is parametrized by a set of weights and biases denoted by $\phi$. The encoder network is illustrated in Figure~\ref{fig:VAE} (left). As for the VAE generative model, for the sake of consistency with the presentation of the other models, we denote
\begin{align}
[\boldsymbol{\mu}_{\phi}(\myx), \boldsymbol{\sigma}_{\phi}(\myx)] = e_{\myz}(\myx), \label{eq:VAE-encoder-a-bis}
\end{align}
where $e_{\myz}$ is the nonlinear function implemented by the encoder DNN.

\section{VAE training}
\label{sec:vae_training}

In the VAE methodology \citep{Kingma2014, rezende2014stochastic}, the VLB in \eqref{eq:VLB-a} is optimized using stochastic gradient-based optimization techniques to learn the generative and inference model parameters. For training the VAE, the encoder and decoder networks are cascaded, as illustrated in Figure~\ref{fig:VAE}, and the sets of parameter $\theta$ and $\phi$ are jointly estimated from the training data $\myX$. This is different from an EM algorithm strategy, which would alternatively optimize the VLB w.r.t. $\phi$ and $\theta$ in the E- and M-steps, respectively. \citet{he2018lagging} showed that this joint encoder-decoder training of the VAE can, however, be suboptimal. 

The VLB  in \eqref{eq:VLB-a} can be reshaped as \citep{Kingma2014}
\begin{align}
\mathcal{L}(\theta, \phi ; \myX)  &= \underbrace{\sum_{n=1}^{N_{}} \mathbb{E}_{q_{\phi}(\myz[n] | \myx[n])}\big[ \log p_{\theta_{\myx}}(\myx[n] | \myz[n]) \big]}_{\text{Reconstruction accuracy}}  \nonumber \\   & \qquad \qquad \underbrace{- \sum_{n=1}^{N_{}} D_{\textit{KL}}\big(q_{\phi}(\myz[n] | \myx[n]) \parallel p_{\theta_{\myz}}(\myz[n])\big) }_{\text{Regularization}}. \label{eq:VLB-b}
\end{align} 
The first term on the right-hand side of \eqref{eq:VLB-b} is a reconstruction term that represents the average accuracy of the chained encoding-decoding process. For instance, if the generative model $p_{\theta_{\myx}}(\myx | \myz)$ is chosen to be Gaussian with an identity covariance matrix, the reconstruction term is equal to the opposite of the mean-squared error (MSE) between the original data and decoder output, up to additive constants. The second term is a regularization one, which enforces the approximate posterior distribution $q_{\phi}(\myz | \myx)$ to be close to the prior distribution $p_{\theta_{\myz}}(\myz)$. Provided that an independent Gaussian prior is used, this term forces $\myz$ to be a disentangled data representation; that is, the $\myz$ entries tend to be independent and encode a different characteristic (or factor of variation) of the data.

For usual distributions, the regularization term has an analytical expression as a function of $\theta$ and $\phi$. However, the expectation taken with respect to $q_\phi(\myz[n]|\myx[n])$ in the reconstruction accuracy term is analytically intractable. Therefore, in practice, it is approximated using a Monte Carlo estimate with $R$ samples $\myzr[n]$ independently and identically drawn from $q_\phi(\myz[n]|\myx[n])$ (for each index $n$):
\begin{equation} 
\mathbb{E}_{q_\phi(\myz[n]|\myx[n])} [\textnormal{log}\,p_\theta(\myx[n]|\myz[n])] \approx  \frac{1}{R} \sum_{r=1}^{R} \textnormal{log}\,p_\theta(\myx[n]|\myzr[n]).
\end{equation}  
The resulting Monte Carlo estimate of the VLB is given by
\begin{align}
\hat{\mathcal{L}}(\theta, \phi ; \myX) &= \sum_{n=1}^{N} \frac{1}{R} \sum_{r=1}^{R} \log p_{\theta}(\myx[n] | \myzr[n]) \ - \sum_{n=1}^{N} D_{\textit{KL}}\big(q_{\phi}(\myz[n] | \myx[n]) \parallel p(\myz[n])\big). \label{eq:VLB-c}
\end{align} 

To optimize this objective function, we can typically resort to the (variants of) stochastic or mini-batch gradient descent (on the negative VLB) \citep{Bottou2004}. While the gradient of $\hat{\mathcal{L}}(\theta, \phi ; \myX)$ w.r.t.~$\theta$ can be easily computed using the standard backpropagation algorithm, that w.r.t.~$\phi$ is problematic because the sampling operation from $q_\phi(\myz[n]|\myx[n])$ is not differentiable w.r.t.~$\phi$. The solution to this problem, proposed by \citet{Kingma2014} and referred to as the \emph{reparameterization trick}, consists in reparametrizing the sample $\myzr[n]$ using a differentiable transformation of a sample $\boldsymbol{\epsilon}^{(r)}$ drawn from a standard Gaussian distribution, which does not depend on $\phi$:
\begin{equation}
\myzr[n] = \boldsymbol{\mu}_{\phi}(\myx[n]) +  \text{diag}\{\boldsymbol{\sigma}_{\phi}^2(\myx[n])\}^{\frac{1}{2}} \boldsymbol{\epsilon}^{(r)}, \qquad \boldsymbol{\epsilon}^{(r)} \sim \mathcal{N}(\mathbf{0}_L, \mathbf{I}_L).
\end{equation}
Using this reparameterization trick, $\hat{\mathcal{L}}(\theta, \phi ; \myX)$ is now differentiable w.r.t.~$\phi$. This differentiable Monte Carlo approximation of the VLB is referred to as the stochastic
gradient variational Bayes (SGVB) estimator  \citep{Kingma2014}. The gradient of $\hat{\mathcal{L}}(\theta, \phi ; \myX)$ w.r.t.~$\phi$ is an unbiased estimate of the gradient of the exact VLB $\mathcal{L}(\theta, \phi ; \myX)$ \citep{kingma2019introduction}. This property allows using very few samples to compute the SGVB estimator, which however impacts the variance of the estimator. \citet{Kingma2014} suggested setting $R=1$ provided that sufficiently large mini-batches are used for the gradient descent. This training procedure of a VAE model is now considered routine within deep learning toolkits, such as TensorFlow \citep{abadi2016tensorflow} and PyTorch \citep{paszke2019pytorch}.

\chapter{Recurrent Neural Networks and State Space Models}
\label{sec:RNNs-SSMs}
As mentioned earlier, DVAEs are formed of combinations of a VAE and temporal models. Most of these temporal models rely on RNNs and/or SSMs. We thus briefly present the basics of RNNs and SSMs in this chapter before moving on to DVAEs in the next chapters. An extended technical overview of RNNs and SSMs, as well as their applications, is beyond of the scope of the present paper.

\section{Recurrent Neural Networks}
\label{sec:RNNs}

\subsection{Principle and definition}
\label{sec:RNNs-principle}

RNNs have been and are still widely used for data sequence modeling and generation and sequence-to-sequence mapping. An RNN is a neural network that processes ordered vector sequences and uses a memory of past input/output data to condition the current output \citep{sutskever2013training, graves2013speech}. This is achieved using an additional vector that recursively encodes the internal state of the network. 

We denote by $\myx[t_1:t_2]=\{\myx[t]\}_{t=t_1}^{t_2}$ a sequence of vectors $\myx[t]$ indexed from $t_1$ to $t_2$, where $t_1 \leq t_2$. When $t_1 > t_2$, we assume $\myx[t_1:t_2]=\emptyset$.  
We present RNNs in the general framework of nonlinear systems, which transform an input vector sequence $\myu[1:T]$ into an output vector sequence $\myx[1:T]$, possibly through an internal state vector sequence $\myh[1:T]$. The input, output, and internal state vectors can have arbitrary (different) dimensions. If $\myu[1:T]$ is an ``external'' input sequence, the network can be considered as a ``system,'' as is usual in control theory ($\myu[1:T]$ being considered as a command to the system). If $\myu[t] = \emptyset$, the RNN is in the \emph{undriven} mode. In contrast, if $\myu[t] = \myx[t-1]$, the RNN is in the \emph{predictive} mode, or \emph{sequence generation} mode, which is a usual mode when we are interested in modeling the evolution of a data sequence $\myx[1:T]$ ``alone'' (i.e., independently of any external input; in this case, $\myx[1:T]$ can be seen both as an input and an output sequence).

A basic single-layer RNN model is defined by
\begin{align}
\myh[t] &= d_{\hid}(\mathbf{W}_{\inp}\myu[t] + \mathbf{W}_{\rec}\myh[t-1] + \mathbf{b}_{\hid}), \\
\myx[t] &= d_\out(\mathbf{W}_\out\myh[t] + \mathbf{b}_\out),
\end{align}
where $\mathbf{W}_{\inp}$, $\mathbf{W}_{\rec}$ and $\mathbf{W}_\out$ are weight matrices of appropriate dimensions; $\mathbf{b}_{\hid}$ and $\mathbf{b}_\out$ are bias vectors; and  $d_{\hid}$ and $d_\out$ are nonlinear activation functions. We also define the initial internal state vector $\myh[0]$. This model is extendable to more complex recurrent architectures:
\begin{align}
\myh[t] &= d_{\myh}(\myu[t], \myh[t-1]), \label{eq:RNN-a} \\
\myx[t] &= d_{\myx}(\myh[t]), \label{eq:RNN-b}
\end{align}
where $d_{\myh}$ and $d_{\myx}$ denote any arbitrary complex nonlinear functions implemented with a DNN. We assume that this representation includes long short-term memory (LSTM) networks \citep{hochreiter1997long} and gated recurrent unit (GRU) networks \citep{cho2014learning}, which comprise additional internal variables called gates. For simplicity of presentation, these additional internal gates are not formalized in \eqref{eq:RNN-a} and \eqref{eq:RNN-b}. The same is true for multi-layer RNNs, where several recursive layers are stacked on top of each other \citep{graves2013speech} (in this case, for the same reason, we do not report layer indexes in \eqref{eq:RNN-a} and \eqref{eq:RNN-b}). This is also true for combinations of multi-layer RNNs and LSTMs (i.e., multi-layer LSTM networks). In summary, we assume that \eqref{eq:RNN-a} and \eqref{eq:RNN-b} are a ``generic'' or ``high-level'' representation of an RNN of arbitrary complexity.

\vspace{0.3cm}
\noindent \textbf{Notation remark}: To clarify the presentation and links between the different models, we use the same generic notation $d_{\myx}$ for the generating function in \eqref{eq:VAE-decoder-a-bis} and \eqref{eq:RNN-b}, and will do that throughout the paper (and the same for $d_{\myh}$ and for $d_{\myz}$ later in the paper). 
\vspace{0.3cm}

So far, the above RNNs are deterministic: given $\myu[1:T]$ and $\myh[0]$, $\myx[1:T]$ is completely determined.
Such networks are trained by optimizing a deterministic criterion, e.g. the MSE between the target output sequences from a training dataset and the corresponding actual output sequences obtained by the network. The training set of i.i.d.~vectors used for VAE training is replaced with that of vector sequences, and consecutive vectors within a training sequence are generally correlated, which is the point of using a dynamical model.

\subsection{Generative recurrent neural networks}

 Deterministic RNNs can easily be transformed into \emph{generative} RNNs (GRNNs) by adding stochasticity at the output level. We just have to define a probabilistic observation model and replace the output data sequence with an output sequence of distribution parameters, similar to the VAE decoder:
\begin{align}
\myh[t] &= d_{\myh}(\myu[t], \myh[t-1]), \label{eq:GRNN-a} \\
[\boldsymbol{\mu}_{\theta_{\myx}}(\myh[t]),\boldsymbol{\sigma}_{\theta_{\myx}}(\myh[t])] &= d_{\myx}(\myh[t]), \label{eq:GRNN-b} \\
p_{\theta_{\myx}}(\myx[t] | \myh[t]) &= \mathcal{N}\big(\myx[t];\boldsymbol{\mu}_{\theta_{\myx}}(\myh[t]),\text{diag}\{\boldsymbol{\sigma}_{\theta_{\myx}}^2(\myh[t])\}\big). \label{eq:GRNN-c} 
\end{align}
Eq.~\eqref{eq:GRNN-a} is the same recursive internal state model as \eqref{eq:RNN-a}. Eqs.~\eqref{eq:GRNN-b} and \eqref{eq:GRNN-c} constitute the observation model. In \eqref{eq:GRNN-c}, we use the Gaussian distribution for its generality and for the convenience of illustration, although any distribution can be used, just as for the VAE decoder. Again, one may choose a distribution that is more appropriate for the nature of the data. For example, \citet{graves2013generating} proposed using mixture distributions. 
The complete set of model parameters $\theta$ here includes $\theta_{\myh}$ and $\theta_{\myx}$, the parameters of the networks implementing $d_{\myh}$ and $d_{\myx}$, respectively. Because the output of $d_{\myx}$ in \eqref{eq:GRNN-b} is now two vectors of pdf parameters instead of a data vector in \eqref{eq:RNN-b}, its size is twice that of the deterministic RNN. When the internal state vector $\myh[t]$ is of (much) lower dimension than the output vector $\myx[t]$, the GRNN observation model becomes similar to the VAE decoder, except that, again, $\myh[t]$ has a deterministic evolution through time, whereas the latent state $\myz$ of the VAE is stochastic and i.i.d., which is a fundamental difference.

Even if the generation of $\myx[t]$ is now stochastic, the evolution of the internal state is still deterministic. Let us denote $\myh[t]$ as a function $\myh[t] = \myh[t](\myu[1:t])$ to make the deterministic relation between $\myu[1:t]$ and $\myh[t]$ explicit (for each time index $t$).\footnote{$\myh[t](\myu[1:t])$ also depends on the initial internal state vector $\myh[0]$, but we omit this term as an argument of the function for conciseness.} We thus have $p_{\theta_{\myx}}(\myx[t] | \myh[t]) = p_{\theta_{\myx}}(\myx[t] | \myh[t](\myu[1:t]))$. In the predictive mode, we have $p_{\theta_{\myx}}(\myx[t] | \myh[t]) = p_{\theta_{\myx}}(\myx[t] | \myh[t](\myx[0:t-1]))$.\footnote{Here,  the first ``input'' $\myx[0]$ has to be set arbitrarily, just like $\myh[0]$. Alternately, one can directly start the generation process from an arbitrary internal state vector $\myh[1]$.}
Such stochastic version of the RNN can be trained with a statistical criterion (e.g., maximum likelihood). As for the VAE training, we search for the maximization of the observed data log-likelihood w.r.t.~$\theta$ over a set of training sequences. For one sequence, with the conditional independence of successive data vectors, the data log-likelihood is given by
\begin{align}
\log p_{\theta_{\myx}}(\myx[1:T] | \myu[1:T]) 
&= \sum_{t=1}^{T}  \log  p_{\theta_{\myx}}\big(\myx[t] |\myh[t](\myu[1:t])\big).
\end{align}

\section{State Space Models}
\label{sec:SSMs}

\subsection{Principle and definition}
\label{sec:SSM-principle}

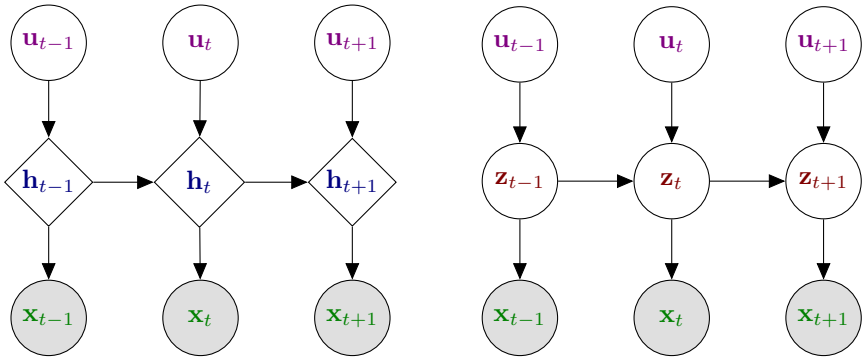
\begin{figure}
\centering
\begin{tikzpicture}[->]
    \node[latent,minimum size=10mm] (um) {$\myu[t-1]$};
    \node[latent,minimum size=10mm,right=of um] (u) {$\myu[t]$};
    \node[latent,minimum size=10mm,right=of u] (up) {$\myu[t+1]$};
    \node[det,minimum size=10mm,below=of um,yshift=2.5mm] (hm) {$\myh[t-1]$};
    \node[det,minimum size=12mm,right=of hm,xshift=-2mm] (h) {$\myh[t]$};
    \node[det,minimum size=10mm,right=of h,xshift=-1.7mm] (hp) {$\myh[t+1]$};
    \node[obs,minimum size=10mm,below=of hm,yshift=3mm] (xm) {$\myx[t-1]$};
    \node[obs,minimum size=10mm,right=of xm] (x) {$\myx[t]$};
    \node[obs,minimum size=10mm,right=of x] (xp) {$\myx[t+1]$};
    \edge {um} {hm};
    \edge {hm} {xm,h};
    \edge{u} {h};
    \edge{h} {x,hp};
    \edge{up} {hp};
    \edge{hp} {xp};
\end{tikzpicture}\hspace{10mm}
\begin{tikzpicture}[->]
    \node[latent,minimum size=10mm] (um) {$\myu[t-1]$};
    \node[latent,minimum size=10mm,right=of um] (u) {$\myu[t]$};
    \node[latent,minimum size=10mm,right=of u] (up) {$\myu[t+1]$};
    \node[latent,minimum size=10mm,below=of um, yshift=2mm] (zm) {$\myz[t-1]$};
    \node[latent,minimum size=10mm,right=of zm] (z) {$\myz[t]$};
    \node[latent,minimum size=10mm,right=of z] (zp) {$\myz[t+1]$};
    \node[obs,minimum size=10mm,below=of zm, yshift=2mm] (xm) {$\myx[t-1]$};
    \node[obs,minimum size=10mm,right=of xm] (x) {$\myx[t]$};
    \node[obs,minimum size=10mm,right=of x] (xp) {$\myx[t+1]$};
    \edge {um} {zm};
    \edge {zm} {xm,z};
    \edge{u} {z};
    \edge{z} {x,zp};
    \edge{up} {zp};
    \edge{zp} {xp};
\end{tikzpicture}
\caption{GRNN (left) against SSM (right): The two models have an identical structure, though the internal state of the GRNN is deterministic (represented with a diamond), whereas that of the SSM is stochastic (represented with a circle).}
\label{fig:GRNN-SSM}
\end{figure}

SSMs are a rich family of models that are widely used to model dynamical systems (e.g., in statistical signal processing, time-series analysis, and control theory) \citep{durbin2012time}. Here, we focus on discrete-time, continuous-valued SSMs of the form 
\begin{align}
[\boldsymbol{\mu}_{\theta_{\myz}}(\myz[t-1],\myu[t]), \boldsymbol{\sigma}_{\theta_{\myz}}(\myz[t-1],\myu[t])] &= d_{\myz}(\myz[t-1],\myu[t]), \label{eq:SSM-a} \\
p_{\theta_{\myz}}(\myz[t] | \myz[t-1],\myu[t]) &= \mathcal{N}\big(\myz[t]; \boldsymbol{\mu}_{\theta_{\myz}}(\myz[t-1],\myu[t]), \text{diag}\{\boldsymbol{\sigma}_{\theta_{\myz}}^2(\myz[t-1],\myu[t])\}\big), \label{eq:SSM-b} \\
[\boldsymbol{\mu}_{\theta_{\myx}}(\myz[t]),\boldsymbol{\sigma}_{\theta_{\myx}}(\myz[t])] &= d_{\myx}(\myz[t]),  \label{eq:SSM-c} \\
p_{\theta_{\myx}}(\myx[t] | \myz[t]) &= \mathcal{N}\big(\myx[t];\boldsymbol{\mu}_{\theta_{\myx}}(\myz[t]),\text{diag}\{\boldsymbol{\sigma}_{\theta_{\myx}}^2(\myz[t])\}\big),  \label{eq:SSM-d} 
\end{align}
where $d_{\myz}$ and $d_{\myx}$ are functions of arbitrary complexity, each being parameterized by a set of parameters denoted $\theta_{\myz}$ and $\theta_{\myx}$, respectively. As for the complete generative model, we have $\theta = \theta_{\myx} \cup \theta_{\myz}$, and we retain this notation hereinafter. At this point, $d_{\myz}$ and $d_{\myx}$ can be linear or nonlinear functions, and we will differentiate the two cases later. The observation model \eqref{eq:SSM-c}--\eqref{eq:SSM-d} is very similar to the GRNN observation model \eqref{eq:GRNN-b}--\eqref{eq:GRNN-c}. However, $\myz[t]$ is here a stochastic internal state vector in contrast to the deterministic internal state $\myh[t]$ of the (G)RNN. The distribution of $\myz[t]$, known as the \textit{state model} or the \textit{dynamical model}, is given by \eqref{eq:SSM-a}--\eqref{eq:SSM-b}. It follows a first-order Markov model; that is, a temporal dependency is introduced where $\myz[t]$ depends on the previous state $\myz[t-1]$ and the corresponding input $\myu[t]$ through the function $d_{\myz}$. In short, the above SSM can be considered a GRNN in which the deterministic internal state $\myh[t]$ is replaced with a stochastic internal state $\myz[t]$, as illustrated in Figure~\ref{fig:GRNN-SSM}.

\vspace{0.3cm}
\noindent \textbf{Notation remark}: In the control theory literature, the input corresponding to the generation of $\myz[t]$ is often denoted as $\myu[t-1]$, or equivalently, $\myu[t]$ is used to generate the next state $\myz[t+1]$. This notation is arbitrary. In the present paper, we prefer to realign the temporal indices, so that the input $\myu[t]$ is used to generate $\myz[t]$, which in turn is used to generate $\myx[t]$, to maintain better consistency through all presented models. 

\vspace{0.3cm}
As for the complete sequence, given the dependencies represented in Figure~\ref{fig:GRNN-SSM}, the joint distribution of all variables can be expressed as
\begin{align}
p_{\theta}(\myx[1:T], \myz[1:T], \myu[1:T]) &= \prod_{t=1}^{T} p_{\theta_{\myx}}(\myx[t] | \myz[t])p_{\theta_{\myz}}(\myz[t] | \myz[t-1], \myu[t])p(\myu[t]), \label{eq:SSM-joint-seq}
\end{align}
from which we can deduce
\begin{align}
p_{\theta_{\myx}}(\myx[1:T] | \myz[1:T]) &= \prod_{t=1}^{T} p_{\theta_{\myx}}(\myx[t] | \myz[t]), \label{eq:SSM-seq-a}
\end{align}
and
\begin{align}
p_{\theta_{\myz}}(\myz[1:T] | \myu[1:T]) &= \prod_{t=1}^{T} p_{\theta_{\myz}}(\myz[t] | \myz[t-1], \myu[t]). \label{eq:SSM-seq-b}
\end{align}
Given the state sequence $\myz[1:T]$, the observation vectors at different time frames are mutually independent. The prior distribution of $\myu[t]$ also factorizes across time frames, but this is of limited interest here. To be complete, we should specify the model ``initialization'': At $t=1$, we need to define $\myz[0]$, which can be set to an arbitrary deterministic value, or defined through a prior distribution $p_{\theta_{\myz}}(\myz[0])$ (which then must be added to the right-hand side of \eqref{eq:SSM-joint-seq} and \eqref{eq:SSM-seq-b}), or we can set $\myz[0] = \emptyset$, in which case the first term of the state model in these equations is $p_{\theta_{\myz}}(\myz[1] | \myu[1])$. 

Solving the above SSM means that we run the inference process; that is, we estimate the state vector sequence $\myz[1:T]$ from an observed data vector sequence $\myx[1:T]$. The use of Gaussian distribution in \eqref{eq:SSM-b} and \eqref{eq:SSM-d} is a convenient choice that generally facilitates inference. More generally, these distributions are within the exponential family, so either exact or approximate inference algorithms can be applied, depending on the nature of $d_{\myz}$ and $d_{\myx}$. In the next subsection, we provide an example of a closed-form inference solution when $d_{\myz}$ and $d_{\myx}$ are linear functions.

\subsection{Kalman filters}
\label{subsec:KF}

Some classical SSMs have been successfully used for decades for a wide set of applications. For example, when $d_{\myx}$ and $d_{\myz}$ are linear functions of the form
\begin{align}
\boldsymbol{\mu}_{\theta_{\myz}}(\myz[t-1],\myu[t]) &= \mathbf{A}_t \myz[t-1] + \mathbf{B}_t \myu[t] + \mathbf{m}_t, \quad \boldsymbol{\sigma}_{\theta_{\myz}}^2(\myz[t-1],\myu[t]) = \boldsymbol{\Lambda}_t,  \label{eq:LDS-a} \\
\boldsymbol{\mu}_{\theta_{\myx}}(\myz[t]) &= \mathbf{C}_t  \myz[t] +  \mathbf{n}_t, \quad \boldsymbol{\sigma}^2_{\theta_{\myx}}(\myz[t]) = \boldsymbol{\Sigma}_t, \label{eq:LDS-b}
\end{align}
where $\mathbf{A}_t$, $\mathbf{B}_t$, $\mathbf{m}_t$, $\boldsymbol{\Lambda}_t$, $\mathbf{C}_t$, $\mathbf{n}_t$,  and $\boldsymbol{\Sigma}_t$ are matrices and vectors of appropriate size, the SMM transforms into a linear-Gaussian linear dynamical system (LG-LDS). In this case, the inference has a very popular closed-form solution, known as a \textit{Kalman filter} \citep{moreno2009kalman}. More precisely, a Kalman filter is the solution obtained when the past and present observations (outputs and inputs) are used at each time $t$ (i.e., causal inference). When a complete sequence of observations is used at each time $t$ (i.e., noncausal inference), the solution is referred to as a \textit{Kalman smoother}, also obtainable in closed form. In practical problems, $\myx[1:T]$ is generally noisy, and the terms ``filter'' and ``smoother'' refer to the estimation of a ``clean'' state vector trajectory $\myz[1:T]$ from noisy observed data.

The Kalman filter is an iterative solution that alternates between a prediction step and an update step. The prediction step involves computing the \textit{predictive distribution}, which is the posterior distribution of $\myz[t]$ given the observations up to time $t-1$. Starting from the joint distribution of all variables and exploiting the dependencies in the generative model, the predictive distribution can be expressed as (we omit the input $\myu$ for simplicity of presentation)
\begin{equation}
    p(\myz[t] | \myx[1:t-1]) = \int p(\myz[t]|\myz[t-1]) p(\myz[t-1] | \myx[1:t-1]) d\myz[t-1]. 
\end{equation}
The update step involves integrating the new (current) observation $\myx[t]$ using Bayes' rule to obtain the so-called \textit{filtering distribution} (up to some normalizing factor that does not depend on $\myz[t]$):
\begin{equation}
p(\myz[t] | \myx[1:t]) \propto  p(\myx[t]|\myz[t]) \int p(\myz[t]|\myz[t-1]) p(\myz[t-1] | \myx[1:t-1]) d\myz[t-1]. 
\end{equation}
The filtering distribution at time $t$ can be computed recursively from the filtering distribution at time $t-1$ (inside the integral). In the case of linear-Gaussian generative distributions, the filtering distribution is Gaussian, with parameters that can be computed recursively from the parameters at time $t-1$ and the generative model parameters with basic matrix/vector operations. In practice, these parameters are computed in the following two steps: prediction step and update step. Finally, the mean vector of the filtering distribution, which is often used as the state estimate, is a linear form of the observation vector. 

In the noncausal case, a similar two-step predictive/update recursive process can be computed, except that the recursion is processed in both forward (causal) and backward (anticausal) directions, leading to the \textit{smoothing distribution}. A more detailed presentation of the Kalman filter and Kalman smoother is beyond the scope of the present paper.

\subsection{Nonlinear Kalman filters}
\label{subsec:NL-KF}

Nonlinear dynamical systems (NDS), sometimes abusively referred to as nonlinear Kalman filters, have also been extensively studied, well before the deep learning era. Principled extensions to the Kalman Filter have been proposed to deal with the nonlinearities (e.g., the extended Kalman filter and the unscented Kalman filter) \citep{wan2000unscented, daum2005nonlinear}. The review of nonlinear Kalman filters is beyond the scope of the present paper, to retain the focus on DVAEs.

\chapter{Definition of Dynamical VAEs}
\label{sec:DVAE}
In this section, we describe a general methodology for defining and training dynamical VAEs. Our goal is to encompass different models proposed in the literature, which we will describe in detail later. These models can be considered particular instances of this general definition, given simplifying assumptions. This section will prepare the readers to understand well the commonalities and differences among all models that we will review and may motivate future developments.
We first define a DVAE in terms of a generative model and then present the general lines of inference and training in the DVAE framework.

\section{Generative model}

As already mentioned, DVAEs consider a sequence of observed random vectors $\myx[1:T] = \{\myx[t] \in \mathbb{R}^F\}_{t=1}^T$ and that of latent random vectors $\myz[1:T]= \{\myz[t] \in \mathbb{R}^L\}_{t=1}^T$. As opposed to the a ``static'' VAE and similarly to SSMs, these two data sequences are assumed to be temporally correlated and can have somewhat complex (cross-)dependencies across time. Defining a DVAE generative model involves specifying the joint distribution of the observed and latent sequential data, $p_{\theta}(\myx[1:T], \myz[1:T])$, the parameters of which are provided by DNNs, which themselves depend on a set of parameters $\theta$.

When the model works in the so-called \emph{driven mode}, one additionally considers an input sequence of observed random vectors $\myu[1:T] = \{\myu[t] \in \mathbb{R}^U\}_{t=1}^T$, and in that case, $\myx[1:T]$ is considered the output sequence.
In this case, to define the full generative model, we need to specify the joint distribution $p_{\theta}(\myx[1:T], \myz[1:T], \myu[1:T])$. However, in practice, we are usually only interested in modeling the generative process of $\myx[1:T]$ and $\myz[1:T]$ given the input sequence $\myu[1:T]$. Loosely speaking, the input sequence is assumed deterministic, while $\myx[1:T]$ and $\myz[1:T]$ are stochastic. Therefore, as is commonly observed in the DVAE literature \citep{krishnan2015deep,fraccaro2016sequential,fraccaro2017disentangled}, we will only focus on modeling the distribution $p_{\theta}(\myx[1:T], \myz[1:T] | \myu[1:T])$. 

In the following section, we will first omit $\theta$ when defining the general structure of dependencies in the generative model. We will specify the parameter notation later when introducing how RNNs are used to parametrize the model. In addition, we will consider the model in the driven mode (i.e., with $\myu[1:T]$ as input) as it is more general than that in the undriven mode (i.e., with no ``external'' input). The undriven mode equations can be obtained from the driven mode equations by simply removing $\myu[1:T]$.  

\subsection{Structure of dependencies in the generative model}
\label{sec:DVAE-structure}

As we will discuss in detail in Section~\ref{sec:VAE-improvements}, a DVAE can be considered a structured or hierarchical VAE in which both observed and latent variables are a set of ordered vectors, and the ordering is imposed by time. However, the natural order present in the data does not imply a unique possible structure of variable dependencies for a DVAE generative (or inference) model.
In fact, in DVAEs, the joint distribution of the observed and latent vector sequences is usually defined using the chain rule; that is, it is written as a product of conditional distributions over the vectors at different time indices. When writing the chain rule, different orderings of the random vectors can be arbitrarily chosen. This is an important point because the choice of ordering when applying the chain rule yields different practical implementations, which result in different sampling processes. 

A natural choice for ordering dependencies at generation is to use a \emph{causal} model. In the present context, a generation (or inference) model is said to be causal if the distribution of the generated (or inferred) variable at time $t$ depends only on its values at previous time indices and/or on the values of the other variables at time $t$ and at previous time indices. If the dependency is only over future time indices, the model is said to be \emph{anticausal}, and if the dependency combines the past, present, and future of the conditioning variables, the model is said to be \emph{noncausal}. 

Let us consider the following simple example:
\begin{align}
    p(\myx[1],\myx[2], \myz[1], \myz[2]) &= p(\myx[2] | \myx[1], \myz[1], \myz[2])p(\myz[2] | \myx[1], \myz[1]) p(\myx[1] | \myz[1])p(\myz[1]) \label{chain_rule_1} \\
    &= p(\myx[2] | \myx[1], \myz[1], \myz[2]) p(\myx[1] | \myz[1], \myz[2]) p(\myz[2] | \myz[1]) p(\myz[1]). \label{chain_rule_2}
\end{align}
In \eqref{chain_rule_1}, the sampling is causal because we alternate between sampling $\myz[t]$ and $\myx[t]$ from their past value or their past and present values, from $t=1$ to $2$. In contrast, in \eqref{chain_rule_2}, the sampling is not causal because we first have to sample the complete sequence of latent vectors $\myz[1:2]$ before sampling $\myx[1]$, and then $\myx[2]$. This principle generalizes to much longer sequences. 

In the DVAE literature, causal modeling is the most popular approach. In what follows, we will therefore focus on causal modeling, but the general methodology is similar for noncausal modeling. To the best of our knowledge, only one noncausal model has been proposed in the literature: the RVAE model \citep{leglaive2020recurrent}. In fact, both causal and noncausal versions of RVAE were proposed in this paper, and both versions will be presented in Section~\ref{sec:RVAE}. 

In (causal) DVAEs, the joint distribution of the latent and observed sequences is first factorized according to the time indices using the chain rule:
\begin{align}
p(\myx[1:T], \myz[1:T]|\myu[1:T]) &= \prod_{t=1}^{T} p(\myx[t], \myz[t] | \myx[1:t-1],\myz[1:t-1],\myu[1:t]).
\label{time_slide_ordered_model_tmp}
\end{align}
The only assumption made in \eqref{time_slide_ordered_model_tmp} is the causal dependence of $\myx[t]$ and $\myz[t]$ on the input sequence $\myu[1:T]$. Then, at each time index $p(\myx[t], \myz[t] | \myx[1:t-1],\myz[1:t-1] ,\myu[1:t])$ is again factorized using the chain rule, so that
\begin{align}
p(\myx[1:T], \myz[1:T] | \myu[1:T]) &= \prod_{t=1}^{T} p(\myx[t] | \myx[1:t-1],\myz[1:t],\myu[1:t] ) p(\myz[t] | \myx[1:t-1],\myz[1:t-1],\myu[1:t]).
\label{time_slide_ordered_model}
\end{align}
This equation is a generalization of \eqref{chain_rule_1}, and again, it exhibits the alternate sampling of $\myz[t]$ and $\myx[t]$. Similarly to our remark in Section~\ref{sec:SSM-principle}, for $t=1$, the first terms of the products in \eqref{time_slide_ordered_model_tmp} and \eqref{time_slide_ordered_model} are $p(\myx[1], \myz[1] | \myu[1])$ and $p(\myx[1] | \myz[1], \myu[1])p(\myz[1] | \myu[1])$, respectively. This is consistent with our notation choice of $\myx[1:0] = \myz[1:0] = \emptyset$. Alternatively, we can define $\myz[0]$ as the initial state vector and consider $p(\myz[0])$, $p(\myz[1]|\myz[0],\myu[1])$, and so on, in these equations. Hereinafter, for each detailed model, we will present the joint distribution in the general form of a product over frames from $t=1$ to $T$, and for conciseness, will not detail the model ``initialization.''

As will be detailed later, the different models proposed in the literature make different conditional independence assumptions to simplify the dependencies in the conditional distributions of \eqref{time_slide_ordered_model}. For instance, the SSM family presented in Section~\ref{sec:SSMs} is based on the following conditional independence assumptions:
\begin{align}
    p(\myx[t] | \myx[1:t-1],\myz[1:t], \myu[1:t] ) &= p(\myx[t] | \myz[t] ), \\
    p(\myz[t] | \myx[1:t-1],\myz[1:t-1],\myu[1:t]) &= p(\myz[t] | \myz[t-1],\myu[t]).
    \label{DVAE_SSM_simplifications}
\end{align}

We have already introduced the concept of the driven mode. In the causal context, we say that a DVAE is in the driven mode if $\myu[1:t]$ is used to generate either $\myx[1:t]$, $\myz[1:t]$, or both. A DVAE is in \emph{predictive} mode if $\myx[1:t-1]$, or part of this sequence, typically $\myx[t-1]$, is used to generate either $\myx[t]$ or $\myz[t]$, or both. This corresponds to feedback or closed-loop control in control theory. This is also strongly related to the concept of \emph{autoregressive process}, jointly found in the control theory, machine learning, signal processing, or time-series analysis literature \citep{papoulis1977signal, frey98graphical, durbin2012time, hamilton2020time}. Therefore, in what follows, we indifferently use the terms \emph{predictive DVAE} or \emph{autoregressive DVAE} to qualify a DVAE in the predictive mode.   

In its most general form~\eqref{time_slide_ordered_model}, a DVAE is both in the driven and predictive modes; however, it can also be in only one of the two modes (e.g., the above SSM is in the driven mode but not in the predictive mode), or even in none of them. In the literature, we did not encounter any DVAE in both modes at the same time. Moreover, there are models in the driven and nonpredictive modes that are converted to the undriven and predictive modes by replacing the control input $\myu[t]$ with the previously generated output $\myx[t-1]$, see~\citep{fraccaro2016sequential}. Note that a model's behavior can be quite different under the various modes. This is consistent with the concept of using a model in an open loop or in a closed loop in control theory. The principle of these different modes has been poorly discussed in the DVAE literature, and it is interesting to clarify it at an early stage of the DVAE presentation. 

\subsection{Parameterization with (R)NNs}
\label{subsec:parametrization}

The factorization in \eqref{time_slide_ordered_model} is a general umbrella for all (causal) DVAEs. As discussed above, each DVAE model will make different conditional independence assumptions, which will simplify the general factorization in various ways. Once the conditional assumptions are made, one can easily determine if there is a need to accumulate the past information (e.g., $\myz[t]$ or $\myx[t]$ depends on past observations $\myx[1:t]$) or if a first-order Markovian relationship holds (e.g., $\myz[t]$ and $\myx[t]$ depend at most on $\myz[t-1]$ and $\myx[t-1]$). Usually, the former is implemented using RNNs, whereas feed-forward DNNs can be used to implement first-order Markovian dependency. Moreover, once the conditional assumptions are made, the remaining dependencies can be implemented in different ways. Therefore, the final family of distributions depends not only on the conditional independence assumptions but also on the networks that are used to implement the remaining dependencies. 

\begin{figure}[t]
\centering
\resizebox{\textwidth}{!}{\begin{tikzpicture}[->]
    \node[obs,minimum size=9mm] (um) {$\myu[t-1]$};
    \node[obs,minimum size=9mm,right=of um] (u) {$\myu[t]$};
    \node[obs,right=of u] (up) {$\myu[t+1]$};
    \node[latent,minimum size=9mm,below=of um, yshift=2mm] (zm) {$\myz[t-1]$};
    \node[latent,minimum size=9mm,right=of zm] (z) {$\myz[t]$};
    \node[latent,minimum size=9mm,right=of z] (zp) {$\myz[t+1]$};
    \node[obs,minimum size=9mm,below=of zm, yshift=2mm] (xm) {$\myx[t-1]$};
    \node[obs,minimum size=9mm,right=of xm] (x) {$\myx[t]$};
    \node[obs,minimum size=9mm,right=of x] (xp) {$\myx[t+1]$};
    \edge {um} {zm};
    \edge {zm} {xm};
    \edge{u} {z};
    \edge{z} {x};
    \edge{up} {zp};
    \edge{zp} {xp};
    \edge{xm} {x,z,zp};
    \edge{x} {zp,xp};
    \path(xm) edge[bend right=30] (xp);
\end{tikzpicture}\hspace{5mm}
\begin{tikzpicture}[->]
    \node[obs,minimum size=9mm] (um) {$\myu[t-1]$};
    \node[obs,minimum size=9mm,right=of um] (u) {$\myu[t]$};
    \node[obs,minimum size=9mm,right=of u] (up) {$\myu[t+1]$};
    \node[latent,minimum size=9mm,below=of um,yshift=3.5mm] (zm) {$\myz[t-1]$};
    \node[latent,minimum size=9mm,right=of zm] (z) {$\myz[t]$};
    \node[latent,minimum size=9mm,right=of z] (zp) {$\myz[t+1]$};
    \node[det,minimum size=9mm,below=of zm,yshift=-2.5mm] (hm) {$\myh[t-1]$};
    \node[det,minimum size=11mm,right=of hm,xshift=-2mm] (h) {$\myh[t]$};
    \node[det,minimum size=9mm,right=of h,xshift=-2.5mm] (hp) {$\myh[t+1]$};
    \node[obs,minimum size=9mm,below=of hm, yshift=-2.5mm] (xm) {$\myx[t-1]$};
    \node[obs,minimum size=9mm,right=of xm] (x) {$\myx[t]$};
    \node[obs,minimum size=9mm,right=of x] (xp) {$\myx[t+1]$};
    \edge {um} {zm};
    \edge {hm} {xm,h,zm};
    \edge{u} {z};
    \edge{h} {x,hp,z};
    \edge{up} {zp};
    \edge{hp} {xp,zp};
    \edge{xm} {h};
    \edge{x} {hp};
    \path(zm) edge[bend left=30] (xm);
    \path(z) edge[bend left=30] (x);
    \path(zp) edge[bend left=30] (xp);
\end{tikzpicture}\hspace{5mm}
\begin{tikzpicture}[->]
    \node[obs] (um) {$\myu[t-1]$};
    \node[obs,minimum size=9mm,right=of um] (u) {$\myu[t]$};
    \node[obs,minimum size=9mm,right=of u] (up) {$\myu[t+1]$};
    \node[latent,minimum size=9mm,below=of um,yshift=3.5mm] (zm) {$\myz[t-1]$};
    \node[latent,minimum size=9mm,right=of zm] (z) {$\myz[t]$};
    \node[latent,minimum size=9mm,right=of z] (zp) {$\myz[t+1]$};
    \node[det,minimum size=9mm,below=of zm,yshift=5mm] (hzm) {$\myh[t-1]$};
    \node[det,minimum size=11mm,right=of hzm,xshift=-2mm] (hz) {$\myh[t]$};
    \node[det,minimum size=9mm,right=of hz,xshift=-2.5mm] (hzp) {$\myh[t+1]$};
    \node[det,minimum size=9mm,below=of hzm,yshift=7mm] (hxm) {$\myk[t-1]$};
    \node[det,minimum size=11mm,right=of hxm,xshift=-2mm] (hx) {$\myk[t]$};
    \node[det,minimum size=9mm,right=of hx,xshift=-2.5mm] (hxp) {$\myk[t+1]$};
    \node[obs,minimum size=9mm,below=of hxm, yshift=5mm] (xm) {$\myx[t-1]$};
    \node[obs,minimum size=9mm,right=of xm] (x) {$\myx[t]$};
    \node[obs,minimum size=9mm,right=of x] (xp) {$\myx[t+1]$};
    \edge {um} {zm};
    \edge {hzm} {hz,zm};
    \edge {hxm} {hx,xm};
    \edge{u} {z};
    \edge{hz} {hzp,z};
    \edge{hx} {hxp,x};
    \edge{up} {zp};
    \edge{hzp} {zp};
    \edge{hxp} {xp};
    \edge{xm} {hx,hz};
    \edge{x} {hxp,hzp};
    \path(zm) edge[bend left=30] (xm);
    \path(z) edge[bend left=30] (x);
    \path(zp) edge[bend left=30] (xp);
\end{tikzpicture}}
\caption{Two different implementations of a given factorization. The probabilistic graphical model (left) shows the dependencies between random variables and corresponds to the factorization in~\eqref{eq:DVAE_example_z} and \eqref{eq:DVAE_example_x}. Two possible implementations based on RNNs are shown: sharing the internal state variables (middle) or with two different internal state variables (right). We refer to the \textit{compact} representation (left) and to the \textit{developed} representations (middle and right). This terminology holds true for both the graphical representations and model formulations.}
\label{fig:DVAE_example}
\end{figure}
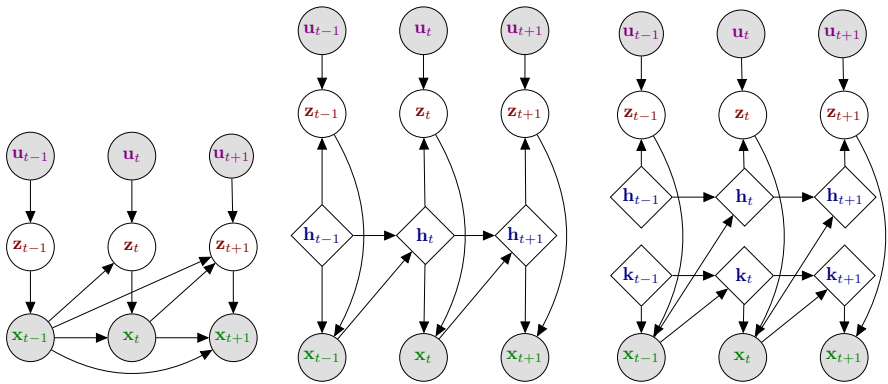

Let us showcase this with a concrete example in which we have the following conditional independence assumptions:
\begin{align}
    p(\myz[t] | \myx[1:t-1],\myz[1:t-1],\myu[1:t]) &= p(\myz[t] | \myx[1:t-1],\myu[t]),
    \label{eq:DVAE_example_z}\\
    p(\myx[t] | \myx[1:t-1],\myz[1:t], \myu[1:t] ) &= p(\myx[t] | \myx[1:t-1],\myz[t]). \label{eq:DVAE_example_x}
\end{align}
Here, we assume that the generation of both $\myx[t]$ and $\myz[t]$ depends on $\myx[1:t-1]$. In addition, the generation of $\myx[t]$ also depends on $\myz[t]$ and that of $\myz[t]$ also depends on $\myu[t]$. To accumulate the information of all past outputs $\myx[1:t-1]$, one can use an RNN. In practice, the past information is accumulated in the internal state variable of the RNN, namely $\myh[t]$, computed recurrently at each frame $t$. Among the many possible implementations, we consider two in this example: in the first implementation, illustrated in Figure~\ref{fig:DVAE_example}~(middle), a single RNN internal state variable $\myh[t]$ is used to generate both $\myx[t]$ and $\myz[t]$, while in the second implementation, illustrated in Figure~\ref{fig:DVAE_example}~(right), two different internal state variables, $\myh[t]$ and $\myk[t]$, are used to generate $\myx[t]$ and $\myz[t]$ separately.

Assuming that all probability distributions are Gaussian, the first implementation can be expressed as
\begin{align}
&\hspace{-4mm}    \myh[t] = d_{\myh}(\myx[t-1],\myh[t-1];\theta_{\myh}), \label{eq:mod1-rech}\\
&\hspace{-4mm}    [\boldsymbol{\mu}_{\theta_{\myz}}(\myx[1:t-1],\myu[t]), \boldsymbol{\sigma}_{\theta_{\myz}}(\myx[1:t-1],\myu[t])] = d_{\myz}(\myh[t],\myu[t];\theta_{\myh\myz}), \label{eq:mod1-parz}\\
&\hspace{-4mm}    p_{\theta_{\myz}}(\myz[t] | \myx[1:t-1],\myu[t]) = \mathcal{N}\big(\myz[t]; \boldsymbol{\mu}_{\theta_{\myz}}(\myx[1:t-1],\myu[t]), \text{diag}\{\boldsymbol{\sigma}_{\theta_{\myz}}^2(\myx[1:t-1],\myu[t])\}\big), \label{eq:mod1-z}\\
&\hspace{-4mm}    [\boldsymbol{\mu}_{\theta_{\myx}}(\myx[1:t-1],\myz[t]), \boldsymbol{\sigma}_{\theta_{\myx}}(\myx[1:t-1],\myz[t])] = d_{\myx}(\myh[t],\myz[t];\theta_{\myh\myx}),\label{eq:mod1-parx}\\
&\hspace{-4mm}    p_{\theta_{\myx}}(\myx[t] | \myx[1:t-1],\myz[t]) = \mathcal{N}\big(\myx[t]; \boldsymbol{\mu}_{\theta_{\myx}}(\myx[1:t-1],\myz[t]), \text{diag}\{\boldsymbol{\sigma}_{\theta_{\myx}}^2(\myx[1:t-1],\myz[t])\}\big),\label{eq:mod1-x}
\end{align}
where $d_{\myh}$, $d_{\myz}$, and $d_{\myx}$ are nonlinear functions implemented with DNNs. It is now clear that the parameters of the conditional distribution of $\myz[t]$ are $\theta_{\myz}=\theta_{\myh}\cup\theta_{\myh\myz}$, whereas those of the conditional distribution of $\myx[t]$ are $\theta_{\myx}=\theta_{\myh}\cup\theta_{\myh\myx}$. Thus, the two conditional distributions share the recurrent parameters $\theta_{\myh}$. Regarding the second implementation, the generative process can be expressed as
\begin{align}
&\hspace{-4mm} \myh[t] = d_{\myh}(\myx[t-1],\myh[t-1];\theta_{\myh}),\label{eq:mod2-rech}\\
&\hspace{-4mm} [\boldsymbol{\mu}_{\theta_{\myz}}(\myx[1:t-1],\myu[t]), \boldsymbol{\sigma}_{\theta_{\myz}}(\myx[1:t-1],\myu[t])] = d_{\myz}(\myh[t],\myu[t];\theta_{\myh\myz}),\label{eq:mod2-parz}\\
&\hspace{-4mm}    p_{\theta_{\myz}}(\myz[t] | \myx[1:t-1],\myu[t]) = \mathcal{N}\big(\myz[t]; \boldsymbol{\mu}_{\theta_{\myz}}(\myx[1:t-1],\myu[t]), \text{diag}\{\boldsymbol{\sigma}_{\theta_{\myz}}^2(\myx[1:t-1],\myu[t])\}\big),\label{eq:mod2-z} \\
&\hspace{-4mm} \myk[t] = d_{\myk}(\myx[t-1],\myk[t-1];\theta_{\myk}),\label{eq:mod2-reck}\\
&\hspace{-4mm} [\boldsymbol{\mu}_{\theta_{\myx}}(\myx[1:t-1],\myz[t]), \boldsymbol{\sigma}_{\theta_{\myx}}(\myx[1:t-1],\myz[t])] = d_{\myx}(\myk[t],\myz[t];\theta_{\myk\myx}),\label{eq:mod2-parx}\\
&\hspace{-4mm}    p_{\theta_{\myx}}(\myx[t] | \myx[1:t-1],\myz[t]) = \mathcal{N}\big(\myx[t]; \boldsymbol{\mu}_{\theta_{\myx}}(\myx[1:t-1],\myz[t]), \text{diag}\{\boldsymbol{\sigma}_{\theta_{\myx}}^2(\myx[1:t-1],\myz[t])\}\big).\label{eq:mod2-x}
\end{align}
We have an additional DNN-based nonlinear function $d_{\myk}$, and analogously, it is clear that the parameters of the conditional distribution of $\myz[t]$ are $\theta_{\myz}=\theta_{\myh}\cup\theta_{\myh\myz}$, whereas those of the conditional distribution of $\myx[t]$ are $\theta_{\myx}=\theta_{\myk}\cup\theta_{\myk\myx}$. In this case, the two conditional distributions do not share any parameter. To ease the notation, hereinafter, we will denote the parameters of $p(\myz[t] | \myx[1:t-1],\myz[1:t-1],\myu[1:t])$ and $p(\myx[t] | \myx[1:t-1],\myz[1:t],\myu[1:t] )$ as $\theta_{\myz}$ and $\theta_{\myx}$, respectively, in \eqref{time_slide_ordered_model}, irrespective of whether or not they share some parameters. We will also use $\theta$ to denote $\theta_{\myz}\cup\theta_{\myx}$. 

In the equations above, the operators $d_{\myh}$, $d_{\myk}$, $d_{\myx}$ and $d_{\myz}$ are nonlinear mappings parametrized by DNNs of arbitrary architecture. How to choose and design these architectures is beyond the scope of this paper, as it largely depends on the target application. To fix ideas, in the present example, $d_{\myh}$ and $d_{\myk}$ are RNNs, and $d_{\myx}$ and $d_{\myz}$ are feed-forward DNNs. In this paper, we will not discuss how to select the hyper-parameters of these networks, such as the number of layers, or the number of units per layer.

Note that \eqref{eq:mod1-z} and \eqref{eq:mod2-z} are exactly the same, meaning that the conditional distributions of $\myz[t]$ are the same for both models. The same remark holds for \eqref{eq:mod1-x} and \eqref{eq:mod2-x}, defining the conditional distribution of $\myx[t]$. However, the computations performed to obtain the parameters $\theta_{\myz}$ and $\theta_{\myx}$ differ depending on the model. Clearly, we need to make a distinction between the two forms. We propose to call the form of a DVAE model or its graphical representation \textit{compact} when only random variables appear (e.g., \eqref{eq:mod1-z} and \eqref{eq:mod1-x}, and Figure~\ref{fig:DVAE_example}~(left)). In addition, we propose to call the form of a DVAE or its graphical representation \emph{developed} when both random and deterministic variables appear (e.g., \eqref{eq:mod1-rech}--\eqref{eq:mod1-x}, \eqref{eq:mod2-rech}--\eqref{eq:mod2-x} and Figure~\ref{fig:DVAE_example}~(middle) and~(right)). Each compact form can have different developed forms corresponding to different implementations. The distinction between the compact and developed forms is important as the optimization occurs on the parameters of the developed form, which is only a subgroup of all possible models satisfying the compact form. It is thus important to present the developed form of a model. However, the temporal dependencies of order higher than one are not directly visible in the developed graphical form, as they might be implicitly encoded in the internal state variables. Therefore, when reviewing DVAE models in the following chapters, we will always present both the compact and developed graphical representations.

\section{Inference model}
\label{sec:DVAE-inference}

In the present DVAE context, the posterior distribution of the state sequence $\myz[1:T]$ is $p_{\theta}(\myz[1:T] | \myx[1:T],\myu[1:T])$ in the driven mode or $p_{\theta}(\myz[1:T] | \myx[1:T])$ in the undriven mode. As for the standard VAE, this posterior distribution is intractable because of the nonlinearities in the generative model. In fact, having temporal dependencies only makes things even more complicated. Therefore, we  also need to define an inference model $q_{\phi}(\myz[1:T] | \myx[1:T], \myu[1:T])$, which is an approximation of the intractable  posterior distribution $p_{\theta}(\myz[1:T] | \myx[1:T],\myu[1:T])$. As for the standard VAE, this model is required not only for performing inference of the latent sequence $\myz[1:T]$ from the observed sequences $\myx[1:T]$ and $\myu[1:T]$ but also for estimating the parameters of the generative model, as will be seen below. As for the standard VAE again, the inference model also uses DNNs to generate its parameters.

\subsection{Exploiting D-separation}
\label{subsec:D-separation}

In a Bayesian network, and in a DVAE in particular, even though the computation of the posterior distribution is often intractable, there exists a general methodology to express its general form (i.e., to specify the dependencies between the variables of a generative model \emph{at inference time}). This methodology is based on the so-called \emph{D-separation} property of Bayesian networks
\citetext{\citealp{geiger1990identifying}; \citealp[Chapter~8]{BishopBook}}. The general principle is that some of the conditioning variables in the expression of the posterior distribution of a given variable can vanish depending on whether the nodes between these conditioning variables and the given variable represent variables that are observed or unobserved and depending on the direction of the dependencies (i.e., the direction of the arrows of the graphical representation).

In detail, D-separation is based on the three principles derived for a Bayesian network with three random variables $a$, $b$, and $c$:
\begin{itemize}
\item A \emph{tail-to-tail} (or common parent) node $c$ corresponding to the structure $a \leftarrow c \rightarrow b$ makes the two other nodes $a$ and $b$ \emph{conditionally independent} when it is observed. In short, we have $p(a, b | c ) = p(a | c ) p(b | c)$.
\item A \emph{head-to-tail} (or cascade) node $c$ corresponding to the structure $a \rightarrow c \rightarrow b$ or $a \leftarrow c \leftarrow b$ makes the two other nodes $a$ and $b$ \emph{conditionally independent} when it is observed. In short, we have $p(a, b | c ) = p(a | c ) p(b | c)$.
\item A \emph{head-to-head} (or V-structure) node $c$ corresponding to the structure $a \rightarrow c \leftarrow b$ makes $a$ and $b$ \emph{conditionally dependent} when it is observed, hence $p(a, b | c ) \neq p(a | c ) p(b | c)$.
\end{itemize}
D-separation consists in applying these three principles recursively to analyze larger Bayesian networks with any arbitrary structure. Let us consider a Bayesian network in which $A$, $B$, and $C$ are arbitrary nonintersecting node sets. \emph{$A$ and $B$ are D-separated given $C$} if all possible paths that connect any node in $A$ to any node in $B$ are blocked given $C$. A path is said to be \emph{blocked} given a set of observed nodes $O$ if it includes a node $c$ such that either
\begin{itemize}
\item $c$ is a tail-to-tail node and $c \in O$ (i.e., it is observed) \emph{or}
\item $c$ is a head-to-tail node and $c \in O$ (i.e., it is observed) \emph{or}
\item $c$ is a head-to-head node and $c \notin O$ (i.e., it is \emph{not} observed).
\end{itemize}
Equivalently, $A$ and $B$ are D-separated given $C$ if they are not connected by any active path (i.e., a path that is not blocked). Finally, if $A$ and $B$ are D-separated given $C$, we have $p(A , B | C) = p(A |C) p(B | C)$.

D-separation is helpful even for more conventional (i.e., nondeep) models because the algebraic derivation of a posterior distribution from a joint distribution is not always easy. In the present variational framework, we can exploit the above methodology to design the approximate posterior distribution $q_{\phi}$. It is reasonable to assume that a good candidate for $q_{\phi}$ will have the same structure as the exact posterior distribution in terms of variable dependency. In other words, if we cannot derive the exact posterior distribution, let us at least use an approximation that exhibits the same dependencies between variables so that it is fed with the same information.
Yet, it is quite surprising to see that a significant proportion of the DVAE papers we have reviewed, especially the early papers, neither refer to this methodology nor consider looking at the form of the exact posterior distribution when designing an approximate distribution. In the early studies in particular, the formulation of $q_{\phi}$ is chosen quite arbitrarily and with no reference to the structure of the exact posterior distribution. In more recent papers however, the structure of $q_{\phi}$ generally follows that of the exact posterior distribution. We will come back on this point on a case-by-case basis when presenting the DVAE models of the literature in the next chapters.

\subsection{Noncausal and causal inference}
\label{subsec:non-causal-vs-causal}

Being aware of this problem, we can now go back to the general form of the exact posterior distribution and factorize it as follows, applying again the chain rule the same way as we did for the generative model:
\begin{equation}
p_{\theta}(\myz[1:T] | \myx[1:T], \myu[1:T]) = \prod_{t=1}^T p_{\theta_{\myz}}(\myz[t] | \myz[1:t-1], \myx[1:T], \myu[1:T]).
\label{posterior_DVAE_chain_rule}
\end{equation}
For the most general generative model defined in \eqref{time_slide_ordered_model}, the dependencies in each conditional distribution $p_{\theta_{\myz}}(\myz[t] | \myz[1:t-1], \myx[1:T], \myu[1:T])$ cannot be simplified. In other words, $\myz[t]$ depends on the past latent vectors $\myz[1:t-1]$ and on the complete sequences of observed vectors $\myx[1:T]$ and $\myu[1:T]$ (past, current, and future time steps). The exact inference is thus a noncausal process, even if the generation is causal. 
This is reminiscent of the Kalman smoother (i.e., the noncausal solution to inference in LG-LDS, see Section~\ref{subsec:KF}). 
As discussed in the previous subsection, the inference model $q_{\phi}$ should here have the same most general structure as the exact posterior distribution of \eqref{posterior_DVAE_chain_rule}: 
\begin{equation}
q_{\phi}(\myz[1:T] | \myx[1:T], \myu[1:T]) = \prod_{t=1}^T q_{\phi}(\myz[t] | \myz[1:t-1], \myx[1:T], \myu[1:T]).
\label{q_dist_chaine_rule}
\end{equation}
Similar to the generative model, each conditional posterior distribution $q_{\phi}(\myz[t] | \myz[1:t-1], \myx[1:T], \myu[1:T])$ should  accumulate information from past latent variables and past observations, but in contrast to the generative model, it should also accumulate information from present and future observations. Typically, this process is implemented with a bidirectional recurrent network. 

Depending on the conditional independence assumptions made when defining the generative model, the posterior dependencies in $p_{\theta}(\myz[t] | \myz[1:t-1], \myx[1:T], \myu[1:T])$ can be simplified using the  D-separation property of Bayesian networks described in the previous subsection. Thus, the posterior dependencies in $q_{\phi}(\myz[t] | \myz[1:t-1], \myx[1:T], \myu[1:T])$ can be simplified similarly. Of course, it is always possible to use an approximate posterior $q_{\phi}$ that does not follow the structure of the exact posterior distribution. In fact, it makes sense to use a simplified version if one wants to decrease the computational cost or satisfy other constraints. In particular, for online or incremental data processing, the inference can be forced to be a causal process by removing the dependencies of $q_{\phi}$ on the future observations (and future inputs). This is similar to the Kalman filter for an LG-LDS, see again Section~\ref{subsec:KF}. This will generally be at risk of degrading the inference performance. Again, we will return to these points when reviewing the DVAE models proposed in the literature.

\subsection{Sharing variables and parameters at generation and inference}
\label{sec:sharing}

We can note a similarity between the (most general causal) generative distribution $p_{\theta_{\myz}}(\myz[t] | \myx[1:t-1],\myz[1:t-1],\myu[1:t])$ and the corresponding inference model $q_{\phi}(\myz[t] | \myz[1:t-1], \myx[1:T], \myu[1:T])$ in terms of random variable dependencies. For instance, the general form of the dependency of $\myz[t]$ on past latent vectors is the same at inference and generation: in both cases, $\myz[t]$ depends on the complete past sequence $\myz[1:t-1]$. Implementing this recurrence at inference and at generation can be made either with a single unique RNN or with two different RNNs, in line with what we discussed in Section~\ref{subsec:parametrization}. The same principle applies to $\myu[1:t]$ and $\myx[1:t]$, which are both used at generation and inference. 
Depending on which variables we consider, it can make sense to use the same RNN at generation and inference, meaning that the deterministic link between the realizations of random variables is the same at generation and at inference.
If this is the case, the decoder and encoder share some network modules and thus $\theta$ and $\phi$ share some parameters. Note that this is not the case in standard VAEs. 

Hereinafter, we will use $\myh[t]$ to denote the internal state of the decoder and $\myg[t]$ to denote that of the encoder if it is different from the internal state of the decoder. Otherwise, we will use $\myh[t]$ for the encoder as well.

\section{VLB and training of DVAEs}
\label{sec:DVAE-VLB-and-training}

As for the standard VAE, training a DVAE is based on the maximization of the VLB. In the case of DVAEs, the VLB initially defined in \eqref{varFreeEnergyParam} is extended to data sequences as follows:
\begin{align}
\mathcal{L}(\theta, \phi ; \myx[1:T], \myu[1:T]) &= \mathbb{E}_{q_{\phi}(\myz[1:T] | \myx[1:T], \myu[1:T])}\big[ \log p_{\theta}(\myx[1:T], \myz[1:T] | \myu[1:T])\big] \nonumber \\
& \hspace{-.5cm} - \mathbb{E}_{q_{\phi}(\myz[1:T] | \myx[1:T], \myu[1:T])}\big[ \log q_{\phi}(\myz[1:T] | \myx[1:T], \myu[1:T]) \big].
\label{VLB1}
\end{align}
With the factorization in \eqref{q_dist_chaine_rule}, the expectation in \eqref{VLB1} can be expressed as a cascade of expectations taken with respect to conditional distributions over individual latent vectors at different time indices:
\begin{align}
\mathbb{E}_{q_{\phi}(\myz[1:T] | \myx[1:T], \myu[1:T])}[\psi(\myz[1:T])] =& \ 
\mathbb{E}_{q_{\phi}(\myz[1] | \myx[1:T], \myu[1:T])}\bigg[ \mathbb{E}_{q_{\phi}(\myz[2] | \myz[1], \myx[1:T], \myu[1:T])}\Big[ \ldots \nonumber \\
&\hspace{-5mm} \mathbb{E}_{q_{\phi}(\myz[T] | \myz[1:T-1], \myx[1:T], \myu[1:T])}\big[\psi(\myz[1:T])\big]\ldots \Big]\bigg],
\end{align}
where $\psi(\myz[1:T])$ denotes any function of $\myz[1:T]$. Then, by injecting \eqref{time_slide_ordered_model} and \eqref{q_dist_chaine_rule} into \eqref{VLB1}, and using the above cascade, we can develop the VLB as follows:
\begin{align}
& \mathcal{L}(\theta, \phi ; \myx[1:T], \myu[1:T]) = \mathbb{E}_{q_{\phi}(\myz[1:T] | \myx[1:T], \myu[1:T])}\big[ \log p_{\theta}(\myx[1:T], \myz[1:T] | \myu[1:T])  \nonumber \\
& \hspace{6.3cm} - \log q_{\phi}(\myz[1:T] | \myx[1:T], \myu[1:T]) \big] \nonumber \\
&= \sum_{t=1}^{T} \mathbb{E}_{q_{\phi}(\myz[1:t] | \myx[1:T], \myu[1:T])}\big[  \log p_{\theta_{\myx}}(\myx[t] | \myx[1:t-1], \myz[1:t], \myu[1:t]) \big] \nonumber \\
& \hspace{0.5cm} - \sum_{t=1}^{T}  \mathbb{E}_{q_{\phi}(\myz[1:t-1] | \myx[1:T], \myu[1:T])}\left[ D_{\text{KL}}\left(q_{\phi}(\myz[t] | \myz[1:t-1], \myx[1:T], \myu[1:T]) \parallel \right.\right. \nonumber\\
&\hspace{5.5cm} \left.\left. p_{\theta_{\myz}}(\myz[t] | \myx[1:t-1],\myz[1:t-1], \myu[1:t]) \right)\right].
\label{VLB1_dev}
\end{align}
To the best of our knowledge, this is the first time that the VLB is presented in this most general form, which is valid for the entire class of (causal) DVAE models.

As for the standard VAE, the VLB contains a reconstruction accuracy term and a regularization term. However, in contrast to the standard VAE, where the regularization term has an analytical form for usual distributions, here, both the reconstruction accuracy and regularization term require the computation of Monte Carlo estimates (i.e., empirical averages) using samples drawn from $q_{\phi}(\myz[1:\tau] | \myx[1:T], \myu[1:T])$, where $\tau \in \{1,...,T\}$ is an arbitrary index. Using the chain rule in \eqref{q_dist_chaine_rule}, we sample from the joint distribution $q_{\phi}(\myz[1:\tau] | \myx[1:T], \myu[1:T])$ by sampling recursively from $q_{\phi}(\myz[t] | \myz[1:t-1], \myx[1:T], \myu[1:T])$, going from $t=1$ to $t=\tau$. Sampling each random vector $\myz[t]$ at a given time instant is straightforward, as $q_{\phi}(\myz[t] | \myz[1:t-1], \myx[1:T], \myu[1:T])$ is analytically specified by the chosen inference model (e.g., Gaussian with mean and variance provided by an RNN). We have to use a similar reparameterization trick as for standard VAEs, so the sampling-based VLB estimator remains differentiable with respect to $\phi$. The VLB can then be maximized with respect to both $\phi$ and $\theta = \theta_{\myz}\cup\theta_{\myx}$ using gradient-ascent-based algorithms. We recall that for DVAEs, $\phi$ and $\theta$ can share parameters, which is different from the ``static'' VAE, but perfectly alright for the optimization. Finally, the VLB is here defined here for a single data sequence, but a common practice is to average the VLB over a mini-batch of training data sequences before updating the model parameters with gradient ascent.

\section{Additional dichotomy for autoregressive DVAE models}
\label{sec:additional-dichotomy}

A DVAE can be used to generate new data, for analysis-synthesis (by chaining the encoder and decoder), or for data transformation, by modifying the latent vector sequence in between analysis and synthesis. In the case of DVAE models functioning in the predictive mode (i.e., autoregressive DVAEs, see Section~\ref{sec:DVAE-structure}), these tasks can be processed in different manners, leading to an additional dichotomy of functioning modes. We describe these functioning modes in the next subsection before we see the implications for model training in the following subsection. Because these additional different modes concern the recursive part of the models, nonpredictive DVAEs are not concerned here. 

\subsection{Teacher forcing against generation mode}
\label{subsec:TF-vs-GM}

In practice, for autoregressive DVAE models, we have two generation modes, for the generation of both $\myx[t]$ and $\myz[t]$. A mode in which we assume that the ground-truth past observed vectors $\myx[1:t-1]$ are used for generating the current vector ($\myx[t]$ or $\myz[t]$), and a mode in which the generated past observed vectors are used for generating the current vector. At this point, it is important to distinguish between the notation for the ground-truth value of the observed data vector $\myx[t]$ and that for its modeled version produced by a DVAE, which we denote by $\myxh[t]$. In practice, $\myx[1:T]$ is a given data sequence that we want to model with a DVAE (or that we use for model training, as shown below), and $\myxh[1:T]$ is the actual output of the DVAE.

This issue of either using the ground-truth past observed data vectors $\myx[1:t-1]$ or reinjecting the previously generated vectors $\myxh[1:t-1]$ at the input of a generative model is a classical problem of recursive models and, in particular, of RNNs. Yet it is poorly discussed in the DVAE literature. In the RNN literature, the first configuration is sometimes referred to as \emph{teacher-forcing} \citep{williams1989learning}, as it is assumed that a teacher (or oracle) can provide the model with ground-truth values. Hereinafter, we will use this terminology. This is a classical configuration at training time, when the whole sequence $\myx[1:T]$ is available and the model is tuned so that $\myxh[1:T]$ fits $\myx[1:T]$. However, this is unrealistic at generation time, when the model produces a new sequence $\myxh[1:T]$. Here, the second configuration must be used. We refer to this second configuration as the \emph{generation mode}.
Regarding the generation of $\myz[1:T]$, the teacher-forcing concept is irrelevant as the concept of ground-truth values for latent vectors is questionable in essence. In practice, $\myz[1:T]$ is either the output of the inference model (this is the case during DVAE training or in analysis-synthesis) or any arbitrary latent vector sequence (generated with $p_{\theta_{\myz}}$ or predefined). Note also that because of the recursivity of the generative process, data generation with a DVAE (strongly) depends on the initialization of the generative process. We do not detail this aspect in the present review. 

If we now focus on the analysis-synthesis task, we have to chain the encoder and decoder. The encoder takes $\myx[1:T]$ as the input and produces a sequence of latent vectors $\myz[1:T]$. Then, the decoder uses $\myz[1:T]$ to generate $\myxh[1:T]$. Decoding can be performed with either teacher-forcing or generation mode. The former case is expected to produce a sequence $\myxh[1:T]$ that is closer to $\myx[1:T]$ than in the latter case, as it uses ground-truth values, whereas the generation mode uses approximate values. However, it suffers from the same ``unrealistic'' aspect as that used for data generation. This configuration can be used to evaluate the prediction power of DVAE models in an ideal (oracle) setting. In contrast, analysis-synthesis with the generation mode is expected to yield lower performance but is the natural configuration from an information-theoretic viewpoint. Here, we test the capability of the model to encode the information of a (generally high-dimensional) data sequence $\myx[1:T]$ into a (generally low-dimensional) latent sequence $\myz[1:T]$. From an application point of view, this corresponds to telecommunication or storage applications, where $\myx[1:T]$ would be encoded into $\myz[1:T]$, $\myz[1:T]$ would be transmitted or stored, and then $\myxh[1:T]$ would be decoded from $\myz[1:T]$. In short, we apply DVAEs to source coding, and the DVAE turns into a codec, apart from quantization issues. Such coding/decoding scheme can be applied offline by using a noncausal inference model (with an optimal structure following that of the exact posterior distribution) or online with a (suboptimal) causal inference model. If some amount of latency is tolerated, one can also use a noncausal inference model with a suitable lookahead. The interest of DVAE models for source coding is further discussed in Section~\ref{sec:discussion-coding}. 

\subsection{Train/test matching}
\label{subsec:train-test-match}

As mentioned in the previous subsection, teacher-forcing is a conventional strategy used for training recursive models. However, in practice, when using autoregressive DVAEs for data generation or compression, the generation mode must be used. This leads to a mismatch between the training and testing conditions, a general problem in machine learning that leads to performance degradation compared to the case in which the same configuration is used for training and testing. Therefore, if the generation mode is used in a practical DVAE use-case, it can be beneficial to use the generation mode during model training as well, so that the training configuration matches the practical use-case configuration. In practice, this implies replacing $\myx[1:t-1]$ with $\myxh[1:t-1]$ in the conditioning variables in the VLB equations of Section~\ref{sec:DVAE-VLB-and-training}. 
In Chapter~13, we illustrate this strategy in our experimental benchmark. We observe in our experiments that using the generation mode at both model training and testing leads to a significant gain in performance compared to the mismatched configuration, though the performance remains slightly lower than that in the case where teacher-forcing is used at both training and testing. More details are given in Section~\ref{subsec:SRNN-TF-GM-experiments-speech}.

\section{DVAE summary}
\label{sec:DVAE-summary}

Dynamical VAEs are constructed with various stochastic relationships amonng the control variables $\myu[1:T]$, latent variables $\myz[1:T]$, and observed variables $\myx[1:T]$. We recall that a random variable $\mathbf{a}$ is called a parent of another random variable $\mathbf{b}$ when the realization of $\mathbf{a}$ is used to compute the parameters of the distribution of $\mathbf{b}$. These parameters can be obtained with a linear or a nonlinear mapping of the realization of $\mathbf{a}$ (and possibly  of other random variables). A DVAE model must contain two types of relationships:
\begin{itemize}
    \item \textbf{Decoding link:}~~$\myz[t]$ is always a parent of $\myx[t]$. Graphically, there is always an arrow from $\myz[t]$ to $\myx[t]$ in the compact graphical representation. This is a fundamental characteristic inherited from the standard VAE.
    \item \textbf{Temporal link:}~~At least one element in $\myz[1:t-1]$ or in $\myx[1:t-1]$ is parent to either $\myz[t]$ or $\myx[t]$. One of the simplest forms of a temporal link, namely $\myz[t-1]$ is a parent of $\myz[t]$, is a fundamental characteristic of first-order SSMs.
\end{itemize}
In a way, the ``minimal DVAE'' is the straightforward combination of a first-order SSM and a VAE, which is the DKF model that we will detail in Section~\ref{sec:DKF}. Other DVAEs include additional temporal links. Moreover, temporal links such as ``$\myz[t-1]$ is a parent of $\myx[t]$'' can be considered additional decoding links, in that $\myx[t]$ is generated from $\myz[t]$ \emph{and} $\myz[t-1]$. As for temporal links, in the papers that we detail in this overview, $\myz[t]$ and/or $\myx[t]$ depend either on $\myz[t-1]$ and/or $\myx[t-1]$, or on $\myz[1:t-1]$ and/or $\myx[1:t-1]$. In other words, the order of temporal dependencies is either 1 (implemented with a basic feed-forward neural network, such as a Multi-Layer Perceptron (MLP)) or infinity (implemented with an RNN). However, one can, in principle, use $N$-order temporal dependencies with $1 < N < \infty$, relying, for instance, on convolutional neural networks (CNNs) with finite-length receptive fields. In particular, temporal convolutional networks (TCNs) \citep{lea2016temporal}, which are based on dilated convolutions, are competitive with RNNs on several sequence modeling tasks, including generative modeling \citep{aksan2018stcn}. 

Finally, as discussed before, a DVAE can be in the \textit{driven mode}, in the \textit{predictive mode}, in both, or in none of these modes. This is also modeled by parenthood relationships that may or may not exist.
\begin{itemize}
    \item \textbf{Driving link:}~~A DVAE is said to be in the driven mode if $\myu[t]$ is a parent of either $\myz[t]$ or $\myx[t]$, or both.
    \item \textbf{Predictive link:}~~A DVAE is said to be in the predictive mode if $\myx[1:t-1]$, or part of this sequence, is a parent of either $\myz[t]$ or $\myx[t]$, or both. In practice, a predictive link is implemented either in the teacher-forcing mode (using the ground-truth past vector sequence) or in the generation mode (using the previously generated vector sequence). Generally, better performance is achieved if the same mode is used at training time and test time. 
\end{itemize}

\chapter{Deep Kalman Filters}
\label{sec:DKF}
Continuing from the previous section, we start our DVAE tour by combining SSMs with neural networks. Such a combination is not recent, see, e.g., \citep{haykin2004kalman,raiko2009variational}, but has been recently investigated under the VAE angle in two papers by the same authors \citep{krishnan2015deep, krishnan2017structured}. The resulting deep SSM is referred to as a DKF \citep{krishnan2015deep} or a deep Markov model (DMM) \citep{krishnan2017structured}.\footnote{The same generative model is considered in both papers, but as we will detail later, the second paper proposes notable improvements regarding the inference model. The authors change the model name from DKF to DMM, maybe because the second denomination appears more general. In the present review, we retain the denomination DKF.} Therefore, these papers do not provide a new concept in terms of models, but they provide a solution to the joint problem of inference and model parameter estimation in the VAE methodological framework applied to the SSM model architecture. In other words, this is, to the best of our knowledge, the first example of unsupervised training of a deep SSM by chaining an approximate inference model with a generative model and using the VLB maximization methodology. This training leads to an unsupervised discovery of the latent space that encodes the temporal dynamics of the data. The VAE methodology circumvents the difficulties encountered in previous approaches \citep{haykin2004kalman,raiko2009variational} concerning the computational complexity and practicability of model parameter estimation, particularly allowing to move directly from single-layer neural networks to DNNs. 

Hereinafter, for all detailed DVAE models, we first present the generative equations, then the inference model (and we discuss its choice by referring to the  exact posterior distribution structure as deduced from the generative model), and finally the detailed form of the VLB used for model training together with clues about the optimization algorithm. 

\section{Generative model}

We have already seen the generative equations of the DKF, as they are the same as  \eqref{eq:SSM-a}--\eqref{eq:SSM-d}, with the specificity that $d_{\myz}$ and $d_{\myx}$ are DNNs here.\footnote{In fact, a Bernouilli distribution was considered for $\myx[t]$ by \citet{krishnan2015deep, krishnan2017structured}, but we have already mentioned that different pdfs can be considered for $p_{\theta_{\myx}}(\myx[t]|\myz[t])$ depending on the data, without affecting the fundamental issues of this review. Hence, we consider here a Gaussian distribution for a better comparison with the other models.} \citet{krishnan2015deep} did not specify these DNNs; we can assume that basic feed-forward neural networks (i.e., MLPs) were used. In their second paper, \citet{krishnan2017structured} implemented $d_{\myx}$  with a two-layer MLP and used a slightly more refined model for $d_{\myz}$: a gated linear combination of a linear model and an MLP for the mean parameter, where the gate is itself provided by an MLP, and the chaining of MLP, rectified linear unit (ReLU) activation, and Softmax activation layers for the variance parameter (see Section~\ref{sec:DKF-implementation-speech}). According to the authors, this allows ``the model have the flexibility to choose a linear transition for some dimensions while having a nonlinear transition for the others.'' 

Even if we are still at an early point in our presentation of the different DVAE models, we can make a first series of remarks to clarify the links between the models we have discussed so far.
\begin{itemize}[leftmargin=*]
\item The stochastic state $\myz[t]$ of an SSM is similar in essence to the latent state of the VAE. In the present DKF case, where $d_{\myx}$ is implemented with a DNN, if $\myz[t]$ is of reduced dimension compared to $\myx[t]$, \emph{the DKF observation model is identical to the VAE decoder}. 
\item Consequently, a DKF can be viewed as a VAE decoder with a temporal (Markovian) model of the latent variable $\myz$.
\item A (deep) SSM can also be viewed as a ``fully stochastic'' version of a (deep) RNN, where stochasticity is introduced at both the observation model level (like a GRNN) \emph{and} internal state level. As mentioned before, in an SSM, the deterministic internal state $\myh[t]$ of the (G)RNN is simply replaced with a stochastic state $\myz[t]$.
\item In summary, DKF = deep SSM = ``Markovian'' VAE decoder = ``fully stochastic'' RNN. The graphical model of the DKF is given by the right-hand schema in Figure~\ref{fig:GRNN-SSM} (which does not make the DNNs apparent). 
\end{itemize}

\section{Inference model}
\label{sec:DKF-inference}

Following the general line of Section~\ref{sec:DVAE-inference}, we first identify the structure of the SSM/DKF posterior distribution $p_{\theta}(\myz[1:T] | \myx[1:T],\myu[1:T])$. Let us first recall that applying the chain rule enables us to rewrite this distribution as follows:
\begin{align}
p_{\theta}(\myz[1:T] | \myx[1:T], \myu[1:T]) &= \prod_{t=1}^{T} p_{\theta}(\myz[t] | \myz[1:t-1], \myx[1:T],  \myu[1:T]).
\end{align}
Then, D-separation can be used to simplify each term of the product. The structure presented in Figure~\ref{fig:GRNN-SSM} (right) shows that the $\myz[t-1]$ node ``blocks'' all information coming from the past and flowing to $\myz[t]$ (i.e., $\myz[1:t-2]$, $\myx[1:t-1]$, and $\myu[1:t-1]$). In other words, $\myz[t-1]$ has accumulated this past information or is a summary of this information. We thus have $p_{\theta}(\myz[t] | \myz[1:t-1], \myx[1:T],  \myu[1:T]) = p_{\theta}(\myz[t] | \myz[t-1], \myx[t:T],  \myu[t:T])$, and therefore (with $\myz[0]$ being arbitrarily set)
\begin{align}
p_{\theta}(\myz[1:T] | \myx[1:T], \myu[1:T]) &= \prod_{t=1}^{T} p_{\theta}(\myz[t] | \myz[t-1], \myx[t:T],  \myu[t:T]).
\end{align}
At each time $t$, the posterior distribution of $\myz[t]$ depends on the previous latent state $\myz[t-1]$ and on the present and \emph{future} observations $\myx[t:T]$ and inputs $\myu[t:T]$ (it is thus a first-order Markovian causal process on $\myz[t]$ combined with an anticausal process on $\myx[t]$ and $\myu[t]$). 

\citet{krishnan2015deep} indicated this structure and the fact that we should inspire from it to design the approximate posterior $q_{\phi}$. However, somewhat surprisingly, they proposed the following four different models: 
\begin{itemize}[leftmargin=*]
\item an instantaneous model: $q_{\phi}(\myz[t] | \myx[t], \myu[t])$ parameterized by an MLP;
\item a model with local past and future context: $q_{\phi}(\myz[t] | \myx[t-1:t+1], \myu[t-1:t+1])$ parameterized by an MLP;
\item a model with the complete past context: $q_{\phi}(\myz[t] | \myx[1:t], \myu[1:t])$ parameterized by an RNN;
\item a model with the whole sequence: $q_{\phi}(\myz[t] | \myx[1:T], \myu[1:T])$ parameterized by a bidirectional RNN.
\end{itemize}
We do not detail these implementations of $q_{\phi}$ here, as we will provide other detailed examples in the next chapters. One might wonder why $\myz[t-1]$ is not present in the conditional variables of the approximate posterior, but this might just be an oversight from the authors. This is difficult to be determined from the paper, as the implementation is not detailed. The authors mention an RNN to model the dependencies on $\myx$ and $\myu$, but the implementation of the dependency on $\myz[t-1]$ is not specified. The point is that, although the authors pointed out the dependency of the exact posterior on the present and future observations and inputs, they did not propose a corresponding approximate model.  

\vspace{0.3cm}
\noindent \textbf{Notation remark}: \citet{krishnan2015deep} denoted by $\myu[t-1]$ the input at time $t$ in the generative model, as in many control theory papers published on SSMs, and not $\myu[t]$, as we do (as pointed out in a previous footnote). However, when defining the four approximate posterior models, they did it exactly as we report here (i.e., with $\myu[t]$ being synchronous to $\myz[t]$ and $\myx[t]$). We conjecture that this problem is just a notation mistake made by \citet{krishnan2015deep}, which we have implicitly corrected by using $\myu[t]$ instead of $\myu[t-1]$ as the input at time $t$ in the generative model.
\vspace{0.3cm}

 \citet{krishnan2017structured} proposed the same generative model, renamed it DMM and presented it in the undriven mode, in which $\myu[1:T]$ was simply removed. However, they largely clarified and improved on their previous paper regarding the inference model. They proposed a new series of inference models that clearly do or do not depend on $\myz[t-1]$ and varied the dependency on the observed data. They again considered the case of dependency on the past and present data sequence $\myx[1:t]$ and on the complete data sequence $\myx[1:T]$. More importantly, they now also consider the case of an inference model with a functional form that corresponds exactly to the form of the exact posterior distribution, namely $q_{\phi}(\myz[t] | \myz[t-1], \myx[t:T])$. In this case, for a complete data sequence, we have
\begin{align}
q_{\phi}(\myz[1:T] | \myx[1:T]) &=  \prod_{t=1}^T q_{\phi}(\myz[t] | \myz[t-1], \myx[t:T]). \label{eq:DKF-inference-seq-a}
\end{align}
This model is referred to as the deep Kalman smoother (DKS), as it combines information from the past, through $\myz[t-1]$, and information from the present and future observations.
 
For conciseness, we report the detailed inference equations only for the DKS and will comment on their extension to the other proposed inference models. The DKS is implemented with a backward RNN on $\myx[t]$, followed by an additional layer for combining the RNN output with $\myz[t-1]$:      
\begin{align}
\mygbw[t] &= e_{\mygbw}(\mygbw[t+1],\myx[t]), \label{eq:DKF-inference-a} \\
\myg[t] &= \frac{1}{2}\big(\tanh(\mathbf{W}\myz[t-1] + \mathbf{b}) + \mygbw[t]\big), \label{eq:DKF-inference-b}\\
[\boldsymbol{\mu}_{\phi}(\myg[t]), \boldsymbol{\sigma}_{\phi}(\myg[t])] &= e_{\myz}(\myg[t]), \label{eq:DKF-inference-c} \\
q_{\phi}(\myz[t] | \myg[t]) &= \mathcal{N}\big(\myz[t]; \boldsymbol{\mu}_{\phi}(\myg[t]), \text{diag}\{\boldsymbol{\sigma}_{\phi}^2(\myg[t])\}\big). \label{eq:DKF-inference-d}
\end{align}
In the above equations, $e_{\myz}$ is a basic combining network, parameterized by $\phi_{\myz}$, $\boldsymbol{\mu}_{\phi}(\myg[t])$ is an affine function of $\myg[t]$, and  $\boldsymbol{\sigma}_{\phi}^2(\myg[t])$ is a Softplus of an affine function of $\myg[t]$. We thus have $\phi = \phi_{\mygbw} \cup \phi_{\myz}$, assuming $\{\mathbf{W},\mathbf{b}\} \in \phi_{\myz}$ for simplicity.

Because of the recursivity in \eqref{eq:DKF-inference-a}, we can see $\myg[t]$ as an unfolded deterministic function of $\myz[t-1]$ and $\myx[t:T]$, which we can rewrite as $\myg[t] = \myg[t](\myz[t-1], \myx[t:T])$,\footnote{This function is also a function of the initial state $\mygbw[T]$ of the backward RNN.} and we have $q_{\phi}(\myz[t] | \myg[t]) = q_{\phi}(\myz[t] | \myg[t](\myz[t-1], \myx[t:T]))$. 
For a complete data sequence, we have 
\begin{align}
q_{\phi}(\myz[1:T] | \myx[1:T]) &=  \prod_{t=1}^T q_{\phi}(\myz[t] | \myg[t]) = \prod_{t=1}^T q_{\phi}\big(\myz[t] | \myg[t](\myz[t-1], \myx[t:T])\big),
\end{align}
which is just a rewriting of \eqref{eq:DKF-inference-seq-a}. In short, the inference of $\myz[t]$ with a DKS requires a first backward pass from $\myx[T]$ up to $\myx[t]$ to compute $\mygbw[t]$, which is then combined with $\myz[t-1]$. 

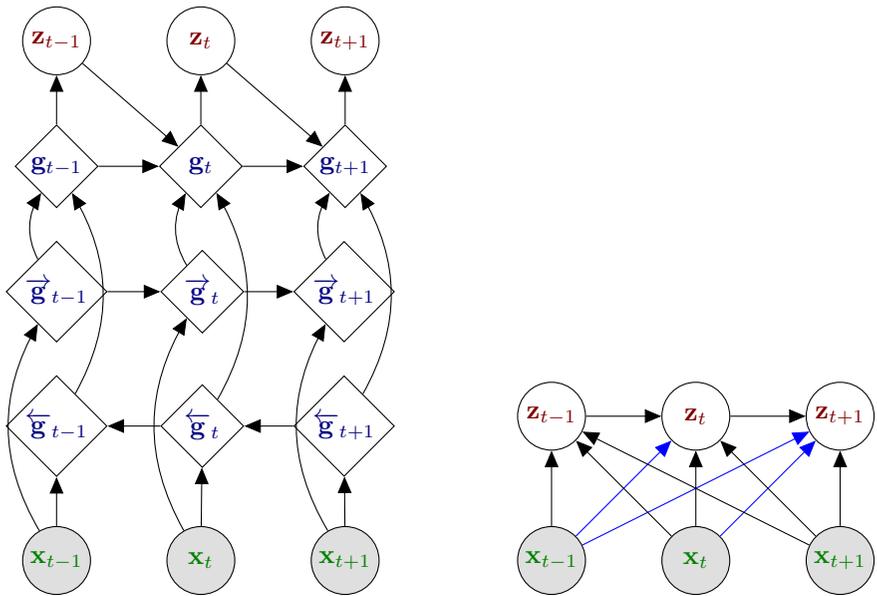
\begin{figure}
\centering
    \begin{tikzpicture}[->]
    \node[latent,minimum size=9mm] (zm) {$\myz[t-1]$};
    \node[latent,minimum size=9mm,right=of zm] (z) {$\myz[t]$};
    \node[latent,minimum size=9mm,right=of z] (zp) {$\myz[t+1]$};
    \node[det,minimum size=11mm,below=of zm, yshift=3.5mm] (gm) {$\myg[t-1]$};
    \node[det,minimum size=11mm,right=of gm, xshift=-2mm] (g) {$\myg[t]$};
    \node[det,minimum size=11mm,right=of g, xshift=-2mm] (gp) {$\myg[t+1]$};
    \node[det,minimum size=11mm,below=of gm, yshift=5.7mm] (gfm) {$\mygfw[t-1]$};
    \node[det,minimum size=11mm,right=of gfm, xshift=-3mm] (gf) {$\mygfw[t]$};
    \node[det,minimum size=11mm,right=of gf, xshift=-3.5mm] (gfp) {$\mygfw[t+1]$};
    \node[det,minimum size=11mm,below=of gfm, yshift=5.7mm] (gbm) {$\mygbw[t-1]$};
    \node[det,minimum size=11mm,right=of gbm, xshift=-3mm] (gb) {$\mygbw[t]$};
    \node[det,minimum size=11mm,right=of gb, xshift=-3.5mm] (gbp) {$\mygbw[t+1]$};
    \node[obs,minimum size=9mm,below=of gbm, yshift=3.5mm] (xm) {$\myx[t-1]$};
    \node[obs,minimum size=9mm,right=of xm] (x) {$\myx[t]$};
    \node[obs,minimum size=9mm,right=of x] (xp) {$\myx[t+1]$};
    \edge{zm} {g};
    \edge{z} {gp};
    \edge{gm} {zm};
    \edge{g} {z};
    \edge{gp} {zp};
    \edge{xm} {gbm};
    \edge{x} {gb};
    \edge{xp} {gbp};
    \edge{gm} {g};
    \edge{g} {gp};
    \edge{gfm} {gf};
    \edge{gf} {gfp};
    \edge{gbp} {gb};
    \edge{gb} {gbm};
    \path (gbm) edge[bend right] (gm);
    \path (gb) edge[bend right] (g);
    \path (gbp) edge[bend right] (gp);
    \path (xm) edge[bend left] (gfm);
    \path (x) edge[bend left] (gf);
    \path (xp) edge[bend left] (gfp);
    \path (gfm) edge[bend left] (gm);
    \path (gf) edge[bend left] (g);
    \path (gfp) edge[bend left] (gp);
\end{tikzpicture}\hspace{14mm}
\begin{tikzpicture}[->]
    \node[latent,minimum size=9mm] (zm) {$\myz[t-1]$};
    \node[latent,minimum size=9mm,right=of zm] (z) {$\myz[t]$};
    \node[latent,minimum size=9mm,right=of z] (zp) {$\myz[t+1]$};
    \node[obs,minimum size=9mm,below=of zm] (xm) {$\myx[t-1]$};
    \node[obs,minimum size=9mm,right=of xm] (x) {$\myx[t]$};
    \node[obs,minimum size=9mm,right=of x] (xp) {$\myx[t+1]$};
    \edge{zm} {z};
    \edge{z} {zp};
    \edge{xm} {zm};
    \edge[blue]{xm} {z};
    \edge[blue]{xm} {zp};
    \edge{x} {zm};
    \edge{x} {z};
    \edge[blue]{x} {zp};
    \edge{xp} {zm};
    \edge{xp} {z};
    \edge{xp} {zp};
\end{tikzpicture}
    \caption{Graphical model of DKF at inference time corresponding to the inference model $q_{\phi}(\myz[t] | \myz[t-1], \myx[1:T])$, in developed form (left) and compact form (right). The specific DKS model, which functional form $q_{\phi}(\myz[t] | \myz[t-1], \myx[t:T])$ perfectly corresponds to the form of the exact posterior distribution, is obtained by removing the forward RNN (and removing the blue arrows on the right-hand schema).}
    \label{fig:DKS-inference}
\end{figure}

As mentioned above, this inference model is extended to a noncausal (bidirectional) model regarding the observations, $q_{\phi}(\myz[t] | \myz[t-1], \myx[1:T])$. This is done by adding a forward RNN on $\myx[t]$ and sending its output $\mygfw[t]$ to the combining network, in addition to $\myz[t-1]$ and $\mygbw[t]$. For conciseness, we do not report the corresponding detailed equations, but this more general model is represented in Figure~\ref{fig:DKS-inference}. The other models proposed by \citet{krishnan2017structured} can be deduced from this general model by removing some elements. In particular, DKS is obtained by simply removing the forward RNN. Moreover, models that do not depend on $\myz[t-1]$ are obtained by removing the arrows between $\myz[t-1]$ and $\myg[t]$ for all $t$.

\section{Training}
\label{sec:DKF-training}

A comparison of the compact form of the DKF model in \eqref{eq:SSM-joint-seq} with the general compact form of a DVAE in \eqref{time_slide_ordered_model} shows that the DKF model makes the following conditional independence assumptions:
\begin{align}
    p_{\theta_{\myx}}(\myx[t] | \myx[1:t-1],\myz[1:t],\myu[1:t] ) &= p_{\theta_{\myx}}(\myx[t] | \myz[t] );  \nonumber \\
    p_{\theta_{\myz}}(\myz[t] | \myx[1:t-1],\myz[1:t-1],\myu[1:t]) &= p_{\theta_{\myz}}(\myz[t] | \myz[t-1],\myu[t]).
\end{align}
Using these two simplifications along with the inference model \eqref{eq:DKF-inference-seq-a} (extended to be in the driven mode for the sake of generality), the VLB in its most general form \eqref{VLB1_dev} can be simplified as follows:
\begin{align}
& \mathcal{L}(\theta, \phi ; \myx[1:T], \myu[1:T]) = \sum_{t=1}^{T} \mathbb{E}_{q_{\phi}(\myz[t] | \myx[1:T], \myu[1:T])}\big[  \log p_{\theta_{\myx}}(\myx[t] | \myz[t] ) \big] \nonumber \\
& \hspace{0.2cm} - \sum_{t=1}^{T}  \mathbb{E}_{q_{\phi}(\myz[t-1] | \myx[1:T], \myu[1:T])}\left[ D_{\text{KL}}\left(q_{\phi}(\myz[t] | \myz[t-1], \myx[t:T], \myu[t:T]) \parallel  p_{\theta_{\myz}}(\myz[t] | \myz[t-1],\myu[t]) \right)\right].
\label{VLB_DKF}
\end{align}
The KL divergence in \eqref{VLB_DKF} can be computed analytically, while the two expectations are intractable. \citep{krishnan2015deep,krishnan2017structured} provided no detailed information regarding how to approximate these expectations; they only mentioned that ``stochastic backpropagation'' is used, referring the reader to the papers of \citet{Kingma2014} and \citet{rezende2014stochastic}, who introduced the reparameterization trick for standard ``static'' VAEs. However, due to the dynamical nature of the model, the sampling procedure required for stochastic backpropagation in DKF is more complicated than in standard VAEs. In particular, we do not have an analytical form for $q_{\phi}(\myz[t] | \myx[1:T], \myu[1:T])$, and only have one for $q_{\phi}(\myz[\tau] | \myz[\tau-1], \myx[\tau:T], \myu[\tau:T])$. Therefore, we need to exploit the chain rule and the ``cascade trick'' to develop and then approximate the intractable expectations in \eqref{VLB_DKF}.
The first expectation in this VLB expression can be developed as follows:
\begin{align}
\mathbb{E}_{q_{\phi}(\myz[t] | \myx[1:T], \myu[1:T])}[f(\myz[t])] &= \mathbb{E}_{q_{\phi}(\myz[1:t] | \myx[1:T], \myu[1:T])}[f(\myz[t])] \nonumber \\
&=
\mathbb{E}_{q_{\phi}(\myz[1] | \myx[1:T], \myu[1:T])}[\nonumber \\
& \hspace{.75cm} \mathbb{E}_{q_{\phi}(\myz[2] | \myz[1], \myx[2:T], \myu[2:T])}[
\, ... \,\nonumber \\
& \hspace{1cm} \mathbb{E}_{q_{\phi}(\myz[t] | \myz[t-1], \myx[t:T], \myu[t:T])}[f(\myz[t])] \, ... \,]],
\end{align}
where $f(\myz[t])$ denotes an arbitrary function of $\myz[t]$. A similar procedure can be used to develop the second expectation in \eqref{VLB_DKF}. Each intractable expectation in this cascade of expectations can then be approximated with a Monte Carlo estimate. It requires to sample $q_{\phi}(\myz[\tau] | \myz[\tau-1], \myx[\tau:T], \myu[\tau:T])$ iteratively from $\tau=1$ to $t$, using the same reparametrization trick as in standard VAEs. Doing so, the VLB becomes differentiable w.r.t.~$\theta = \theta_{\myx} \cup \theta_{\myz}$ and $\phi = \phi_{\mygbw} \cup \phi_{\myz}$, and it can be optimized with gradient-ascent-based techniques.

\chapter{Kalman Variational Autoencoders}
\label{sec:KVAE}

\begin{figure}[t]
\centering
\begin{tikzpicture}[->]
    \node[latent,minimum size=9mm] (um) {$\myu[t-1]$};
    \node[latent,minimum size=9mm,right=of um] (u) {$\myu[t]$};
    \node[latent,minimum size=9mm,right=of u] (up) {$\myu[t+1]$};
    \node[latent,minimum size=9mm,below=of um, yshift=4mm] (zm) {$\myz[t-1]$};
    \node[latent,minimum size=9mm,right=of zm] (z) {$\myz[t]$};
    \node[latent,minimum size=9mm,right=of z] (zp) {$\myz[t+1]$};
    \node[latent,minimum size=9mm,below=of zm, yshift=4mm] (am) {$\mya[t-1]$};
    \node[latent,minimum size=9mm,right=of am] (a) {$\mya[t]$};
    \node[latent,minimum size=9mm,right=of a] (ap) {$\mya[t+1]$};
   \node[obs,minimum size=9mm,below=of am, yshift=4mm] (xm) {$\myx[t-1]$};
    \node[obs,minimum size=9mm,right=of xm] (x) {$\myx[t]$};
    \node[obs,minimum size=9mm,right=of x] (xp) {$\myx[t+1]$};
    \edge {um} {zm};
    \edge {zm} {am,z};
    \edge{u} {z};
    \edge{z} {a,zp};
    \edge{up} {zp};
    \edge{zp} {ap};
    \edge{am} {xm};
    \edge{a} {x};
    \edge{ap} {xp};
\end{tikzpicture}
\caption{KVAE’s graphical model.}
\label{fig:KVAE}
\end{figure}
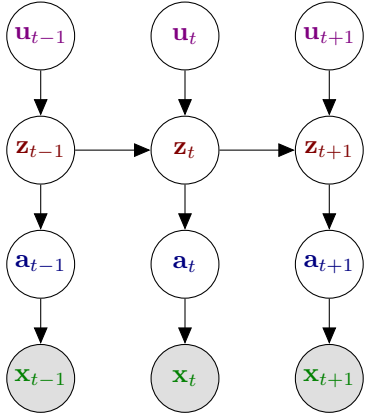

The KVAE model was presented by \citet{fraccaro2017disentangled}. This model can be considered a variant of the DKF model, and hence as another deep SSM, where an additional random variable, denoted $\mya[t]$, is inserted between the latent vector $\myz[t]$ and the observed vector $\myx[t]$, as illustrated in Figure~\ref{fig:KVAE}. This enables us to separate the model into two parts: A deep feature extractor linking $\mya[t]$ and $\myx[t]$ and the dynamical model on $\myz[t]$ with ``new observations'' $\mya[t]$. As we will see below, this provides the model with interesting properties for inference and training.

\section{Generative model}
\label{sec:KVAE-generative-model}

The general formulation of the KVAE model is given by
\begin{align}
[\boldsymbol{\mu}_{\theta_{\myz}}(\myz[t-1],\myu[t]), \boldsymbol{\sigma}_{\theta_{\myz}}(\myz[t-1],\myu[t])] &= d_{\myz}(\myz[t-1],\myu[t]), \label{eq:KVAE-a} \\
p_{\theta_{\myz}}(\myz[t] | \myz[t-1],\myu[t]) &= \mathcal{N}\big(\myz[t]; \boldsymbol{\mu}_{\theta_{\myz}}(\myz[t-1],\myu[t]), \text{diag}\{\boldsymbol{\sigma}_{\theta_{\myz}}^2(\myz[t-1],\myu[t])\}\big), \label{eq:KVAE-b} \\
[\boldsymbol{\mu}_{\theta_{\mya}}(\myz[t]),\boldsymbol{\sigma}_{\theta_{\mya}}(\myz[t])] &= d_{\mya}(\myz[t]),  \label{eq:KVAE-c} \\
p_{\theta_{\mya}}(\mya[t] | \myz[t]) &= \mathcal{N}\big(\mya[t];\boldsymbol{\mu}_{\theta_{\mya}}(\myz[t]),\text{diag}\{\boldsymbol{\sigma}_{\theta_{\mya}}^2(\myz[t])\}\big),  \label{eq:KVAE-d} \\
[\boldsymbol{\mu}_{\theta_{\myx}}(\mya[t]),\boldsymbol{\sigma}_{\theta_{\myx}}(\mya[t])] &= d_{\myx}(\mya[t]),  \label{eq:KVAE-e} \\
p_{\theta_{\myx}}(\myx[t] | \mya[t]) &= \mathcal{N}\big(\myx[t];\boldsymbol{\mu}_{\theta_{\myx}}(\mya[t]),\text{diag}\{\boldsymbol{\sigma}_{\theta_{\myx}}^2(\mya[t])\}\big).  \label{eq:KVAE-f} 
\end{align}
In \citeauthor{fraccaro2017disentangled}'s \citeyearpar{fraccaro2017disentangled} paper, \eqref{eq:KVAE-a} and \eqref{eq:KVAE-c} are linear equations; that is, they are given as \eqref{eq:LDS-a} and \eqref{eq:LDS-b}, respectively (with $\mya$ in place of $\myx$, null bias vectors $\mathbf{n}_t$ and $\mathbf{m}_t$, and time-invariant covariance matrices $\boldsymbol{\Lambda}$ and $\boldsymbol{\Sigma}$). Therefore, we have $\theta_{\myz} = \{\mathbf{A}_t, \mathbf{B}_t  \}_{t=1}^T\cup\{\boldsymbol{\Lambda}\}$ and $\theta_{\mya} = \{\mathbf{C}_t  \}_{t=1}^T\cup\{\boldsymbol{\Sigma}\}$, and the submodel on $\{\myu[t],\myz[t],\mya[t]\}$ is a classical (non-deep) LG-LDS. In contrast, $d_{\myx}$ in \eqref{eq:KVAE-e} is implemented with a DNN, with parameter set $\theta_{\myx}$ (e.g., a basic MLP or a CNN for video sequence modeling). This network plays the role of a deep feature extractor, with the dimension of $\mya[t]$ being possibly much smaller than that of $\myx[t]$. The feature vector $\mya[t]$ is expected to encode the properties of the ``object(s)'' present in the observation $\myx[t]$, whereas $\myz[t]$ is expected to encode the dynamics of these objects, which is an important application of the previously described disentanglement concept. As we will see, in the KVAE case, this can be a great advantage for solving the dynamical part of the model. 

The joint distribution of all random variables can be factorized as follows:
\begin{align}
p_{\theta}(\myx[1:T], \mya[1:T], \myz[1:T], \myu[1:T]) &= \prod_{t=1}^{T} p_{\theta_{\myx}}(\myx[t] | \mya[t]) p_{\theta_{\mya}}(\mya[t] | \myz[t]) p_{\theta_{\myz}}(\myz[t] | \myz[t-1], \myu[t])p(\myu[t]), \label{eq:KVAE-joint-seq}
\end{align}
and we also have
\begin{align}
p_{\theta_{\myx}}(\myx[1:T] | \mya[1:T]) &= \prod_{t=1}^{T} p_{\theta_{\myx}}(\myx[t] | \mya[t]), \label{eq:KVAE-seq-a} \\
p_{\theta_{\mya}}(\mya[1:T] | \myz[1:T]) &= \prod_{t=1}^{T} p_{\theta_{\mya}}(\mya[t] | \myz[t]), \label{eq:KVAE-seq-b} \\
p_{\theta_{\myz}}(\myz[1:T] | \myu[1:T]) &= \prod_{t=1}^{T} p_{\theta_{\myz}}(\myz[t] | \myz[t-1], \myu[t]). \label{eq:SSM-seq-c}
\end{align}
Given the state sequence $\myz[1:T]$, the features $\mya[1:T]$ are independent, and given the sequence of features, the observations $\myx[1:T]$ are independent. 

\citet{fraccaro2017disentangled} mentioned the classical limitation of LDS for modeling abrupt changes in trajectories. A classical solution to this problem is to include in the model a ``switching strategy'' between different models or different parameterizations of the model, see, e.g., the switching Kalman filter \citep{murphy1998switching, fox2011bayesian}. \citet{fraccaro2017disentangled} proposed to define each LDS parameter at each time $t$ (e.g., matrix $\mathbf{A}_t$) as a linear combination of predefined matrices/vectors from a parameter bank, and the coefficients of the linear combination were estimated at each time $t$ from the past features $\mya[1:t-1]$ using an LSTM network. Although it is an interesting contribution on its own, we do not further consider this part of the KVAE model here, as it is loosely relevant to our model review. A similar transition model was proposed independently by \citet{karl2016deep} as an instance of a deep variational Bayesian filter (DVBF), an extended class of SSM-based DVAE models enriched with stochastic transition parameters (see also \citet{watter2015embed}).

\section{Inference model}

For the KVAE model, the posterior distribution of all latent variables given the observed variables, $p_{\theta}(\mya[1:T],\myz[1:T] | \myx[1:T],\myu[1:T])$, can be factorized as follows:
\begin{align}
p_{\theta}(\mya[1:T],\myz[1:T] | \myx[1:T],\myu[1:T]) &=  p_{\theta}(\myz[1:T] | \mya[1:T], \myx[1:T], \myu[1:T])p_{\theta}(\mya[1:T] | \myx[1:T],\myu[1:T]), \nonumber \\
&=  p_{\theta}(\myz[1:T] | \mya[1:T], \myu[1:T])p_{\theta}(\mya[1:T] | \myx[1:T],\myu[1:T]),
\label{eq:KVAE-inference-true}
\end{align}
where the simplification of the first term on the right-hand side results from D-separation. 
A keypoint that appears here is that, if the sequence of features $\mya[1:T]$ is known, then $p_{\theta}(\myz[1:T] | \mya[1:T],\myu[1:T])$ has a closed-form solution, which is a Kalman filter or a Kalman smoother (see Section~\ref{subsec:KF}). This Kalman solution is classic, and it is not detailed here for conciseness. We can simply mention that it depends only on $\theta_{\mya} \cup \theta_{\myz}$ but not on $\theta_{\myx}$. The other factor, $p_{\theta}(\mya[1:T] | \myx[1:T],\myu[1:T])$, is more problematic. 

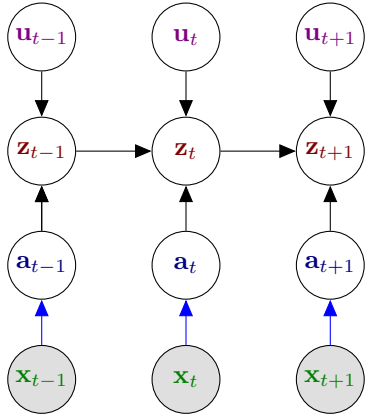
\begin{figure}[t]
\centering
\begin{tikzpicture}[->]
    \node[latent,minimum size=9mm] (um) {$\myu[t-1]$};
    \node[latent,minimum size=9mm,right=of um] (u) {$\myu[t]$};
    \node[latent,minimum size=9mm,right=of u] (up) {$\myu[t+1]$};
    \node[latent,minimum size=9mm,below=of um, yshift=4mm] (zm) {$\myz[t-1]$};
    \node[latent,minimum size=9mm,right=of zm] (z) {$\myz[t]$};
    \node[latent,minimum size=9mm,right=of z] (zp) {$\myz[t+1]$};
    \node[latent,minimum size=9mm,below=of zm, yshift=4mm] (am) {$\mya[t-1]$};
    \node[latent,minimum size=9mm,right=of am] (a) {$\mya[t]$};
    \node[latent,minimum size=9mm,right=of a] (ap) {$\mya[t+1]$};
   \node[obs,minimum size=9mm,below=of am, yshift=4mm] (xm) {$\myx[t-1]$};
    \node[obs,minimum size=9mm,right=of xm] (x) {$\myx[t]$};
    \node[obs,minimum size=9mm,right=of x] (xp) {$\myx[t+1]$};
    \edge {um} {zm};
    \edge{u} {z};
    \edge{up} {zp};
    \edge{am} {zm};
    \edge{a} {z};
    \edge{ap} {zp};
    \edge{am} {zm};
    \edge{zm} {z};
    \edge{z} {zp};
    \edge[blue] {xm} {am};
    \edge[blue]{x} {a};
    \edge[blue]{xp} {ap};
\end{tikzpicture}
\caption{KVAE’s graphical model at inference time. The black arrows represent the Kalman filter solution (causal solution) of LG-LDS on $\{\myu[1:T]$, $\myz[1:T]$, $\mya[1:T]\}$. The blue arrows represent the VAE encoder from $\myx[t]$ to $\mya[t]$. The inference solution for the complete KVAE model is a combination of these two.}
\label{fig:KVAE-inference}
\end{figure}

\citet{fraccaro2017disentangled} did not discuss the form of the exact posterior distribution, yet they proposed the following factorized inference model, which exploits the Kalman solution:
\begin{align}
q_{\phi}(\mya[1:T],\myz[1:T] | \myx[1:T],\myu[1:T]) &=  p_{\theta}(\myz[1:T] | \mya[1:T],\myu[1:T]) q_{\phi}(\mya[1:T] | \myx[1:T]) \\
&=  p_{\theta}(\myz[1:T] | \mya[1:T],\myu[1:T]) \prod_{t=1}^{T} q_{\phi}(\mya[t] | \myx[t]),   \label{eq:KVAE-inference-seq-a}
\end{align}
where 
\begin{align}
q_{\phi}(\mya[t] | \myx[t]) = \mathcal{N}\big(\mya[t];\boldsymbol{\mu}_{\phi}(\myx[t]),\text{diag}\{\boldsymbol{\sigma}_{\phi}^2(\myx[t])\}\big) \label{eq:KVAE-inference-feature-a}
\end{align}
is implemented with a fully-connected DNN:
\begin{align}
[\boldsymbol{\mu}_{\phi}(\myx[t]), \boldsymbol{\sigma}_{\phi}(\myx)] = e_{\mya}(\myx[t]). \label{eq:KVAE-inference-feature-b}
\end{align}
Eqs. \eqref{eq:KVAE-e}, \eqref{eq:KVAE-f}, \eqref{eq:KVAE-inference-feature-a}, and \eqref{eq:KVAE-inference-feature-b} are identical to \eqref{eq:VAE-decoder-a-bis}, \eqref{eq:VAE-decoder-a}, \eqref{eq:VAE-encoder-a}, and \eqref{eq:VAE-encoder-a-bis}, respectively, with $\mya[t]$ substituting $\myx$ and $\myz[t]$ substituting $\myz$. This means that, with the proposed inference model, a KVAE is composed of a VAE modeling the relationship between $\mya[t]$ and $\myx[t]$, placed on top of an LG-LDS on $\myu[t]$, $\myz[t]$ and $\mya[t]$. Therefore, the inference solution of the VAE and that of LG-LDS can be combined for the solution of this combined model. This is illustrated in Figure~\ref{fig:KVAE-inference}.

This inference model is thus designed to benefit from both the VAE methodology and the well-known efficiency of the ``simple'' LG-LDS model for tracking data dynamics. The Kalman solution, which requires inverse matrix calculation, greatly benefits from the notable $\myx[t]$-to-$\mya[t]$ dimension reduction. In addition, although this was not exactly discussed in these terms by \citet{fraccaro2017disentangled}, one possible motivation for designing the KVAE model is that the joint learning of all parameters (see the next subsection) can encourage the feature extractor to provide $\mya[t]$ features that are well-suited to a linear dynamical model; that is, the nonlinear relations between observations $\myx[t]$ and dynamics $\myz[t]$ are (at least largely) captured by the VAE. 

\section{Training}
The VLB for the KVAE model writes as follows:
\begin{align}
    &\!\!\! \mathcal{L}(\phi,\theta;\myx[1:T],\myu[1:T]) = \mathbb{E}_{q(\myz[1:T],\mya[1:T]|\myx[1:T],\myu[1:T])} \Big[ \log p_{\theta}(\myx[1:T]|\myz[1:T],\mya[1:T],\myu[1:T])\Big] \nonumber \\
    & - D_{\textit{KL}} \big(q_\phi(\myz[1:T],\mya[1:T]|\myx[1:T],\myu[1:T]) \parallel p_{\theta}(\myz[1:T],\mya[1:T]|\myu[1:T])\big) \\
    &\!\!\!= \mathbb{E}_{q_{\phi}(\mya[1:T]|\myx[1:T])p_{\theta}(\myz[1:T]|\mya[1:T],\myu[1:T])} \Big[ \log p_{\theta}(\myx[1:T]|\mya[1:T])\Big] \nonumber \\
    & - D_{\textit{KL}} \big(q_{\phi}(\mya[1:T]|\myx[1:T])p_{\theta}(\myz[1:T]|\mya[1:T],\myu[1:T]) \parallel p_{\theta}(\myz[1:T],\mya[1:T]|\myu[1:T])\big) \\
    &\!\!\! = \mathbb{E}_{q_{\phi}(\mya[1:T]|\myx[1:T])} \Big[ \log p_{\theta}(\myx[1:T]|\mya[1:T])/q_{\phi}(\mya[1:T]|\myx[1:T]) \nonumber \\
    & - D_{\textit{KL}} \big(p_{\theta}(\myz[1:T]|\mya[1:T],\myu[1:T]) \parallel p_{\theta}(\mya[1:T]|\myz[1:T])p_{\theta}(\myz[1:T]|\myu[1:T])\big) \Big],
\end{align}
where we use the proposed decomposition of both the generative and inference models.

In practice, one first samples from $q_{\phi}(\mya[t]|\myx[t])$ for each $t$. These samples are fed to a standard Kalman smoother that computes $p_{\theta}(\myz[1:T]|\mya[1:T],\myu[1:T])$. We can then easily sample from this distribution. This allows for jointly learning the parameters of the VAE (both the encoder and decoder) and those of the LG-LDS.

\chapter{STOchastic Recurrent Networks}
\label{sec:STORN}
To the best of our knowledge, STORN \citep{bayer2014learning} is the first DVAE model to combine an internal deterministic state $\myh[t]$ and an internal stochastic state $\myz[t]$. \citet{bayer2014learning} presented STORN in the undriven and predictive modes (i.e., with $\myu[t] = \myx[t-1]$). Hereinafter, we retain this mode for STORN, VRNN, and SRNN, for an easier comparison, but these models can also be easily set up in the driven mode with an external input $\myu[t]$. 

\section{Generative model}

The STORN observation model is given by  
\begin{align}
\myh[t] &= d_{\hid}(\mathbf{W}_{\inp}\myx[t-1] + \mathbf{W}_{\text{lat}} \myz[t] + \mathbf{W}_{\rec} \myh[t-1] + \mathbf{b}_{\hid}), \label{eq:STORN-1} \\
[\boldsymbol{\mu}_{\theta_{\myx}}(\myh[t]),\boldsymbol{\sigma}_{\theta_{\myx}}(\myh[t])] &= d_\out(\mathbf{W}_\out \myh[t] + \mathbf{b}_\out), \label{eq:STORN-2} \\
p_{\theta_{\myx}}(\myx[t] | \myh[t])  &= \mathcal{N}\big(\myx[t];\boldsymbol{\mu}_{\theta_{\myx}}(\myh[t]),\text{diag}\{\boldsymbol{\sigma}_{{\theta}_{\myx}}^2(\myh[t])\}\big).   \label{eq:STORN-3}
\end{align}
Therefore, \eqref{eq:STORN-2} and \eqref{eq:STORN-3} are the same as those for a basic single-layer GRNN, but in STORN, $\myz[t]$ forms an input additional to the internal state $\myh[t]$. Moreover, STORN was originally presented in the above framework of a single-layer RNN, but it can be easily generalized to a deep RNN, defined by \eqref{eq:GRNN-a}--\eqref{eq:GRNN-c}, by inserting $\myz[t]$ as an additional input to the network (and setting $\myu[t] = \myx[t-1]$):
\begin{align}
\myh[t] &= d_{\myh}(\myx[t-1], \myz[t], \myh[t-1]), \label{eq:deep-STORN-a} \\
[\boldsymbol{\mu}_{\theta_{\myx}}(\myh[t]),\boldsymbol{\sigma}_{\theta_{\myx}}(\myh[t])] &= d_{\myx}(\myh[t]), \label{eq:deep-STORN-b} \\
p_{\theta_{\myx}}(\myx[t] | \myh[t]) &= \mathcal{N}\big(\myx[t];\boldsymbol{\mu}_{\theta_{\myx}}(\myh[t]),\text{diag}\{\boldsymbol{\sigma}_{\theta_{\myx}}^2(\myh[t]\})\big). \label{eq:deep-STORN-c}
\end{align}
In the following, we retain this latter more general formulation for easier comparison with the other models. We denote by $\theta_{\myh}$ and $\theta_{\myh\myx}$ the set of parameters of the networks implementing $d_{\myh}$ and $d_{\myx}$, respectively, and have $\theta_{\myx} = \theta_{\myh} \cup \theta_{\myh\myx}$.

\begin{figure}
\centering
\begin{tikzpicture}
        \node[latent,minimum size=9mm] (zm) {$\myz[t-1]$};
        \node[latent,minimum size=9mm,right=of zm] (z) {$\myz[t]$};
        \node[latent,minimum size=9mm,right=of z] (zp) {$\myz[t+1]$};
        \node[det,minimum size=11mm,below=of zm, yshift=3.5mm] (hm) {$\myh[t-1]$};
        \node[det,minimum size=11mm,right=of hm,xshift=-2.3mm] (h) {$\myh[t]$};
        \node[det,minimum size=11mm,right=of h,xshift=-2.5mm] (hp) {$\myh[t+1]$};
        \node[obs,minimum size=9mm,below=of hm, yshift=3.5mm] (xm) {$\myx[t-1]$};
        \node[obs,minimum size=9mm,right=of xm] (x) {$\myx[t]$};
        \node[obs,minimum size=9mm,right=of x] (xp) {$\myx[t+1]$};
        \edge {zm} {hm};
        \edge {hm} {xm};
        \edge{hm,z,xm} {h};
        \edge{h} {x};
        \edge{zp} {hp};
        \edge{h,zp,x} {hp};
        \edge{hp} {xp};
    \end{tikzpicture}\hspace{15mm}
    \begin{tikzpicture}[->]
        \node[latent,minimum size=9mm] (zm) {$\myz[t-1]$};
        \node[latent,minimum size=9mm,right=of zm] (z) {$\myz[t]$};
        \node[latent,minimum size=9mm,right=of z] (zp) {$\myz[t+1]$};
        \node[obs,minimum size=9mm,below=of zm] (xm) {$\myx[t-1]$};
        \node[obs,minimum size=9mm,right=of xm] (x) {$\myx[t]$};
        \node[obs,minimum size=9mm,right=of x] (xp) {$\myx[t+1]$};
        \edge {zm} {xm,x,xp};
        \edge{z} {x,xp};
        \edge{x,zp} {xp};
        \edge{xm} {x};
        \path (xm) edge[bend right] (xp);
    \end{tikzpicture}
\caption{STORN's graphical model in developed form (left) and compact form (right).}
\label{fig:STORN}
\end{figure}
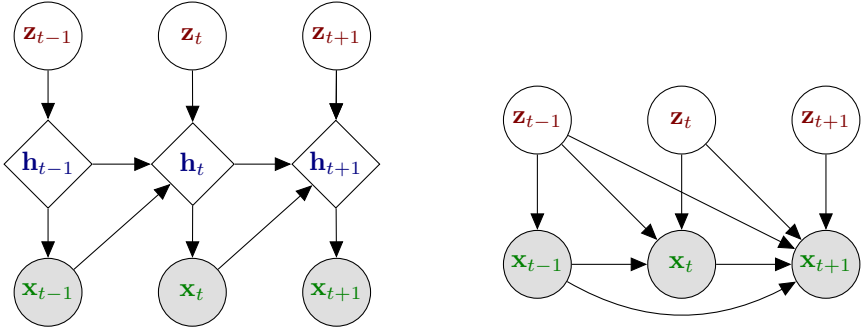

In STORN, $\myz[t]$ is assumed i.i.d. with a standard Gaussian distribution:
\begin{align}
p_{\theta_{\myz}}(\myz[1:T]) = \prod_{t=1}^T p_{\theta_{\myz}}(\myz[t]) \qquad \text{with} \qquad p_{\theta_{\myz}}(\myz[t]) = \mathcal{N}(\myz[t];\mathbf{0},\mathbf{I}_L). \label{eq:STORN-prior-z}
\end{align}
In short, there is no temporal model on the prior distribution of $\myz[t]$ and $\theta_{\myz} = \emptyset$ (and therefore $\theta = \theta_{\myx}$). Here, it is the temporal recursion on $\myh[t]$ and the use of $\myh[t]$ to generate $\myx[t]$ that makes STORN a member of the DVAE family. The graphical model of STORN is shown in Figure~\ref{fig:STORN} (left). 

\vspace{0.3cm}
\noindent \textbf{Notation remark}: To ensure homogeneous notations across models, we slightly change the notation used by \citet{bayer2014learning} by ``synchronizing'' $\myx[t]$ and $\myh[t]$; that is, in our presentation of STORN, $\myx[t]$ is generated from $\myh[t]$ (which is generated from $\myx[t-1]$, $\myh[t-1]$, and $\myz[t]$). In contrast, in Eq.~(4) in \citeauthor{bayer2014learning}'s \citeyearpar{bayer2014learning} paper, $\myx[t+1]$ is generated from $\myh[t]$ (which is generated from $\myx[t]$, $\myh[t-1]$, and $\myz[t]$). In other words, we replace $\myx[t]$ with $\myx[t-1]$. This change of notation does not change the model in essence but makes the comparison with other models easier.
\vspace{0.3cm}

Eq. \eqref{eq:deep-STORN-a}  shows that the two states, $\myh[t]$ and $\myz[t]$, are intricate, and the interpretation of $\myh[t]$ as a deterministic state is now an issue. In fact, $\myh[t]$ is now a random variable, but it is not a ``free'' one, as it is a deterministic function of the latent random variables $\myz[t]$ and $\myx[t-1]$ and of its previous value $\myh[t-1]$. We present a proper treatment of this issue in Appendix~\ref{Appendix-A}. The recurrence on $\myh[t]$ is unfolded to consider $\myh[t]$ as a deterministic function of $\myz[1:t]$ and $\myx[1:t-1]$, which we denote $\myh[t] = \myh[t](\myx[1:t-1], \myz[1:t])$,\footnote{This function also depends on the initial vectors $\myx[0]$ and $\myh[0]$ (and we can set $\myx[0] = \emptyset$), but we omit them for clarity of presentation.} and we consider a Dirac probability distribution over $\myh[t]$, positioned at that function's output value. These points were briefly discussed by \citet{bayer2014learning}. In Appendix~\ref{Appendix-A}, it is shown that marginalizing the joint density $p_{\theta}(\myx[1:T], \myz[1:T], \myh[1:T])$ w.r.t.~$\myh[1:T]$ leads to
\begin{align}
p_{\theta}(\myx[1:T], \myz[1:T]) &=  \prod_{t=1}^{T} p_{\theta_{\myx}}\big(\myx[t] | \myh[t](\myx[1:t-1],\myz[1:t])\big)p(\myz[t]). \label{eq:STORN-seq-a}
\end{align}
From the above equation and \eqref{eq:STORN-prior-z}, we deduce the conditional distribution
\begin{align}
p_{\theta_{\myx}}(\myx[1:T] | \myz[1:T]) &= \prod_{t=1}^{T} p_{\theta_{\myx}}\big(\myx[t] | \myh[t](\myx[1:t-1],\myz[1:t])\big). \label{eq:STORN-seq-b}
\end{align}
Note that these data sequence densities factorize across time frames, but the whole history of the present and past latent variables and past outputs is necessary to generate the present output. This history is summarized in $\myh[t](\myx[1:t-1],\myz[1:t])$. 

In line with the discussion in Section~\ref{subsec:parametrization}, an alternate description of STORN can be written, where we remove the internal deterministic state $\myh$ and express the model only in terms of the free random variables $\myx[1:T]$ and $\myz[1:T]$:
\begin{align}
p_{\theta}(\myx[1:T], \myz[1:T]) &=  \prod_{t=1}^{T} p_{\theta_{\myx}}(\myx[t] | \myx[1:t-1],\myz[1:t])p(\myz[t]), \label{eq:STORN-gen-a}
\end{align}
and
\begin{align}
p_{\theta_{\myx}}(\myx[1:T] | \myz[1:T]) &= \prod_{t=1}^{T} p_{\theta_{\myx}}(\myx[t] | \myx[1:t-1],\myz[1:t]). \label{eq:STORN-gen-b}
\end{align}
The above two equations are more general than \eqref{eq:STORN-seq-a} and \eqref{eq:STORN-seq-b}, but they lose some information on the deterministic link between $\myx[1:t-1]$ and $\myz[1:t]$ in the process of generating $\myx[t]$. The compact graphical representation corresponding to this alternate formulation is given in Figure~\ref{fig:STORN} (right).

\section{Inference model}

Following Section~\ref{sec:DVAE-inference}, it is easy to show that the exact posterior distribution of STORN takes the following form:
\begin{align}
p_{\theta}(\myz[1:T] | \myx[1:T]) =  \prod_{t=1}^T p_{\theta}(\myz[t] | \myz[1:t-1], \myx[1:T]). \label{eq:STORN-inference-seq-true}
\end{align}
In fact, this expression is obtained by the chain rule, and it cannot be simplified by applying D-separation. This is because any vector in $\myz[1:t-1]$ and $\myx[1:T]$ is either a child, a parent, or a co-parent of $\myz[t]$ in the graphical representation of STORN. In other words, each product term at time $t$ in \eqref{eq:STORN-inference-seq-true} depends on the past observed and latent state vectors that propagate through the internal state, and it depends on the present and future observed vectors, as $\myz[t]$ propagates to them through the internal hidden states. 

To complement the discussion on the form of the exact posterior distribution, we note the following:
\begin{align}
p_{\theta}(\myz[1:T] | \myx[1:T]) &\propto  p_{\theta}(\myx[1:T], \myz[1:T]) = p_{\theta_{\myx}}(\myx[1:T] | \myz[1:T])p(\myz[1:T]).
\end{align}
Thus, combining the above equation with \eqref{eq:STORN-seq-a}, we get
\begin{align}
p_{\theta}(\myz[1:T] | \myx[1:T]) &\propto \prod_{t=1}^{T} p_{\theta_{\myx}}\big(\myx[t] | \myh[t](\myx[1:t-1], \myz[1:t])\big)p(\myz[t]). \label{eq:STORN-pos-alternate}
\end{align}
As $\myz[t]$ is present in all terms of $\myh[t](\myx[1:t-1], \myz[1:t])$ from $t$ to $T$, it is confirmed that \eqref{eq:STORN-pos-alternate}, considered as a function of $\myz[t]$, depends not only on $\myx[1:t-1]$ but also on $\myx[t:T]$.

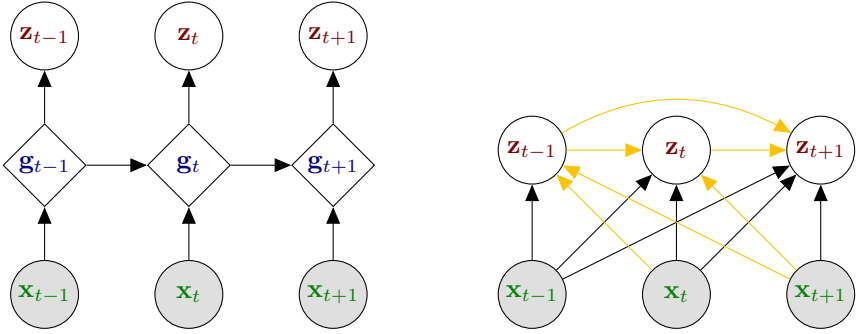
\begin{figure}
\centering
\begin{tikzpicture}
        \node[latent,minimum size=9mm] (zm) {$\myz[t-1]$};
        \node[latent,minimum size=9mm,right=of zm] (z) {$\myz[t]$};
        \node[latent,minimum size=9mm,right=of z] (zp) {$\myz[t+1]$};
        \node[det,minimum size=11mm,below=of zm, yshift=3mm] (hm) {$\myg[t-1]$};
        \node[det,minimum size=11mm,right=of hm, xshift=-2mm] (h) {$\myg[t]$};
        \node[det,minimum size=11mm,right=of h,xshift=-2mm] (hp) {$\myg[t+1]$};
        \node[obs,minimum size=9mm,below=of hm, yshift=3mm] (xm) {$\myx[t-1]$};
        \node[obs,minimum size=9mm,right=of xm] (x) {$\myx[t]$};
        \node[obs,minimum size=9mm,right=of x] (xp) {$\myx[t+1]$};
        \edge{hm} {zm};
        \edge{h} {z};
        \edge{hp} {zp};
        \edge{xm} {hm};
        \edge{x,hm} {h};
        \edge{xp,h} {hp};
    \end{tikzpicture}\hspace{15mm}
    \begin{tikzpicture}[->]
        \node[latent,minimum size=9mm] (zm) {$\myz[t-1]$};
        \node[latent,minimum size=9mm,right=of zm] (z) {$\myz[t]$};
        \node[latent,minimum size=9mm,right=of z] (zp) {$\myz[t+1]$};
        \node[obs,minimum size=9mm,below=of zm] (xm) {$\myx[t-1]$};
        \node[obs,minimum size=9mm,right=of xm] (x) {$\myx[t]$};
        \node[obs,minimum size=9mm,right=of x] (xp) {$\myx[t+1]$};
        \edge{xm} {z,zp};
        \edge{x} {zp};
        \path (xm) edge (zm);
        \path (x) edge (z);
        \path (xp) edge (zp);
        \path (x) edge[color=goldenpoppy] (zm);
        \path (xp) edge[color=goldenpoppy] (zm);
        \path (xp) edge[color=goldenpoppy] (z);
        \path (zm) edge[color=goldenpoppy] (z);
        \path (z) edge[color=goldenpoppy] (zp);
        \path (zm) edge[color=goldenpoppy,bend left] (zp);
    \end{tikzpicture}
\caption{STORN's graphical model at inference time in developed form (left) and compact form (right). Golden arrows correspond to missing links on the proposed probabilistic dependencies (compared to the exact inference dependencies).}
\label{fig:STORN-inference}
\end{figure}

As for the practical inference in STORN, the approximate posterior distribution $q_{\phi}$ was chosen by \citet{bayer2014learning} as generated by an additional \emph{forward} RNN. Little information is available on the implementation. The parameters of $q_{\phi}$ are said to be generated from $\myx[1:t]$, which is slightly odd as $\myx[t+1]$ was generated from $\myz[t]$ (with their notations; see our remark in the previous subsection). There is thus a one-step lag between generation and inference, which is difficult to justify (in practice, we have found that this leads to significantly inferior inference performance). With our change in notation at generation, we somehow automatically compensate for this gap and assume that the ``correct'' detailed inference equations are given as 
\begin{align}
\myg[t] &= e_{\myg}(\mathbf{W}_\inp^\enc\myx[t] + \mathbf{W}_\rec^\enc\myg[t-1] + \mathbf{b}_\hid^\enc),   \label{eq:STORN-inference-1} \\
[\boldsymbol{\mu}_{\phi}(\myg[t]),\boldsymbol{\sigma}_{\phi}(\myg[t])] &= e_{\myz}(\mathbf{W}_\out^\enc\myg[t] + \mathbf{b}_\out^\enc),  \label{eq:STORN-inference-2}\\
q_{\phi}(\myz[t] | \myg[t]) &=  \mathcal{N}\big(\myz[t];\boldsymbol{\mu}_{\phi}(\myg[t]),\text{diag}\{\boldsymbol{\sigma}_{\phi}^2(\myg[t])\}\big), \label{eq:STORN-inference-3}
\end{align}
where $\myg[t]$ denotes the inference RNN internal state,\footnote{In \citeauthor{bayer2014learning}'s \citeyearpar{bayer2014learning} paper, it is denoted as $\mathbf{h}_t^r$.} and $e_{\myg}$ and $e_{\myz}$ are nonlinear activation functions. Similarly to the generative model, because of the recursivity in \eqref{eq:STORN-inference-1}, $\myg[t]$ can be considered as an unfolded deterministic function of $\myx[1:t]$, which we note $\myg[t] = \myg[t](\myx[1:t])$.\footnote{This function is also a function of $\myg[0]$.} For a complete data sequence, we have 
\begin{align}
q_{\phi}(\myz[1:T] | \myx[1:T]) &=  \prod_{t=1}^T q_{\phi}(\myz[t] | \myg[t]) = \prod_{t=1}^T q_{\phi}\big(\myz[t] | \myg[t](\myx[1:t])\big).
\label{inference_model_storn}
\end{align}
This inference model can be rewritten in compact form as
\begin{align}
q_{\phi}(\myz[1:T] | \myx[1:T]) = \prod_{t=1}^T q_{\phi}(\myz[t] | \myx[1:t]). \label{eq:STORN-inference-model-seq-general}
\end{align}
The corresponding graphical model is shown in Figure~\ref{fig:STORN-inference}. Note that this inference model is inconsistent with the exact posterior distribution $p_{\theta}(\myz[1:T] | \myx[1:T])$ at several points: At each time $t$, it neither considers future observations $\myx[t+1:T]$ nor past latent states $\myz[1:t-1]$. As discussed in Section~\ref{sec:sharing}, the internal states of the encoder and of decoder can be identical, or they can be different. In STORN, given the choice of the inference model, these internal states depend on different variables and, therefore, are different.

\section{Training}

A comparison of the compact form of STORN in \eqref{eq:STORN-gen-a} with the general compact form of a DVAE in \eqref{time_slide_ordered_model} (simplified without $\myu[1:T]$) shows that STORN makes the following conditional independence assumption:
\begin{align}
    p_{\theta_{\myz}}(\myz[t] | \myx[1:t-1],\myz[1:t-1]) &= p(\myz[t]).
\end{align}
Using this single simplification, the VLB given in its most general form in \eqref{VLB1_dev} becomes
\begin{align}
\mathcal{L}(\theta, \phi ; \myx[1:T]) = & \sum_{t=1}^{T} \mathbb{E}_{q_{\phi}(\myz[1:t] | \myx[1:T] )}\big[  \log p_{\theta_{\myx}}(\myx[t] | \myx[1:t-1], \myz[1:t]) \big] \nonumber \\
& - \sum_{t=1}^{T}  \mathbb{E}_{q_{\phi}(\myz[1:t-1] | \myx[1:T])}\left[ D_{\text{KL}}\left(q_{\phi}(\myz[t] | \myz[1:t-1], \myx[1:T]) \parallel p(\myz[t]) \right)\right].
\label{VLB_STORN_1}
\end{align}
This expression of the VLB relies on an inference model that is consistent with the exact posterior distribution \eqref{eq:STORN-inference-seq-true}. However, as discussed above, STORN assumes an inference model of the form: $q_{\phi}(\myz[t] | \myz[1:t-1], \myx[1:T]) = q_{\phi}(\myz[t] | \myx[1:t])$. Consequently, the VLB in \eqref{VLB_STORN_1} can be simplified as follows:
\begin{align}
\mathcal{L}(\theta, \phi ; \myx[1:T]) = & \sum_{t=1}^{T} \mathbb{E}_{q_{\phi}(\myz[1:t] | \myx[1:T] )}\big[  \log p_{\theta_{\myx}}(\myx[t] | \myx[1:t-1], \myz[1:t]) \big] \nonumber \\
& - \sum_{t=1}^{T}  D_{\text{KL}}\left(q_{\phi}(\myz[t] | \myx[1:t]) \parallel p(\myz[t]) \right).
\label{VLB_STORN_2}
\end{align}
The KL divergence in this expression can be computed analytically, while the expectation is intractable and should be approximated by a Monte Carlo estimate, using samples drawn recursively from $q_{\phi}(\myz[1:t] | \myx[1:T])$ based on the inference model \eqref{inference_model_storn}. As for DKF, using the reparameterization trick for this recursive sampling leads to an objective function, which is differentiable w.r.t.~$\theta$ and $\phi$.

\chapter{Variational Recurrent Neural Networks}
\label{sec:VRNN}
The VRNN model was proposed by \citet{chung2015recurrent} as a combination of a VAE and an RNN. 

\section{Generative model}

The VRNN observation model is given by  
\begin{align}
\myh[t] &= d_{\myh}(\varphi_{\myx}(\myx[t-1]), \varphi_{\myz}(\myz[t-1]), \myh[t-1]), \label{eq:VRNN-a} \\
[\boldsymbol{\mu}_{\theta_{\myx}}(\myz[t], \myh[t]),\boldsymbol{\sigma}_{\theta_{\myx}}(\myz[t], \myh[t])] &= d_{\myx}(\varphi_{\myz}(\myz[t]), \myh[t]), \label{eq:VRNN-b} \\
p_{\theta_{\myx}}(\myx[t] | \myz[t], \myh[t]) &= \mathcal{N}\big(\myx[t];\boldsymbol{\mu}_{\theta_{\myx}}(\myz[t],\myh[t]),\text{diag}\{\boldsymbol{\sigma}_{\theta_{\myx}}^2(\myz[t],\myh[t])\}\big), \label{eq:VRNN-c}
\end{align}
where $\varphi_{\myz}$ and $\varphi_{\myx}$ are feature extractors, which were mentioned by \citet{chung2015recurrent} to be important in practice. These feature extractors are DNNs parameterized by a set of weights and biases, denoted as $\tau$.

The generative distribution of $\myz[t]$ is given by
\begin{align}
[\boldsymbol{\mu}_{\theta_{\myz}}( \myh[t]), \boldsymbol{\sigma}_{\theta_{\myz}}( \myh[t])] &= d_{\myz}( \myh[t]), \label{eq:VRNN-d} \\
p_{\theta_{\myz}}( \myz[t] | \myh[t]) &= \mathcal{N}\big( \myz[t]; \boldsymbol{\mu}_{\theta_{\myz}}(\myh[t]), \text{diag}\{\boldsymbol{\sigma}_{\theta_{\myz}}^2(\myh[t])\}\big). \label{eq:VRNN-e}
\end{align}

\vspace{0.3cm}
\noindent \textbf{Notation remark}: In \citeauthor{chung2015recurrent}'s \citeyearpar{chung2015recurrent} paper, $\tau$ denotes the set of parameters for both feature extractors, which are denoted as $\varphi_{\tau}^{\mathbf{x}}$ and $\varphi_{\tau}^{\mathbf{z}}$, respectively, as well as for $d_{\myx}$ and $d_{\myz}$, which are denoted $\varphi_{\tau}^{\text{dec}}$ and $\varphi_{\tau}^{\text{prior}}$, respectively. Moreover, $d_{\myh}$ is denoted $d_{\theta}$. We find this a bit confusing and prefer to distinguish among $\tau$, $\theta_{\myx} = \theta_{\myh} \cup \theta_{\myh\myx}$, $\theta_{\myz} = \theta_{\myh} \cup \theta_{\myh\myz}$, and $\theta = \tau \cup \theta_{\myx} \cup \theta_{\myz}$. Moreover, one may also want to distinguish between $\tau_x$ and $\tau_z$ to clarify that the two feature extractors are different. We retain $\tau$ for simplicity. Besides, we replace $\myh[t-1]$ in the paper by \citet{chung2015recurrent} with $\myh[t]$ to synchronize $\myh[t]$, $ \myz[t]$, and $\myx[t]$; that is, $\myx[t]$ is generated from $\myh[t]$ and $\myz[t]$. This arbitrary reindexing of $\myh[t]$ does not change the model conceptually.
\vspace{0.3cm}

In VRNN, we thus have multiple intrications of $\myh[t]$ and  $\myz[t]$ in both the observation model and the distribution of $\myz[t]$. The graphical model of VRNN is given in Figure~\ref{fig:VRNN} (left). The generative process starts with an initial internal state $\myh[1]$, from which we generate $\myz[1]$. From $\myh[1]$ and $\myz[1]$, we generate $\myx[1]$. Then, $\myh[2]$ is deterministically calculated from $\myh[1]$, $\myz[1]$, and $\myx[1]$, and so on, except that hereinafter, $\myz[t]$ is generated from  $\myz[t-1]$, $\myh[t-1]$, \emph{and} $\myx[t-1]$. Using the ``unfolding the recurrence'' trick mentioned in the previous sections, we can here denote $\myh[t] = \myh[t](\myx[1:t-1], \myz[1:t-1])$\footnote{This function also depends on $\myh[1]$, which is omitted for clarity of presentation.} and have $p_{\theta_{\myz}}(\myz[t] | \myh[t]) = p_{\theta_{\myz}}(\myz[t] | \myh[t](\myx[1:t-1], \myz[1:t-1]))$ and $p_{\theta_{\myx}}(\myx[t] | \myz[t], \myh[t]) = p_{\theta_{\myx}}(\myx[t] | \myz[t], \myh[t](\myx[1:t-1], \myz[1:t-1]))$. This provides both $\myz[t]$ and $\myx[t]$ with an implicit temporal model.

\begin{figure}
\centering
\begin{tikzpicture}[->]
    \node[latent,minimum size=9mm] (zm) {$\myz[t-1]$};
    \node[latent,minimum size=9mm,right=of zm] (z) {$\myz[t]$};
    \node[latent,minimum size=9mm,right=of z] (zp) {$\myz[t+1]$};
    \node[det,minimum size=11mm,below=of zm, yshift=3mm] (hm) {$\myh[t-1]$};
    \node[det,minimum size=11mm,right=of hm,xshift=-2.3mm] (h) {$\myh[t]$};
    \node[det,minimum size=11mm,right=of h,xshift=-2.5mm] (hp) {$\myh[t+1]$};
    \node[obs,minimum size=9mm,below=of hm, yshift=3mm] (xm) {$\myx[t-1]$};
    \node[obs,minimum size=9mm,right=of xm] (x) {$\myx[t]$};
    \node[obs,minimum size=9mm,right=of x] (xp) {$\myx[t+1]$};
    \edge {hm} {zm};
    \edge {hm} {xm};
    \edge{hm,zm,xm} {h};
    \edge{h} {x,z};
    \edge{h,x,z} {hp};
    \edge{hp} {zp,xp};
    \path (zm) edge[bend left] (xm);
    \path (z) edge[bend left] (x);
    \path (zp) edge[bend left] (xp);
\end{tikzpicture}\hspace{15mm}
\begin{tikzpicture}[->]
    \node[latent,minimum size=9mm] (zm) {$\myz[t-1]$};
    \node[latent,minimum size=9mm,right=of zm] (z) {$\myz[t]$};
    \node[latent,minimum size=9mm,right=of z] (zp) {$\myz[t+1]$};
    \node[obs,minimum size=9mm,below=of zm] (xm) {$\myx[t-1]$};
    \node[obs,minimum size=9mm,right=of xm] (x) {$\myx[t]$};
    \node[obs,minimum size=9mm,right=of x] (xp) {$\myx[t+1]$};
    \edge {zm} {xm,x,xp,z};
    \edge{z} {x,xp,zp};
    \edge{x,zp} {xp};
    \edge{x} {zp};
    \edge{xm} {x,z,zp};
    \path (xm) edge[bend right] (xp);
    \path (zm) edge[bend left] (zp);
\end{tikzpicture}
\caption{VRNN's graphical model in developed (left) and compact (right) forms.}
\label{fig:VRNN}
\end{figure}

As for a data sequence, when marginalizing the joint distribution of all variables w.r.t. $\myh[1:T]$ following the line of Appendix~\ref{Appendix-A}, we get
\begin{align}
p_{\theta}(\myx[1:T], \myz[1:T]) &= \prod_{t=1}^{T} p_{\theta_{\myx}}\big(\myx[t] | \myz[t], \myh[t](\myx[1:t-1],\myz[1:t-1])\big)p_{\theta_{\myz}}\big(\myz[t] | \myh[t](\myx[1:t-1],\myz[1:t-1])\big) \label{eq:VRNN-joint-x-z} \\
&= \prod_{t=1}^{T} p_{\theta}\big(\myx[t], \myz[t]  |  \myh[t](\myx[1:t-1],\myz[1:t-1])\big).
\end{align}
Again, we have a factorization of the conditional densities over time frames. However, we do \emph{not} have conditional independence of  $\myx[t]$ and $\myz[t]$ conditionally to the state $\myh[t](\myx[1:t-1],\myz[1:t-1]$) due to the direct link from $\myz[t]$ to $\myx[t]$.

As for STORN, we can provide a more general alternate expression for $p_{\theta}(\myx[1:T], \myz[1:T])$ that does not make the internal state explicit but only represents the general dependencies among the ``free'' random variables:
\begin{align}
p_{\theta}(\myx[1:T], \myz[1:T]) &= \prod_{t=1}^{T} p_{\theta_{\myx}}(\myx[t] | \myx[1:t-1],\myz[1:t])p_{\theta_{\myz}}(\myz[t] | \myx[1:t-1],\myz[1:t-1]) \label{eq:VRNN-joint-alternate} \\
&= \prod_{t=1}^{T} p_{\theta}(\myx[t], \myz[t]  |  \myx[1:t-1],\myz[1:t-1]).
\end{align}
The corresponding compact graphical model is shown in Figure~\ref{fig:VRNN} (right).

\section{Inference model}
\label{sec:VRNN-inference}
The general form of the exact posterior distribution of VRNN is identical to the one of STORN; that is, it factorizes into
\begin{align}
p_{\theta}(\myz[1:T] | \myx[1:T]) =  \prod_{t=1}^T p_{\theta}(\myz[t] | \myz[1:t-1], \myx[1:T]),
\end{align}
and, here also, no further simplification can be obtained from D-separation. 

\begin{figure}
\centering
\begin{tikzpicture}[->]
    \node[latent,minimum size=9mm] (zm) {$\myz[t-1]$};
    \node[latent,minimum size=9mm,right=of zm] (z) {$\myz[t]$};
    \node[latent,minimum size=9mm,right=of z] (zp) {$\myz[t+1]$};
    \node[det,minimum size=11mm,below=of zm, yshift=3mm] (hm) {$\myh[t-1]$};
    \node[det,minimum size=11mm,right=of hm,xshift=-2.3mm] (h) {$\myh[t]$};
    \node[det,minimum size=11mm,right=of h,xshift=-2.5mm] (hp) {$\myh[t+1]$};
    \node[obs,minimum size=9mm,below=of hm, yshift=4mm] (xm) {$\myx[t-1]$};
    \node[obs,minimum size=9mm,right=of xm] (x) {$\myx[t]$};
    \node[obs,minimum size=9mm,right=of x] (xp) {$\myx[t+1]$};
    \edge {hm} {zm};
    \edge{hm,zm,xm} {h};
    \edge{h} {z};
    \edge{h,x,z} {hp};
    \edge{hp} {zp};
    \path (xm) edge[bend right] (zm);
    \path (x) edge[bend right] (z);
    \path (xp) edge[bend right] (zp);
\end{tikzpicture}\hspace{15mm}
\begin{tikzpicture}[->]
    \node[latent,minimum size=9mm] (zm) {$\myz[t-1]$};
    \node[latent,minimum size=9mm,right=of zm] (z) {$\myz[t]$};
    \node[latent,minimum size=9mm,right=of z] (zp) {$\myz[t+1]$};
    \node[obs,minimum size=9mm,below=of zm] (xm) {$\myx[t-1]$};
    \node[obs,minimum size=9mm,right=of xm] (x) {$\myx[t]$};
    \node[obs,minimum size=9mm,right=of x] (xp) {$\myx[t+1]$};
    \edge {zm} {z};
    \edge{z} {zp};
    \edge{x} {zp,z};
    \edge{xm} {z,zp,zm};
    \edge{xp} {zp};
    \path (zm) edge[bend left] (zp);
    \path (x) edge[color=goldenpoppy] (zm);
    \path (xp) edge[color=goldenpoppy] (zm);
    \path (xp) edge[color=goldenpoppy] (z);
\end{tikzpicture}
\caption{VRNN's graphical model at inference time in developed form (left) and compact form (right). Golden arrows correspond to missing links on the proposed probabilistic dependencies (compared to the exact inference dependencies).}
\label{fig:VRNN-inference_v2}
\end{figure}
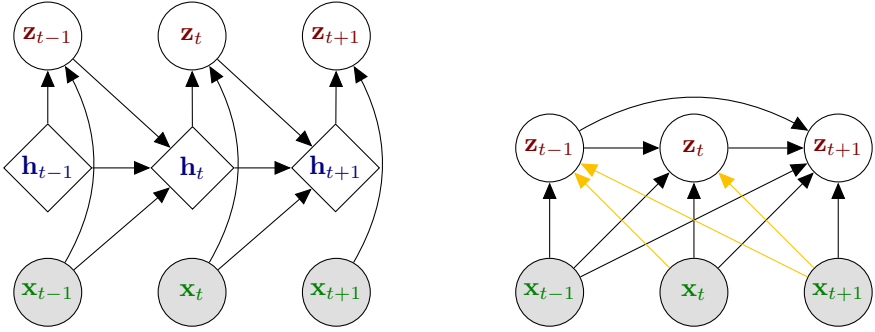

The approximate posterior distribution $q_{\phi}$ was chosen by \citet{chung2015recurrent} as
\begin{align}
[\boldsymbol{\mu}_{\phi}(\myx[t], \myh[t]), \boldsymbol{\sigma}_{\phi}(\myx[t], \myh[t])] &= e_{\myz}\big(\varphi_{\myx}(\myx[t]),\myh[t]\big), \label{eq:VRNN-inference-a}\\
q_{\phi}(\myz[t] | \myx[t], \myh[t]) &= \mathcal{N}\big(\myz[t]; \boldsymbol{\mu}_{\phi}(\myx[t], \myh[t]), \text{diag}\{\boldsymbol{\sigma}_{\phi}^2(\myx[t], \myh[t])\}\big), \label{eq:VRNN-inference-b}
\end{align}
where $e_{\myz}$ is the encoder DNN, parameterized by $\phi_{\myz}$. 
As for data sequence inference, we have 
\begin{align}
q_{\phi}(\myz[1:T] | \myx[1:T]) = \prod_{t=1}^T q_{\phi}\big(\myz[t] | \myx[t], \myh[t](\myx[1:t-1], \myz[1:t-1])\big). \label{eq:VRNN-inference-model-seq}
\end{align}
In contrast to STORN, the same internal state $\myh[t]$ is here shared by the VRNN encoder and decoder, which, in our opinion, makes the approximate model more consistent with the exact posterior distribution. This makes the set of parameters $\theta_{\myh}$ common to the encoder and decoder. Because the feature extractor $\varphi_{\myx}(\myx[t])$ is also used at the encoder, the same remark applies to its parameter set $\tau_{\myx}$. In addition, the inference at time $t$ depends on past outputs \emph{and} past latent states, which also makes the inference model closer to the exact posterior distribution. However, compared to the exact posterior, the future observations (from $t+1$ to $T$) are missing again. In short, here also, the approximate inference is causal, whereas the exact posterior distribution is noncausal. 
The graphical model corresponding to the VRNN approximate inference process is shown in Figure~\ref{fig:VRNN-inference_v2}. The inference model can be rewritten in the following general form:
\begin{align}
q_{\phi}(\myz[1:T] | \myx[1:T]) = \prod_{t=1}^T q_{\phi}(\myz[t] | \myz[1:t-1], \myx[1:t]). \label{eq:VRNN-inference-model-seq-general}
\end{align}

\section{Training}

A comparison of the compact form of VRNN in \eqref{eq:VRNN-joint-alternate} with the general compact form of a DVAE in \eqref{time_slide_ordered_model} (simplified without $\myu[1:T]$) shows that VRNN does not make any conditional independence assumption in the generative model. In this sense, VRNN is the most general DVAE model we have seen so far. The expression of the VLB for VRNN should therefore be the one given in \eqref{VLB1_dev}. However, as discussed above, the inference model in VRNN is inconsistent with the exact posterior distribution, as the following conditional independence assumption is made: $q_{\phi}(\myz[t] | \myz[1:t-1], \myx[1:T]) = q_{\phi}(\myz[t] | \myz[1:t-1], \myx[1:t])$. Consequently, the VLB in \eqref{VLB1_dev} can be simplified as follows:
\begin{align}
& \mathcal{L}(\theta, \phi ; \myx[1:T]) = \sum_{t=1}^{T} \mathbb{E}_{q_{\phi}(\myz[1:t] | \myx[1:T])}\big[  \log p_{\theta_{\myx}}(\myx[t] | \myx[1:t-1], \myz[1:t]) \big] \nonumber \\
& \hspace{0.5cm} - \sum_{t=1}^{T}  \mathbb{E}_{q_{\phi}(\myz[1:t-1] | \myx[1:T])}\left[ D_{\text{KL}}\left(q_{\phi}(\myz[t] | \myz[1:t-1], \myx[1:t]) \parallel p_{\theta_{\myz}}(\myz[t] | \myx[1:t-1],\myz[1:t-1]) \right)\right].
\label{VLB_VRNN}
\end{align}
As for the previously presented DVAE models, the KL divergence can be computed analytically, and intractable expectations are approximated by Monte Carlo estimates.

\section{Improved VRNN and VRNN applications}

To complement this VRNN section, we report that an improved version of VRNN was presented by \citet{goyal2017z}. The authors pointed out the difficulty in learning meaningful latent variables when coupled with a strong autoregressive decoder. We further discuss this point and we provide a series of references in Chapter~\ref{sec:discussion}. \citet{goyal2017z} proposed to improve the inference and training of VRNN with the following three features. 

First, possibly inspired by \citet{krishnan2017structured}, as well as \citep{fraccaro2016sequential} (see Section~\ref{sec:SRNN-inference}), they introduced a backward RNN on $\myx[t]$ to feed the approximate posterior distribution $q_{\phi}(\myz[1:T] | \myx[1:T])$, in line with  DKS (see Section~\ref{sec:DKF-inference}). Therefore, they accounted for the future observations in the inference process, as opposed to the original VRNN, moving toward a better compliance with the structure of the exact posterior distribution $p_{\theta}(\myz[1:T] | \myx[1:T])$. 

Second, they forced the latent variable $\myz[t]$ to contain relevant information about the future of the sequence by connecting $\myz[t]$ with the internal state of the inference backward network (denoted $\mathbf{b}_t$ by \citet{goyal2017z}). This was achieved by introducing  an additional conditional model $p_{\xi}(\mathbf{b}_t | \myz[t])$ and adding it in the VLB. Similarly, they also considered an additional conditional model $p_{\xi}(\myx[t] | \mathbf{b}_t)$. 

Finally, they slightly modified the VRNN model itself by removing the direct link between $\myz[t]$ and $\myx[t]$; that is, they replaced \eqref{eq:VRNN-b}--\eqref{eq:VRNN-c} with \eqref{eq:deep-STORN-b}--\eqref{eq:deep-STORN-c}, while all other equations remained identical to the VRNN equations. The authors reported that ``[they] observed better performance by avoiding the latent variables from directly producing the next output.''

An adaptation of VRNN to automatic language translation was proposed by \citet{su2018variational}. This is doubly interesting because this paper considers a sequence of discrete inputs and outputs, which contrasts with the ``all continuous'' models we focus on. This paper also contrasts with the previously proposed VAE-based models for text/language processing, which, as mentioned in the Introduction, usually consider a single latent vector to encode the whole input sequence (a full sentence). In \citeauthor{su2018variational}'s \citeyearpar{su2018variational} paper, it is the sequence of latent vectors $\myz[1:T]$ that encodes the semantic content of the sequence to translate ``over time.'' 

Finally, we can also mention the study by \citet{lee2018acoustic}, which uses VRNN for speech synthesis and adopts adversarial training. Besides VRNN, all these papers illustrate the flexibility of the DVAE class of models.

\chapter{Stochastic Recurrent Neural Networks}
\label{sec:SRNN}
The SRNN model was proposed by \citet{fraccaro2016sequential}, with an objective to ``glue (or stack) a deterministic recurrent neural network and a state space model together to form a stochastic and sequential neural generative model.''  

\vspace{0.3cm}
\noindent \textbf{Notation remark}: In \citeauthor{fraccaro2016sequential}'s \citeyearpar{fraccaro2016sequential} paper, $\myh[t]$ is denoted as $\mathbf{d}_{t}$, and the model is presented in the driven mode. We replace the external input $\myu[t]$ with $\myx[t-1]$ (i.e., predictive mode) for a better comparison with VRNN and STORN.

\section{Generative model}

The SRNN observation model is given by  
\begin{align}
{\color{blue}\myh[t]} &= d_{\myh}(\myx[t-1], \myh[t-1]), \label{eq:SRNN-a} \\
[\boldsymbol{\mu}_{\theta_{\myx}}(\myz[t], \myh[t]),\boldsymbol{\sigma}_{\theta_{\myx}}(\myz[t], \myh[t])] &= d_{\myx}(\myz[t], \myh[t]),   \label{eq:SRNN-b}\\
p_{\theta_{\myx}}(\myx[t] | \myz[t], \myh[t]) &= \mathcal{N}\big(\myx[t];\boldsymbol{\mu}_{\theta_{\myx}}(\myz[t], \myh[t]),\text{diag}\{\boldsymbol{\sigma}_{\theta_{\myx}}^2(\myz[t],\myh[t])\}\big).  \label{eq:SRNN-c}  
\end{align}
Eq. \eqref{eq:SRNN-a} is identical to \eqref{eq:RNN-a} (with $\myx[t-1]$ instead of $\myu[t]$) and thus refers to the usual deterministic RNN. Eqs. \eqref{eq:SRNN-b} and \eqref{eq:SRNN-c} are quite similar to \eqref{eq:VRNN-b} and \eqref{eq:VRNN-c}, respectively. Thus, the internal state $\myh[t]$ remains deterministic here, and the latent state $\myz[t]$ is integrated at the $d_{\myx}$ level. This justifies the ``clear(er) separation of deterministic and stochastic layers'' claimed by \citet{fraccaro2016sequential}, compared to VRNN.

\citet{fraccaro2016sequential} also introduced an explicit temporal model on the distribution of $\myz[t]$ (as opposed to implicit temporal dependency through $\myh[t]$ in VRNN), in addition to the dependency on the internal state $\myh[t]$:
\begin{align}
&[\boldsymbol{\mu}_{\theta_{\myz}}(\myz[t-1],\myh[t]), \boldsymbol{\sigma}_{\theta_{\myz}}(\myz[t-1],\myh[t])] = d_{\myz}(\myz[t-1],\myh[t]), \label{eq:SRNN-d}\\
&p_{\theta_{\myz}}(\myz[t] | \myz[t-1],\myh[t]) = \mathcal{N}\big(\myz[t]; \boldsymbol{\mu}_{\theta_{\myz}}(\myz[t-1],\myh[t]), \text{diag}\{\boldsymbol{\sigma}_{\theta_{\myz}}^2(\myz[t-1],\myh[t])\}\big). \label{eq:SRNN-e}
\end{align}
Compared to VRNN, the arrow from $\myz[t-1]$ to $\myh[t]$ is replaced with an arrow from $\myz[t-1]$ to $\myz[t]$, leading to an explicit  first-order Markovian dependency for $\myz[t]$ (which is combined with the dependency on $\myh[t]$). In addition, compared to VRNN, no feature extractor is mentioned in SRNN, so we have here $\theta = \theta_{\myx} \cup \theta_{\myz}$ (and again $\theta_{\myx} = \theta_{\myh} \cup \theta_{\myh\myx}$ and $\theta_{\myz} = \theta_{\myh} \cup \theta_{\myh\myz}$).
The graphical model of SRNN is shown in Figure~\ref{fig:SRNN} (left). In \citeauthor{fraccaro2016sequential}'s \citeyearpar{fraccaro2016sequential} paper, both $d_{\myx}$ and $d_{\myz}$ are two-layer feed-forward networks. The function $d_{\myh}$ is a GRU RNN, so that, according to the authors, ``the SSM can therefore utilize long-term information captured by the RNN.'' 

\begin{figure}
\centering
\begin{tikzpicture}[->]
    \node[latent,minimum size=9mm] (zm) {$\myz[t-1]$};
    \node[latent,minimum size=9mm,right=of zm] (z) {$\myz[t]$};
    \node[latent,minimum size=9mm,right=of z] (zp) {$\myz[t+1]$};
    \node[det,minimum size=11mm,below=of zm, yshift=3mm] (hm) {$\myh[t-1]$};
    \node[det,minimum size=11mm,right=of hm,xshift=-2mm] (h) {$\myh[t]$};
    \node[det,minimum size=11mm,right=of h,xshift=-2.5mm] (hp) {$\myh[t+1]$};
    \node[obs,minimum size=9mm,below=of hm, yshift=3mm] (xm) {$\myx[t-1]$};
    \node[obs,minimum size=9mm,right=of xm] (x) {$\myx[t]$};
    \node[obs,minimum size=9mm,right=of x] (xp) {$\myx[t+1]$};
    \edge {hm} {zm};
    \edge {hm} {xm};
    \edge{hm,xm} {h};
    \edge{zm} {z};
    \edge{z} {zp};
    \edge{h} {x,z};
    \edge{h,x} {hp};
    \edge{hp} {zp,xp};
    \path (zm) edge[bend left] (xm);
    \path (z) edge[bend left] (x);
    \path (zp) edge[bend left] (xp);
\end{tikzpicture}\hspace{15mm}
\begin{tikzpicture}[->]
    \node[latent,minimum size=9mm] (zm) {$\myz[t-1]$};
    \node[latent,minimum size=9mm,right=of zm] (z) {$\myz[t]$};
    \node[latent,minimum size=9mm,right=of z] (zp) {$\myz[t+1]$};
    \node[obs,minimum size=9mm,below=of zm] (xm) {$\myx[t-1]$};
    \node[obs,minimum size=9mm,right=of xm] (x) {$\myx[t]$};
    \node[obs,minimum size=9mm,right=of x] (xp) {$\myx[t+1]$};
    \edge {zm} {xm,z};
    \edge{z} {x,zp};
    \edge{x,zp} {xp};
    \edge{x} {zp};
    \edge{xm} {x,z,zp};
    \path (xm) edge[bend right] (xp);
\end{tikzpicture}
\caption{SRNN's graphical model in developed (left) and compact (right) forms.}
\label{fig:SRNN}
\end{figure}

Using the same ``unfolding the recurrence'' trick as in the previous sections, we here denote $\myh[t]$ as $\myh[t](\myx[1:t-1])$\footnote{Again, we omit the initial term $\myh[1]$ for clarity of presentation.} and have $p_{\theta_{\myz}}(\myz[t] | \myz[t-1], \myh[t]) = p_{\theta_{\myz}}(\myz[t] | \myz[t-1], \myh[t](\myx[1:t-1]))$ and $p_{\theta_{\myx}}(\myx[t] | \myz[t], \myh[t]) = p_{\theta_{\myx}}(\myx[t] | \myz[t], \myh[t](\myx[1:t-1]))$. Again, if we follow the line of Appendix~\ref{Appendix-A}, marginalizing the joint distribution of all variables w.r.t.~$\myh[1:T]$ leads to
\begin{align}
p_{\theta}(\myx[1:T], \myz[1:T]) &= \prod_{t=1}^{T} p_{\theta_{\myx}}\big(\myx[t] | \myz[t], \myh[t](\myx[1:t-1])\big)p_{\theta_{\myz}}\big(\myz[t] | \myz[t-1], \myh[t](\myx[1:t-1])\big).  \label{eq:SRNN-joint}
\end{align}
As for VRNN, we have a factorization over time frames, but no independence of $\myx[t]$ and $\myz[t]$ conditionally to the state $\myh[t](\myx[1:t-1])$.
As for STORN and VRNN,  \eqref{eq:SRNN-joint} can be reshaped into the following more general expression:
\begin{align}
p_{\theta}(\myx[1:T], \myz[1:T]) &= \prod_{t=1}^{T} p_{\theta_{\myx}}(\myx[t] | \myx[1:t-1],\myz[t])p_{\theta_{\myz}}(\myz[t] | \myz[t-1],\myx[1:t-1]). \label{eq:SRNN-joint-alternate} 
\end{align}
The corresponding compact graphical model is shown in Figure~\ref{fig:SRNN} (right).

\section{Inference model}
\label{sec:SRNN-inference}

For SRNN, because of the dependencies in the generative model, the general form of the exact posterior distribution is given by
\begin{align}
p_{\theta}(\myz[1:T] | \myx[1:T]) =  \prod_{t=1}^T p_{\theta}(\myz[t] | \myz[t-1], \myx[1:T]).
\end{align}
At each time $t$, the posterior distribution of $\myz[t]$ depends on the previous latent state and the whole observation sequence.

\citet{fraccaro2016sequential} indicated this structure and proposed an approximate posterior distribution $q_{\phi}$ with the same structure. This is the second time this proper methodology is considered in the present review after the DKS in Section~\ref{sec:DKF-inference}, but in the publication chronology, to the best of our knowledge, this was the first time. The dependency of $q_{\phi}$ on future observations, as well as on past observations through the future internal states, is implemented with a gated backward RNN. This network is denoted by $e_{\mygbw}$ in the equations below, it is parameterized by $\phi_{\mygbw}$, and $\mygbw[t]$ denotes its internal state, where the right-to-left arrow highlights the backward nature of the process. This backward RNN is followed by a basic feed-forward network $e_{\myz}$, which is parameterized by $\phi_{\myz}$. Formally, $q_{\phi}$ can be written as
\begin{align}
q_{\phi}(\myz[1:T] | \myx[1:T]) = \prod_{t=1}^T q_{\phi}(\myz[t] | \myz[t-1],\mygbw[t]), \label{eq:SRNN-inference-a}
\end{align}
with
\begin{align}
\mygbw[t] &= e_{\mygbw}([\myh[t], \myx[t]],\mygbw[t+1]), \label{eq:SRNN-inference-b} \\
[\boldsymbol{\mu}_{\phi}(\myz[t-1],\mygbw[t]), \boldsymbol{\sigma}_{\phi}(\myz[t-1],\mygbw[t])] &= e_{\myz}(\myz[t-1],\mygbw[t]), \label{eq:SRNN-inference-c} \\
q_{\phi}(\myz[t] |\myz[t-1],\mygbw[t]) &= \mathcal{N}\big(\myz[t]; \boldsymbol{\mu}_{\phi}(\myz[t-1],\mygbw[t]), \text{diag}(\boldsymbol{\sigma}_{\phi}^2(\myz[t-1],\mygbw[t]))\big). \label{eq:SRNN-inference-d}
\end{align}
We thus have here $\phi = \phi_{\mygbw} \cup \phi_{\myz}$.

\vspace{0.3cm}
\noindent \textbf{Notation remark}: In \citeauthor{fraccaro2016sequential}'s \citeyearpar{fraccaro2016sequential} paper, $\myh[t]$ is denoted by $\mathbf{d}_{t}$, $\mygbw[t]$ is denoted by $\mathbf{a}_t$, and  $q_{\phi}(\myz[1:T] | \myx[1:T])$ is denoted by $q_{\phi}(\myz[1:T] | \mathbf{d}_{1:T}, \myx[1:T])$. Because we have $\myh[t] = \myh[t](\myx[1:t-1])$, we can stick to $q_{\phi}(\myz[1:T] | \myx[1:T])$.  

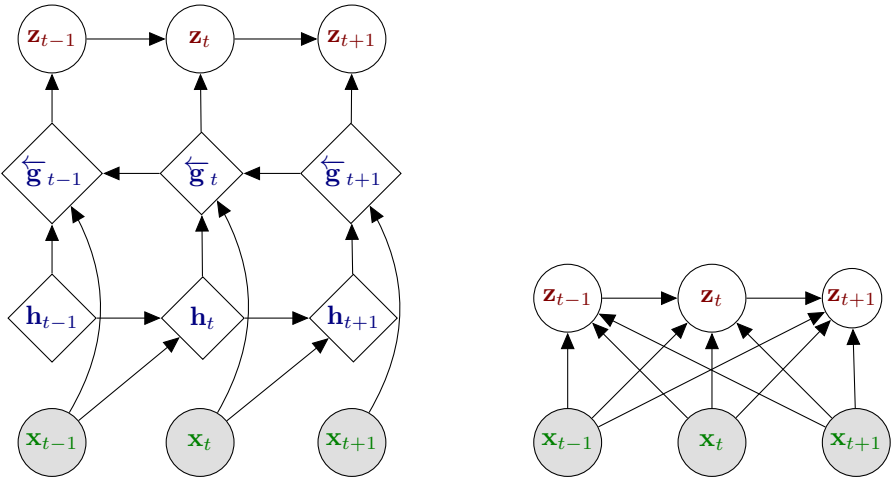
\begin{figure}
\centering
\begin{tikzpicture}[->]
    \node[latent,minimum size=9mm] (zm) {$\myz[t-1]$};
    \node[latent,minimum size=9mm,right=of zm, xshift=0.5mm] (z) {$\myz[t]$};
    \node[latent,minimum size=9mm,right=of z, xshift=1mm] (zp) {$\myz[t+1]$};
    \node[det,minimum size=11mm,below=of zm, yshift=3.5mm] (am) {$\mygbw[t-1]$};
    \node[det,minimum size=11mm,right=of am, xshift=-2.5mm] (a) {$\mygbw[t]$};
    \node[det,minimum size=11mm,right=of a,xshift=-2.5mm] (ap) {$\mygbw[t+1]$};
    \node[det,minimum size=11mm,below=of am, yshift=3.5mm] (hm) {$\myh[t-1]$};
    \node[det,minimum size=11mm,right=of hm, xshift=-1.5mm] (h) {$\myh[t]$};
    \node[det,minimum size=11mm,right=of h,xshift=-1.5mm] (hp) {$\myh[t+1]$};
    \node[obs,minimum size=9mm,below=of hm, yshift=4mm] (xm) {$\myx[t-1]$};
    \node[obs,minimum size=9mm,right=of xm, xshift=0.5mm] (x) {$\myx[t]$};
    \node[obs,minimum size=9mm,right=of x, xshift=1mm] (xp) {$\myx[t+1]$};
    \edge{zm,a} {z};
    \edge{z,ap} {zp};
    \edge{hm,xm} {h};
    \edge{h,x} {hp};
    \edge{ap,h} {a};
    \edge{a,hm} {am};
    \edge{hp} {ap};
    \edge{am} {zm};
    \path (xm) edge[bend right] (am);
    \path (x) edge[bend right] (a);
    \path (xp) edge[bend right] (ap);
\end{tikzpicture}\hspace{15mm}
\begin{tikzpicture}[->]
    \node[latent,minimum size=9mm] (zm) {$\myz[t-1]$};
    \node[latent,minimum size=9mm,right=of zm] (z) {$\myz[t]$};
    \node[latent,right=of z] (zp) {$\myz[t+1]$};
    \node[obs,minimum size=9mm,below=of zm] (xm) {$\myx[t-1]$};
    \node[obs,minimum size=9mm,right=of xm] (x) {$\myx[t]$};
    \node[obs,minimum size=9mm,right=of x] (xp) {$\myx[t+1]$};
    \edge{zm} {z};
    \edge{z} {zp};
    \edge{xm,x,xp} {zm,z,zp};
\end{tikzpicture}
\caption{SRNN's graphical model at inference time in developed form (left) and compact form (right). In this case, there are no missing links on the proposed probabilistic dependencies (compared to the exact inference dependencies). In \citeauthor{fraccaro2016sequential}'s \citeyearpar{fraccaro2016sequential} paper, the dependencies of $\myh[t]$ were omitted in the inference graphical model for clarity. We make them explicit here to recall that $\myh[t]$ follows the deterministic update \eqref{eq:SRNN-a}.} 
\label{fig:SRNN-inference_v2}
\end{figure}

\vspace{0.3cm}
The above equations show that inference requires a forward pass on the internal state $\myh[t]$ (which is shared by the encoder and decoder), its combination with $\myx[t]$, and a backward pass on the inference RNN, which makes $\mygbw[t]$ a deterministic function of the whole data sequence $\myx[1:T]$. Similarly to $\myh[t]$, we can denote this function by $\mygbw[t](\myx[1:T])$ to make this latter point explicit; however, this would be poorly informative about the way $\mygbw[t]$ depends on $\myx[1:T]$. The graphical model corresponding to the inference process in SRNN is shown in Figure~\ref{fig:SRNN-inference_v2}. The inference model can be rewritten in the following general form:
\begin{align}
q_{\phi}(\myz[1:T] | \myx[1:T]) = \prod_{t=1}^T q_{\phi}(\myz[t] | \myz[t-1], \myx[1:T]). \label{eq:SRNN-inference-model-seq-general}
\end{align}
\citet{fraccaro2016sequential} stated that this \emph{smoothing} process (combination of forward and backward RNNs on $\myx[t]$) can be replaced with a \emph{filtering} process, by replacing \eqref{eq:SRNN-inference-b}--\eqref{eq:SRNN-inference-c} with an ``instantaneous'' DNN $e_{\myz}(\myz[t-1],\myh[t], \myx[t])$.

\section{Training}

A comparison of the compact form of SRNN in \eqref{eq:SRNN-joint-alternate} with the general compact form of a DVAE in \eqref{time_slide_ordered_model} (simplified without $\myu[1:T]$) shows that the SRNN model makes the following conditional independence assumptions:
\begin{align}
    p_{\theta_{\myx}}(\myx[t] | \myx[1:t-1],\myz[1:t] ) &= p_{\theta_{\myx}}(\myx[t] | \myx[1:t-1],\myz[t]);  \nonumber \\
    p_{\theta_{\myz}}(\myz[t] | \myx[1:t-1],\myz[1:t-1]) &= p_{\theta_{\myz}}(\myz[t] | \myx[1:t-1],\myz[t-1]).
\end{align}
Using these two simplifications, along with the inference model \eqref{eq:SRNN-inference-model-seq-general} (which we recall is consistent with the exact posterior distribution), the VLB in its most general form \eqref{VLB1_dev} can be simplified as follows:
\begin{align}
& \mathcal{L}(\theta, \phi ; \myx[1:T]) = \sum_{t=1}^{T} \mathbb{E}_{q_{\phi}(\myz[t] | \myx[1:T])}\big[  \log p_{\theta_{\myx}}(\myx[t] | \myx[1:t-1],\myz[t]) \big] \nonumber \\
& \hspace{0.5cm} - \sum_{t=1}^{T}  \mathbb{E}_{q_{\phi}(\myz[t-1] | \myx[1:T])}\left[ D_{\text{KL}}\left(q_{\phi}(\myz[t] | \myz[t-1], \myx[1:T]) \parallel p_{\theta_{\myz}}(\myz[t] | \myx[1:t-1],\myz[t-1]) \right)\right].
\label{VLB_SRNN}
\end{align}
Again, the KL divergence can be computed analytically, and intractable expectations are approximated by Monte Carlo estimates. The procedure to sample from $q_{\phi}(\myz[t] | \myx[1:T])$ and $q_{\phi}(\myz[t-1] | \myx[1:T])$ relies on the ``cascade trick,'' as for DKF (see Section~\ref{sec:DKF-training}).

\chapter{Recurrent Variational Autoencoders}
\label{sec:RVAE}
The RVAE model was introduced by \citet{leglaive2020recurrent} to represent clean speech signals in a speech enhancement application. It was combined with a Gaussian noise model with nonnegative Matrix factorization of the variance within a Bayesian framework. The RVAE parameters were estimated offline on a large dataset of clean speech signals using the VAE methodology (maximization of the VLB). A variational expectation-maximization (VEM) algorithm was used for estimating the remaining parameters from a noisy speech signal, and probabilistic Wiener filters were then derived for speech enhancement. Here, we present only the RVAE model.

\section{Generative model}
\label{sec:RVAE-generative-model}

The RVAE model was designed to model speech signals in the short-term Fourier transform (STFT) domain. This implies that the model applies to a sequence of \emph{complex-valued} vectors. Therefore, the observation model uses a multivariate zero-mean circular complex Gaussian distribution \citep{neeser1993}, denoted $\mathcal{N}_c$, instead of the usual multivariate real-valued Gaussian distribution.  
This observation model has the following generic form: 
\begin{align}
\boldsymbol{\sigma}_{\theta_{\myx}}(\myz[\mathcal{T}]) &= d_{\myx}(\myz[\mathcal{T}]), \label{eq:Simon-a} \\
p_{\theta_{\myx}}(\myx[t] | \myz[\mathcal{T}]) &=  \mathcal{N}_c\big(\myx[t]; \mathbf{0}, \text{diag}\{ \boldsymbol{\sigma}_{\theta_{\myx}}^2(\myz[\mathcal{T}]) \}\big), \label{eq:Simon-b}
\end{align}
where $\mathcal{T}$ denotes a set of time frames, and the following three cases are considered: i)~an instantaneous model:  $\mathcal{T}= \{t\}$, which only considers the current latent state vector to model the observation at time $t$; ii)~a causal model: $\mathcal{T}= \{1:t\}$, which considers the sequence of past and present latent state vectors; and iii)~a noncausal model: $\mathcal{T}= \{1:T\}$, which considers the complete sequence of latent state vectors. 

This model can be adapted to real-valued observations with a usual Gaussian distribution:\footnote{Or any other distribution for real-valued vectors, as already mentioned. In fact, under some conditions, the complex proper Gaussian distribution applied on STFT coefficients corresponds to a Gamma distribution on the squared magnitude of those coefficients \citep{girin2019notes}.} we just have to replace $\mathcal{N}_c$ with $\mathcal{N}$, replace $\mathbf{0}$ with a mean parameter $\boldsymbol{\mu}_{\theta}(\myz[\mathcal{T}])$ in $\eqref{eq:Simon-b}$, and add this mean parameter to the left-hand side of $\eqref{eq:Simon-a}$, as usual. This is what we have done hereinafter for easier comparison with the other models.

As in STORN, $\myz[t]$ is assumed i.i.d. with a standard Gaussian distribution:
\begin{align}
p(\myz[1:T]) = \prod_{t=1}^T p(\myz[t]) \qquad \text{with} \qquad p(\myz[t]) = \mathcal{N}(\myz[t];\mathbf{0},\mathbf{I}_L). \label{eq:Simon-c}
\end{align}
Therefore, there is no explicit temporal model on $\myz[t]$, and $\myx[t]$ possibly depends on the past and future values of the latent state through \eqref{eq:Simon-b}. We have here $\theta_{\myz} = \emptyset$ and $\theta = \theta_{\myx}$. As case i) is strictly equivalent to the original VAE of Section~\ref{sec:VAEs}, with no temporal model at all, we will now focus on cases ii) and iii).

\citet{leglaive2020recurrent} only mentioned that cases ii) and iii) are implemented using a forward RNN and a bidirectional RNN, respectively, which take as input the sequence $\myz[1:t]$ or $\myz[1:T]$, respectively. The authors did not provide detailed implementation equations (though they provided a link to some supplementary material, including informative schemas). Let us write them now for easier comparison with the other models (for the same reason, we consider real-valued observations). 

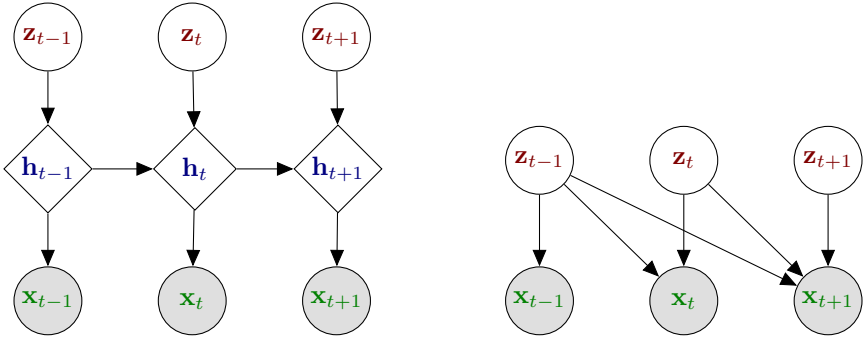
\begin{figure}
 \centering
     \begin{tikzpicture}[->]
        \node[latent,minimum size=9mm] (zm) {$\myz[t-1]$};
        \node[latent,minimum size=9mm,right=of zm] (z) {$\myz[t]$};
        \node[latent,minimum size=9mm,right=of z] (zp) {$\myz[t+1]$};
        \node[det,minimum size=11mm,below=of zm, yshift=3mm] (hm) {$\myh[t-1]$};
        \node[det,minimum size=11mm,right=of hm, xshift=-2mm] (h) {$\myh[t]$};
        \node[det,minimum size=11mm,right=of h, xshift=-2.5mm] (hp) {$\myh[t+1]$};
        \node[obs,minimum size=9mm,below=of hm, yshift=3mm] (xm) {$\myx[t-1]$};
        \node[obs,minimum size=9mm,right=of xm] (x) {$\myx[t]$};
        \node[obs,minimum size=9mm,right=of x] (xp) {$\myx[t+1]$};
        \edge {zm} {hm};
        \edge {hm} {xm,h};
        \edge{z} {h};
        \edge{h} {x};
        \edge{h,zp} {hp};
        \edge{hp} {xp};
    \end{tikzpicture}\hspace{15mm}
    \begin{tikzpicture}[->]
        \node[latent,minimum size=9mm] (zm) {$\myz[t-1]$};
        \node[latent,minimum size=9mm,right=of zm] (z) {$\myz[t]$};
        \node[latent,minimum size=9mm,right=of z] (zp) {$\myz[t+1]$};
        \node[obs,minimum size=9mm,below=of zm, yshift=0.5mm] (xm) {$\myx[t-1]$};
        \node[obs,minimum size=9mm,right=of xm] (x) {$\myx[t]$};
        \node[obs,minimum size=9mm,right=of x] (xp) {$\myx[t+1]$};
        \edge {zm} {xm,x,xp};
        \edge{z} {x,xp};
        \edge{zp} {xp};
    \end{tikzpicture}
    \caption{Causal RVAE's graphical model in developed form (left) and compact form (right). \label{fig:RVAE-RNN}}
\end{figure}

\vspace{0.2cm}
\noindent \textbf{Causal case:} Let us start with the causal case, for which we have 
\begin{align}
\myh[t] &= d_{\myh}(\myz[t],\myh[t-1]), \label{eq:Simon-detail-1} \\
[\boldsymbol{\mu}_\theta(\myh[t]), \boldsymbol{\sigma}_\theta(\myh[t])] &= d_{\myx}(\myh[t]), \label{eq:Simon-detail-2} \\
p_\theta(\myx[t] | \myh[t]) &= \mathcal{N}\big(\myx[t] ; \boldsymbol{\mu}_\theta(\myh[t]), \text{diag}\{\boldsymbol{\sigma}_\theta^2(\myh[t])\} \big).  \label{eq:Simon-detail-3}
\end{align}
Eq. \eqref{eq:Simon-detail-1} is similar to the RNN internal state update \eqref{eq:RNN-a} or \eqref{eq:SRNN-a}, with the major difference being that the latent state $\myz[t]$ is used as an input instead of an external input $\myu[t]$ or previous observation vector $\myx[t-1]$. Alternatively, \eqref{eq:Simon-detail-1} can be viewed as a simplified version of the STORN or VRNN internal state updates \eqref{eq:STORN-1} or \eqref{eq:VRNN-a}, where only $\myz[t]$ and $\myh[t-1]$ (and not $\myx[t-1]$) are used as inputs. Considering both the observation model and the prior latent-state model, the causal RVAE model is quite close to STORN. The two differences with STORN are that here $\myx[t-1]$ is not reinjected as input to the internal state $\myh[t]$ and an LSTM network is used instead of a single-layer RNN in the original STORN formulation.

The graphical model of RVAE (causal case) is shown in Figure~\ref{fig:RVAE-RNN}. As is now usual in our developments, we rewrite $\myh[t] = \myh[t](\myz[1:t])$\footnote{This is also a function of $\myh[0]$, which we omit for clarity.} and have $p_\theta(\myx[t] | \myh[t]) = p_\theta(\myx[t] | \myh[t](\myz[1:t]))$.
For a complete data sequence, we have
\begin{align}
p_\theta(\myx[1:T], \myz[1:T]) &= \prod_{t=1}^{T} p_\theta\big(\myx[t] | \myh[t](\myz[1:t])\big) p(\myz[t]),
\end{align}
which, as for the other models, can be reshaped into the following more general expression:
\begin{align}
p_{\theta}(\myx[1:T], \myz[1:T]) &= \prod_{t=1}^{T} p_{\theta_{\myx}}(\myx[t] | \myz[1:t])p(\myz[t]). \label{eq:RVAE-joint-alternate}
\end{align}

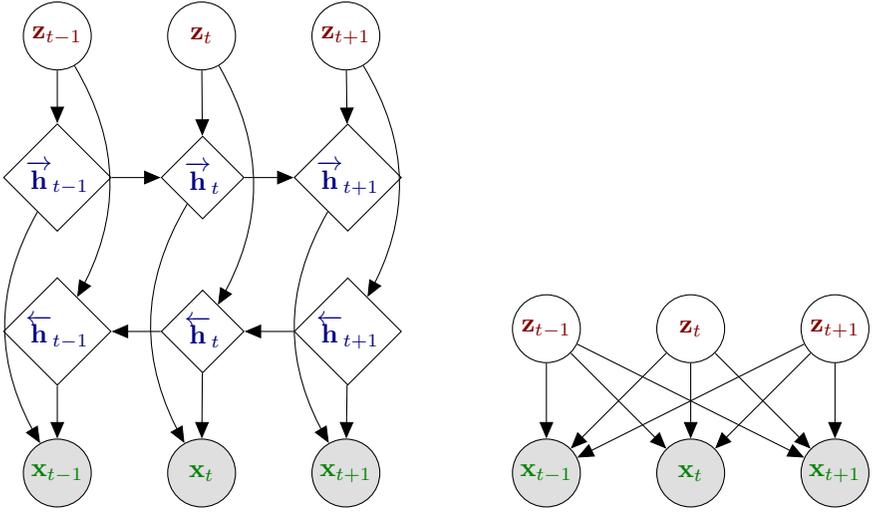
\begin{figure}
 \centering
    \begin{tikzpicture}[->]
        \node[latent,minimum size=9mm] (zm) {$\myz[t-1]$};
        \node[latent,minimum size=9mm,right=of zm] (z) {$\myz[t]$};
        \node[latent,minimum size=9mm,right=of z] (zp) {$\myz[t+1]$};
        \node[det,minimum size=11mm,below=of zm, yshift=3mm] (hmfw) {$\myhfw[t-1]$};
        \node[det,minimum size=11mm,right=of hmfw, xshift=-3.5mm] (hfw) {$\myhfw[t]$};
        \node[det,minimum size=11mm,right=of hfw, xshift=-3.5mm] (hpfw) {$\myhfw[t+1]$};
        \node[det,minimum size=11mm,below=of hmfw, yshift=4mm] (hmbw) {$\myhbw[t-1]$};
        \node[det,minimum size=11mm,right=of hmbw, xshift=-3.5mm] (hbw) {$\myhbw[t]$};
        \node[det,minimum size=11mm,right=of hbw, xshift=-3.5mm] (hpbw) {$\myhbw[t+1]$};
        \node[obs,minimum size=9mm,below=of hmbw, yshift=3mm] (xm) {$\myx[t-1]$};
        \node[obs,minimum size=9mm,right=of xm] (x) {$\myx[t]$};
        \node[obs,minimum size=9mm,right=of x] (xp) {$\myx[t+1]$};
        \edge {zm} {hmfw};
        \edge {hmfw} {hfw};
        \edge{z} {hfw};
        \path (hfw) edge[bend right] (x);
        \path (hmfw) edge[bend right] (xm);
        \edge{hfw,zp} {hpfw};
        \path (hpfw) edge[bend right] (xp);
        \edge{hpbw} {hbw};
        \edge{hbw} {hmbw};
        \path (zm) edge[bend left] (hmbw);
        \path (z) edge[bend left] (hbw);
        \path (zp) edge[bend left] (hpbw);
        \edge{hmbw} {xm};
        \edge{hbw} {x};
        \edge{hpbw} {xp};
    \end{tikzpicture}\hspace{12mm}
    \begin{tikzpicture}[->]
        \node[latent,minimum size=9mm] (zm) {$\myz[t-1]$};
        \node[latent,minimum size=9mm,right=of zm] (z) {$\myz[t]$};
        \node[latent,minimum size=9mm,right=of z] (zp) {$\myz[t+1]$};
        \node[obs,minimum size=9mm,below=of zm] (xm) {$\myx[t-1]$};
        \node[obs,minimum size=9mm,right=of xm] (x) {$\myx[t]$};
        \node[obs,minimum size=9mm,right=of x] (xp) {$\myx[t+1]$};
        \edge {zm} {xm,x,xp};
        \edge{z} {xm,x,xp};
        \edge{zp} {xm,x,xp};
    \end{tikzpicture}
    \caption{Noncausal RVAE's graphical model in developed form (left) and compact form (right). \label{fig:RVAE-BRNN}}
\end{figure}

\vspace{0.2cm}
\noindent \textbf{Noncausal case:} All DVAE models we have seen so far are causal (at generation). The noncausal case presented by \citet{leglaive2020recurrent} is the first noncausal DVAE model found in the literature. It is implemented with a combination of a forward RNN and a backward RNN on $\myz[t]$: 
\begin{align}
\myhfw[t] &= d_{\myhfw}(\myz[t], \myhfw[t-1]), \label{eq:Simon-detail-4} \\
\myhbw[t] &= d_{\myhbw}(\myz[t], \myhbw[t+1]), \label{eq:Simon-detail-4bis} \\
\myh[t] &= [\myhfw[t], \myhbw[t]] \label{eq:Simon-detail-5}, \\
[\boldsymbol{\mu}_{\theta_{\myx}}(\myh[t]), \boldsymbol{\sigma}_{\theta_{\myx}}(\myh[t])] &= d_{\myx}(\myh[t]), \label{eq:Simon-detail-6} \\
p_{\theta_{\myx}}(\myx[t] | \myh[t]) &= \mathcal{N}\big(\myx[t] ; \boldsymbol{\mu}_{\theta_{\myx}}(\myh[t]), \text{diag}\{\boldsymbol{\sigma}_{\theta_{\myx}}^2(\myh[t])\}\big).  \label{eq:Simon-detail-7}
\end{align}
We thus have $\myh[t] = \myh[t](\myz[1:T])$.\footnote{This function is also a function of the initial internal states $\myhfw[0]$ and $\myhbw[T+1]$, which we omit for clarity.} The graphical representation of the noncausal RVAE model is shown in Figure~\ref{fig:RVAE-BRNN}. For a complete data sequence, we have
\begin{align}
p_\theta(\myx[1:T], \myz[1:T]) &= \prod_{t=1}^{T} p_\theta\big(\myx[t] | \myh[t](\myz[1:T])\big) p(\myz[t]),
\end{align}
which can be reshaped into
\begin{align}
p_{\theta}(\myx[1:T], \myz[1:T]) &= \prod_{t=1}^{T} p_{\theta_{\myx}}(\myx[t] | \myz[1:T])p(\myz[t]). \label{eq:NC-RVAE-joint-alternate}
\end{align}

\section{Inference model}

As for the inference model, \citet{leglaive2020recurrent} first remarked that, using the chain rule and D-separation, the posterior distribution of the latent vectors can be expressed as follows: 
\begin{equation}
p_\theta(\myz[1:T] | \myx[1:T]) = \prod_{t=1}^{T} p_\theta(\myz[t] | \myz[1:t-1], \myx[\mathcal{T}']),
\label{eq:Simon-inference-1}
\end{equation}
where in the causal case, $\mathcal{T}'= \{t:T\}$, and in the noncausal case, $\mathcal{T}'= \{1:T\}$. For the causal generative model, the latent vector at a given time step depends (a posteriori) on past latent vectors and on present and future observations, whereas for the noncausal generative model, it also depends on the past observations.
Therefore, the authors chose to define the variational distribution $q_\phi$ with the same form:
\begin{equation}
q_\phi(\myz[1:T] | \myx[1:T]) = \prod_{t=1}^{T} q_\phi(\myz[t] | \myz[1:t-1],\myx[\mathcal{T'}]).
\label{eq:Simon-inference-2}
\end{equation}
As for the generative model, we now detail the implementation of the inference model.

\vspace{0.2cm}
\noindent \textbf{Causal case:} The inference corresponding to the causal generative model is implemented by combining a forward RNN on the latent vectors and a backward RNN on the observations: 
\begin{align}
\mygfw[t] &= e_{\mygfw}(\myz[t-1], \mygfw[t-1]), \label{eq:causal-RVAE-inference-model-a}\\
\mygbw[t] &= e_{\mygbw}(\myx[t], \mygbw[t+1]), \label{eq:causal-RVAE-inference-model-b}\\
\myg[t] &= [\mygfw[t], \mygbw[t]], \label{eq:causal-RVAE-inference-model-c}\\
[\boldsymbol{\mu}_\phi(\myg[t]), \boldsymbol{\sigma}_\phi(\myg[t]) ] &= e_{\myz}(\myg[t]), \label{eq:causal-RVAE-inference-model-d}\\
q_\phi(\myz[t] | \myz[1:t-1], \myx[t:T]) &= \mathcal{N}\big(\myz[t]; \boldsymbol{\mu}_\phi(\myg[t]), \text{diag}\{ \boldsymbol{\sigma}_\phi^2(\myg[t])\}\big), \\
q_\phi(\myz[1:T] | \myx[1:T]) &= \prod_{t=1}^{T}q_\phi(\myz[t] | \myz[1:t-1], \myx[t:T]).
\label{eq:causal-RVAE-inference-model}
\end{align}
This inference model is shown in Figure~\ref{fig:RVAE-inference-causal}.

\begin{figure}
 \centering
    \begin{tikzpicture}[->]
    \node[latent,minimum size=9mm] (zm) {$\myz[t-1]$};
    \node[latent,minimum size=9mm,right=of zm] (z) {$\myz[t]$};
    \node[latent,minimum size=9mm,right=of z] (zp) {$\myz[t+1]$};
    \node[det,minimum size=11mm,below=of zm, yshift=3mm] (gfm) {$\mygfw[t-1]$};
    \node[det,minimum size=11mm,right=of gfm, xshift=-3mm] (gf) {$\mygfw[t]$};
    \node[det,minimum size=11mm,right=of gf, xshift=-3.5mm] (gfp) {$\mygfw[t+1]$};
    \node[det,minimum size=11mm,below=of gfm, yshift=5mm] (gbm) {$\mygbw[t-1]$};
    \node[det,minimum size=11mm,right=of gbm, xshift=-3mm] (gb) {$\mygbw[t]$};
    \node[det,minimum size=11mm,right=of gb,xshift=-3.5mm] (gbp) {$\mygbw[t+1]$};
    \node[obs,minimum size=9mm,below=of gbm, yshift=3mm] (xm) {$\myx[t-1]$};
    \node[obs,minimum size=9mm,right=of xm] (x) {$\myx[t]$};
    \node[obs,minimum size=9mm,right=of x] (xp) {$\myx[t+1]$};
    \edge{zm} {gf};
    \edge{z} {gfp};
    \edge{gfm} {zm};
    \edge{gf} {z};
    \edge{gfp} {zp};
    \edge{xm} {gbm};
    \edge{x} {gb};
    \edge{xp} {gbp};
    \edge{gfm} {gf};
    \edge{gf} {gfp};
    \edge{gbp} {gb};
    \edge{gb} {gbm};
    \path (gbm) edge[bend right] (zm);
    \path (gb) edge[bend right] (z);
    \path (gbp) edge[bend right] (zp);
\end{tikzpicture}\hspace{15mm}
\begin{tikzpicture}[->]
    \node[latent,minimum size=9mm] (zm) {$\myz[t-1]$};
    \node[latent,minimum size=9mm,right=of zm] (z) {$\myz[t]$};
    \node[latent,minimum size=9mm,right=of z] (zp) {$\myz[t+1]$};
    \node[obs,minimum size=9mm,below=of zm] (xm) {$\myx[t-1]$};
    \node[obs,minimum size=9mm,right=of xm] (x) {$\myx[t]$};
    \node[obs,minimum size=9mm,right=of x] (xp) {$\myx[t+1]$};
    \edge{zm} {z};
    \edge{z} {zp};
    \path(zm) edge[bend left=30] (zp);
    \edge{xm} {zm};
    \edge{x} {zm,z};
    \edge{xp} {zm,z,zp};
\end{tikzpicture}
    \caption{Graphical model of causal RVAE at inference time in developed form (left) and compact form (right). 
    \label{fig:RVAE-inference-causal}}
\end{figure}

\begin{figure}
\centering
    \begin{tikzpicture}[->]
    \node[latent,minimum size=9mm] (zm) {$\myz[t-1]$};
    \node[latent,minimum size=9mm,right=of zm, xshift=2.5mm] (z) {$\myz[t]$};
    \node[latent,minimum size=9mm,right=of z, xshift=2.5mm] (zp) {$\myz[t+1]$};
    \node[det,minimum size=11mm,below=of zm, yshift=3mm] (gfm) {$\mygfwz[t-1]$};
    \node[det,minimum size=11mm,right=of gfm, xshift=-0.5mm] (gf) {$\mygfwz[t]$};
    \node[det,minimum size=11mm,right=of gf, xshift=-1mm] (gfp) {$\mygfwz[t+1]$};
    \node[det,minimum size=11mm,below=of gfm, yshift=6mm] (gfxm) {$\mygfwx[t-1]$};
    \node[det,minimum size=11mm,right=of gfxm, xshift=-0.5mm] (gfx) {$\mygfwx[t]$};
    \node[det,minimum size=11mm,right=of gfx, xshift=-1mm] (gfxp) {$\mygfwx[t+1]$};
    \node[det,minimum size=11mm,below=of gfxm, yshift=6mm] (gbm) {$\mygbwx[t-1]$};
    \node[det,minimum size=11mm,right=of gbm, xshift=-0.5mm] (gb) {$\mygbwx[t]$};
    \node[det,minimum size=11mm,right=of gb, xshift=-1mm] (gbp) {$\mygbwx[t+1]$};
    \node[obs,minimum size=9mm,below=of gbm, yshift=3mm] (xm) {$\myx[t-1]$};
    \node[obs,minimum size=9mm,right=of xm, xshift=2.5mm] (x) {$\myx[t]$};
    \node[obs,minimum size=9mm,right=of x, xshift=2.5mm] (xp) {$\myx[t+1]$};
    \edge{zm} {gf};
    \edge{z} {gfp};
    \edge{gfm} {zm};
    \edge{gf} {z};
    \edge{gfp} {zp};
    \edge{xm} {gbm};
    \edge{x} {gb};
    \edge{xp} {gbp};
    \edge{gfm} {gf};
    \edge{gf} {gfp};
    \edge{gfxm} {gfx};
    \edge{gfx} {gfxp};
    \edge{gbp} {gb};
    \edge{gb} {gbm};
    \path (gbm) edge[bend right] (zm);
    \path (gb) edge[bend right] (z);
    \path (gbp) edge[bend right] (zp);
    \path (xm) edge[bend left] (gfxm);
    \path (x) edge[bend left] (gfx);
    \path (xp) edge[bend left] (gfxp);
    \path (gfxm) edge[bend left] (zm);
    \path (gfx) edge[bend left] (z);
    \path (gfxp) edge[bend left] (zp);
\end{tikzpicture}\hspace{5mm}
\begin{tikzpicture}[->]
    \node[latent,minimum size=9mm] (zm) {$\myz[t-1]$};
    \node[latent,minimum size=9mm,right=of zm] (z) {$\myz[t]$};
    \node[latent,minimum size=9mm,right=of z] (zp) {$\myz[t+1]$};
    \node[obs,minimum size=9mm,below=of zm] (xm) {$\myx[t-1]$};
    \node[obs,minimum size=9mm,right=of xm] (x) {$\myx[t]$};
    \node[obs,minimum size=9mm,right=of x] (xp) {$\myx[t+1]$};
    \edge{zm} {z};
    \edge{z} {zp};
    \path(zm) edge[bend left=30] (zp);
    \edge{xm,x,xp} {zm,z,zp};
\end{tikzpicture}
    \caption{Graphical model of noncausal RVAE at inference time in developed form (left) and compact form (right). 
    \label{fig:RVAE-inference-non-causal}}
\end{figure}
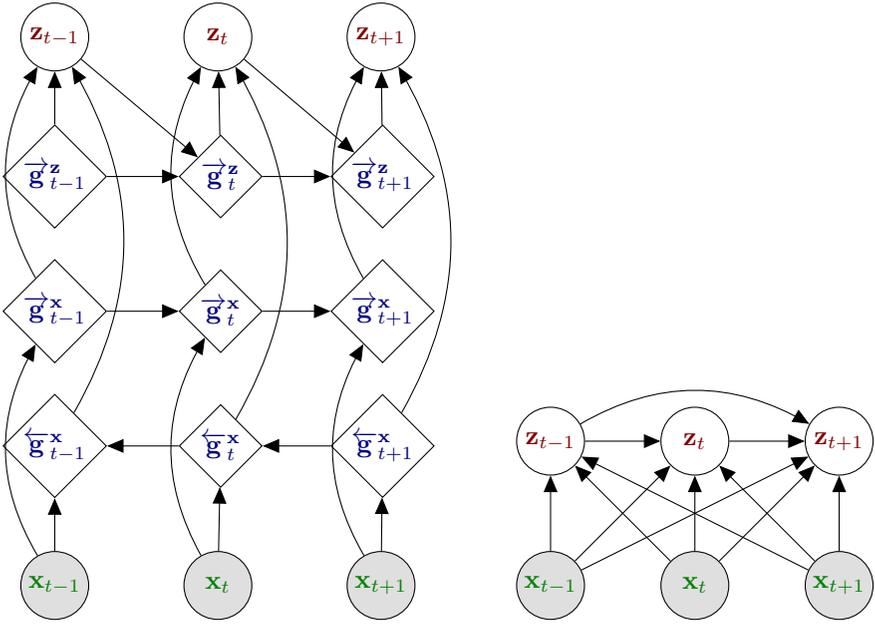

\vspace{0.2cm}
\noindent \textbf{Noncausal case:} The inference corresponding to the noncausal generative model is similar to the causal case, except that the RNN on the observations is bidirectional: 
\begin{align}
\mygfwz[t] &= e_{\mygfwz}(\myz[t-1], \mygfwz[t-1]), \label{eq:non-causal-RVAE-inference-model-a}\\
\mygfwx[t] &= e_{\mygfwx}(\myx[t], \mygfwx[t-1]), \label{eq:non-causal-RVAE-inference-model-b}\\
\mygbwx[t] &= e_{\mygbwx}(\myx[t], \mygbwx[t+1]), \label{eq:non-causal-RVAE-inference-model-c}\\
\myg[t] &= [\mygfwz[t], \mygfwx[t], \mygbwx[t]], \label{eq:non-causal-RVAE-inference-model-d}\\
[\boldsymbol{\mu}_\phi(\myg[t]), \boldsymbol{\sigma}_\phi(\myg[t]) ] &= e_{\myz}(\myg[t]), \label{eq:non-causal-RVAE-inference-model-e}\\
q_\phi(\myz[t] | \myz[1:t-1], \myx[1:T]) &= \mathcal{N}\big(\myz[t]; \boldsymbol{\mu}_\phi(\myg[t]), \text{diag}\{ \boldsymbol{\sigma}_\phi^2(\myg[t])\}\big), \\
q_\phi(\myz[1:T] | \myx[1:T]) &= \prod_{t=1}^{T}q_\phi(\myz[t] | \myz[1:t-1], \myx[1:T]).
\end{align}
This inference model is shown in Figure~\ref{fig:RVAE-inference-non-causal}.

\section{Training}

For conciseness, and because we focus on reviewing causal DVAEs, we only describe in this section the VLB for the causal RVAE model. The methodology to derive the VLB in the noncausal case is similar. 

A comparison of the compact form of causal RVAE in \eqref{eq:RVAE-joint-alternate} with the general compact form of a DVAE in \eqref{time_slide_ordered_model} (simplified without $\myu[1:T]$) shows that the causal RVAE model makes the following conditional independence assumptions:
\begin{align}
    p_{\theta_{\myx}}(\myx[t] | \myx[1:t-1],\myz[1:t] ) &= p_{\theta_{\myx}}(\myx[t] | \myz[1:t] );  \nonumber \\
    p_{\theta_{\myz}}(\myz[t] | \myx[1:t-1],\myz[1:t-1]) &= p(\myz[t]).
\end{align}
Using these two simplifications, along with the inference model \eqref{eq:causal-RVAE-inference-model} (which we recall is consistent with the exact posterior distribution), the VLB can be simplified as follows:
\begin{align}
\mathcal{L}(\theta, \phi ; \myx[1:T]) = & \sum_{t=1}^{T} \mathbb{E}_{q_{\phi}(\myz[1:t] | \myx[1:T])}\big[  \log p_{\theta_{\myx}}(\myx[t] | \myz[1:t]) \big] \nonumber \\
& \hspace{0.5cm} - \sum_{t=1}^{T}  \mathbb{E}_{q_{\phi}(\myz[1:t-1] | \myx[1:T])}\left[ D_{\text{KL}}\left(q_{\phi}(\myz[t] | \myz[1:t-1], \myx[t:T]) \parallel p(\myz[t]) \right)\right].
\label{VLB_RVAE}
\end{align}
As for the previous models, the KL divergence can be computed analytically, while the two intractable expectations are approximated by Monte Carlo estimates using samples drawn from $q_{\phi}(\myz[1:t] | \myx[1:T])$ and $q_{\phi}(\myz[1:t-1] | \myx[1:T])$ in a recursive manner.

\chapter{Disentangled Sequential Autoencoders}
\label{sec:DSAE}
\citet{yingzhen2018disentangled} proposed a hierarchical model called DSAE. This model introduces the idea of adding to the usual sequence of latent variables $\myz[1:T]$ a sequence-level latent vector $\myv$ (denoted $\mathbf{f}$ by \citet{yingzhen2018disentangled}), which is assumed to encode the sequence-level characteristics of the data. Therefore, $\myz[t]$ is assumed to encode time-dependent data features (e.g., the dynamics of an object in a video clip), and $\myv$ is assumed to encode ``everything else'' (e.g., object characteristics in a video clip). 

\section{Generative model}
\label{sec:DSAE-generative}

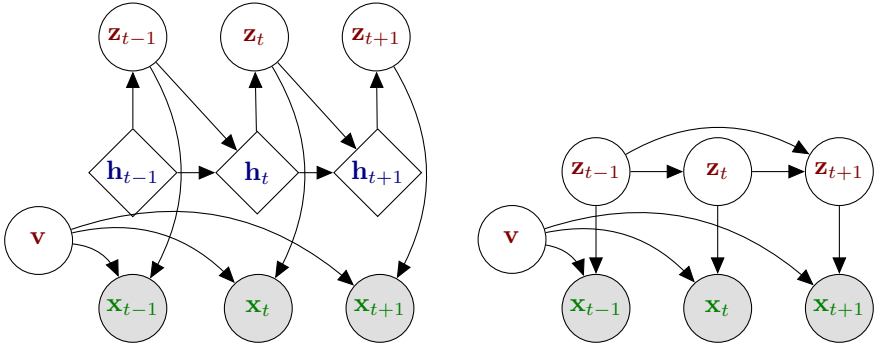
\begin{figure}
\centering
\begin{tikzpicture}[->]
    \node[latent,minimum size=9mm] (zm) {$\myz[t-1]$};
    \node[latent,minimum size=9mm,right=of zm, xshift=-3mm] (z) {$\myz[t]$};
    \node[latent,minimum size=9mm,right=of z, xshift=-3mm] (zp) {$\myz[t+1]$};
    \node[det,minimum size=11mm,below=of zm, yshift=2.5mm] (hm) {$\myh[t-1]$};
    \node[det,minimum size=11mm,right=of hm, xshift=-5mm] (h) {$\myh[t]$};
    \node[det,minimum size=11mm,right=of h, xshift=-5.5mm] (hp) {$\myh[t+1]$};
    \node[obs,minimum size=9mm,below=of hm, yshift=2.5mm] (xm) {$\myx[t-1]$};
    \node[obs,minimum size=9mm,right=of xm, xshift=-2.5mm] (x) {$\myx[t]$};
    \node[obs,minimum size=9mm,right=of x, xshift=-3mm] (xp) {$\myx[t+1]$};
    \node[latent,minimum size=9mm,left=of hm,yshift=-9mm, xshift=8mm] (v) {$\myv$};
    \edge {hm} {zm};
    \edge{hm,zm} {h};
    \edge{h} {z};
    \edge{h,z} {hp};
    \edge{hp} {zp};
    \path (v) edge[bend left] (xm);
    \path (v) edge[bend left] (x);
    \path (v) edge[bend left] (xp);
    \path (zm) edge[bend left] (xm);
    \path (z) edge[bend left] (x);
    \path (zp) edge[bend left] (xp);
\end{tikzpicture}\hspace{4mm}
\begin{tikzpicture}[->]
    \node[latent,minimum size=9mm] (zm) {$\myz[t-1]$};
    \node[latent,minimum size=9mm,right=of zm, xshift=-3mm] (z) {$\myz[t]$};
    \node[latent,minimum size=9mm,right=of z, xshift=-3mm] (zp) {$\myz[t+1]$};
    \node[obs,minimum size=9mm,below=of zm, yshift=1mm] (xm) {$\myx[t-1]$};
    \node[obs,minimum size=9mm,right=of xm, xshift=-3mm] (x) {$\myx[t]$};
    \node[obs,minimum size=9mm,right=of x, xshift=-3mm] (xp) {$\myx[t+1]$};
    \node[latent,minimum size=9mm,left=of zm,yshift=-9mm, xshift=8mm] (v) {$\myv$};
    \edge {zm} {xm,z};
    \edge{z} {x,zp};
    \edge{zp} {xp};
    \path (zm) edge[bend left] (zp);
    \path (v) edge[bend left] (xm);
    \path (v) edge[bend left] (x);
    \path (v) edge[bend left] (xp);
\end{tikzpicture}
\caption{DSAE's graphical model in developed form (left) and compact form (right).}
\label{fig:DSAE}
\end{figure}

\citet{yingzhen2018disentangled} only provided the following general form of the generative DSAE model for a complete data sequence:
\begin{align}
p_{\theta}(\myx[1:T], \myz[1:T], \myv) &= p_{\theta_{\myv}}(\myv) \prod_{t=1}^{T} p_{\theta_{\myx}}(\myx[t] | \myz[t], \myv)p_{\theta_{\myz}}(\myz[t] | \myz[1:t-1]). \label{eq:DSAE-joint-seq}
\end{align}
More detailed information about the pdfs and implementation issues are found in annexes from the ArXiv version of the paper. The authors used different variants for different datasets according to the nature of the data (e.g., video clips or speech signals). Here, we only report the model implemented for speech signals. The dynamical model $p_{\theta_{\myz}}(\myz[t] | \myz[1:t-1])$ is a Gaussian distribution whose parameters are provided by an LSTM network. The observation model $p_{\theta_{\myx}}(\myx[t] | \myz[t], \myv)$ is a Gaussian distribution whose parameters are provided by a feed-forward DNN. With the simplified generic RNN formalism used for LSTM (see Section~\ref{sec:RNNs-principle}), we can thus write 
\begin{align}
\myh[t] &= d_{\myh}(\myz[t-1], \myh[t-1]), \label{eq:DSAE-a} \\
[\boldsymbol{\mu}_{\theta_{\myz}}(\myh[t]),\boldsymbol{\sigma}_{\theta_{\myz}}(\myh[t])] &= d_{\myz}(\myh[t]), \label{eq:DSAE-b} \\
p_{\theta_{\myz}}( \myz[t] | \myh[t]) &= \mathcal{N}\big( \myz[t]; \boldsymbol{\mu}_{\theta_{\myz}}(\myh[t]), \text{diag}\{\boldsymbol{\sigma}_{\theta_{\myz}}^2(\myh[t])\}\big), \label{eq:DSAE-c} \\
[\boldsymbol{\mu}_{\theta_{\myx}}(\myz[t], \myv), \boldsymbol{\sigma}_{\theta_{\myx}}(\myz[t], \myv)] &= d_{\myx}(\myz[t], \myv), \label{eq:DSAE-d} \\
p_{\theta_{\myx}}(\myx[t] | \myz[t], \myv) &= \mathcal{N}\big(\myx[t];\boldsymbol{\mu}_{\theta_{\myx}}(\myz[t],\myv),\text{diag}\{\boldsymbol{\sigma}_{\theta_{\myx}}^2(\myz[t],\myv)\}\big). \label{eq:DSAE-e} 
\end{align}
The graphical representation of DSAE is shown in Figure~\ref{fig:DSAE}.
This model is similar to a DKF in the undriven mode conditioned on variable $\myv$, except that, in addition to this conditioning, the first-order Markov temporal model of DKF is replaced with a virtually infinite-order model owing to the LSTM. Although this is poorly discussed in the DVAE papers in general, it is an example of interesting model extensions that are easy to implement in the deep learning and VAE frameworks. As stated by \citet{krishnan2015deep}, ``using deep neural networks, we can enhance Kalman filters with arbitrarily complex transition dynamics and emission distributions. [...] we can tractably learn such models by optimizing a bound on the likelihood of the data.''

\section{Inference model}
\label{sec:DSAE-inference}

For DSAE, the posterior distribution of latent variables 
is given by
\begin{align}
p_\theta(\myz[1:T], \myv | \myx[1:T]) &= p_{\theta}(\myv | \myx[1:T]) p_{\theta}(\myz[1:T] | \myv, \myx[1:T]) \label{eq:DSAE-true-inference-a}
\\
&= p_{\theta}(\myv | \myx[1:T]) \prod_{t=1}^{T} p_{\theta}(\myz[t] | \myz[1:t-1], \myv, \myx[1:T]) \label{eq:DSAE-true-inference-b}
\\
&= p_{\theta}(\myv | \myx[1:T]) \prod_{t=1}^{T} p_{\theta}(\myz[t] | \myz[1:t-1], \myv, \myx[t:T]).
\label{eq:DSAE-true-inference-c}
\end{align}
The simplification in the last line results from D-separation. This decomposition can be interpreted as follows: The whole sequence of observations $\myx[1:T]$ is used to estimate the ``object'' representation $\myv$, and then $\myv$, the present and future observations $\myx[t:T]$, and previous latent state vectors $\myz[1:t-1]$ are used to update the object dynamics. 

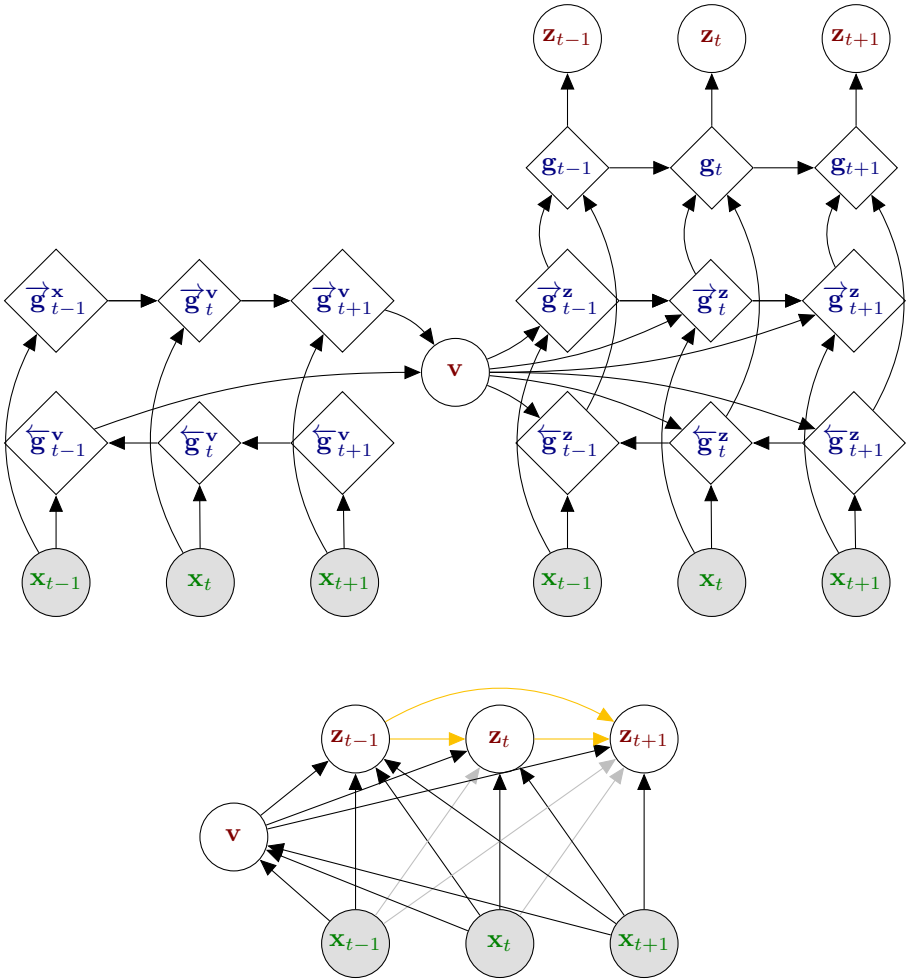
\begin{figure}
    \begin{tikzpicture}[->]
    \node[det,minimum size=11mm] (gfm) {$\mygfwx[t-1]$};
    \node[det,minimum size=11mm,right=of gfm, xshift=-3.5mm] (gf) {$\mygfwv[t]$};
    \node[det,minimum size=11mm,right=of gf, xshift=-3.5mm] (gfp) {$\mygfwv[t+1]$};
    \node[det,minimum size=11mm,below=of gfm, yshift=5mm] (gbm) {$\mygbwv[t-1]$};
    \node[det,minimum size=11mm,right=of gbm, xshift=-3.5mm] (gb) {$\mygbwv[t]$};
    \node[det,minimum size=11mm,right=of gb, xshift=-3.5mm] (gbp) {$\mygbwv[t+1]$};
    \node[obs,minimum size=9mm,below=of gbm, yshift=3mm] (xm) {$\myx[t-1]$};
    \node[obs,minimum size=9mm,right=of xm] (x) {$\myx[t]$};
    \node[obs,minimum size=9mm,right=of x] (xp) {$\myx[t+1]$};
    \node[latent,minimum size=9mm,right=of gfp, yshift=-9.5mm, xshift=-6.5mm] (v) {$\myv$};
    \edge{xm} {gbm};
    \edge{x} {gb};
    \edge{xp} {gbp};
    \edge{gfm} {gf};
    \edge{gf} {gfp};
    \edge{gfm} {gf};
    \edge{gf} {gfp};
    \edge{gbp} {gb};
    \edge{gb} {gbm};
    \path (gfp) edge[bend left=20] (v);
    \path (gbm) edge[bend left=10] (v);
    \path (xm) edge[bend left] (gfm);
    \path (x) edge[bend left] (gf);
    \path (xp) edge[bend left] (gfp);
    \node[det,right=of gfp, xshift=6mm] (gfzm) {$\mygfwz[t-1]$};
    \node[det,minimum size=11mm,right=of gfzm, xshift=-3.5mm] (gfz) {$\mygfwz[t]$};
    \node[det,minimum size=11mm,right=of gfz, xshift=-3.5mm] (gfzp) {$\mygfwz[t+1]$};
    \node[det,minimum size=11mm,below=of gfzm, yshift=5mm] (gbzm) {$\mygbwz[t-1]$};
    \node[det,minimum size=11mm,right=of gbzm, xshift=-3.5mm] (gbz) {$\mygbwz[t]$};
    \node[det,minimum size=11mm,right=of gbz, xshift=-3.5mm] (gbzp) {$\mygbwz[t+1]$};
    \node[obs,minimum size=9mm,below=of gbzm, yshift=3mm] (xzm) {$\myx[t-1]$};
    \node[obs,minimum size=9mm,right=of xzm] (xz) {$\myx[t]$};
    \node[obs,minimum size=9mm,right=of xz] (xzp) {$\myx[t+1]$};
    \edge{xzm} {gbzm};
    \edge{xz} {gbz};
    \edge{xzp} {gbzp};
    \edge{gfzm} {gfz};
    \edge{gfz} {gfzp};
    \edge{gfzm} {gfz};
    \edge{gfz} {gfzp};
    \edge{gbzp} {gbz};
    \edge{gbz} {gbzm};
    \path (v) edge[bend left=10] (gbzm);
    \path (v) edge[bend left=10] (gbz);
    \path (v) edge[bend left=10] (gbzp);
    \path (v) edge[bend right=10] (gfzm);
    \path (v) edge[bend right=10] (gfz);
    \path (v) edge[bend right=10] (gfzp); 
    \path (xzm) edge[bend left] (gfzm);
    \path (xz) edge[bend left] (gfz);
    \path (xzp) edge[bend left] (gfzp);
     \node[det,minimum size=11mm,above=of gbzm,yshift=14mm] (gm) {$\myg[t-1]$};
     \node[det,minimum size=11mm,right=of gm,xshift=-2mm] (g) {$\myg[t]$};
     \node[det,minimum size=11mm,right=of g,xshift=-2mm] (gp) {$\myg[t+1]$};
     \node[latent,minimum size=9mm,above=of gm,yshift=-3mm] (zm) {$\myz[t-1]$};
     \node[latent,minimum size=9mm,right=of zm] (z) {$\myz[t]$};
     \node[latent,minimum size=9mm,right=of z] (zp) {$\myz[t+1]$};
    \path (gbzm) edge[bend right=30] (gm);
    \path (gbz) edge[bend right=30] (g);
    \path (gbzp) edge[bend right=30] (gp);
    \path (gfzm) edge[bend left=30] (gm);
    \path (gfz) edge[bend left=30] (g);
    \path (gfzp) edge[bend left=30] (gp);
    \edge{gm} {zm,g};
    \edge{g} {z,gp};
    \edge{gp} {zp};
\end{tikzpicture} \\
\centering
\vskip 2.5mm
\begin{tikzpicture}[->]
    \node[latent,minimum size=9mm] (zm) {$\myz[t-1]$};
    \node[latent,minimum size=9mm,right=of zm] (z) {$\myz[t]$};
    \node[latent,minimum size=9mm,right=of z] (zp) {$\myz[t+1]$};
    \node[obs,minimum size=9mm,below=of zm, yshift=-8mm] (xm) {$\myx[t-1]$};
    \node[obs,minimum size=9mm,right=of xm] (x) {$\myx[t]$};
    \node[obs,minimum size=9mm,right=of x] (xp) {$\myx[t+1]$};
    \node[latent,minimum size=9mm,left=of zm, yshift=-13mm, xshift=3mm] (v) {$\myv$};
    \edge {xm} {zm};
    \edge{x} {z};
    \path(x) edge[color=silver] (zp);
    \edge{x} {zm};
    \edge{xp} {zp};
    \edge{xp} {z};
    \path(xm) edge[color=silver] (z);
    \path(xm) edge[color=silver] (zp);
    \edge{xp} {zm};
    \edge{xm} {v};
    \edge{x} {v};
    \edge{xp} {v};    
    \edge{v} {zm};
    \edge{v} {z};
    \edge{v} {zp};
    \path (zm) edge[color=goldenpoppy] (z);
    \path (z) edge[color=goldenpoppy] (zp);
    \path (zm) edge[color=goldenpoppy,bend left=30] (zp);
\end{tikzpicture}
    \caption{Graphical model of DSAE at inference time in developed form (top) and compact form (bottom). In addition to the missing arrows shown in gold, we display in silver the arrows that should not be used, as compared to the structure of the exact posterior distribution. To the best of our knowledge, DSAE is the only model that uses probabilistic dependencies that do not appear in the exact posterior distribution.\label{fig:DSAE-inference}}
\end{figure}

As for the approximate posterior distribution $q_\phi$, \citet{yingzhen2018disentangled} proposed two models. The first one, referred to as ``factorized,'' is expressed as
\begin{align}
q_\phi(\myz[1:T], \myv | \myx[1:T]) &= q_{\phi_{\myv}}(\myv | \myx[1:T]) \prod_{t=1}^{T} q_{\phi_{\myz}}(\myz[t] | \myx[t]).
\end{align}
This model thus relies on an instantaneous frame-wise inference model $q_\phi(\myz[t] | \myx[t])$ for encoding the latent vector dynamics. This approach is oversimplistic compared to the exact posterior distribution and yields inferior performance to that of the second inference model. We thus focus on the latter, which is referred to as ``full,'' and is given by
\begin{align}
q_\phi(\myz[1:T], \myv | \myx[1:T]) &= q_{\phi_{\myv}}(\myv | \myx[1:T])q_{\phi_{\myz}}(\myz[1:T] | \myv, \myx[1:T])\label{eq:DSAE-inference-full}.
\end{align}
As the authors mentioned, ``The idea behind [this] structured approximation is that content may affect dynamics.'' So far, this model has been compliant with the exact posterior distribution, as expressed by \eqref{eq:DSAE-true-inference-a}.
From the information given in the annexes of the ArXiv version of the paper, we can write the detailed equations of the full inference model as follows:
\begin{align}
\mygfwv[t] &= e_{\mygfwv}(\myx[t], \mygfwv[t-1]), \label{eq:DSAE-inference-a}\\
\mygbwv[t] &= e_{\mygbwv}(\myx[t], \mygbwv[t+1]), \label{eq:DSAE-inference-b}\\
\mygv &= [\mygfwv[T], \mygbwv[1]], \label{eq:DSAE-inference-c}\\
[\boldsymbol{\mu}_{\phi_{\myv}}(\mygv), \boldsymbol{\sigma}_{\phi_{\myv}}(\mygv) ] &= e_{\myv}(\mygv), \label{eq:DSAE-inference-d}\\
q_{\phi_{\myv}}(\myv | \myx[1:T]) &= \mathcal{N}\big(\myv; \boldsymbol{\mu}_{\phi_{\myv}}(\mygv), \text{diag}\{ \boldsymbol{\sigma}_{\phi_{\myv}}^2(\mygv)\}\big), \label{eq:DSAE-inference-e}\\
\mygfwz[t] &= e_{\mygfwz}([\myx[t], \myv], \mygfwz[t-1]), \label{eq:DSAE-inference-f}\\
\mygbwz[t] &= e_{\mygbwz}([\myx[t], \myv],\mygbwz[t+1]), \label{eq:DSAE-inference-g}\\
\mygz[t] &= [\mygfwz[t], \mygbwz[t]], \label{eq:DSAE-inference-h}\\ 
[\boldsymbol{\mu}_{\phi_{\myz}}(\mygz[t]), \boldsymbol{\sigma}_{\phi_{\myz}}(\mygz[t]) ] &= e_{\myz}(\mygz[t]), \label{eq:DSAE-inference-i}\\
q_{\phi_{\myz}}(\myz[t] | \myv, \myx[1:T]) &= \mathcal{N}\big(\myz[t]; \boldsymbol{\mu}_{\phi_{\myz}}(\mygz[t]), \text{diag}\{ \boldsymbol{\sigma}_{\phi_{\myz}}^2(\mygz[t])\}\big),\label{eq:DSAE-inference-j}
\end{align}
and for the full sequence $\myz[1:T]$ we have
\begin{align}
q_{\phi_{\myz}}(\myz[1:T] | \myv, \myx[1:T]) &= \prod_{t=1}^{T}q_{\phi_{\myz}}(\myz[t] | \myv, \myx[1:T]).
\end{align}
Note that none of the two approximations proposed by the authors (factorized and full) actually follow the dependencies of the exact posterior distribution shown in~(\ref{eq:DSAE-true-inference-c}). The graphical representation of the full inference model is shown in Figure~\ref{fig:DSAE-inference}.\footnote{In \citeauthor{yingzhen2018disentangled}'s \citeyearpar{yingzhen2018disentangled} paper, Appendix A, the schematic representation of the inference model given in Figure~9(b) is not consistent with the following sentence (reported with our notations): 
``Finally the parameters of $q_\phi(\myz[1:T] | \myv, \myx[1:T])$ are computed by a simple RNN with input $[\mygfwz[t], \mygbwz[t]]$ at time $t$.'' It is indeed inconsistent and a bit odd that $\myz[t-1]$ is not mentioned as an input of the $\myz[t]$ inference process, as is well apparent in Figure~9(b). We base the writing of \eqref{eq:DSAE-inference-h}--\eqref{eq:DSAE-inference-j} on their text and not on their figure.}

 \section{Training}
 
To derive the VLB for the DSAE model, we apply the same strategy as that applied for the other models (i.e., inject the generative model and the approximate posterior in the VLB general formulation). We do so for the full inference model (\ref{eq:DSAE-inference-full}) and obtain
\begin{align}
     \mathcal{L}(\theta,\phi,\myx[1:T]) =\;&
     \mathbb{E}_{q_{\phi}(\myv,\myz[1:T]|\myx[1:T])}\Big[\log p_{\theta_{\myx}}(\myx[1:T]|\myz[1:T],\myv)\Big] \nonumber\\
     &- D_{\textit{KL}}\Big( q_{\phi}(\myv,\myz[1:T]|\myx[1:T]) \parallel p_{\theta}(\myv,\myz[1:T]) \Big)\\
     =\;& \mathbb{E}_{q_{\phi_{\myv}}(\myv|\myx[1:T])} \Bigg[\sum_{t=1}^T \mathbb{E}_{q_{\phi_{\myz}}(\myz[t]|\myv,\myx[1:T])}\Big[\log p_{\theta_{\myx}}(\myx[t]|\myz[t],\myv)\Big] \nonumber \\
     & \hspace{-1.3cm} - \sum_{t=1}^T \mathbb{E}_{q_{\phi_{\myz}}(\myz[1:t-1] | \myv, \myx[1:T])}\Big[ D_{\textit{KL}}\big( q_{\phi_{\myz}}(\myz[t]|\myv,\myx[1:T]) \parallel p_{\theta_{\myz}}(\myz[t]|\myz[1:t-1]) \big) \Big]  \Bigg] \label{eq:DSAE-training-KLZ}  \nonumber \\
     &- D_{\textit{KL}}\big( q_{\phi_{\myv}}(\myv|\myx[1:T]) \parallel p_{\theta_{\myv}}(\myv) \big).
 \end{align}
Therefore, one must first compute $q_{\phi_{\myv}}(\myv|\myx[1:T])$ to then sample from it. Once this is achieved, the parameters of $q_{\phi_{\myz}}(\myz[t]|\myv,\myx[1:T])$ for all time-steps $t$ can be computed without sampling from any random variable. Once this is achieved, the samples of $\myz[1:t-1]$ are used to compute the $t$-th KL divergence term in~(\ref{eq:DSAE-training-KLZ}).

\chapter{Brief tour of other models}
\label{sec:other-models}
In this section, we briefly present a few other models that have been recently proposed in the literature and can be considered members of the DVAE family. We choose not to present them in a detailed manner, as in the previous sections, because they are either too far from the scope of the review, which focuses on models associating a sequence of observations with a sequence of latent variables, or too close to the already presented models.

\section{Models related to DKF}

\noindent \textbf{Latent LDS and structured VAE:} \citet{johnson2016composing} considered several models. One of them is a simplified DKF in which the latent variable model (i.e., the dynamical model) is linear-Gaussian; that is, it follows \eqref{eq:LDS-a} (with $\myu[t]$ following a standard Gaussian distribution), while the observation model is a DNN-based nonlinear model similar to the DKF observation model. They called it a latent LDS and extended it to a latent switching LDS model based on a bank of dynamical models (actually the same latent LDS but with different parameters) and an additional \emph{discrete} latent variable that controls the switch between the dynamical models over time to adjust to the observed data dynamics. This model can be considered an extension of the switching Kalman filter \citep{murphy1998switching, fox2011bayesian} to a DNN-based observation model (see also \citet{linderman2016recurrent} for a similar combination). \citet{johnson2016composing} did not provide detailed equations for these models. Rather, they showed how the use of structured mean-field approximation in the inference model, combined with the use of an observation model that is conjugate to the latent variable model, can make the inference and training processes particularly efficient. They called the resulting model a structured VAE (SVAE). As they presented these developments in the general framework of probabilistic graphical models \citep{koller2009probabilistic}, which is more general than the DVAE framework, they did not provide ``temporal equations.'' This makes this paper somewhat poorly connected to our review, although they mentioned that the DKF model \citep{krishnan2015deep} is strongly related to their work (they also implied that using RNN models for implementing time dependencies is restrictive in the general framework that they present).  

\vskip 0.25cm
\noindent \textbf{Black-box deep SSM:} Similarly, \citet{archer2015black} also focused on the structure of the approximate posterior distribution to improve the computational efficiency of the inference. They proposed using a multivariate Gaussian approximate posterior with a block tridiagonal inverse covariance matrix. They also proposed a corresponding fast and scalable inference algorithm. This general approach can be applied with different (deep and nondeep) parameterizations of the inference model and is applicable to a large family of SSMs, hence the ``black box'' denomination in the paper title. The authors mainly focused and experimented on an LG-LDS (to show that their algorithm can efficiently recover the solution of the Kalman filter), an LDS with a linear-Poisson observation model (which has no closed-form inference solution), and a basic one-dimensional nonlinear LDS. This study is well-connected with the DKF model and with deep SSMs in general. Interestingly, in this study, which, again, focuses more on the inference model than on the generative model, only the inference model is deep, whereas the generative models used in the experiments are nondeep. This study was later extended, notably addressing online learning and real-time issues \citep{zhao2020variational, zhao2019streaming}.

\vskip 0.25cm
\noindent \textbf{Deep variational Bayesian filters:} We have already mentioned this class of models in Section~\ref{sec:KVAE-generative-model}. DVBFs, which were proposed by \citep{karl2016deep}, are an extension of the class of SSM-based DVAE models with dynamical models that depend on stochastic parameters. For example, the transition model at time $t$ (i.e., between $\myz[t]$ and $\myz[t+1]$) can be a linear-Gaussian model with matrices and vectors that are a weighted sum of matrices/vectors randomly selected in a predefined set (possibly learned from data) with weights that are provided by a DNN. A similar transition model was applied within the KVAE model in \citeauthor{fraccaro2017disentangled}'s \citeyearpar{fraccaro2017disentangled} paper (see also \citet{watter2015embed}).

\vskip 0.25cm
\noindent \textbf{Disentangled SSM:} \citet{miladinovic2019disentangled} recently proposed a model called the disentangled state-space model (DSSM), in line with the DSAE model and, more generally, with models that attempt to separate the encoding of the content/object at the sequence level from that of its dynamics at the time-frame level. However, in contrast to DSAE, in which the sequence-level latent variable conditions the observation model, in DSSM, this sequence-level variable conditions the dynamical model. It is assumed to model the fact that the dynamics of an object are dependent on the considered applicative domain (e.g., enzyme kinetics or bouncing ball kinematics). In other words, \eqref{eq:DSAE-joint-seq} for DSAE can be reshaped in DSSM as\footnote{In \citeauthor{miladinovic2019disentangled}'s \citeyearpar{miladinovic2019disentangled} paper, the sequence-level variable is denoted as $D$ for  ``domain.'' For consistency, we retain the notation $\myv$ used in DSAE.}
\begin{align}
p_{\theta}(\myx[1:T], \myz[1:T], \myv) &= p_{\theta_{\myv}}(\myv) p_{\theta_{\myz}}(\myz[0]) \prod_{t=1}^{T} p_{\theta_{\myx}}(\myx[t] | \myz[t])p_{\theta_{\myz}}(\myz[t] | \myz[t-1], \myv). \label{eq:DSSM-joint-seq}
\end{align}
Here, the dynamical model is a (conditioned) first-order model. \citet{miladinovic2019disentangled} also proposed a filtering inference model that is implemented in the more general DVBF framework mentioned above.

\section{Models related to STORN, VRNN, and SRNN}
\label{sec:other-models-STORN-VRNN-SRNN}

\noindent \textbf{Variational recurrent autoencoder (VRAE):} The VRAE model, proposed by \citet{fabius2014variational}, can been considered a simplified version of STORN, from which, according to the authors themselves, it took inspiration. Here, the data sequence $\myx[1:T]$ is encoded by a single latent random vector $\myz$, instead of the sequence $\myz[1:T]$. The compact form of the generative model is given by
 \begin{align}
p_{\theta}(\myx[1:T], \myz) &= p_{\theta_{\myz}}(\myz) \prod_{t=1}^{T} p_{\theta_{\myx}}(\myx[t] | \myx[1:t-1],\myz), \label{eq:VRAE-joint} 
\end{align}
\citet{fabius2014variational} provided no information about $p_{\theta_{\myz}}(\myz)$. The generative distribution $p_{\theta_{\myx}}(\myx[t] | \myx[1:t-1],\myz)$ is implemented with a forward RNN that uses $\myz$ to calculate the first hidden state $\myh[1]$ and then iteratively takes $\myx[t-1]$ as input to calculate $\myh[t]$, which provides the parameters of the distribution of $\myx[t]$. Conversely, the inference model $q_{\phi_{\myz}}(\myz | \myx[1:T])$ is based on a forward RNN that takes $\myx[1:T]$ as input and delivers a final internal state $\myg[T]$, from which we obtain the distribution parameters for $\myz$. In short, the VRAE generative model can be described by Figure~\ref{fig:STORN}, where the sequence $\myz[1:T]$ is replaced with a single input $\myz$ for $\myh[1]$, and the VRAE inference model can be described by Figure~\ref{fig:STORN-inference}, where the sequence $\myz[1:T]$ is replaced with a single output $\myz$ for $\myg[T]$. We thus have a sequence-to-one encoding and a one-to-sequence decoding, which evoke the models designed for text/language processing mentioned in the Introduction. As \citeauthor{fabius2014variational}'s \citeyearpar{fabius2014variational} paper was published in 2014 and was part of the early papers on DVAEs, it was probably inspiring for the natural language processing (NLP) community. 

A similar model was proposed by \citet{babaeizadeh2017stochastic}, 
with a difference being that several vectors $\myx[t:T]$ are predicted from the past context $\myx[1:t-1]$ and from the unique latent vector $\myz$. The inference model is also of the form $q_{\phi_{\myz}}(\myz | \myx[1:T])$, as in \citeauthor{fabius2014variational}'s \citeyearpar{fabius2014variational} study. \citet{babaeizadeh2017stochastic} compared this model with a ``baseline'' model in which the latent vector is defined on a frame-by-frame basis (i.e., $\myz[t]$). The observation model of this baseline model is similar to that of SRNN, and the prior over $\myz[t]$ is an i.i.d.~standard Gaussian distribution. 

\vskip 0.25cm
\noindent \textbf{Factorized hierarchical variational autoencoder (FHVAE):} The FHVAE model was proposed by \citet{hsu2017unsupervised}, to learn disentangled and interpretable latent representations from sequential data without supervision. To achieve this aim, FHVAE
explicitly models the multi-scaled aspect of the temporal information contained in the data. This is done by splitting each sequence of data vectors into a set of fixed-size consecutive sub-sequences, called segments, and defining two latent variables $\myz$ and $\myv$ at the segment level.\footnote{In \citeauthor{hsu2017unsupervised}'s \citeyearpar{hsu2017unsupervised} paper, $\myz$ and $\myv$ are denoted $\mathbf{z}_1$ and $\mathbf{z}_2$, respectively. We changed the notation to avoid confusion between the variable index and time index.} The former is dedicated to capturing data information at the segment level, whereas the latter is dedicated to capturing data information across segments (i.e., at the sequence level). This model is particularly appropriate for speech signals: In this case, $200$-ms segments represent the approximate duration of a syllable, and thus, $\myz$ would typically encode phonetic information, whereas $\myv$ would typically encode speaker information at the level of a complete utterance. In essence, FHVAE is strongly related to DSAE, which also contains a sequence-level latent variable $\myv$ but preserves a time-frame resolution for the dynamical latent variable $\myz[t]$ (see Section~\ref{sec:DSAE}). In fact, DSAE was published after FHVAE, from which it was probably inspired.

Even if we do not detail this model, we report a few equations to help better understand how segmental modeling works. Let here $t \in [1,T]$ denote the index of a vector within a segment (each segment has $T$ vectors), and let $n \in [1,N]$ denote the index of a segment within a sequence. The FHVAE observation model \emph{for each individual segment of data} is given by
\begin{align}
\myhn[t] &= d_{\myh}(\myzn, \myvn, \myhn[t-1]), \label{eq:FHVAE-a} \\
[\boldsymbol{\mu}_{\theta_{\myx}}(\myhn[t]),\boldsymbol{\sigma}_{\theta_{\myx}}(\myhn[t])] &= d_{\myx}(\myhn[t]), \label{eq:FHVAE-b} \\
p_{\theta_{\myx}}(\myxn[t] | \myhn[t]) &= \mathcal{N}\big(\myxn[t] ; 
\boldsymbol{\mu}_{\theta_{\myx}}(\myhn[t]),\textrm{diag}\{\boldsymbol{\sigma}_{\theta_{\myx}}^2(\myhn[t])\}\big). \label{eq:FHVAE-c}
\end{align}
In \eqref{eq:FHVAE-a}, $\myzn$ and $\myvn$ respectively denote the latent vectors $\myz$ and $\myv$ for the considered $n$-th segment. There is a single pair of such vectors for each segment, and hence, a many-to-one encoding and one-to-many decoding at the segment level. In practice, these equations are implemented with an LSTM network.
The prior distribution of $\myzn$, $p_{\theta_{\myz}}(\myzn)$, is a centered isotropic Gaussian that is independent of both the segment and the sequence. In contrast, the prior distribution of $\myvn$ depends on a latent variable $\myw$, which is defined at the sequence level and whose prior distribution $p_{\theta_{\myw}}(\myw)$ is also a centered isotropic Gaussian. The distribution of $\myvn$ is then given by $p_{\theta_{\myv}}(\myvn | \myw) = \mathcal{N}(\myvn ; \myw, \sigma_{\theta_{\myv}}^2\mathbf{I}_{L_v})$. For a given sequence, $p_{\theta_{\myv}}(\myvn | \myw)$ depends on the value of $\myw$ drawn for that particular sequence. In practice, all generated $\myvn$ vectors within a sequence are close to $\myw$. This makes $\myvn$ a sequence-dependent latent factor, whereas $\myzn$ behaves as a segment-dependent and sequence-independent latent factor.
The joint density of a sequence is given by
 \begin{align}
\nonumber p_\theta(\myxN[1:T], \myzN, \myvN, \myw) = p_{\theta_{\myw}}(\myw) \prod_{n=1}^{N} &\prod_{t=1}^{T} p_{\theta_{\myx}}\big(\myxn[t] | \myh[t](\myzn,\myvn)\big)  \\
& p_{\theta_{\myz}}(\myzn)p_{\theta_{\myv}}(\myvn | \myw),
\end{align}
where, as in the previous sections, $\myh[t]\big(\myzn,\myvn\big)$ is a shortcut for the function that results from unfolding the recurrence in \eqref{eq:FHVAE-a}.

The inference model is a many-to-one encoder that works at the segment level: each segment  $\myxn[1:T]$ is encoded into a pair $\{\myzn,\myvn\}$ (plus an estimate of $\myw$ for each whole sequence).
As for the variational approximate posterior $q_\phi$,  \citet{hsu2017unsupervised} proposed the following model:
 \begin{align}
q_{\phi}\big(\myzN, \myvN, \myw | \myxN[1:T]\big) &= q_{\phi_{\myw}}(\myw) \prod_{n=1}^{N} q_{\phi_{\myz}}\big(\myzn | \myxn[1:T], \myvn \big) q_{\phi_{\myv}}\big(\myvn | \myxn[1:T] \big),
\end{align}
where $q_{\phi_{\myz}}$ and $q_{\phi_{\myv}}$ are both implemented with a forward LSTM network, whose last state vector is passed to a DNN to provide the distribution parameters. Two encoders are chained here: The first one is used to generate $\myvn$ (by sampling $q_{\phi_{\myv}}\big(\myvn | \myxn[1:T]\big)$), and then, $\myvn$ is injected in the second encoder to generate $\myzn$. As for $q_{\phi_{\myw}}(\myw)$, it is a Gaussian distribution whose mean vector is obtained from a look-up table that is jointly learned with the model parameters (see \citet{hsu2017unsupervised} for details).
Cascading the sequence-to-one encoder with the one-to-sequence decoder results in a sequence-to-sequence neural network architecture that is trained by maximizing the VLB (not detailed here). The model can be optimized at the segment level instead of the sequence level; that is, each data segment can be used as a batch dataset. According to \citet{hsu2017unsupervised}, this can solve scalability issues when the training sequences become too long. 

\vskip 0.25cm
\noindent \textbf{Deep recurrent attentive writer (DRAW):} A somewhat dual model of VRAE was proposed by \citet{gregor2015draw} and called DRAW. This model considers a sequence of latent vectors $\myz[1:T]$ to encode a single static but highly structured data $\myx$ (typically a low-resolution image). The generative model is of the general form $p_{\theta_{\myx}}(\myx | \myz[1:T])$. It involves the iterative construction of a sequence $\myxh[1:T]$ that can be considered the sequence of images resulting from the ``natural'' drawing of $\myx$ over time. The dependency of $\myx$ on $\myz[1:T]$ is implemented by combining the output of a decoder RNN (which takes $\myz[t]$ as the input) and a so-called canvas matrix $\mathbf{c}_{t-1}$, which encodes the difference between the final target image $\myx$ and the current draw $\myxh[t-1]$. Hence, the model combines deep learning and some form of predictive coding \citep{gersho2012vector}. The inference model is of the form $q_{\phi_{\myz}}(\myz[t] | \myz[1:t-1], \myx)$ and is implemented with an encoder RNN. This network takes as inputs a combination of $\myx$, $\myxh[t-1]$, and the decoder output at the previous time step; hence, the predictive coding is implemented in closed-loop mode \citep{gersho2012vector}. As the name indicates, DRAW includes a selective attention model that enables it to focus on the most relevant parts of the observation. The description of such an attention model is beyond the scope of the present review (see \citet{gregor2015draw} and references therein for details).

\vskip 0.25cm
\noindent \textbf{Neural adaptive sequential Monte Carlo (NASMC):} \citet{gu2015neural} proposed the NAMSC generative model, which combines an infinite-order Markovian model on $\myz[t]$ (as that used in DSAE) with the most general observation model, which is that used in STORN and VRNN. In one variant of this model, they replaced the Markovian model on $\myz[t]$ with an i.i.d.~model, and thus, this variant is similar to STORN. The inference model has the same general form as that of VRNN; that is, it follows \eqref{eq:VRNN-inference-model-seq-general} and is parametrized by an RNN. The originality of NASMC lies in connecting the DVAE inference framework with sequential Monte Carlo (SMC) sampling. The inference model was used as a proposal distribution for SMC sampling. In fact, a complete framework to learn the parameters of the generative model, the proposal model, and for sampling from the posterior distribution with SMC was proposed. In other words, \citet{gu2015neural} showed that their sampling-based approach can be used to optimize the observed data marginal likelihood for estimating the generative model parameters in the variational framework. For other applications of SMC methods in the general context of variational approximations, see \citep{maddison2017filtering,le2018auto,naesseth2018variational}.

\vskip 0.25cm
\noindent \textbf{Recurrent state-space model (RSSM):} \citet{hafner2018learning} used the RSSM model, which is very similar, if not identical, to the VRNN model used in the driven mode. In fact, RSSM corresponds to VRNN with an external input $\myu[t]$ (denoted as $\mathbf{a}_{t-1}$ by \citet{hafner2018learning}), which is used in place of $\myx[t-1]$ to compute $\myh[t]$. This is how VRNN was presented in the ``Related Works'' section of the SRNN paper \citep{fraccaro2016sequential} (see in particular Figure~4(b) of this latter paper). \citet{hafner2018learning} used this RSSM model for learning the dynamics and planning the actions of a synthetic agent from image sequences in a reinforcement learning framework. They also presented a way to perform multi-step prediction (i.e., prediction several steps ahead).  

\vskip 0.25cm
\noindent \textbf{Stochastic video generation (SVG):} \citet{denton2018stochastic} presented a model very similar to STORN and applied it to stochastic video generation and multiple-frame video prediction, similarly to \citep{babaeizadeh2017stochastic}. The inference model is of the form $q_{\phi_{\myz}}(\myz[t] | \myx[1:t])$. It is thus also similar to the inference model of STORN (see \eqref{eq:STORN-inference-model-seq-general}). A variant of the generative distribution of $\myz[t]$, called ``learned prior'' in the paper, is also proposed. It is of the form $p_{\theta_{\myz}}(\myz[t]|\myx[1:t-1])$ and can be considered a simplification of the generative distribution of $\myz[t]$ in VRNN or SRNN.  

\section{Other models}

\noindent \textbf{Factorized variational autoencoder (FVAE):} \citet{deng2017factorized} proposed the FVAE model, which combines a VAE with tensor factorization \citep{kuleshov2015tensor, huang2016flexible}. This latter is applied on the latent vector $\myz$. As one of the tensor factorization dimension is discrete time, this model implicitly involves data dynamics modeling. However, the temporal patterns are sampled from a standard log-normal distribution, hence independently over time, and data decoding is also processed independently at each time frame. It is thus unclear how temporal dynamics are actually encoded.

\vskip 0.25cm

\noindent \textbf{Gaussian process variational autoencoder (GP-VAE):} \citet{fortuin2020gp} recently combined a VAE for the observed data dimension reduction and a multivariate Gaussian process (GP) \citep{williams2006gaussian} for modeling the dynamics of the resulting latent vector $\myz[t]$. For the GP model, they used a Cauchy kernel, which is appropriate to model data with multiscale time dynamics. They proposed using an approximate posterior distribution that is also a multivariate GP (here, a first-order one). The resulting overall GP-VAE model was trained with the VAE methodology and then used to efficiently recover missing data in test sequences (in videos and medical data). One interesting property of this model is that it provides interpretable uncertainty estimates.

\chapter{Experiments}
\label{sec:experiments}
In this chapter, we present an experimental benchmark of six of the DVAE models detailed in the previous chapters (DKF, STORN, VRNN, SRNN, RVAE, DSAE). This benchmark is conducted on a dataset of speech signals and a dataset of 3D human motion data. 
We provide a series of quantitative results on the task of analysis-resynthesis; that is encoding with the DVAE encoder followed by decoding with the DVAE decoder. We also provide qualitative results, in the form of examples of data generated by the models. 
We first present the models implementation in Section~\ref{sec:model-architecture}. Then, we describe the experimental protocol, datasets, model training and testing settings, and evaluation metrics in Section~\ref{sec:experimental-protocol}. Finally, we present and discuss the results in Sections~\ref{sec:results-speech} and \ref{sec:results-motion}. We recall that a link to the open-source code and the best-trained models can be found at \url{https://team.inria.fr/robotlearn/dvae/}.

\section{DVAE architectures}
\label{sec:model-architecture}

\begin{figure*}[htbp]
    \centering
    \begin{tabular}{cc}
    \includegraphics[width=0.47\linewidth]{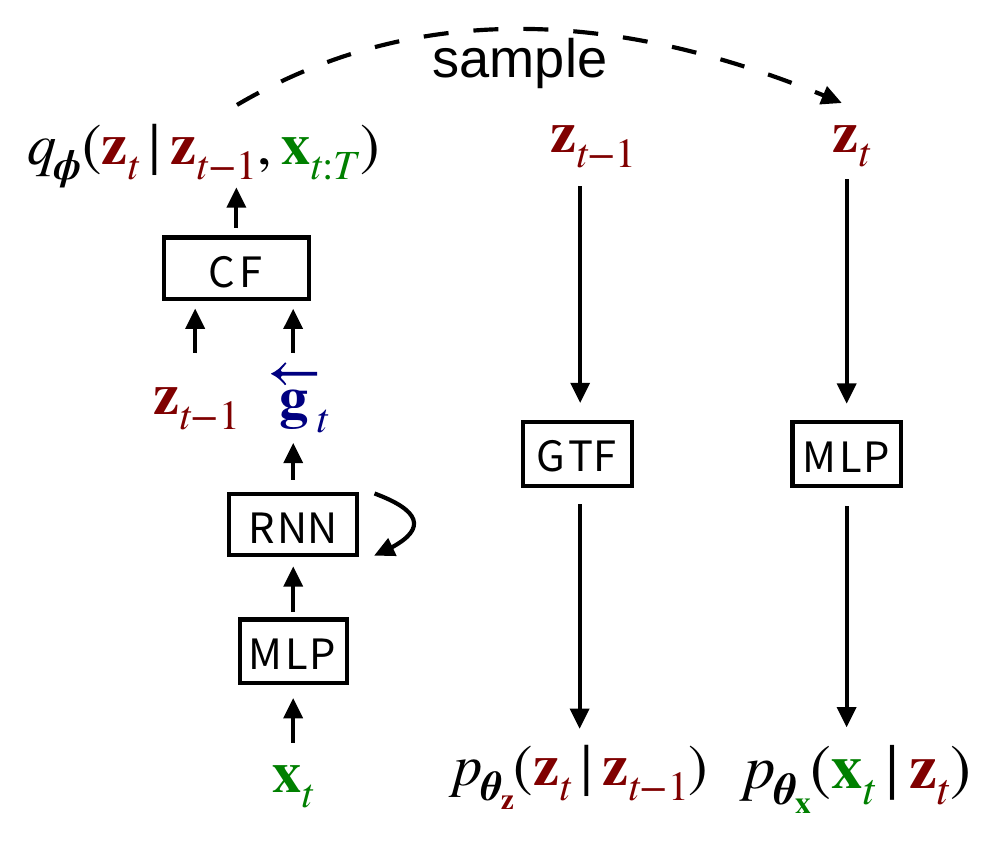} &
    \hspace{-1cm} \includegraphics[width=0.40\linewidth]{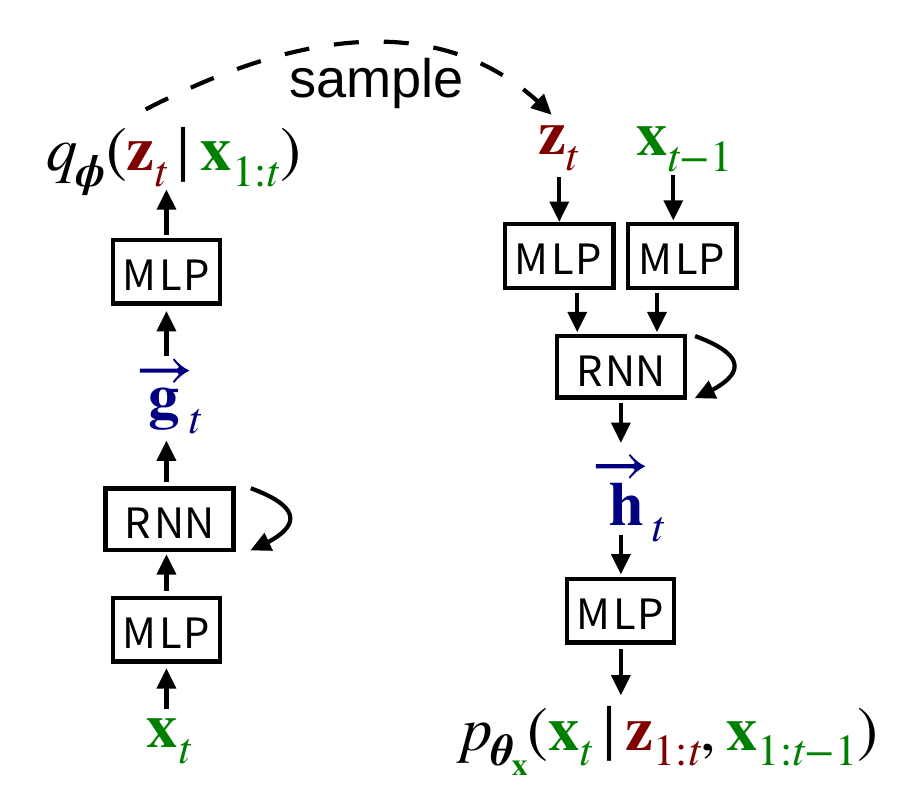} \\
    {(a) DKF} & {(b) STORN}\\
    \includegraphics[width=0.49\linewidth]{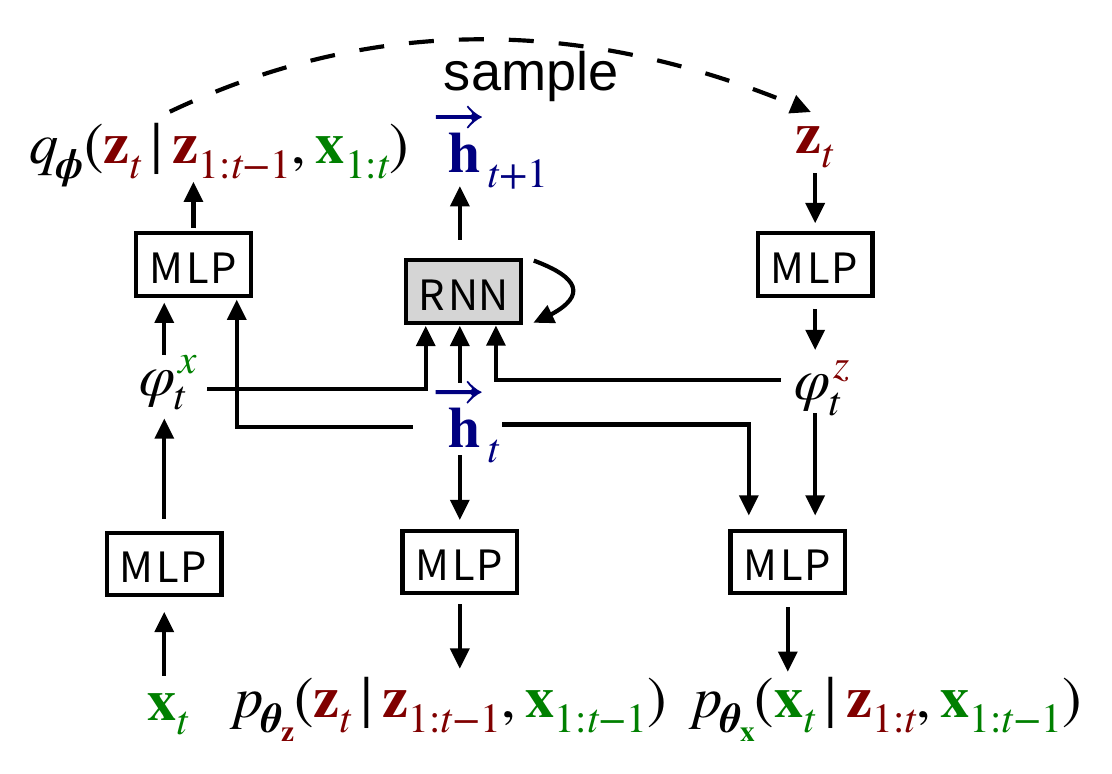} &
    \hspace{-1.05cm} \includegraphics[width=0.49\linewidth]{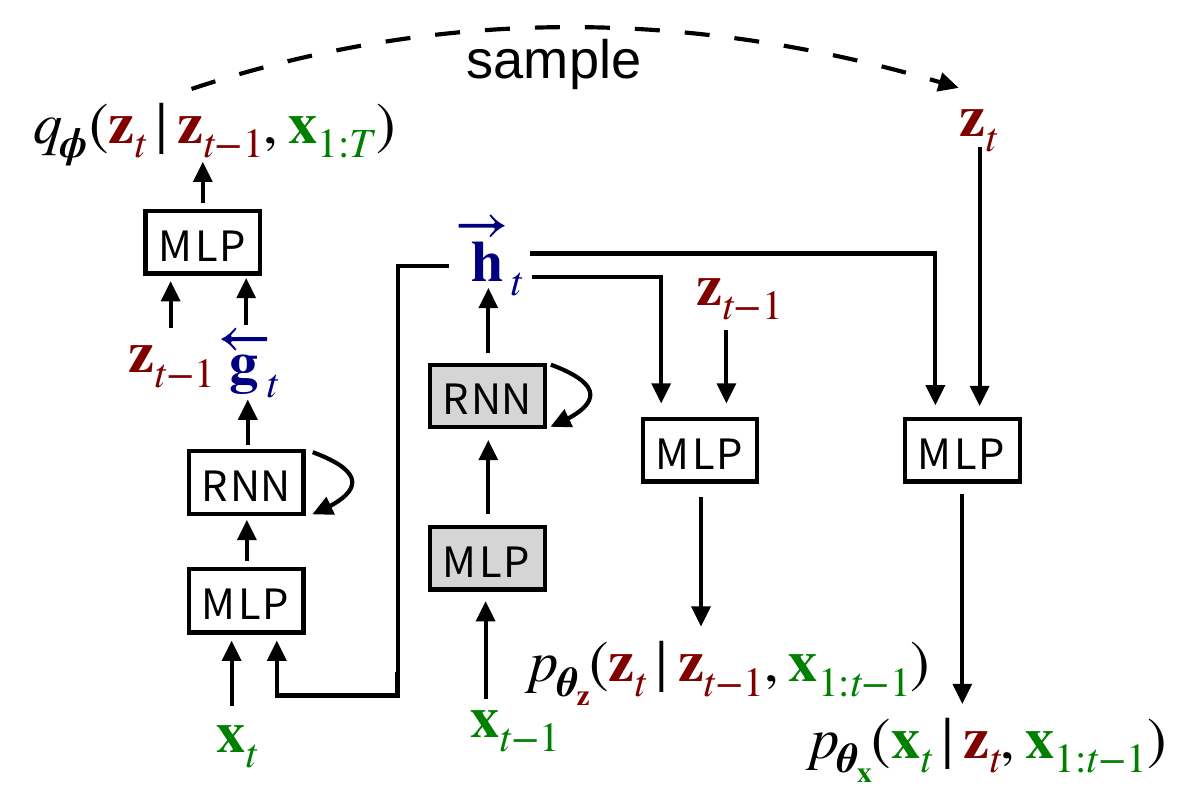} \\
    {(c) VRNN} & {(d) SRNN}\\
    \includegraphics[width=0.40\linewidth]{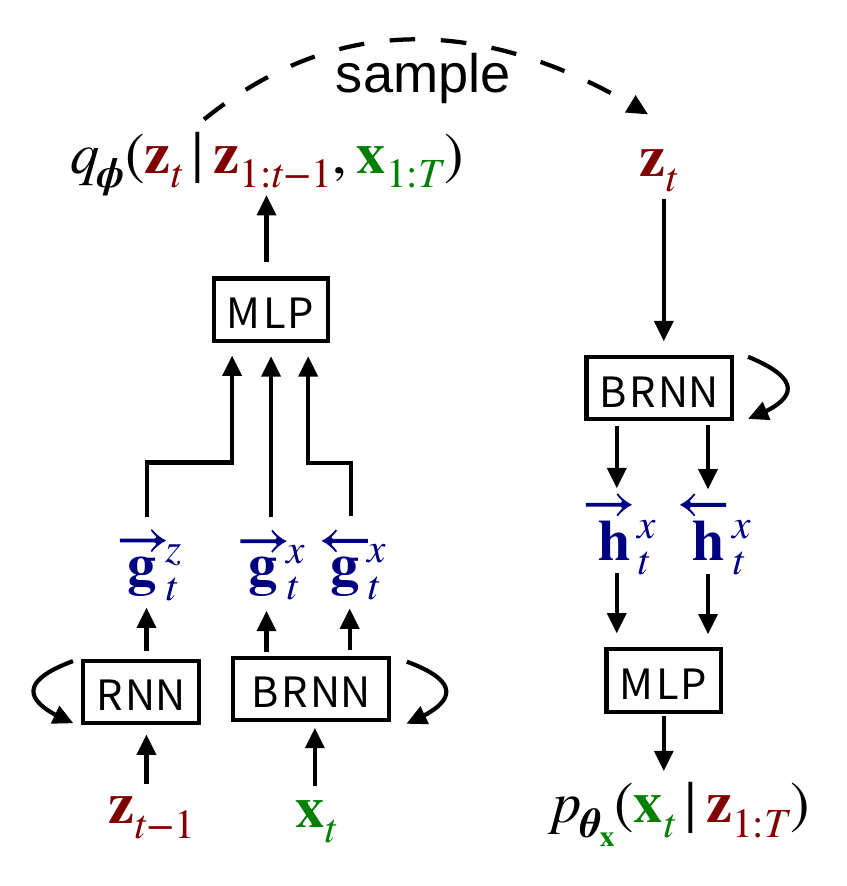} &
    \hspace{-1.05cm} \includegraphics[width=0.55\linewidth]{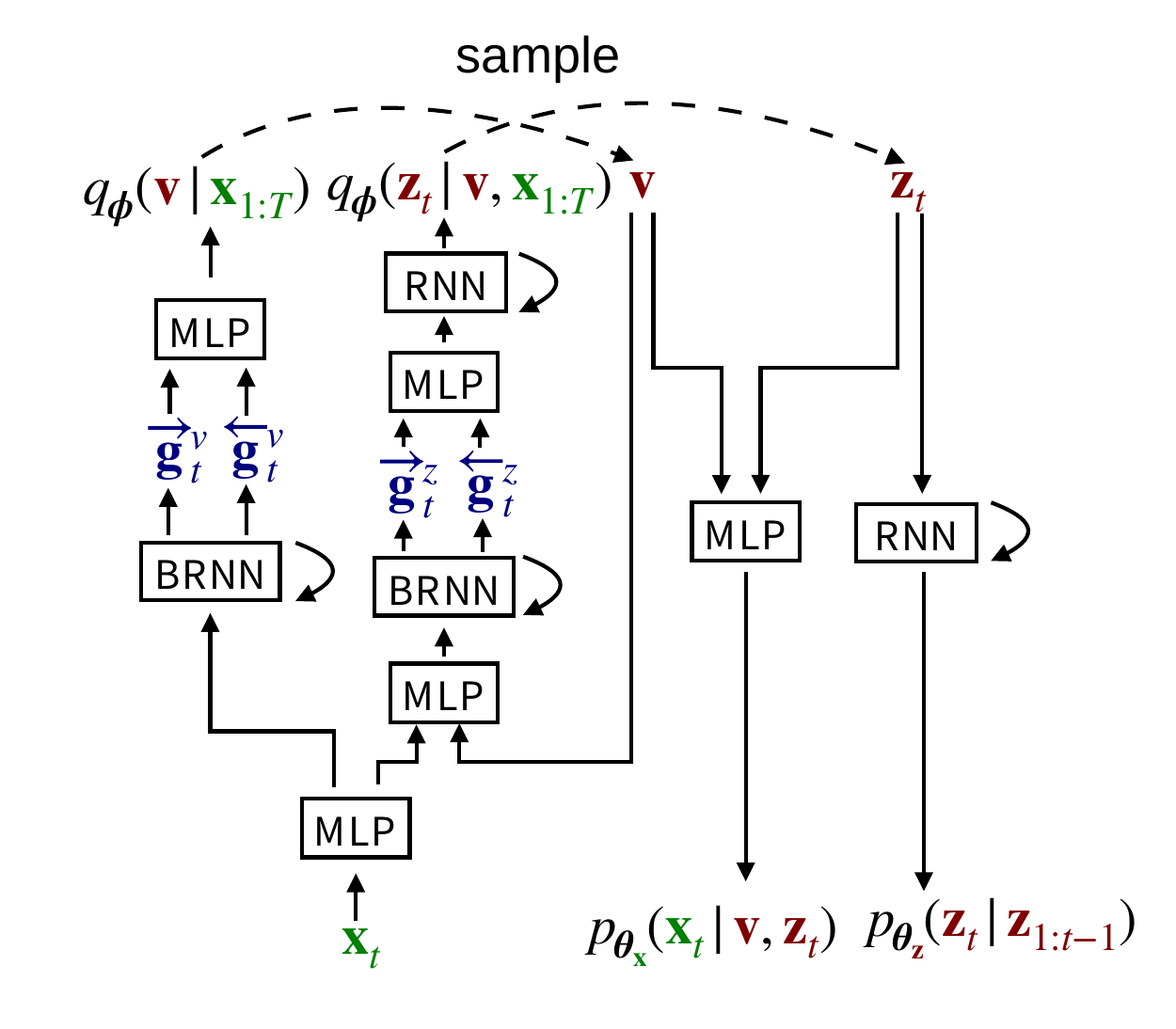} \\
    {(e) RVAE} & {(f) DSAE}\\
    \end{tabular}
    \caption{Model architecture for the six tested DVAE models. For DKF, CF and GTF are a combiner function and a gated transition function, respectively. These functions are described in Appendix~\ref{Appendix-B} (Section~\ref{sec:DKF-implementation-speech}) and in \citeauthor{krishnan2015deep}'s \citeyearpar{krishnan2015deep} paper. For VRNN and SRNN, the gray-shaded boxes are modules shared by the encoder and decoder.
    }
    \label{fig:arc}
\end{figure*}

The architectures of the six DVAE models that we benchmark are summarized in Figure~\ref{fig:arc}. For each model, we represent the high-level computational graph corresponding to the encoding, sampling and decoding processes. In particular, we show the types of layers that compose the encoder and decoder networks. Note that none of the DVAEs is used in the driven mode; that is, none of them feature an external input $\myu[1:T]$. The MLPs are generally used to extract high-level features and/or as a combiner function, whereas the recurrent networks (RNNs and BRNNs) are used to accumulate the information over time. All RNNs are instantiated as LSTM networks (and BRNNs are instantiated as bidirectional LSTM networks). DKF includes a specific combiner function (CF) at the end of the encoder and a gated transition function (GTF) to implement the dynamical model. These functions are described in the original DKF paper by \citet{krishnan2015deep} and we report them in Appendix~\ref{Appendix-B} (Sections~\ref{sec:DKF-implementation-speech}). 
We recall that the internal state vector $\myh[t]$ of VRNN and SRNN is shared between the encoder and decoder (see Sections~\ref{sec:VRNN-inference} and \ref{sec:SRNN-inference}). In Figure~\ref{fig:arc}, we represent the corresponding shared networks with grey-shaded boxes. We also recall that RVAE was presented in two versions: causal and noncausal (see Section~\ref{sec:RVAE-generative-model}). Figure~\ref{fig:arc} only shows the architecture of the noncausal RVAE. The schema of the causal RVAE architecture is obtained by replacing the BRNN in the inference and generative models with a backward and a forward RNN, respectively. Finally, for all output variance parameters, we use log-parameterization (i.e., the output of the network corresponding to a variance parameter $\sigma^2$ is actually $\log \sigma^2$). 

The general architectures shown in Figure~\ref{fig:arc} are common to the two sets of experiments that we conducted on speech data and on human motion data, although with different layer dimensions. For our experiments with speech data, the observed, latent, and RNN internal state vectors are of dimension $513$, $16$, and $128$, respectively, for all DVAE models. For our experiments with human motion data, they are of dimension $96$, $10$, and $64$, respectively, for all DVAE models (see Sections~\ref{subsec:speech-data} and \ref{subsec:3Dhm-data} for the definition of the observed data vectors). Low-level implementation details such as the number of layers and the number of units per layer may also differ between the two sets of experiments on speech and human motion data. These details are provided in Appendices~\ref{Appendix-B} and~\ref{Appendix-C}).

\section{Experimental protocol}
\label{sec:experimental-protocol}

\subsection{Speech data} 
\label{subsec:speech-data}

For our experiments with speech data, we used the Wall Street Journal dataset \citetext{WSJ0; \citealp{WSJ0}}, which comprises speech read from WSJ news. We used the speaker-independent, medium vocabulary (5k words) subset of the corpus. More precisely, the \textit{si\_tr\_s} subset ($\sim$25~h) was used for training, the \textit{si\_dt\_05} subset ($\sim$2~h) was used for validation, and the \textit{si\_et\_05} subset ($\sim$1.5~h) was used for testing. 

The raw speech waveform was sampled at 16~kHz. Analysis-resynthesis was performed with the DVAEs in the time-frequency domain on power spectrograms. Time-domain speech signals were thus preprocessed with the short-time Fourier transform (STFT), using a $64$-ms sine window ($1,024$ samples) with $25$\%-overlap to obtain sequences of $513$-dimensional discrete Fourier spectra (for positive frequencies). Then, we computed the squared magnitude of these STFT spectrograms. For the training dataset, we set $T=50$, meaning that speech utterances of $0.8$~s were extracted from the raw dataset and pre-processed with the STFT. In summary, each training speech sequence is a $513 \times 50$ STFT power spectrogram. 
This data preprocessing resulted in a set of $N_{\rm{tr}} = 46,578$ training sequences (representing about $10.3$ hours of speech signal) and $N_{\rm{val}} = 7,775$ validation sequences ($\sim1.7$~h). For testing, we used the STFT spectrogram of each complete test sequence (with the beginning and ending silence portions removed), which can be of variable length, most often larger than $2.4$~s. 

As discussed by~\citet{leglaive2020recurrent} and mentioned in Section~\ref{sec:RVAE-generative-model}, the complex-valued STFT coefficients are modeled with a zero-mean circular complex Gaussian distribution (see \eqref{eq:Simon-b}), whereas $\myz[t]$ is modeled as usual with a real-valued Gaussian distribution. The data sequence $\myx[1:T]$ processed by the DVAE models is the squared magnitude of the STFT spectrogram (i.e., a real-valued nonnegative power spectrogram). The corresponding phase spectrogram is directly combined with the DVAE output magnitude spectrogram to reconstruct the output speech signal using inverse STFT with overlap-add. Modeling the STFT coefficients with a zero-mean circular complex Gaussian distribution with variance ${\sigma}_{\theta_{\myx},f,t}^2(\cdot)$ amounts to modeling each entry $x_{f,t}$ of the power spectrogram $\myx[1:T]$ with a Gamma distribution with shape parameter $1$ and scale parameter ${\sigma}_{\theta_{\myx},f,t}^2(\cdot)$ (i.e., $x_{f,t} \sim \mathcal{G}(1, 1/{\sigma}_{\theta_{\myx},f,t}^2(\cdot))$).\footnote{Here, we do not specify the variables generating the variance, as they depend on the DVAE model. Instead, the subscripts indicate frequency bin $f$ and time frame $t$.} This also amounts to using the Itakura-Saito divergence between $x_{f,t}$ and ${\sigma}_{\theta_{\myx},f,t}^2(\cdot)$ in the reconstruction term of the VLB \citep{girin2019notes}. We recall that all presented DVAE models are versatile regarding the conditional pdf of $\myx[t]$, and using a Gamma distribution (more appropriate for speech/audio power spectrograms) in place of the Gaussian distribution that was used in the generic presentation of the models does not present any problem. 
The linear layer estimating the parameters of this distribution has $513$ output units corresponding to the log-variance parameters $\{\log {\sigma}_{\theta_{\myx},f,t}^2(\cdot)\}_{f=1}^F$.

\subsection{3D human motion data}
\label{subsec:3Dhm-data}

For our experiments with 3D human motion data, we used the H3.6M dataset \citep{h36m}, which is one of the largest dataset of the kind and has been widely used in video prediction \citep{finn2016unsupervised}, human pose and shape estimation \citep{bogo2016keep}, and human motion prediction \citep{martinez2017human}. This dataset was obtained from multi-view video recordings of 11 professional actors performing 17 various scenarios (e.g., discussing, smoking, taking a picture, or talking on the phone), using four calibrated cameras with 50~Hz resolution. The 3D $\{x,y,z\}$ coordinates of 32 human skeleton joints were extracted from these multi-view recordings. Each set of coordinates was centered w.r.t.~the coordinates of the pelvis joint. 

For our experiments with DVAEs, each data frame is organized as a $96$-dimensional vector $\myx[t]$ by concatenating the 3D coordinates of the 32 joints. We used sequences of $T = 50$ consecutive vectors, which represent a duration of 2~s (the data were previously downsampled by a factor 2). These sequences were obtained by applying a $50$-frame sliding window on the original H3.6M sequences, with a shift of two frames. In summary, each example in our dataset is thus a matrix of 3D coordinates of skeleton joints of size $96 \times 50$, which corresponds to 2~s of human motion. 

In H3.6M, 15 scenarios from 7 actors are provided with the ground-truth annotations. Similarly to \citet{mao2020history}, we used the data of all 15 scenarios from 5 actors (Actors 1, 6, 7, 8, and 9) for training and from 1 actor (Actor 11) for testing. Applying the sequence extraction procedure described above led to $N_{\rm{tr}} = 88,952$ training sequences ($\sim50$~h) and $N_{\rm{test}} = 13,838$ test sequences ($\sim8$~h). For validation, to reduce the computation time, we selected $128$ sequences for each of the 15 scenarios by Actor 5 ($N_{\rm{val}} = 1,920$ sequences, $\sim1$~h). 

Following \citet{bayer2014learning, petrovich21actor}, the 3D human motion data vectors $\myx[t]$ are modeled by a Gaussian generative conditional distribution with a covariance matrix equal to the identity matrix.

\subsection{Training and testing}
 
All tested models were implemented in PyTorch~\citep{paszke2019pytorch}. To train the models, we used the Adam optimizer \citep{Adam} with mini-batches of size $128$. For the speech data, we set the learning rate to $0.002$, whereas we used $0.0001$ for the human motion data. We also used early stopping on the validation set with a patience of 50 epochs for the speech data and 30 epochs for the human motion data. During the training of the models with the human motion data, we applied warm-up to the KL regularization term in the VLB by multiplying it with a factor $\beta$ and linearly increasing this factor from 0 to 1 after each epoch, during the first 50 epochs~\citep{sonderby2016ladder,vahdat2020nvae}. 

Once a model was trained on the training set, with early stopping on the validation set, its weights were fixed and the model was run on the test set. We report the average performance obtained on the test set using the metrics presented in the next subsection.

\subsection{Evaluation metrics} 
 
For the experiments on speech data, we used three metrics to evaluate the resynthesized speech quality and compare the performance of the different DVAE models: the scale-invariant signal-to-distortion ratio (SI-SDR) in dB \citep{le2019sdr}, the perceptual evaluation of speech quality (PESQ) score in $[-0.5, 4.5]$ \citep{rix2001perceptual}, and the extended short-time objective intelligibility (ESTOI) score in $[0, 1]$ \citep{taal2011algorithm}. For all metrics, the higher the better. Note that these metrics are applied on the time-domain signals (i.e., speech waveforms). We combined the reconstructed magnitude spectrogram with the phase spectrogram of the original signal to obtain the analyzed-resynthesized speech waveform (using the inverse STFT).

For 3D human motion data, we can directly compare each original sequence with the corresponding analyzed-resynthesized sequence. We used the mean per joint position error (MPJPE) proposed by \citet{h36m}, which is an averaged Euclidean distance per joint. We report the results in millimeters (mm). Note that this error corresponds to the log-likelihood term (or reconstruction error term) of the VLB, up to a constant factor that is controlled through the setting of the variance of the data conditional distribution.

\section{Results on speech data}
\label{sec:results-speech}

\begin{table} [t]
\center
\tabcolsep=0.11cm
    \begin{tabular}{l c c c}
    \toprule 
    DVAE & SI-SDR (dB) & PESQ & ESTOI \\
    \midrule
    VAE             & 5.3 & 2.97 & 0.83  \\
    DKF             & 9.3 & 3.53 & 0.91 \\ 
    STORN           & 6.9 & 3.42 & 0.90 \\ 
    VRNN            & 10.0 & 3.61 & 0.92 \\ 
    SRNN            & \bf{11.0} & \bf{3.68} & \bf{0.93} \\ 
    RVAE-causal     & 9.0 & 3.49 & 0.90 \\ 
    RVAE-noncausal  & 8.9 & 3.58 & 0.91 \\ 
    DSAE            & 9.2 & 3.55 & 0.91 \\ 
    \midrule
    SRNN-TF-GM      & $-1.0$ & 1.93 & 0.64 \\
    SRNN-GM         & 7.8 & 3.37 & 0.88 \\
    \bottomrule
    \end{tabular}
\caption{Performance of the DVAE models tested in our speech analysis-resynthesis experiment. The SI-SDR, PESQ, and ESTOI scores are averaged over the test subset of the WSJ0 dataset. STORN, SRNN and VRNN were trained and tested in the teacher-forcing mode. SRNN-TF-GM stands for the SRNN model trained in the teacher-forcing mode and tested in the generation mode. SRNN-GM stands for the SRNN model trained and tested in generation mode.}
\label{tab:comparison_speech}
\end{table}

\subsection{Analysis-resynthesis}
\label{subsec:analysis-resynthesis-speech}

\begin{figure}[t]
    \centering
    \begin{tabular}{c}
        \hspace{-.3cm}\small{Original} \\
        \includegraphics[width=\linewidth]{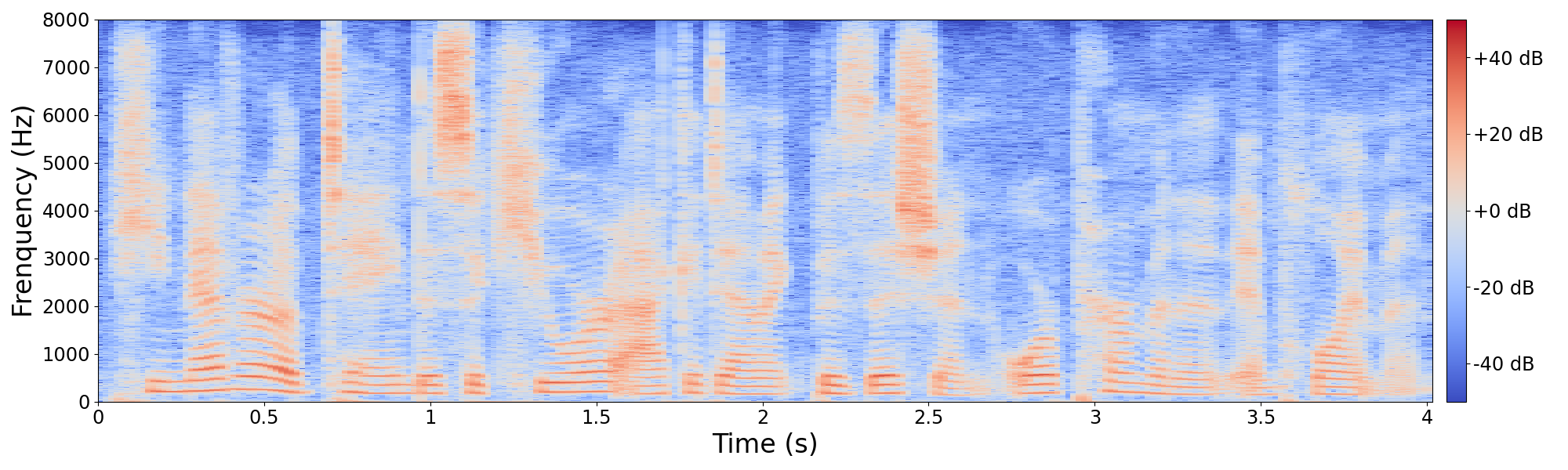} \\
        \hspace{-.3cm}\small{VAE} \\
        \includegraphics[width=\linewidth]{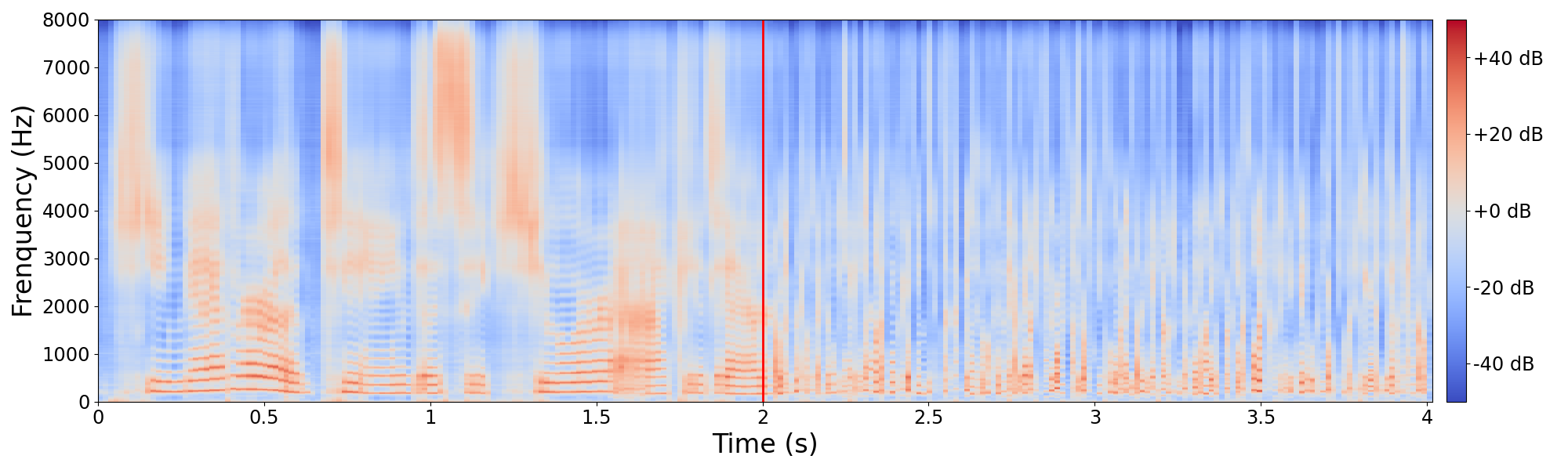} \\
        \end{tabular}
    \caption{Example of power spectrogram for a speech signal uttered by a female speaker. Top: spectrogram of the original signal. Bottom: spectrogram reconstructed (0-2~s) and generated (2-4~s) with a vanilla VAE (the red line indicates the transition between reconstruction and generation).}
    \label{tab:dvae_orig_vae}
\end{figure}

\begin{figure}[!ht]
    \centering
    \begin{tabular}{c}
        \hspace{-.3cm}\small{DKF} \\
        \includegraphics[width=\linewidth]{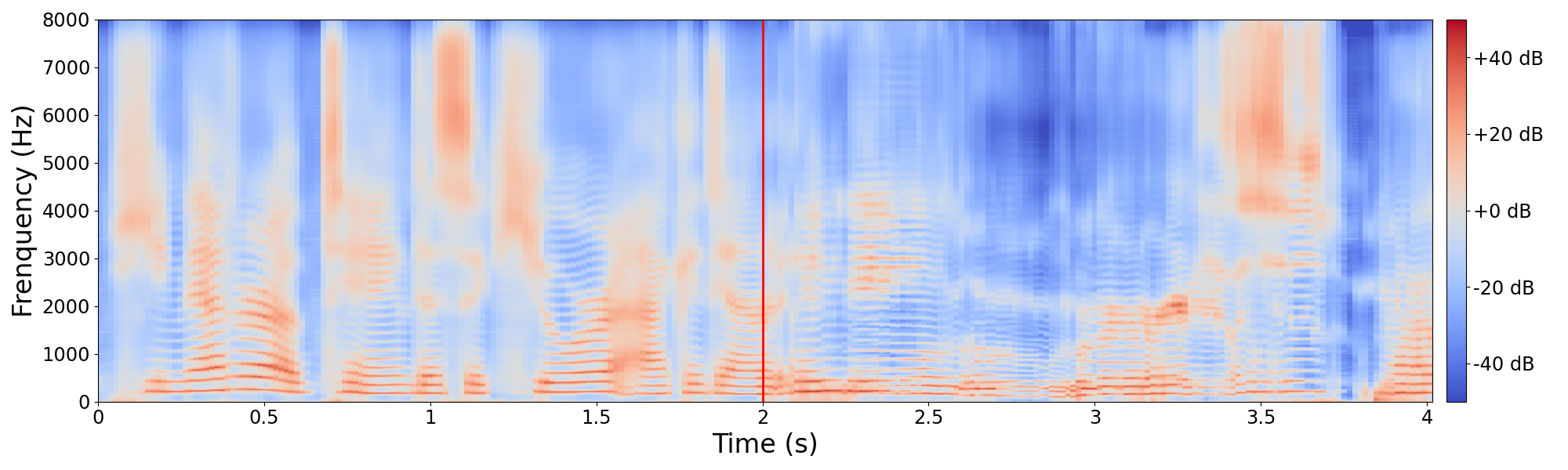} \\
        \hspace{-.3cm}\small{DSAE} \\
        \includegraphics[width=\linewidth]{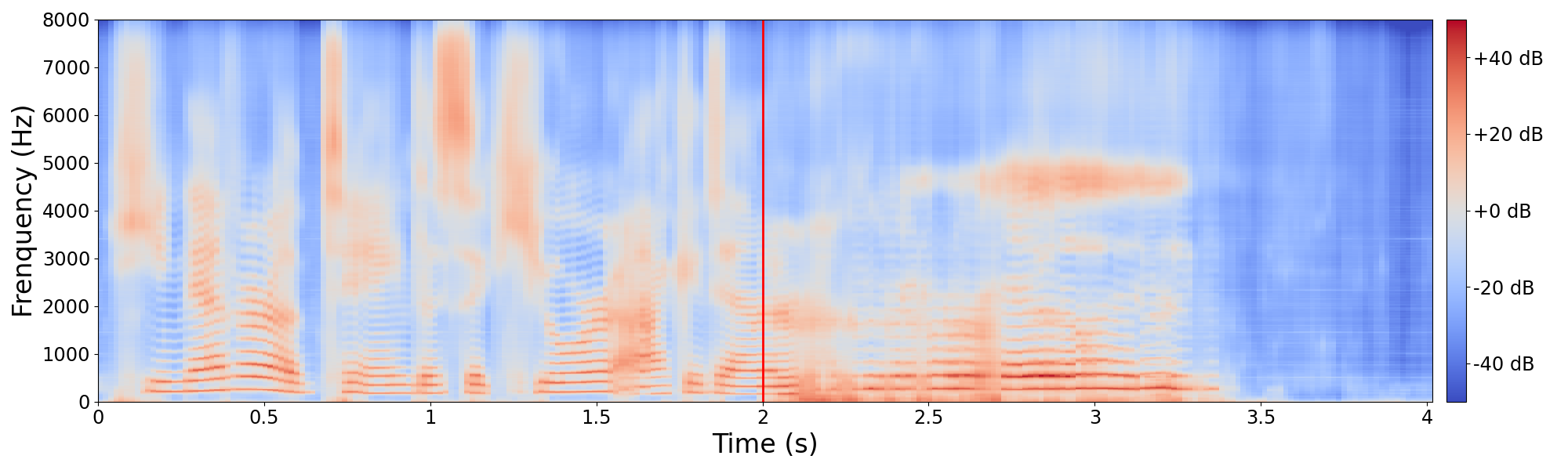} \\
        \hspace{-.3cm}\small{Noncausal RVAE} \\
        \includegraphics[width=\linewidth]{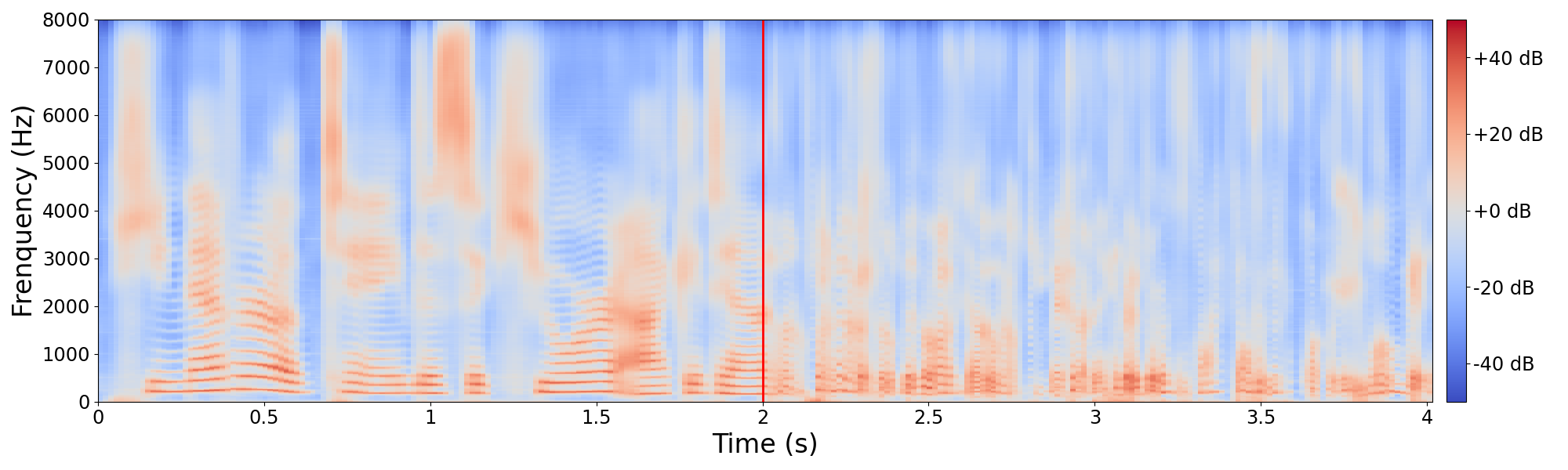} \\
        \end{tabular}
    \caption{Example of speech power spectrogram reconstructed (0-2~s) and generated (2-4~s) by a DVAE model (the original spectrogram is in Figure~\ref{tab:dvae_orig_vae} (top)). Top: DKF; middle: DSAE; bottom: noncausal RVAE. The red line indicates the transition between reconstruction and generation.}
    \label{tab:dvae_pred_nonauto}
\end{figure}

\begin{figure}[!ht]
    \centering
    \begin{tabular}{c}
        \hspace{-.3cm}\small{STORN} \\
        \includegraphics[width=\linewidth]{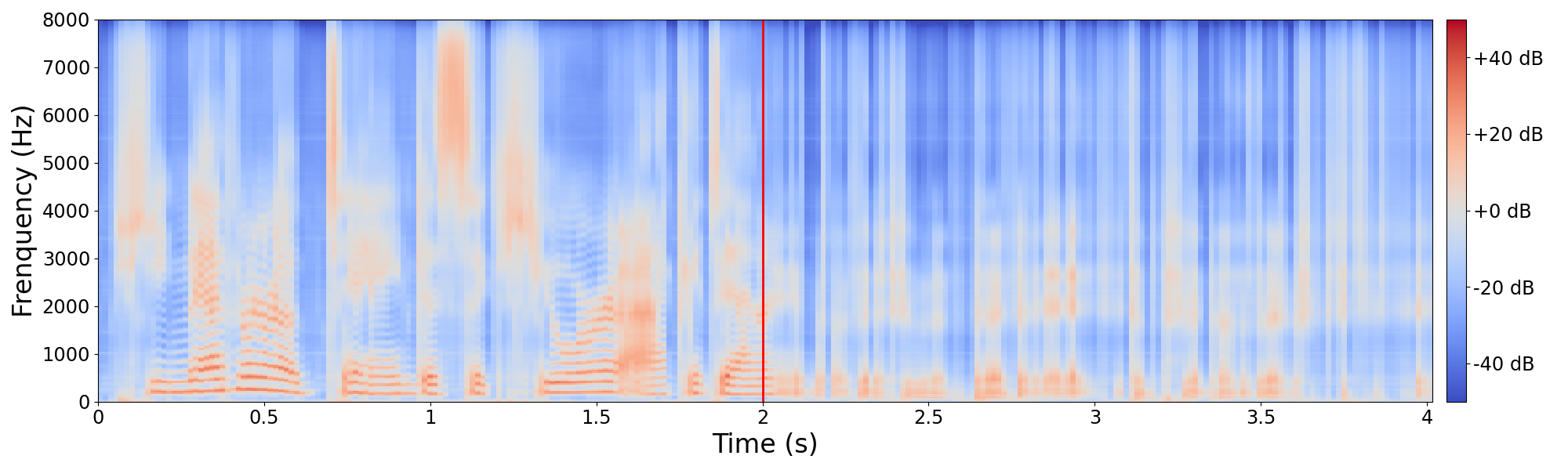}\\
        \hspace{-.3cm}\small{SRNN} \\
        \includegraphics[width=\linewidth]{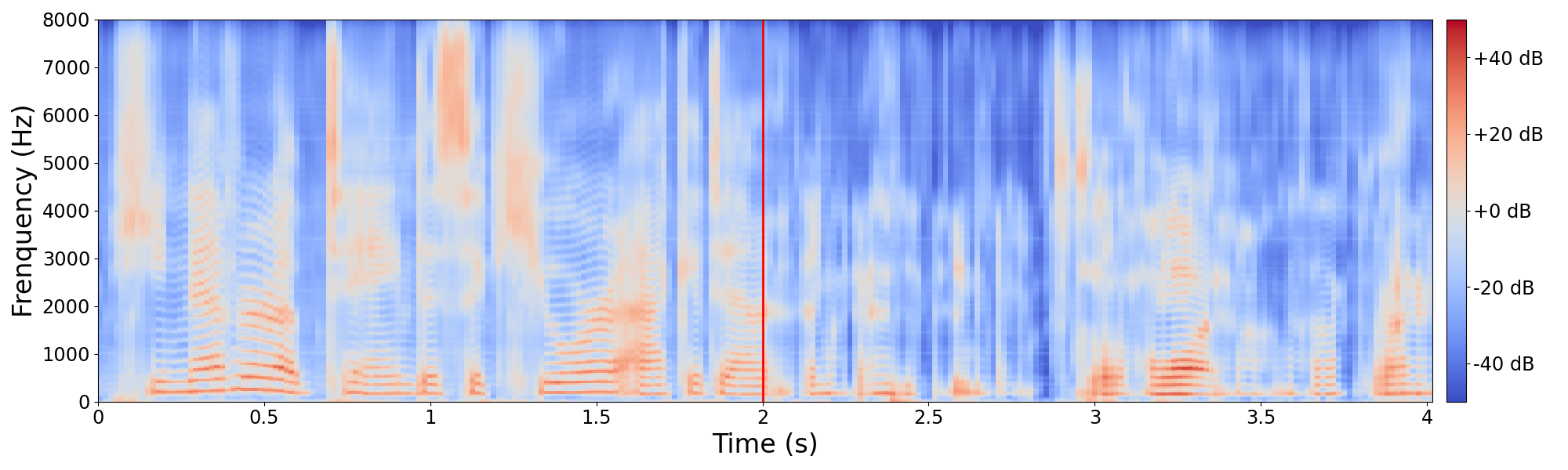}\\
        \hspace{-.3cm}\small{VRNN} \\
        \includegraphics[width=\linewidth]{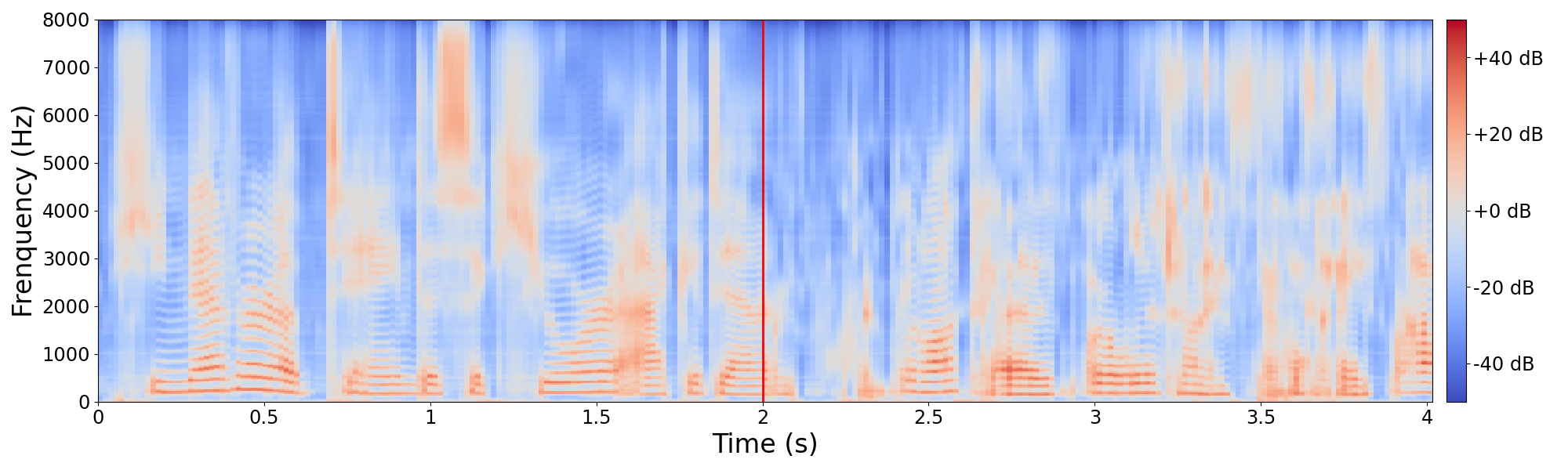} \\
        \end{tabular}
    \caption{Example of speech power spectrogram reconstructed (0-2~s) and generated (2-4~s) by a DVAE model (the original spectrogram is in Figure~\ref{tab:dvae_orig_vae} (top)). Top: STORN; middle: SRNN; bottom: VRNN. The red line indicates the transition between reconstruction and generation. Note that this figure was obtained with the models being trained and evaluated in a slightly different configuration regarding STFT parameters. Here, we have a window length of $512$ points and an overlap of $50\%$. This is to illustrate the robustness of the results w.r.t.~the ``audio parameterization.'' }
    \label{tab:dvae_prediction_auto}
\end{figure}

We first present the results of analysis-resynthesis performed on the speech data. The values of the three metrics described in the previous subsection and averaged over the test dataset are reported in Table~\ref{tab:comparison_speech}. In this experiment, all three autoregressive models (STORN, VRNN, and SRNN) were trained and tested in the teacher-forcing mode (i.e., using the ground-truth values of past observed vectors $\myx[1:t-1]$ when generating $\myx[t]$, see Section~\ref{subsec:TF-vs-GM}). Results of analysis-resynthesis in the generation mode will be discussed in Section~\ref{subsec:SRNN-TF-GM-experiments-speech}.

From Table~\ref{tab:comparison_speech}, we can draw the following conclusions.
First, all tested DVAE models lead to fair signal reconstruction, with an SI-SDR ranging in 6.9--11~dB. This range is in accordance with the fact that we compress each $513$-dimensional data vector into a $16$-latent vector (using also the $128$-dimensional RNN internal state that encodes the past data vectors in the case of autoregressive models). The quality of the reconstructed spectrograms is illustrated in Figures~\ref{tab:dvae_orig_vae}, \ref{tab:dvae_pred_nonauto} and \ref{tab:dvae_prediction_auto}. Figure~\ref{tab:dvae_orig_vae}(top) shows the power spectrogram of a speech signal uttered by a female speaker. Figure~\ref{tab:dvae_orig_vae}(bottom), and Figures~\ref{tab:dvae_pred_nonauto} and \ref{tab:dvae_prediction_auto} show the corresponding spectrogram obtained after analysis-resynthesis by the six tested models. Actually, the first 2~s on the left of the red line are obtained with analysis-resynthesis; the following 2~s are obtained by switching the models into generation mode, as presented in the next subsection. As for the analysis-resynthesis part (the first 2~s), we can see in these plots that the reconstructed spectrograms are all quite close to the original spectrogram. They look like a slightly smoothed or blurred version of the original spectrogram, which is typical of a data compression effect, although retaining most of the speech content characteristics. Regarding the perceptual quality of the reconstructed speech signals, Table~\ref{tab:comparison_speech} shows the PESQ scores that range from fair to good. The STOI scores, generally higher than $0.90$, show good intelligibility. 

Second, all DVAE models outperform the standard VAE model by a large margin (except maybe for STORN, which is ``only'' $1.6$~dB SI-SDR higher). The harmonics in the spectrogram reconstructed with the VAE in Figure~\ref{tab:dvae_orig_vae}(bottom) are slightly noisy and blurrier compared to the harmonics reconstructed with the DVAE models. This demonstrates the merit of including temporal modeling in the VAE framework for modeling sequential data, such as speech signals.
SRNN exhibits the best performance and VRNN comes second. By looking at the associated probabilistic models, we can observe that SRNN and VRNN are the most complex models in terms of dependencies between observed and latent variables. We believe that these dependencies allow SRNN and VRNN to better capture the temporal structure of speech signal than the other models. SRNN performs slightly better than VRNN (e.g., it is $1$~dB SI-SDR above VRNN), although VRNN has richer variable dependencies. This may be because the inference model of SRNN respects the structure of the exact posterior distribution, whereas that of VRNN (as proposed in the original paper and implemented here) does not. In both cases, the exact posterior of $\myz[t]$ at each time $t$ depends on all observations $\myx[1:T]$. However, the inference model of VRNN only takes the causal observations $\myx[1:t]$ into account.

The performance scores of DKF and DSAE are very close to each other, and slightly below those of VRNN. This is an interesting result, as we recall that DKF and DSAE are SSM-like models, with no explicit temporal dependency between $\myx[t-1]$ and $\myx[t]$, but only between $\myz[t-1]$ (or $\myz[1:t-1]$ for DSAE) and $\myz[t]$. Thus, even if they are expected to have less ``predictive'' power compared to SRNN or VRNN, their SSM structure appears quite efficient at encoding the speech dynamics. DSAE can be considered as an improved version of DKF, with an additional sequence-level variable $\myv$ and infinite-order temporal dependency of $\myz[t]$ (as opposed to first-order dependency for DKF). The fact that this more sophisticated structure does not lead to improved performance over DKF might be explained by the structure of the inference model. For DSAE, the inference of $\myz[t]$ depends on $\myx[1:T]$, whereas the exact posterior distribution depends on $\myz[1:t-1]$ and $\myx[t:T]$. Thus, the inference model of DSAE is not only missing some dependencies it should have (previous latent variables) but it is adding dependencies that it should not have (previous observed variables), see Section~\ref{sec:DSAE-inference}. In contrast, the DKF inference model respects the structure of the exact posterior distribution. In the end, all these differences between DKF and DSAE may compensate each other, leading to similar results. One way to improve DSAE may be to design an inference model with the structure of the exact posterior distribution.

The SI-SDR scores obtained by the RVAE model are just below those of DKF and DSAE, whereas the PESQ score of noncausal RVAE is slightly superior to those of DKF and DSAE. This is interesting considering that as DKF and DSAE, RVAE has no predictive link (e.g., no direct dependency between $\myx[t-1]$ and $\myx[t]$), and contrary to DKF and DSAE, RVAE does not have a dynamical model on the latent vector $\myz[t]$ (i.e., it is modeled as an i.i.d.~variable). However, in the analysis-resynthesis framework, the sequence $\myz[1:t-1]$ or $\myz[1:T]$ efficiently encodes $\myx[1:T]$ owing to an efficient inference model, and then, the generative model is able to exploit this whole sequence to regenerate $\myx[t]$. As expected, the noncausal version of RVAE is slightly better than the causal version.

As for STORN, its performance revealed a bit disappointing in our speech analysis-resynthesis experiment. Here also, this can be explained by the fact that the inference model of STORN does not respect the structure of the exact posterior distribution. In particular it does not use $\myz[1:t-1]$ nor $\myx[t+1:T]$, which is doubly penalizing compared to if the structure of the exact posterior distribution was considered. Another reason for the relatively low scores obtained by STORN in these experiments is given in Section~\ref{subsec:visualization-speech}.

\subsection{Generation of speech spectrograms}

In this subsection, we briefly illustrate the ability of the DVAE models to generate ``speech-like'' spectrograms with a qualitative example. This example is the ``continuation'' of the example that we have seen in the previous subsection, provided in Figures~\ref{tab:dvae_orig_vae}, \ref{tab:dvae_pred_nonauto} and \ref{tab:dvae_prediction_auto}. In these figures, the first 2~s of each spectrogram (on the left of the red line) was obtained with analysis-resynthesis; that is, the latent vectors were provided by the inference model, using the ground-truth observed vectors as input, and the output spectrogram was then provided by the generative model using the inferred latent vectors (and the ground-truth past observed vectors for the autoregressive models). After 2~s (on the right of the red line), we turn the models to pure generation mode; that is, the latent vectors and the output spectrogram are now both provided by the generative model, without relying on the inference model and ground-truth past observed vectors anymore. This strategy allows the generation mode to benefit from a good initialization, induced by the analysis-resynthesis part. Indeed, at the time instant corresponding to 2~s, the generation starts with the past latent vectors and current RNN internal state provided by the analysis-resynthesis part, thus encoding the past observed speech data. Therefore, we can expect a smooth transition from the analysized-resynthesized spectrogram to the generated one. 

As expected, we observe in Figure~\ref{tab:dvae_orig_vae} that a vanilla VAE is not able to generate a spectrogram with a realistic speech-like structure. In particular, the successive spectrogram ``chunks'' are too short and with too abrupt transitions to be speech sounds. This is due to the fact that there is no temporal modeling.

Figure~\ref{tab:dvae_pred_nonauto} shows the results obtained with the nonautoregressive DVAE models. We can see that the spectrograms generated by DKF and DSAE, although different, both exhibit a harmonic structure and a variety of different speech-like sounds, which smoothly evolve with time. The smoothness probably comes from the use of a Markov model for the latent vector, which precisely enforces smoothness. In the original DSAE paper, \citet{li2017an} did not provide examples of generated speech spectrograms but they presented good results in voice conversion obtained by exchanging the value of the variable $\myv$ across two sentences spoken by a different speaker. Regarding RVAE, even though the latent vectors are independently and identically sampled from a standard Gaussian distribution, we observe a generated spectrogram with a temporal structure. This means that the RVAE model is able to ``recreate''  correlation in the generated data by combining the vectors of an uncorrelated sequence. However, the energy is mostly concentrated in low frequencies, and the harmonic structure, although present, is not as clearly visible as for DKF and DSAE. Moreover, the segments corresponding to the successive speech sounds are shorter than for DKF and DSE and seem shorter than what is expected in natural speech.

Figure~\ref{tab:dvae_prediction_auto} shows the generation results with autoregressive DVAE models. We see that, as the other models, STORN manages to ensure a smooth transition between the analysis-resynthesis and generation parts, but then the quality of the generated spectrogram becomes much lower than with the other autoregressive DVAE models. SRNN generates a spectrogram with a speech-like structure and a lot of variability. The best result for this example sentence is obtained with VRNN, for which we can observe segments that resemble different phonemes, with smooth transitions between them. The harmonic structure is also clearly visible and the generated data cover the full bandwidth.

\subsection{Training with scheduled sampling}
\label{subsec:SRNN-TF-GM-experiments-speech}

To complement the previous results, we performed an additional analysis-resynthesis experiment with the autoregressive models being trained and tested in the generation mode instead of the teacher-forcing mode; that is, using the previously generated data vectors $\myxh[1:t-1]$ in place of the ground-truth past vectors $\myx[1:t-1]$ when generating $\myx[t]$ (see Section~\ref{subsec:TF-vs-GM}). Note that here, the sequence $\myz[1:T]$ is still provided by the encoder. For conciseness, we present only the results for SRNN, which was the model performing best in the first experiment above. 

Directly training a model in the generation mode was observed to be difficult, so we adopted a \textit{scheduled sampling} approach \citep{bengio2015scheduled}. We started with the SRNN model trained in the teacher-forcing mode used in the previous experiment. 
Then, we fine-tuned this model by randomly replacing $\myx[1:t-1]$ with $\myxh[1:t-1]$ at the input of the encoder-decoder shared module, when estimating $\myx[t]$. We replaced 20\% of the $\myx[1:t-1]$ vectors for the first 50 epochs, and increased to 40\% for the next 50 epochs, and so on, until we completely replaced the ground-truth clean speech signal with generated speech signals. Then we fine-tuned with totally generated speech signals for another 300 epochs. Overall, we fine-tuned the model for 500 epochs. The resulting model is referred to as SRNN-GM hereinafter. Moreover, we also evaluated the initial SRNN model (trained with teacher-forcing) in the generation mode (i.e., we use here $\myx[1:t-1]$ during training and $\myxh[1:t-1]$ during testing when generating $\myx[t]$). We refer to this ``hybrid'' configuration as SRNN-TF-GM. 

We can see from Table~\ref{tab:comparison_speech} that SRNN trained in the teacher-forcing mode and tested in the generation mode (SRNN-TF-GM) obtains very poor results. This illustrates the problem of train/test mismatch discussed in Section~\ref{subsec:train-test-match}. The strategy that consists in training SRNN in the generation mode using scheduled sampling (SRNN-GM) is shown to be effective, as the gap between SRNN-TF-GM and SRNN is largely reduced. Nevertheless, SRNN-GM remains a bit below DKF and RVAE in this experiment, showing that it is more difficult to exploit the predictive links in a practical application where the ground-truth values of the observed vectors are not available, compared to the ``oracle'' configuration of teacher-forcing.

\subsection{Visualization of the latent vector sequence}
\label{subsec:visualization-speech}

\begin{figure}[!ht]
    \centering
    \begin{tabular}{cc}
    \includegraphics[width=0.48\linewidth]{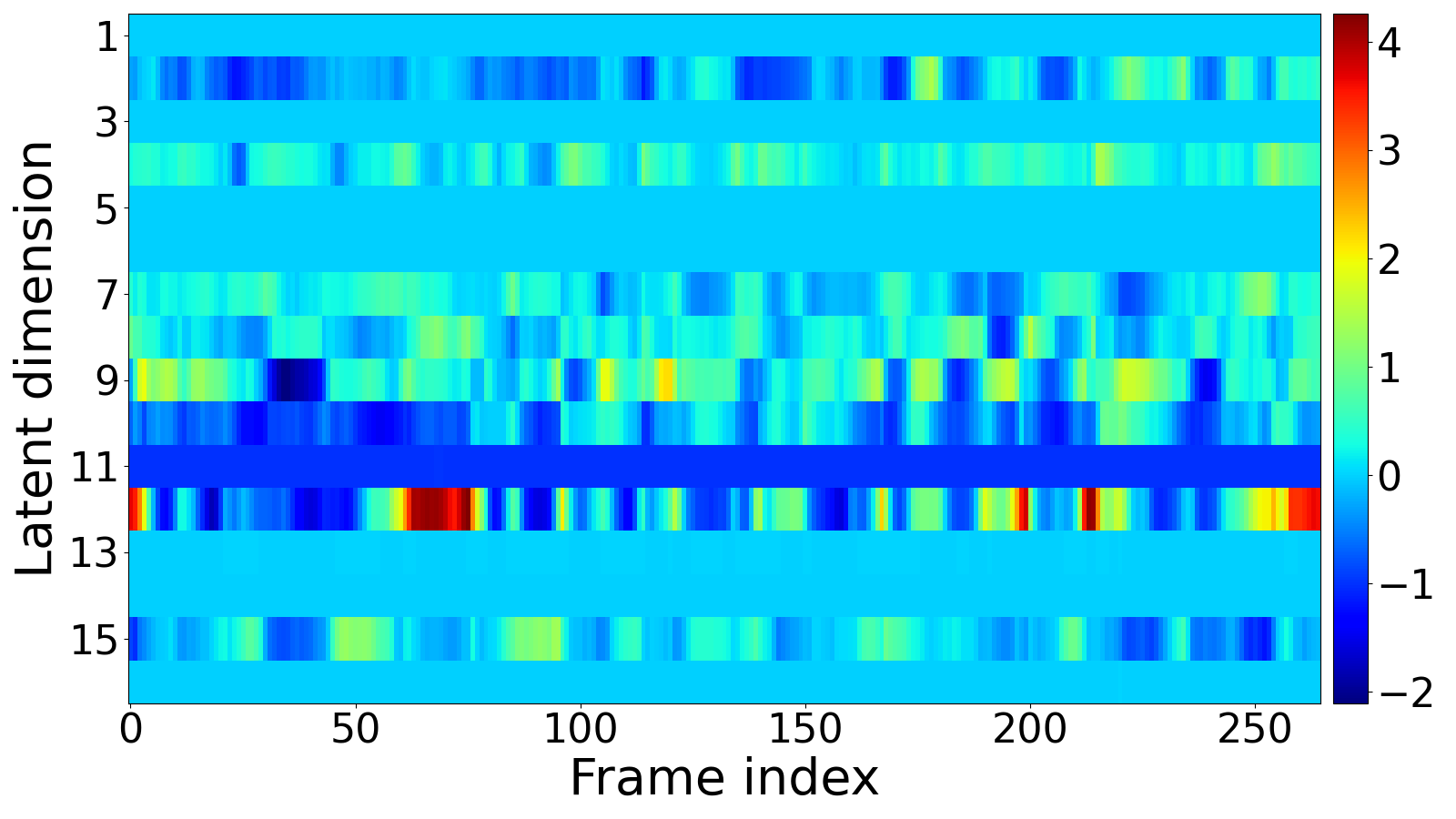} &
    \includegraphics[width=0.48\linewidth]{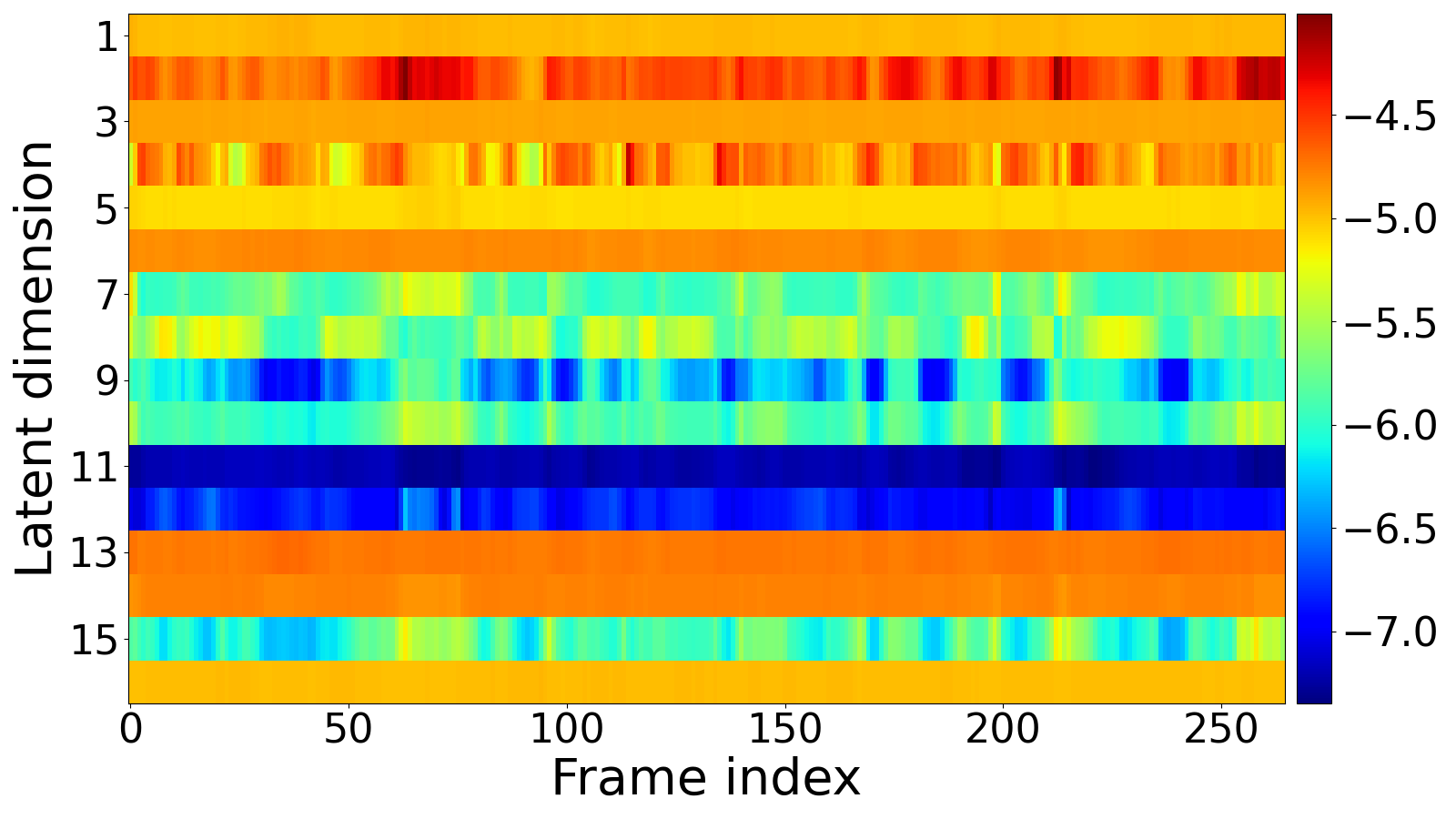} \\
    {(a) mean} & {(b) log-variance}\\
    \includegraphics[width=0.48\linewidth]{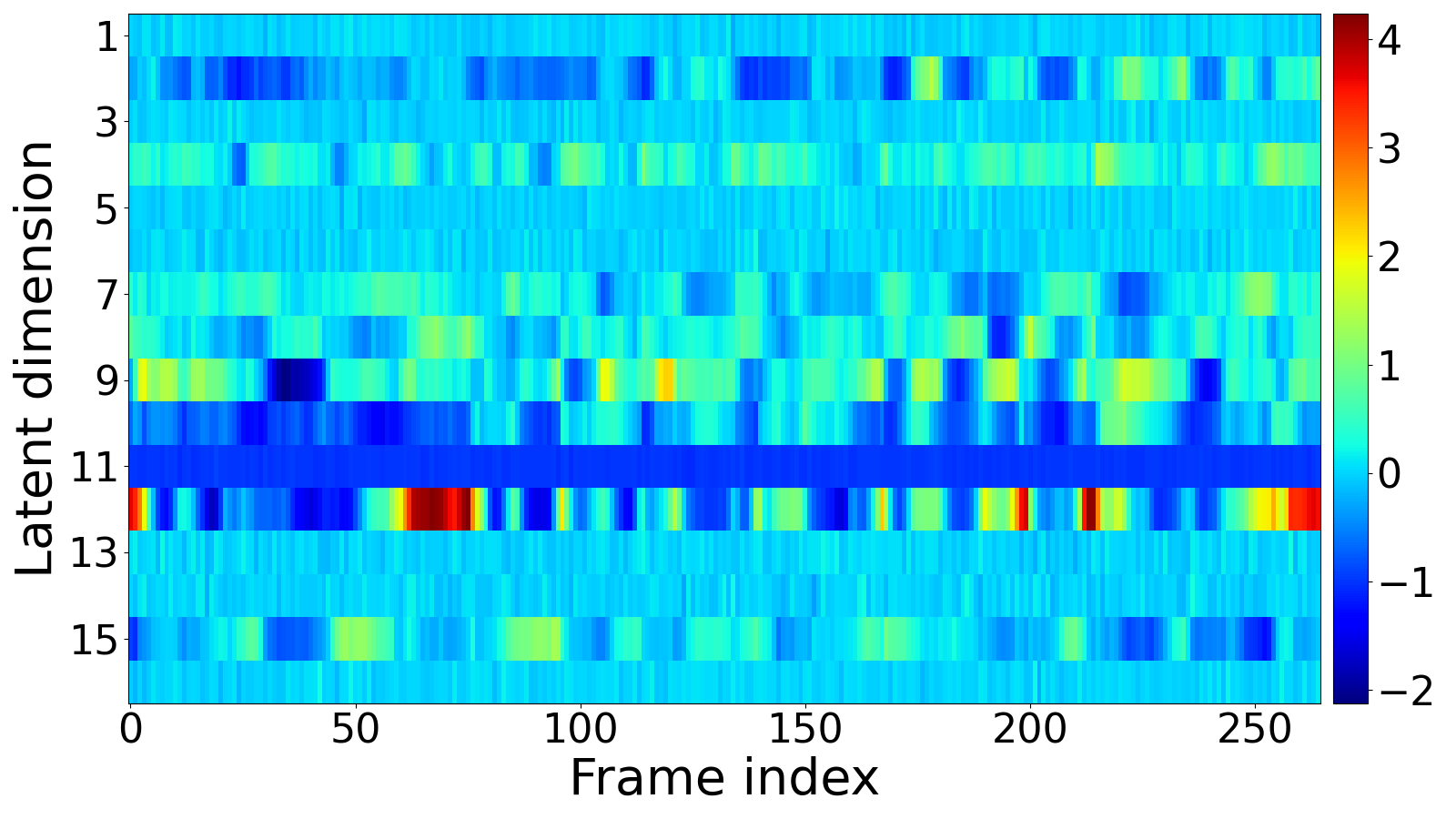} &
    \includegraphics[width=0.48\linewidth]{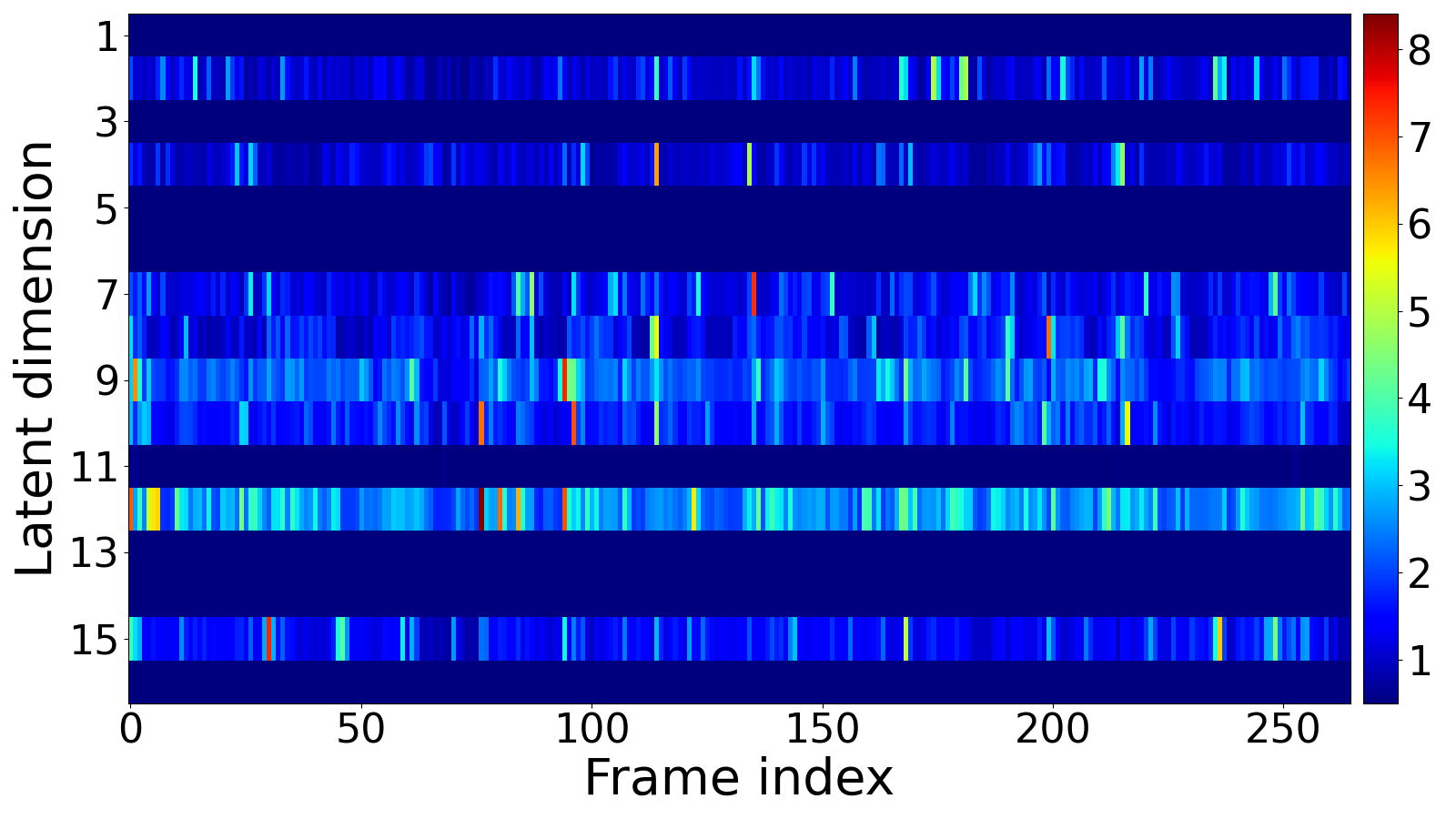} \\
    {(c) sampled latent vector} & {(d) KL divergence}
    \end{tabular}
    \caption{Example of the behavior of the latent vector of a speech spectrogram for SRNN (trained with teacher-forcing). (a) mean value of the posterior distribution (i.e., $\boldsymbol{\mu}_{\phi_\mathbf{z}}(\cdot)$); (b) log-variance of the posterior distribution (i.e., $\boldsymbol{\sigma}^2_{\phi_\mathbf{z}}(\cdot)$); (c) sampled latent vector  $\myz[t]$; (d) KL divergence term of the VLB. }
    \label{fig:vis-z-speech}
\end{figure}

In this subsection, we illustrate the behavior of the learned latent vector. Figure~\ref{fig:vis-z-speech} displays the trajectories of the mean vector of $q_{\phi_{\myz}}$, the log-variance vector of $q_{\phi_{\myz}}$, a vector $\myz[t]$ sampled from $q_{\phi_{\myz}}$, and the KL divergence term of the VLB, for an example sentence of the speech test dataset and for the SRNN model trained and applied in the teacher-forcing mode. We observe that some of the dimensions of the latent vector, for example the first one, show a steady profile of the mean (Figure~\ref{fig:vis-z-speech}~(a)) and log-variance (Figure~\ref{fig:vis-z-speech}~(b)) for these dimensions, with the mean being close to zero and the variance being much lower than 1. As a result, the corresponding entries of the $\myz[t]$ samples shown in~Figure~\ref{fig:vis-z-speech}~(c) look like noise with small fluctuations around zero. In short, those dimensions are noninformative, whereas the other ``active'' dimensions show much larger, and thus informative, fluctuations (note that for these ``active'' dimensions, because the variance is also small, yet not steady, the sampled latent values are close to the mean). This inactivity of certain dimensions is the sign of the \textit{posterior collapse} problem. The latent vector $\myz[t]$, or at least some dimensions of it, becomes noninformative, as its posterior distribution becomes too close to its prior (generative) distribution, as illustrated by the small (and steady) values of the KL divergence term of the VLB, compared to the other ``active'' dimensions (Figure~\ref{fig:vis-z-speech}~(d)). This problem has been well identified and largely discussed in the VAE literature \citep{bowman2015generating, serban2016piecewise,chen2017variational,lucas2019don, razavi2019preventing, dai2020usual}. It remains a largely open topic in the framework of DVAE, as further discussed in Chapter~14.

Overall, the input of the decoder consists of some dimensions containing informative patterns and some other dimensions containing low-variance stationary noise. In the present example of the SRNN model, we identified $8$ dimensions out of $16$ that are active and $8$ dimensions that seem to have collapsed, suggesting that we would obtain similar analysis-resynthesis performance by setting the size of the latent vector to $8$. In these experiments with speech signals, we observed that VRNN has only $2$ collapsed dimensions out of $16$, whereas STORN had $12$ collapsed dimensions out of $16$; that is, only $4$ active dimensions for STORN. This latter point is to be related to the fact that STORN is the less efficient of the three autoregressive models in these experiments, with performance significantly below that of SRNN and VRNN. In contrast, we did not observe posterior collapse of any dimension for the nonautoregressive models (DKF, DSAE, and RVAE) in our experiments; that is, for the nonautoregressive models, all $16$ dimensions are useful to encode $\myx[t]$. This is consistent with the fact that, for these models, there is no predictive link and all the information in $\myx[1:T]$ must be encoded in $\myz[1:T]$.

\begin{figure}[!ht]
    \centering
    \begin{tabular}{cc}
    \includegraphics[width=0.48\linewidth]{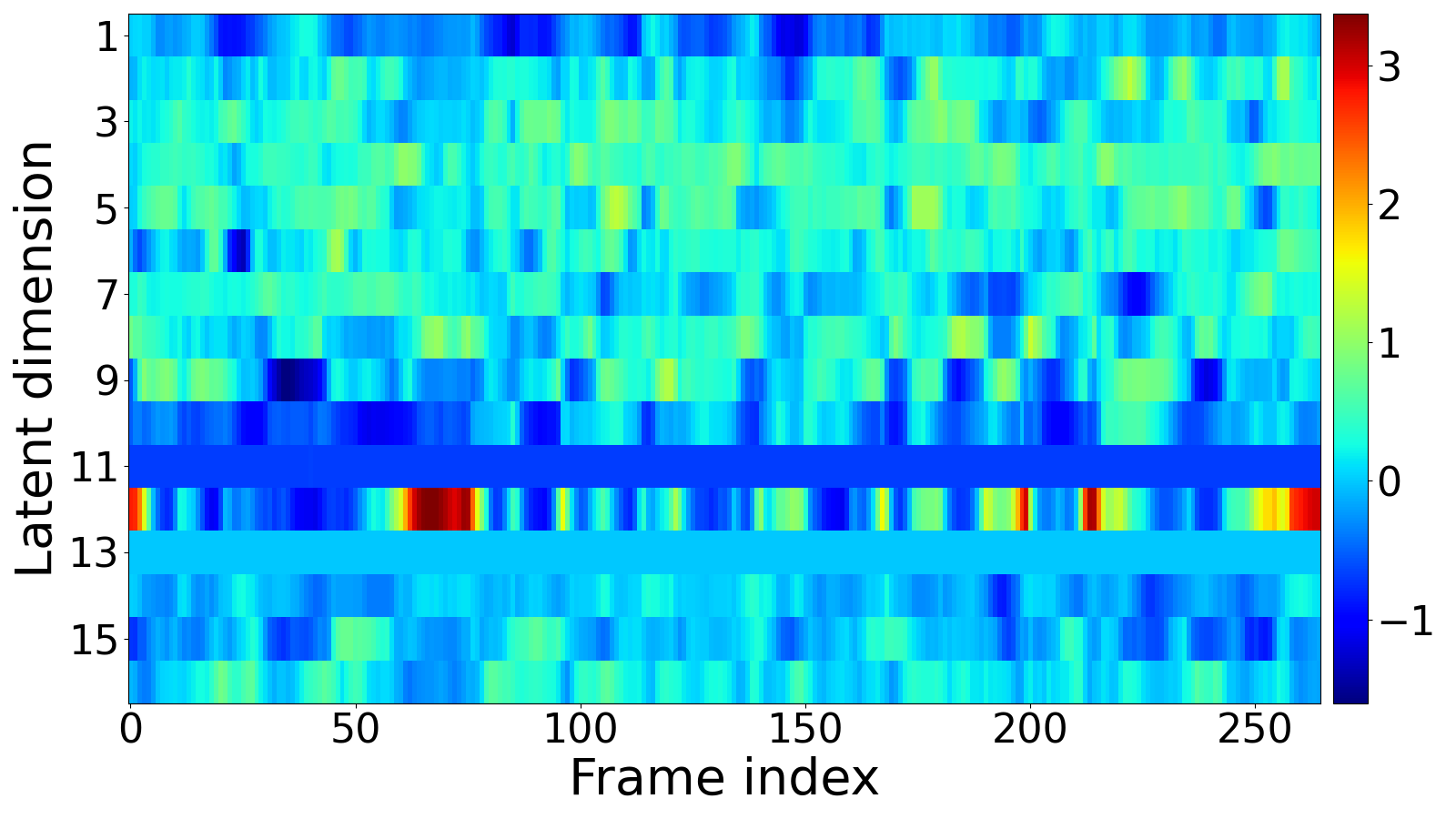} &
    \includegraphics[width=0.48\linewidth]{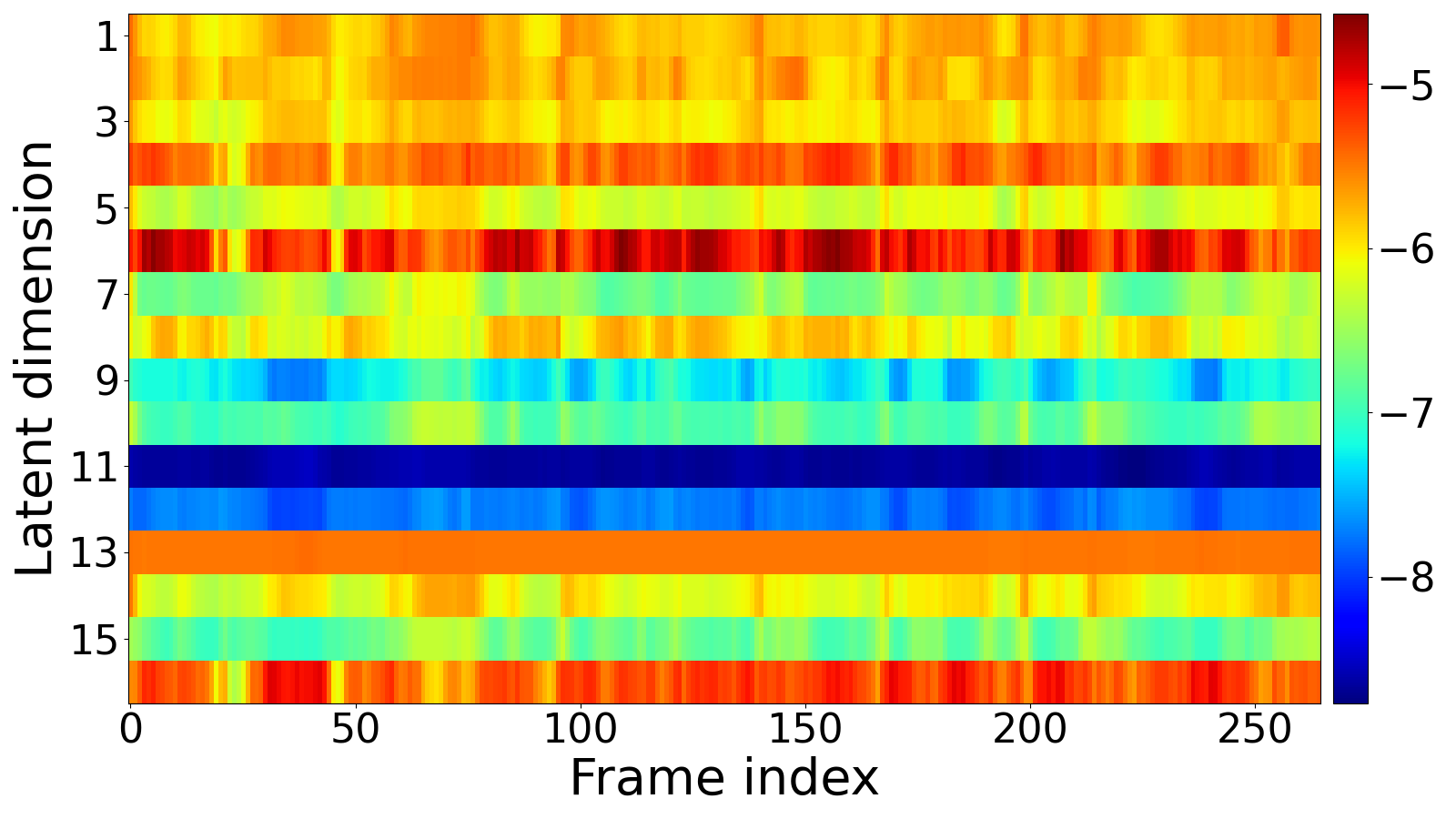} \\
    {(a) mean} & {(b) log-variance}\\
    \includegraphics[width=0.48\linewidth]{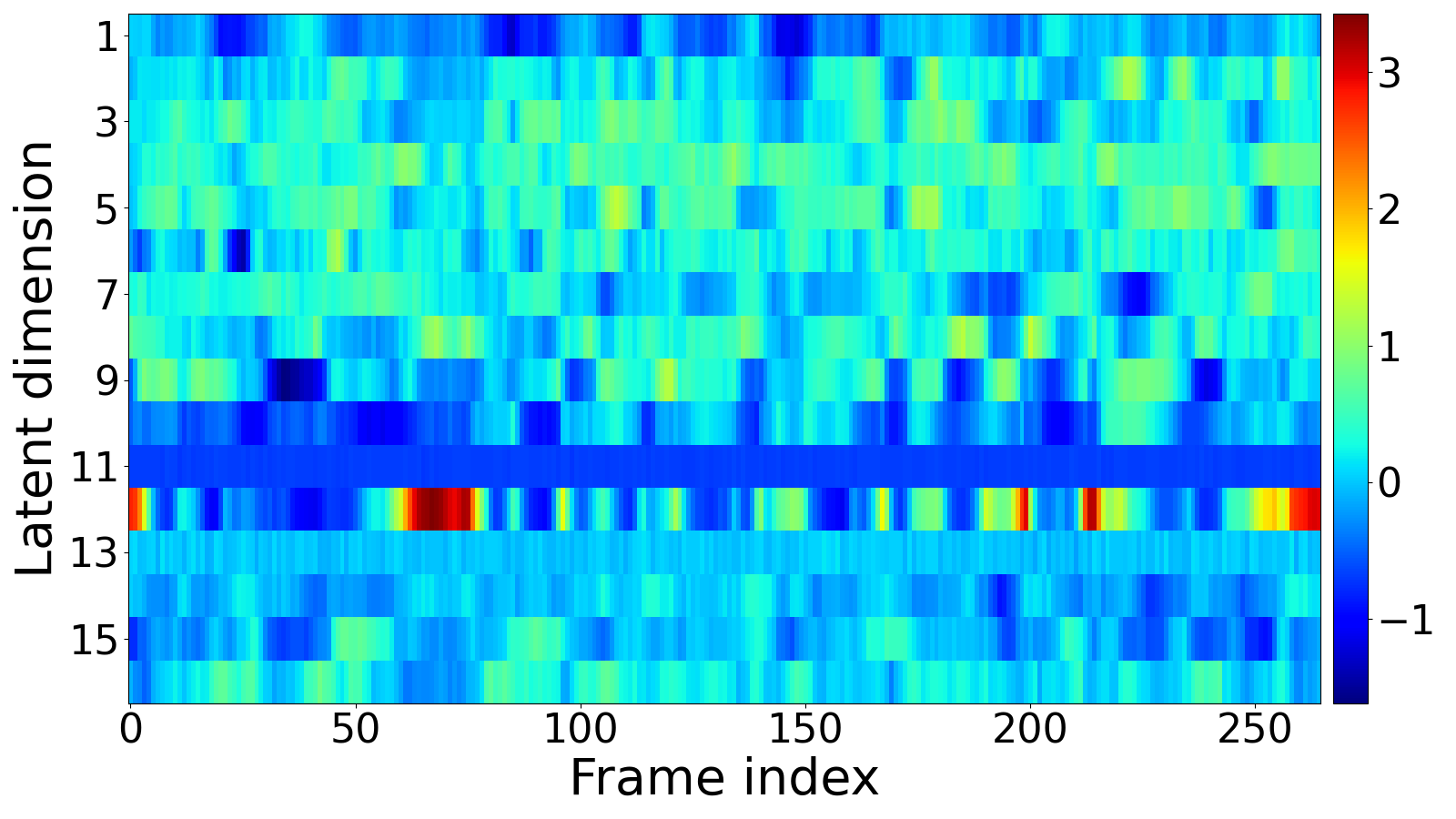} &
    \includegraphics[width=0.48\linewidth]{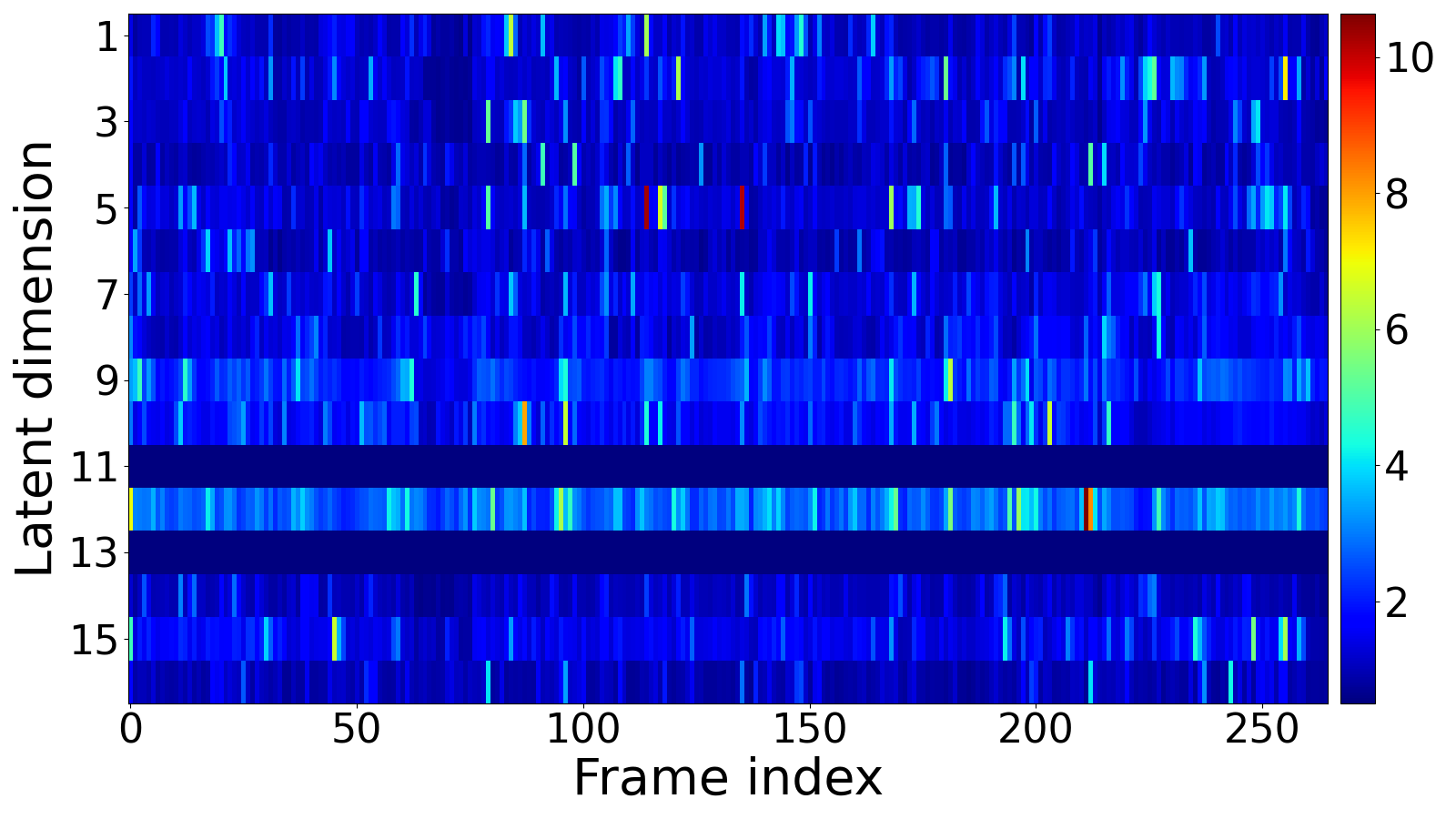} \\
    {(c) sampled latent vector} & {(d) KL divergence}
    \end{tabular}
    \caption{Example of the behavior of the latent vector of a speech spectrogram for SRNN (trained in the generation mode with scheduled sampling). (a) mean value of the posterior distribution (i.e., $\boldsymbol{\mu}_{\phi_\mathbf{z}}(\cdot)$); (b) log-variance of the posterior distribution (i.e., $\boldsymbol{\sigma}^2_{\phi_\mathbf{z}}(\cdot)$); (c) sampled latent vector  $\myz[t]$; (d) KL divergence term of the VLB.}
    \label{fig:vis-z-ss-speech}
\end{figure}

We have checked that, for a given model, the active/inactive dimensions remain the same across different examples. In addition, as shown in the example of Figure~\ref{fig:vis-z-speech}, the dimensions suffering from posterior collapse are the same over time. In principle, a small KL divergence at a certain time frame means that the posterior and the generative distributions of $\myz[t]$ are very close to each other for that time frame. However, this does not guarantee that the generative distribution on two consecutive frames remains the same, and the same for the posterior distribution. In practice, we observe that both the generative and posterior distributions of collapsed dimensions are time-invariant and noninformative. We believe this is due to the way these models are implemented in practice, since the architectures used to implement the DVAEs are time-invariant; that is, the same architecture with the same weights is used at every time step. Combined with the fact that noninformative generative distributions are zero-mean and low-variance Gaussians, this leads us to conjecture that posterior collapse in a given dimension of $\myz$ is associated to very small weights to compute the mean and log-variance of that dimension. We have verified this statement by visualising the weights of the output linear layers (not shown here for conciseness).

Figure~\ref{fig:vis-z-ss-speech} displays the same trajectories as Figure~\ref{fig:vis-z-speech}, for the same example sentence, but for the SRNN model trained in the generation mode (with scheduled sampling) instead of the teacher-forcing mode. By comparing the two figures, we observe that the SRNN model trained in the generation mode exhibits much less posterior collapse, with a larger number of ``active'' dimensions. 
By definition, the generated data vectors are approximate values of the ground-truth vectors, hence the model trained in the generation mode uses less accurate and thus less reliable past values of $\myx[t]$ to generate the current value, compared to the same model trained in the teacher-forcing mode. Therefore, the model trained in the generation mode needs more informative latent dimensions to resynthesize $\myx[t]$.

\section{Results on 3D human motion data}
\label{sec:results-motion}

\subsection{Analysis-resynthesis}
\label{subsec:analysis-resynthesis-motion}

\begin{table} [t]
\center
\tabcolsep=0.11cm
    \begin{tabular}{l c}
    \toprule 
    DVAE & MPJPE ($mm$) \\
    \midrule
    VAE             & 48.69 \\
    DKF             & 42.21 \\ 
    STORN           & 9.47  \\ 
    VRNN            & 9.22  \\ 
    SRNN            & 7.86  \\ 
    RVAE-Causal     & 31.09 \\ 
    RVAE-NonCausal  & 28.59\\ 
    DSAE            & 28.61 \\ 
    \midrule
    SRNN-TF-GM      & 221.87 \\
    SRNN-GM         & 43.98 \\
    \bottomrule
    \end{tabular}
\caption{Performance of the DVAE models tested on 3D human motion data analysis-resynthesis. The MPJPE scores are averaged over the test subset of the H3.6M dataset.}
\label{tab:comparison_motion}
\end{table}

Table~\ref{tab:comparison_motion} shows the results of the analysis-resynthesis experiment with the 3D human motion data. The MPJPE  values are approximately within $9$--$49$mm, which is relatively small compared to the average amplitude of the joint coordinates in a human body, and therefore show a fair to good reconstruction for all models. As for the speech analysis-resynthesis experiment, all DVAE models outperform the vanilla VAE model. This confirms the interest of using DVAE models for modeling sequential data.
    
In this experiment with human motion data, the autoregressive DVAEs largely outperform the nonautoregressive DVAEs. STORN, VRNN, and SRNN have an MPJPE of about 9.5, 9.2 and 7.9~mm, respectively, whereas DKF, RVAE and DSAE (noncausal) obtain about 42.2, 28.6 and 28.6~mm, respectively. Therefore, the performance gap between the autoregressive models (trained and tested in the teacher-forcing mode) and the nonautoregressive models is larger than in our experiment with speech signals. We conjecture that this is because the 3D human motion data has less variability (or, say, smoother trajectories) compared to speech data. Therefore, for such data, knowing the ground-truth values of the previous observation(s) ($\myx[t-1]$ or $\myx[1:t-1]$) is a very strong information for predicting the current observation. 
    
Again, SRNN exhibits the best performance, which is consistent with the analysis-resynthesis results obtained with the speech data. In this new experiment with human motion data, STORN is more efficient than in our experiment with speech data. In contrast, DKF underperforms compared to the other nonautoregressive models and exhibits a quite limited improvement over the vanilla VAE.

\subsection{Generation of 3D human motion sequences}

Example videos of a human ``skeleton'' animated from 3D  motion data sequences generated by the different DVAE models are available at \url{https://team.inria.fr/robotlearn/dvae/}.

\subsection{Training with scheduled sampling}
\label{subsec:SRNN-TF-GM-experiments-motion}

As for the experiment with speech signals, we have tested the influence of training and testing the models in the generation mode (using scheduled sampling for training). Here, we briefly report and comment the results obtained on the 3D human motion data with SRNN. We can see from Table~\ref{tab:comparison_motion} that, similarly to what we observed in our experiment with speech signals, SRNN-TF-GM has very poor performance, whereas training SRNN with scheduled sampling partially addresses this problem, placing SRNN-GM between the vanilla VAE and the nonautoregressive DVAEs. However, the gain in performance of SRNN-GM over the vanilla VAE is here quite limited, and we believe there is room for improvement when designing the model adaptation method. In other words, in these experiments, we adopted a simple scheduled sampling strategy and did not further investigated this issue, but other strategies to bridge the gap between ground-truth and generated data could be investigated. 

\subsection{Visualization of the latent vector sequence}
\label{subsec:visualization-motion}

To conclude this set of experiments, we also provide an example of visualization of the latent space of the 3D human data, as we did for speech signals.

\begin{figure}[!ht]
    \centering
    \begin{tabular}{cc}
    \includegraphics[width=0.48\linewidth]{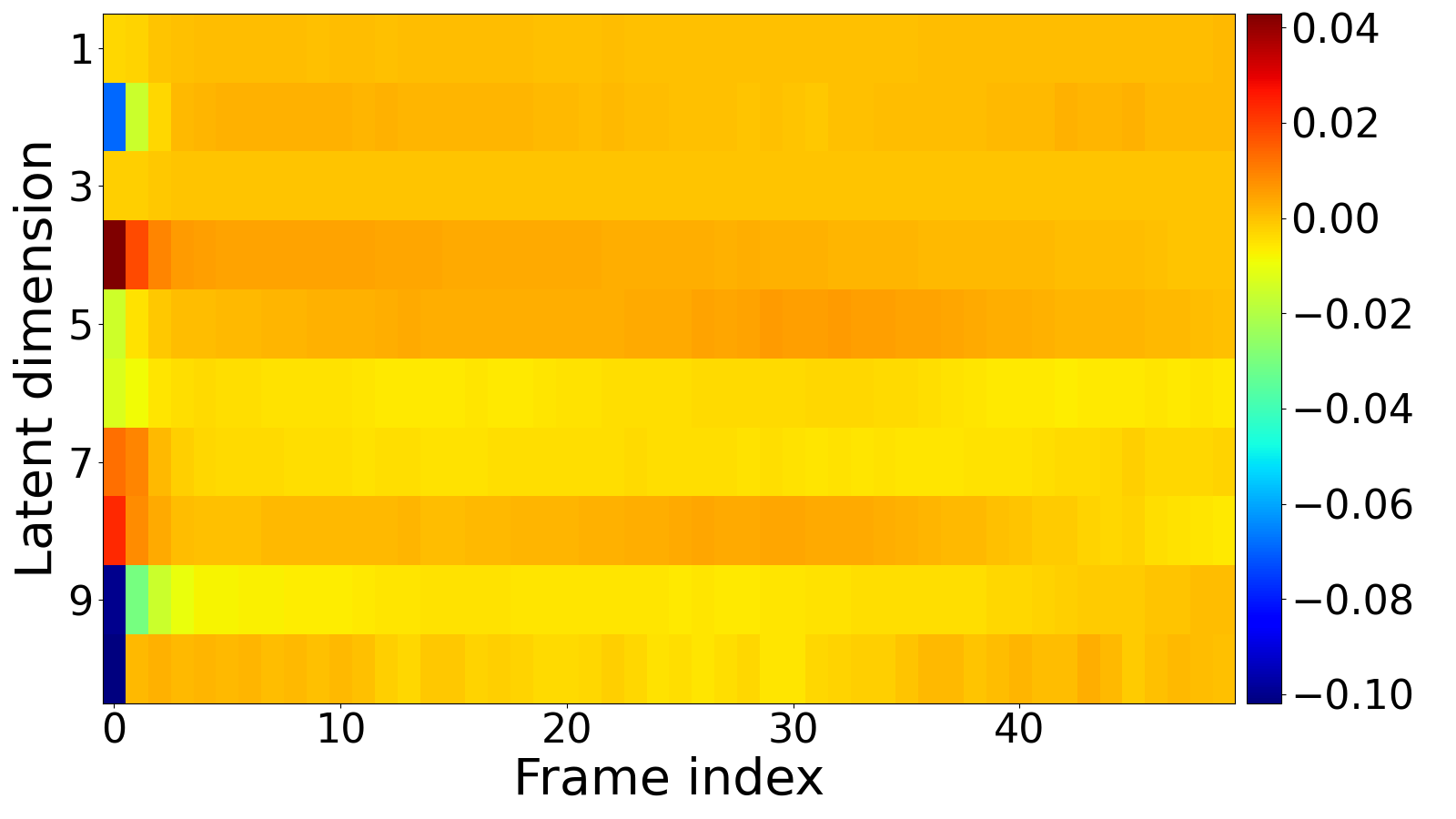} &
    \includegraphics[width=0.48\linewidth]{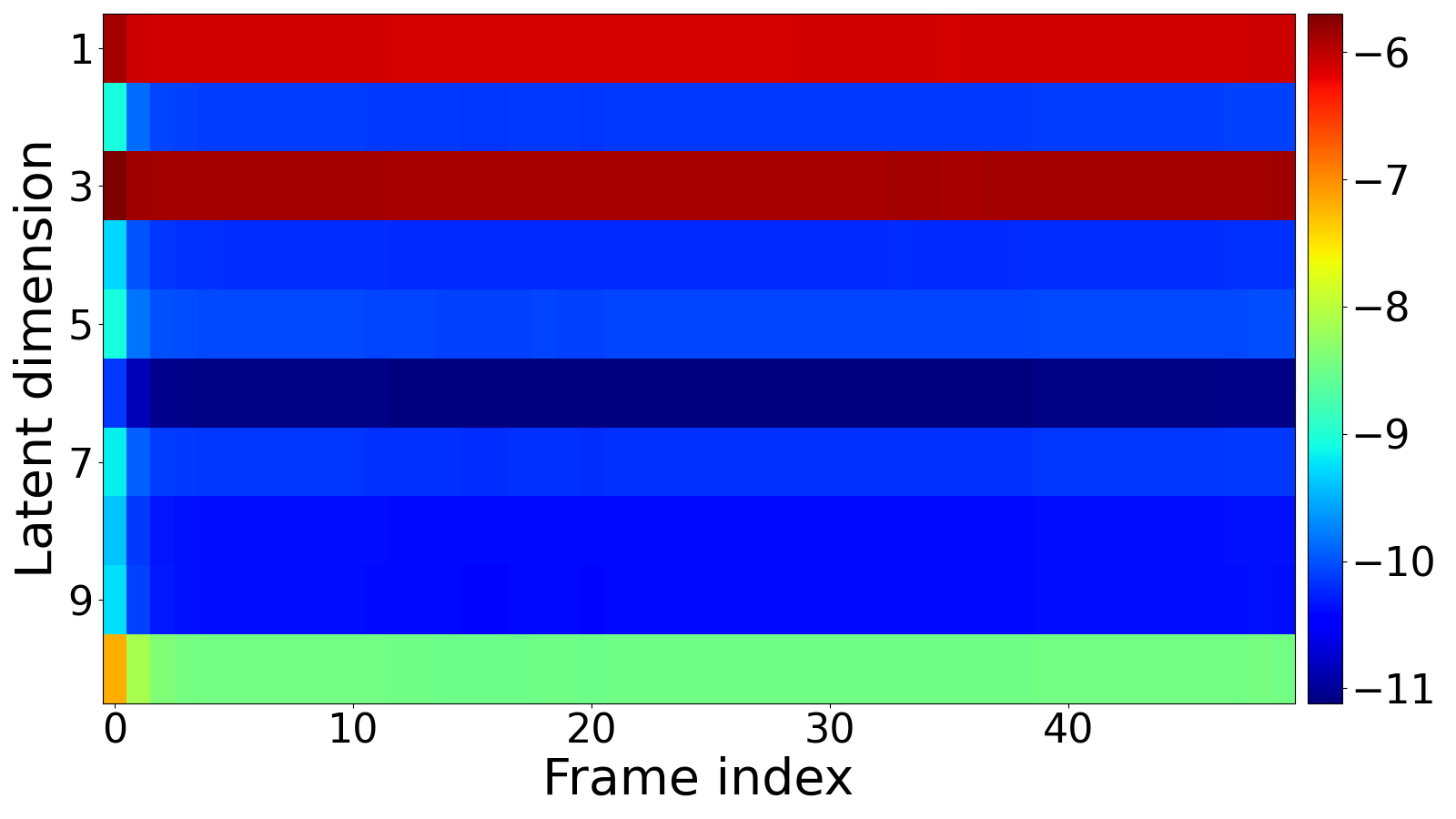} \\
    {(a) mean} & {(b) log-variance}\\
    \includegraphics[width=0.48\linewidth]{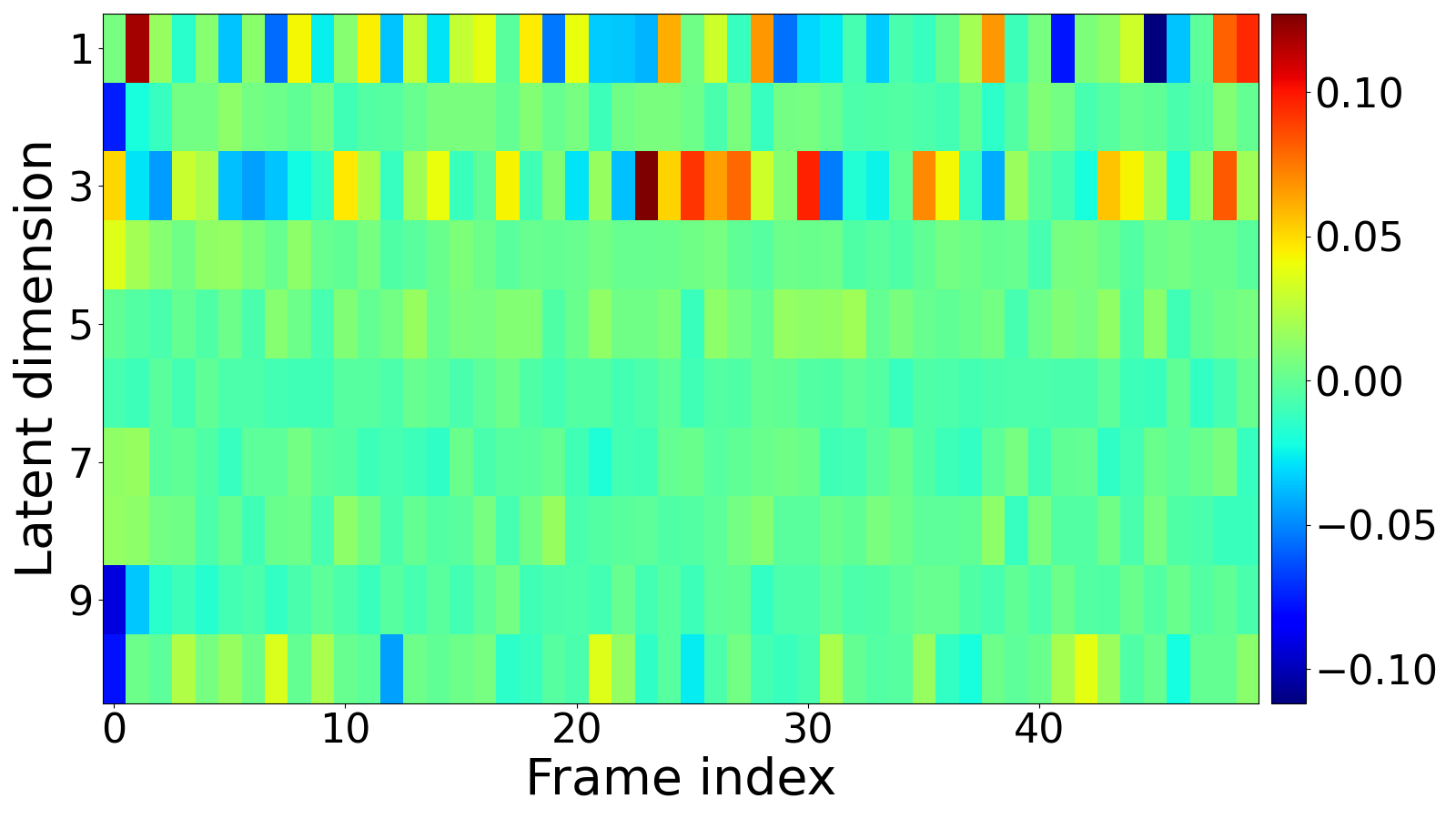} &
    \includegraphics[width=0.48\linewidth]{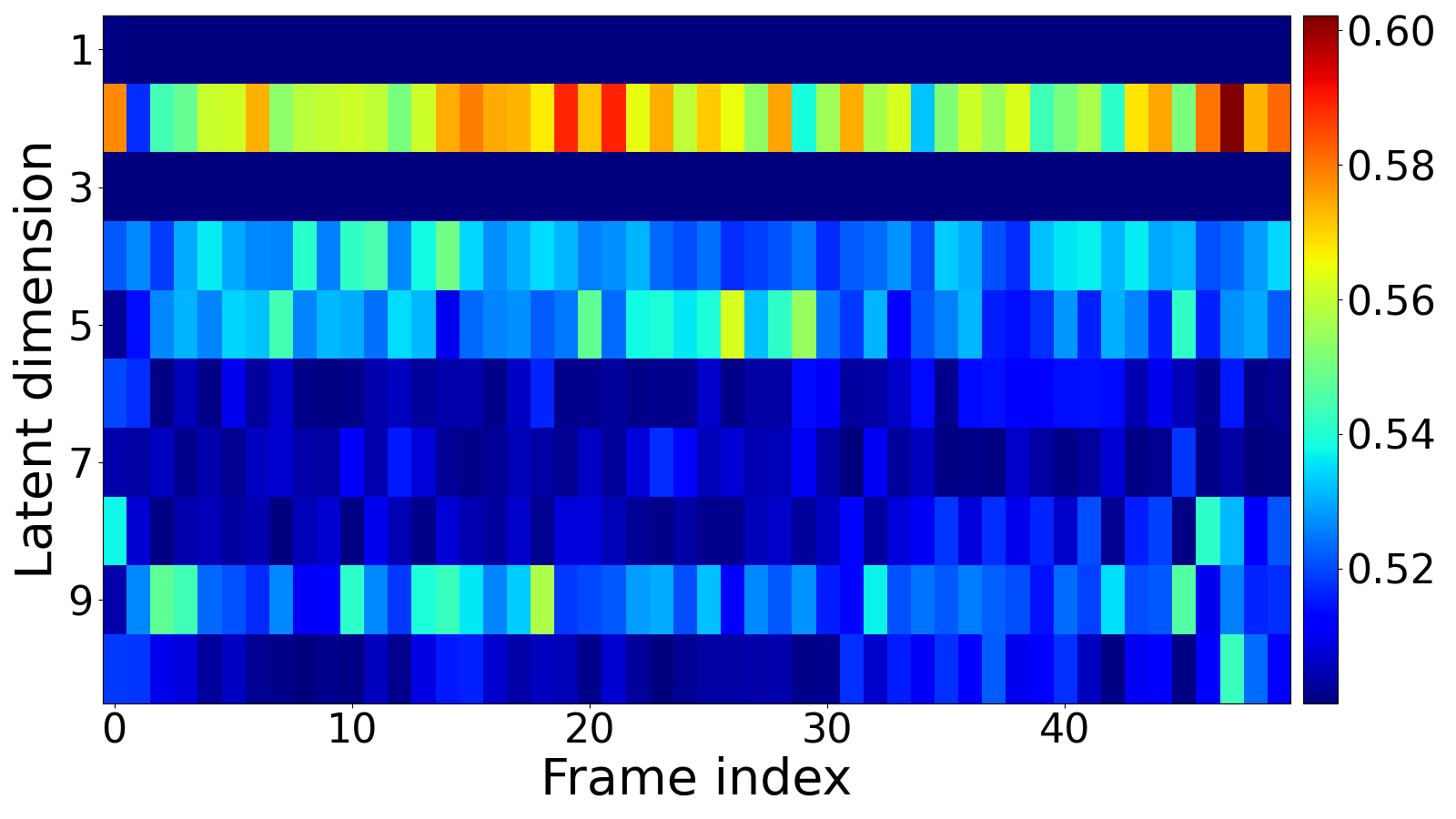} \\
    {(c) sampled latent vector} & {(d) KL divergence}
    \end{tabular}
    \caption{Example of the behavior of the latent vector for a test sequence from the H3.6M dataset and for SRNN (trained with teacher-forcing). (a) mean value of the posterior distribution (i.e., $\boldsymbol{\mu}_{\phi_\mathbf{z}}(\cdot)$); (b) log-variance of the posterior distribution (i.e., $\boldsymbol{\sigma}^2_{\phi_\mathbf{z}}(\cdot)$); (c) sampled latent vector  $\myz[t]$; (d) KL divergence term of the VLB. }
    \label{fig:vis-z-motion}
\end{figure}

\begin{figure}[!ht]
    \centering
    \begin{tabular}{cc}
    \includegraphics[width=0.48\linewidth]{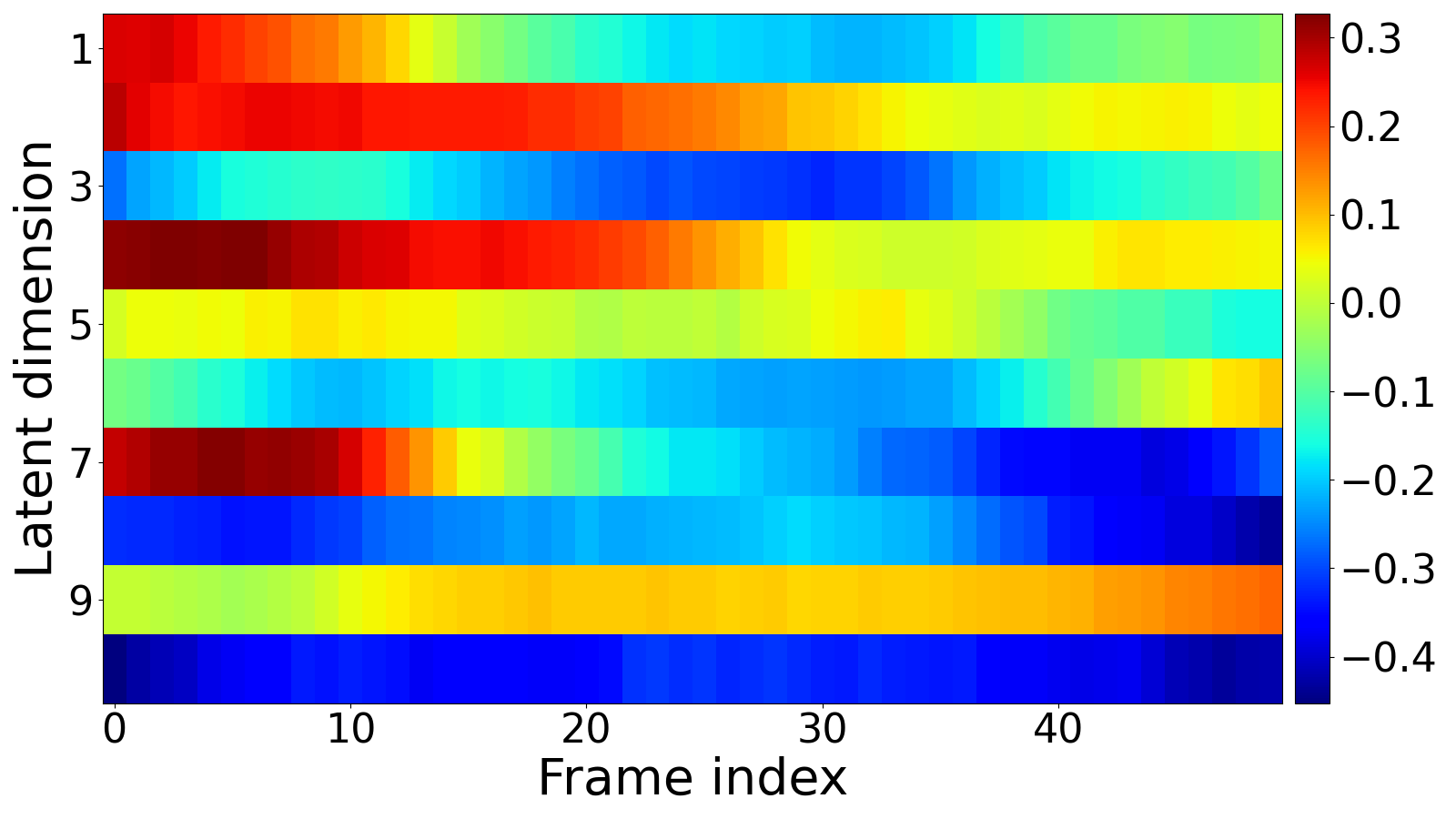} &
    \includegraphics[width=0.48\linewidth]{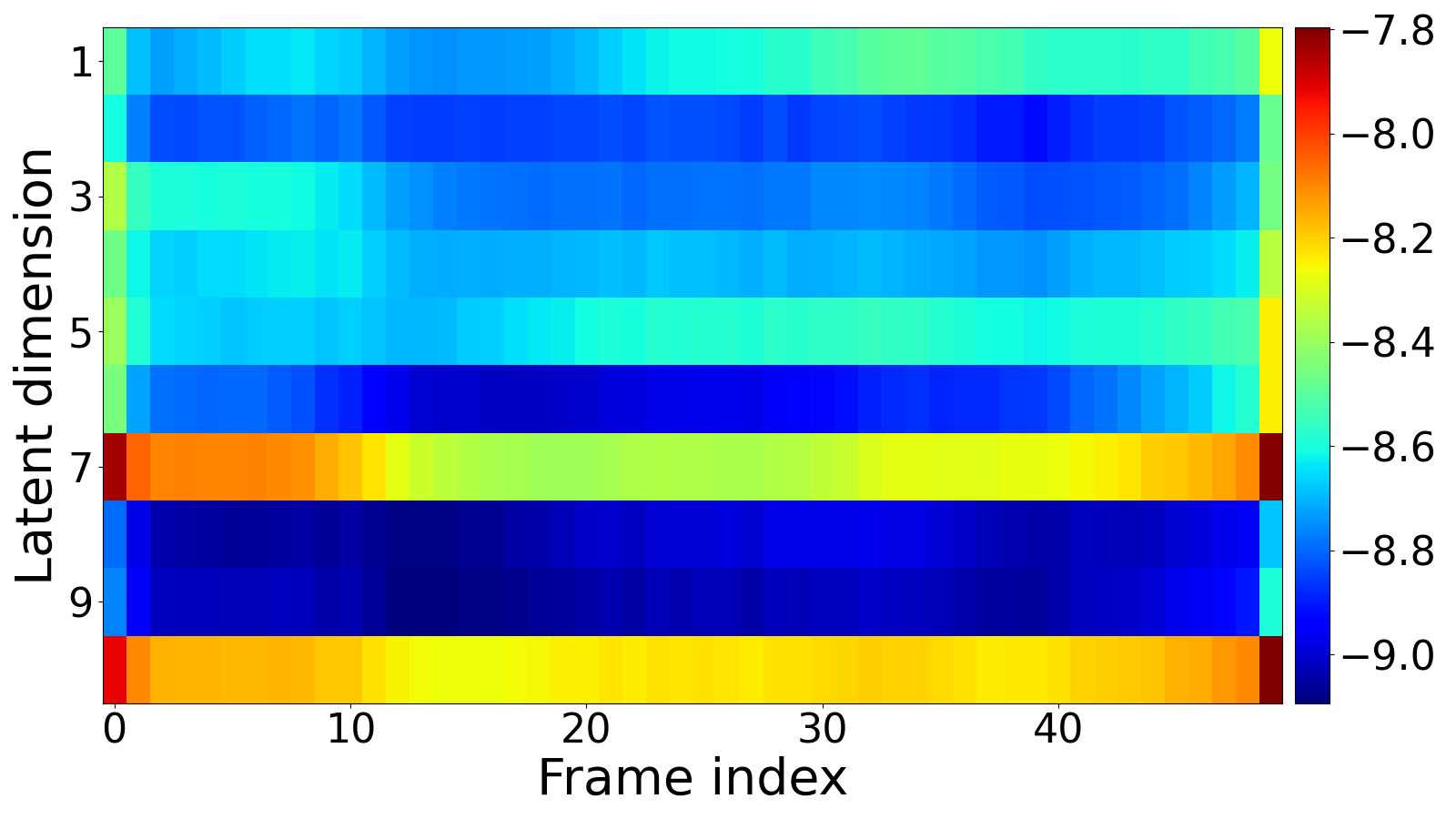} \\
    {(a) mean} & {(b) log-variance}\\
    \includegraphics[width=0.48\linewidth]{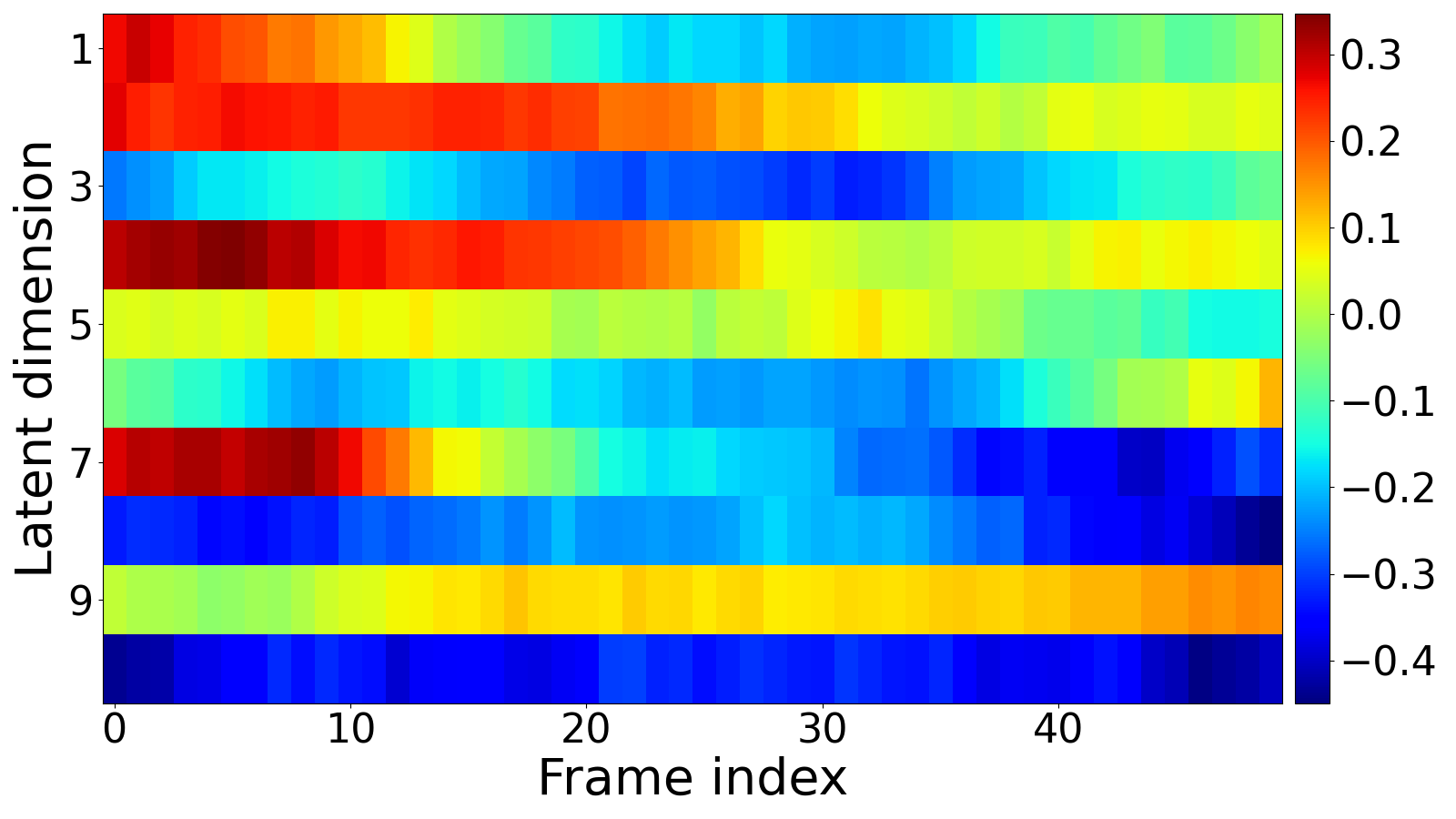} &
    \includegraphics[width=0.48\linewidth]{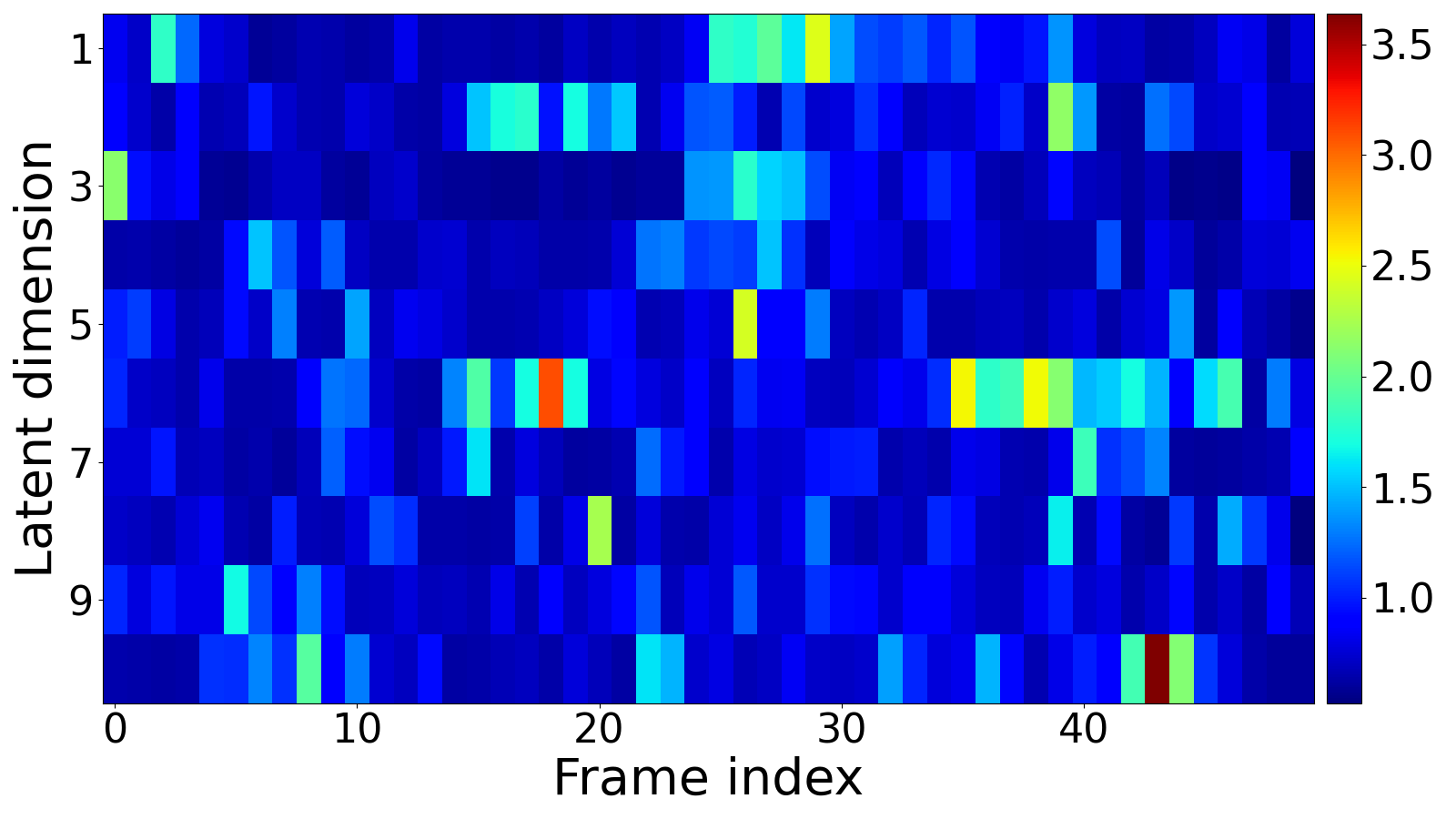} \\
    {(c) sampled latent vector} & {(d) KL divergence}
    \end{tabular}
    \caption{Example of the behavior of the latent vector for a test sequence from the H3.6M dataset and for SRNN (trained in generation mode with scheduled sampling). (a) mean value of the posterior distribution (i.e., $\boldsymbol{\mu}_{\phi_\mathbf{z}}(\cdot)$); (b) log-variance of the posterior distribution (i.e., $\boldsymbol{\sigma}^2_{\phi_\mathbf{z}}(\cdot)$); (c) sampled latent vector  $\myz[t]$; (d) KL divergence term of the VLB.}
    \label{fig:vis-z-ss-motion}
\end{figure}

Figure~\ref{fig:vis-z-motion} displays the trajectories of the mean vector of $q_{\phi_{\myz}}$, the log-variance vector of $q_{\phi_{\myz}}$, a vector $\myz[t]$ sampled from $q_{\phi_{\myz}}$, and the KL divergence term of the VLB, for an example sequence of the 3D human motion test dataset and for the SRNN model trained and applied in the teacher-forcing mode. Figure~\ref{fig:vis-z-ss-motion} displays the corresponding trajectories for the SRNN model trained and applied in the generation mode (trained with scheduled sampling).
We can see in Figure~\ref{fig:vis-z-motion} that the trajectories of the parameters (and of the sampled $\myz[t]$ vector) are (much) smoother than in the case of speech signals, as, again, the 3D human motion data themselves have smoother trajectories compared to speech data. Some dimensions seem more ``active'' than others, even if, without a deeper investigation, it is quite difficult to interpret the range of values covered by the entries of the latent vector. Such a thorough investigation is beyond the scope of the present paper. We simply note here that there are two dimensions, dimensions 1 and 3, that seem to collapse. For these two dimensions, the mean is steady around zero and the variance has a large (and steady) value, hence the sampled trajectory of the corresponding $\myz[t]$ entries looks like noise. In contrast, some other dimensions seem to have an interesting informative profile. For example, dimensions 4 and 5 have an opposite fluctuation, probably encoding an opposite evolution of the corresponding factors of data variation. 

In Figure~\ref{fig:vis-z-ss-motion}, the latent vector generally seems more ``active'' or informative than in Figure~\ref{fig:vis-z-motion}. For most dimensions, the ratio between the range of the mean variation and the range of variance variation is larger. The range of the KL divergence term values is also larger than in Figure~\ref{fig:vis-z-motion}, indicating that the approximate posterior and the ``prior'' (generative distribution of $\myz[t]$) are less close to each other than in the teacher-forcing case. These observations are consistent with those made on the speech signals in Section~\ref{subsec:visualization-speech}. Our conjecture that the latent variable was less important in the teacher-forcing mode than in the generation mode is thus confirmed with the motion data.

\section{Conclusion}

In a practical application requiring the modeling of temporal data such as speech spectrograms or 3D human motion data, using either VRNN or SRNN seems a relevant choice, especially if autoregressive models can be used in the teacher-forcing mode. In addition, considering the above results and associated discussion, we suspect that having an inference model that respects the exact variable dependencies at inference time is important for obtaining the optimal performance. However, this is not always possible, as some applications require a causal inference model for online processing. Finally, in a practical application where only $\myz[1:T]$ needs to be transmitted, data resynthesis from $\myz[1:T]$ with an autoregressive model used in the generation mode was shown to be reasonably robust in our experiment with SRNN and speech data, provided that the model is fine-tuned in the generation mode (using here scheduled sampling). For 3D motion data, SRNN-GM was shown in our experiments to perform more poorly (with a limited gain over the vanilla VAE). 

The performance of some other models, in particular DKF, also seem to depend on the data type.
We thus insist that the above ``model ranking'' is valid only for the presented experiments, which involve pure analysis-resynthesis of speech spectrograms or 3D human motion data. For data generation, we presented only a limited set of qualitative examples to illustrate the behavior of the models (evaluating the quality of generated data is still a very difficult problem and a hot topic of machine learning with generative model). For other tasks, such as signal/data transformation (with modification of the latent vectors between analysis and resynthesis), we do not know if our experimental results would generalize. In particular, it is difficult to know how much of the information contained in $\myx[1:T]$ is encoded into $\myz[1:T]$, or what ``features'' of $\myx[1:T]$ are encoded. In particular, in the presented experiments we did not investigate the disentanglement power of each model and how the disentangled latent dimensions can be interpreted as relevant factors of variation of the data (from a physical point of view for example). Also, we illustrated the posterior collapse problem but we did not proceed to a thorough quantitative investigation of this issue. These points will be further discussed in the next chapter. However, the experiments dedicated to illustrating them are beyond the scope of the present paper.

\chapter{Discussion}
\label{sec:discussion}
In this chapter, we conclude our review of DVAEs with a discussion. First, we recall the fundamental motivation for designing and using DVAEs and then comment on their remarkable flexibility at multiple levels (design of the generation and inference models, high-level and low-level implementation). Then, we return to the crucial point of the disentanglement of latent factors in the present context of sequential data processing. Finally, we present some perspectives on data source coding. 

\section{Fundamental motivation for DVAEs}

The fundamental motivation for designing and using DVAEs is to combine various dynamical models, aimed at modeling the dynamics of sequential data, and various VAEs, aimed at modeling the latent factors of data variations. In doing so, we expect to separate the data dynamics from the other factors of variations (see Section~\ref{sec:discussion-disentanglement} below dedicated to this specific point) and use the latter to augment the expressivity of the models. Another way to express this idea is to point out the superiority of DVAEs over RNNs: Adding a latent variable $\myz[t]$ within an RNN adds considerable flexibility and modeling power to the conditional output density. Let us here quote \citet{chung2015recurrent}: \begin{quote}``We show that the introduction of latent random variables can provide significant improvements in modelling highly structured sequences such as natural speech sequences. We empirically show that the inclusion of randomness into high-level latent space can enable the VRNN to model natural speech sequences with a simple Gaussian distribution as the output function. However, the standard RNN model using the same output function fails to generate reasonable samples. An RNN-based model using more powerful output function such as a GMM can generate much better samples, but they contain a large amount of high-frequency noise compared to the samples generated by the VRNN-based models.''\end{quote}
 
Similarly, we can point out the superiority of DVAE over classical (nondeep) DBNs  and SSMs, owing to the deep nonlinear layers of information processing. Again, let us quote \citet{chung2015recurrent}: \begin{quote}``Drawing inspiration from simpler dynamic Bayesian networks (DBNs) such as HMMs and Kalman filters, the proposed variational recurrent neural network (VRNN) explicitly models the dependencies between latent random variables across subsequent timesteps. However, unlike these simpler DBN models, the VRNN retains the flexibility to model highly non-linear dynamics.''\end{quote} Of course, such a statement applies to the entire DVAE family of models. 

\section{DVAE outcome: A story of flexibility}

\subsection{Flexibility of the generative model(s)}

As seen in this review, various generative models can be derived from the general form \eqref{time_slide_ordered_model} by simplifying variable dependencies. The models we have reviewed (such as STORN, VRNN, and SRNN) are instances of these possible generative models, but there are other possibilities. Moreover, each model has several variants: Driven/undriven mode, predictive/nonpredictive mode, and with one or several feature extractors. 

When designing a generative model, complexity issues can be considered. For example, we can quote \citet{bayer2014learning}:  \begin{quote}``[...] we can restrict ourselves to prior distributions over the latent variables that factorize over time steps, i.e., $p(\myz[1:T]) = \prod_{t=1}^T p(\myz[t])$. This is much easier to handle in practice, as calculating necessary quantities such as the KL-divergence can be done independently over all time steps and components of $\myz[t]$.''\end{quote}
However, at the same time, the systematic aspect of the VAE methodology (and the versatility of the current deep learning toolkits) enables, in principle, to train a model of arbitrary complexity. Hence, if one is not limited by computational cost, this offers new possibilities. For example, we can recall our remark in Section~\ref{sec:DSAE-generative} about the choice of the dynamical model in deep SSMs. In the DVAE framework, not only is it easy to move from a linear dynamical model to a nonlinear one, but also to move from a first-order temporal model to a (much) higher order.  

\subsection{Flexibility of the inference model(s)}

In DVAEs, as in standard VAEs, the exact posterior distribution is usually intractable due to nonlinearities, which is why we have to define an inference model in addition to the generative model (we cannot apply the Bayes rule analytically). However, a key feature of DVAEs with respect to standard VAEs is that we must define an inference model over a \emph{sequence} of latent vectors. Even though the exact posterior distribution over this latent sequence is analytically intractable, we can leverage the chain rule and the D-separation principle to analyze the structure of the exact posterior distribution induced by the chosen generative model. It seems quite natural to exploit this knowledge to design the structure of the inference model so that it is consistent with the structure of dependencies in the exact intractable posterior. Yet, several seminal papers on DVAEs have not followed this ``consistency principle,'' and, more importantly, not justified the chosen structure of the inference model. Nevertheless, it is not mandatory to follow the structure of the exact posterior distribution to design the inference model. For instance, if the structure of the exact posterior distribution implies a noncausal processing of the observations, the anticausal dependencies can be dropped for the purpose of online applications. Simplifying posterior dependencies can also be motivated by a need to reduce the computational complexity of inference.

Another key difference between DVAEs and VAEs relates to how the VLB (or, actually, its estimate) is computed. The VLB involves intractable expectations, which are usually replaced with empirical averages, using samples drawn from the inference model. The sampling procedure in DVAEs has to be recursive due to the dynamical nature of the model, a constraint that standard VAEs do not have. This recursive sampling is related to the use of RNNs and can be costly. As will be discussed below, other neural network architectures can be more computationally efficient than RNNs.

\subsection{Flexibility of the implementation}

As already discussed in Section~\ref{subsec:parametrization}, various possibilities exist for the high-level implementation of DVAEs. We recall that various developed model representations can correspond to the same compact representation. In fact, the compact form describes all parent-child relationships among random variables, regardless of how these relationships are implemented in practice. Therefore, the compact representation is important to understand the probabilistic dependencies between variables. However, one must be aware that the optimization does not search for all possible models satisfying the relationships of the compact representation but only for a specific model corresponding to the developed representation. This representation allows us to understand how the dependencies are implemented in practice, and therefore, over which parameter space the model is optimized. This representation typically involves a recurrent architecture. While such architectures allow the encoding of high-order temporal dependencies, their developed graphical representations generally do not exhibit dependencies higher than first-order. Therefore, the developed representation can be ``visually misleading.'' This duality is important in DVAEs, and we encourage to provide both representations when presenting and discussing DVAEs, as done in this review.

Once the high-level DVAE architecture is chosen, various possibilities exist for the low-level implementation: network type (e.g., LSTM against GRU) and low-level (hyper)parameterization (number of layers in a network, number of units per layer, type of activation function, and classical deep learning modules, such as batch normalization). We choose not to detail these low-level implementation aspects in this review, and instead, considered them a deep learning routine. All these choices (or at least part of them) depend on data nature and datasets and can significantly impact the modeling performance. 

\subsection{Other network architectures for sequential data modeling}

In this review, we focus on deep generative latent-variable models of sequential data using RNNs (or simple feed-forward fully-connected DNNs for first-order temporal dependencies). However, other neural network architectures can deal with sequential data of arbitrary length, the most popular ones probably being convolutional architectures. While RNNs are (virtually) based on infinite-order temporal modeling, CNNs generally have a fixed-length receptive field, which implies a finite-order temporal modeling. In particular, temporal convolutional networks (TCNs) are becoming increasingly popular due to their competitive performance with RNNs (e.g., in speech separation \citep{luo2019conv}) while being more flexible and computationally efficient \citep{bai2018empirical}. A TCN is based on dilated 1D convolutions, sharing similarities with the Wavenet architecture \citep{Oord_wavenet2016}, and just as an RNN, it outputs a sequence of the same length as that of the input sequence. The combination of TCNs with VAEs was explored by \citet{aksan2018stcn}.

Another popular neural network architecture that can deal with sequential data is the transformer \citep{vaswani2017attention}, which is based solely on attention mechanisms, dispensing with recurrence and convolutions entirely. However, only a few studies have considered leveraging transformers for generative modeling in the VAE framework. We found only transformer-based VAEs recently proposed for sentence generation \citep{liu2019transformer}, story completion \citep{wang2019t}, and music representation learning \citep{jiang2020transformer}.

\section{VAE improvements and extensions applicable to DVAEs}
\label{sec:VAE-improvements}

Following the seminal VAE papers by \citet{Kingma2014} and \citet{rezende2014stochastic}, many papers have been proposed for VAE improvements and extensions. In this subsection, we mention some of these improvements and extensions and discuss their relation and possible adaptation to DVAEs. This is a nonexhaustive review; the purpose of the present paper is not to deepen this rich part of VAE literature but rather to show that the DVAE development is still largely open, and one way to improve DVAEs is to get inspired from the recent studies on VAEs. The interested readers can refer to \citeauthor{kingma2019introduction}'s \citeyearpar{kingma2019introduction} paper for a more detailed review of the ``static'' VAE improvements and extensions.  

\subsection{Improved VAE decoders and the posterior collapse problem}
\label{sec:discussion-decoders}

The mathematical formulation of the VAE in the seminal paper by \citet{Kingma2014} considered a 1D data vector framework; that is, $\myx$ is a fixed-size $F$-dimensional vector. What happens if we want to apply the VAE to 2D data, such as images, or more generally, to $N$-dimensional data? What happens if one of the dimensions is variable, like variable-length time sequences? By considering variable-size sequences and, in particular, variable-length time sequences, we take a step towards DVAEs. However, we consider models with many-to-one encoding and one-to-many decoding more as VAEs than DVAEs, following the line announced in the Introduction.

\citet{Kingma2014} considered the application of the VAE to 2D image modeling. However, the correlation between neighboring pixels was poorly exploited, as the conditional generative model (conditioned on the latent variable) was pixelwise independent.\footnote{Technically, an image is arbitrarily reshaped into a vector $\myx$, with each pixel being an entry of this vector, and the conventional ``vector'' VAE model, as described in Section~2, is applied. Therefore, each pixel is modeled independently conditioned on $\myz$, even though all pixels are not assumed marginally independent.} In such a setting, $\myz$ encodes both the local statistics of an image (e.g., local texture) and the global structural information of the image (e.g., objects), whereas it is desirable to separate this information, following the essence of latent factors disentanglement.

Subsequent studies \citep{gulrajani2016pixelvae,gregor2016towards,chen2017variational, lucas2018auxiliary,shang2018channel} considered mixing the VAE latent representation with a more sophisticated decoder exploiting local pixel correlations with either convolutive or autoregressive decoding \citep{van2016conditional, oord2016pixel}, possibly combined with a multilevel or hierarchical latent encoding (see Section~\ref{subsec:structured}). For example, \citet{chen2017variational} considered an autoregressive conditional density of the form  $p_{\theta_{\myx}}(\myx | \myz) = \prod_{i} p_{\theta_{\myx}}(\myx[i] | \myz, \myx[{\text{nb}[i]}])$ with application to 2D image modeling, where $\myx[i]$ is the $i$-th pixel of the image and $\myx[{\text{nb}[i]}]$ are the neighboring pixels. The autoregressive part is typically implemented with an RNN \citep{oord2016pixel}. 
Ideally, the local statistics of an image should be modeled by the autoregressive part, whereas the global structural information of the image should be encoded in $\myz$. 
Another example of a structured VAE for modeling images is VAEs based on convolutional neural networks (CNNs) \citep{gulrajani2016pixelvae, gregor2016towards}, which decompose/recompose an image into/from successive feature maps.

\citet{chen2017variational} discussed the tendency of the autoregressive part of the model to capture all information on the data structure and let the latent variable remain unused. This problem is referred to as \textit{latent variable vanishing} or as \textit{posterior collapse} in the literature, a term that we have already encountered in the preceding chapter. A general strategy to counter this effect (i.e., controlling the data features encoded by the RNN and the data features encoded in $\myz$) is proposed by \citet{chen2017variational} at an early level of the model design: The local autoregressive window is constrained to be small, weakening the modeling power of the decoder. This can also be done with a hierarchical structure of the latent space (see the next subsection), possibly combined with the different levels of image feature maps in CNNs \citep{gulrajani2016pixelvae, gregor2016towards}, or by introducing in the training procedure an auxiliary loss function that controls which information is captured by $\myz$ and what is left to the autoregressive decoder \citep{lucas2018auxiliary}. 

The posterior collapse problem has also been observed and discussed in the context of natural language processing \citep{bowman2015generating, serban2016piecewise}. Here, a sequence of words, individually pre-encoded into word embedding vectors, is encoded into and/or decoded from a single latent vector $\myz$. In \citeauthor{bowman2015generating}'s \citeyearpar{bowman2015generating} paper, both the encoder and decoder are single-layer LSTM RNNs. In this case, the problem is that it is difficult for the latent vector $\myz$ to encode the content of a long input sentence, and again, the RNN internal state vectors tend to encode the whole information, leaving $\myz$ unused. \citet{bowman2015generating} proposed two strategies to address this problem. The first one is applying annealing to the KL term of the VAE: A weighting factor growing from 0 to 1 is applied progressively to this term during training, first forcing $\myz$ to encode the data information and only then forcing $\myz$ to get disentangled. The second strategy is, as above, a deliberate weakening of the decoder, here by masking a part of the word embedding sequence during training. A more complex strategy was proposed by \citet{yeung2017tackling}, where an extra latent variable was added to activate/deactivate certain subvectors of $\myz$. As only a small part of the latent representation is used at each learning step, the VAE does not need to deactivate some of the dimensions of the latent variable. A heuristic approach was proposed by \citet{he2018lagging}, where the encoder is aggressively trained (i.e., trained for many iterations) before each training iteration of the decoder. The main intuition behind this approach is that the encoder has difficulties catching up with the changes in the exact posterior distribution and is lagging behind. Aggressively training the encoder allows it to catch up with the evolution of the posterior distribution at each encoder update. Other more recent solutions to the posterior collapse problem in VAEs have been proposed and discussed by \citet{lucas2019don, razavi2019preventing, dai2020usual}.

Generally, the solutions to the posterior collapse problem proposed in the literature have yielded a more influential, as well as a more disentangled latent representation. Yet, there is still room for improvement. 
The DVAEs focused on in this review do not consider a single latent vector $\myz$ for a data sequence; rather, they consider a latent vector sequence $\myz[1:T]$, which is generally synchronized with the data sequence $\myx[1:T]$ and with the sequence(s) of internal state vectors of the temporal models. This raises new issues and challenges, compared to the studies conducted on, for example, 2D image or language/text modeling. However, an important remark that is worth mentioning, although quite trivial, is that this DVAE configuration first solves the encoding capacity problem for large data sequences. As mentioned by \citet{yingzhen2018disentangled}, 
\begin{quote} ``[the model] keeps track of the time-varying aspects of $\myx[t]$ in $\myz[t]$ for every $t$, making the reconstruction to be time-local and therefore much easier. Therefore, the stochastic model is better suited if the sequences are long and complex.'' \end{quote} 
In short, with DVAEs, it is quite unlikely that a posterior collapse finds its origin in the limited capacity of the latent vector. In fact, we conjecture that it may be the opposite (i.e., a too large capacity of the latent vector sequence $\myz[1:T]$, depending on the size of $\myz[t]$) that leads to posterior collapse. This is suggested by what we observed in our experiments in Sections~\ref{subsec:visualization-speech} and \ref{subsec:visualization-motion} with the autoregressive models, where inactive entries of $\myz[1:T]$ might be considered as ``superfluous'' components.  Therefore, adjusting the dimension of $\myz[t]$ so that it can optimally fit to the content of the observed data sequence (i.e., adjusting the ``coding cost'' of the latent representation) while limiting the computational complexity is a major issue in DVAEs. All that being said, in autoregressive DVAE models such as SRNN or VRNN, even if we have a ``high-capacity'' sequence of latent variables, there is no guarantee that the autoregressive part of the model will not capture most of the information, thus ignoring the sequence of latent vectors. This remains an open problem in the DVAE literature.

\subsection{Improved inference models and algorithms}
\label{subsec:improved-inference}

As shown in Chapter~\ref{sec:VAEs}, learning in the VAE framework relies on amortized variational inference techniques. In fixed-form variational inference \citep{honkela2010approximate,salimans2013fixed} and, in particular, stochastic variational inference \citep{hoffman2013stochastic}, the approximate posterior distribution is fixed to a certain parametric form, say Gaussian for instance, and its mean and variance parameters are ``freely'' optimized through direct maximization of the VLB. In amortized variational inference, the approximate posterior distribution is still Gaussian, but there is an additional constraint imposed by the fact that its parameters are provided by an inference model corresponding to the encoder network in the VAE case. This is an additional constraint in the definition of the variational family, and the resulting amortized approximate posterior distribution is generally less expressive than its counterpart with free parameters. Consequently, the KL divergence between the approximate and exact posterior distributions is generally increased in the amortized variational inference setting, which is referred to as the \emph{amortization gap} \citep{cremer2018inference,krishnan2018challenges}. This issue can also limit the performance of the learned generative model, as the amortization gap is directly related to the gap between the VLB and the intractable log-marginal likelihood of the data, which is the criterion that we would ideally like to optimize to learn the generative model parameters (i.e., the parameters of the VAE decoder).

To reduce this gap, several studies have proposed resorting to more sophisticated inference models. Normalizing flow \citep{rezende2015variational} builds arbitrarily complex approximate posterior distributions with tractable densities by applying a series of invertible transformations to a simple initial distribution. Various normalizing flows have been proposed in the literature, for instance, based on autoregressive models \citep{kingma2016improved, chen2017variational}. Because a normalizing flow consists in chaining multiple transformations of an initial latent variable, it can be considered a particular case of a hierarchical model \citep{kingma2019introduction, kingma2016improved}, a type of model that we will discuss in Section~\ref{subsec:structured}.  An expressive approximate posterior distribution can also be defined as a mixture of simpler distributions by introducing auxiliary latent variables in the approximate posterior itself and then marginalizing \citep{maaloe2016auxiliary, ranganath2016hierarchical, salimans2015markov}.

An alternative to the design of more sophisticated inference models consists in directly modifying the inference algorithm. \citet{marino2018iterative}, proposed an iterative amortized inference technique, which consists in iteratively estimating the approximate posterior parameters using a parametric \emph{iterative inference model} that takes as input the current estimate of the parameters, the approximate gradient of the VLB (w.r.t.~the approximate posterior parameters), and potentially the observed data. This iterative inference model can, for instance, be defined using a neural network. Similarly to the ``learning to learn'' principle \citep{andrychowicz2016learning}, iterative inference models learn to perform optimization of the VLB for approximate posterior estimation. Moreover, through the encoding of the VLB gradient, iterative inference models naturally account for the top-down information obtained from the data and bottom-up information obtained from the prior to estimate the approximate posterior distribution. This feature complies with the fundamental principle of the Bayes rule, in contrast to the standard inference models, which are purely bottom-up, by simply mapping the observed data to the approximate posterior. With the same objective of overcoming the limitations of standard amortized variational inference, the semi-amortized VAEs proposed by \citet{kim2018semi} use a standard inference model (i.e., an encoder network) to provide an initial estimate of the approximate posterior parameters and then run stochastic variational inference \citep{hoffman2013stochastic} to refine them.

Of particular relevance to DVAEs, the amortized variational filtering algorithm proposed by \citet{marino2018general} generalizes iterative inference models \citep{marino2018iterative} to a general class of dynamical latent variable models for sequential data processing. This algorithm is a general method for performing causal variational inference, using only past and present observed data. When combined with DNNs, the considered general class of dynamical latent variable models corresponds to the DVAE class. The proposed inference method is thus applicable to DVAEs, and the authors conducted experiments using VRNN and SRNN (among other models). An interesting feature of this method is its versatility. In the context of standard amortized variational inference, the DVAE inference model should be designed in accordance with the form of the DVAE generative model, following, for instance, the structure of the exact posterior distribution which can be identified using D-separation (see Chapter~\ref{sec:DVAE}). In contrast, the amortized variational filtering algorithm is agnostic of the form of the generative model. Another strength of this algorithm is inherited from iterative inference models \citep{marino2018iterative}, which combine information from both the data and the prior to compute the approximate posterior distribution parameters. This principle is also applicable in the context of the amortized variational filtering algorithm, where the ``prior'' (e.g., $p(\myz[t] | \myx[1:t-1], \myz[1:t-1])$) and the approximate posterior (e.g., $q(\myz[t] | \myx[1:t], \myz[1:t-1])$) vary in time. The resulting amortized variational filtering algorithm thus resembles classical Bayesian inference filtering methods, such as a Kalman filter, where at a given time instant, the posterior distribution is computed by updating the predictive distribution (involving the prior distribution at the current time instant and the posterior distribution at the previous time instant) using the current observation, as discussed in Section~\ref{subsec:KF}.

\subsection{Disentanglement of latent factors}
\label{sec:discussion-disentanglement}

A common crucial issue for VAEs and DVAEs is how to ensure the disentanglement of latent factors. As stated by \citet{chen2017variational}, \begin{quote}``A key goal of representation learning is to identify and disentangle the underlying causal factors of the data, so that it becomes easier to understand the data, to classify it, or to perform other tasks.''\end{quote}  
Such disentanglement is not necessarily natural or efficient in the standard VAE; it somehow has to be ``encouraged,'' either in the model design or in the training procedure (or both).

\subsubsection{Semi-supervised VAEs} 

\citet{siddharth2017learning} proposed forcing the disentanglement of $\myz$ and thus improving its interpretability by using a small amount of supervision during training. This study does not particularly deal with static or dynamical VAEs, and this weak supervision principle can be applied to both. 

\subsubsection{Modification of the loss function}

A second strategy to improve the disentanglement of latent factors is to modify the loss function (i.e., the VLB). In this line, \citet{Higgins2017} introduced a weighting factor, denoted $\beta$, to weight the regularization term in (\ref{eq:VLB-b}), so that the VLB becomes
\begin{align}
\!\! \mathcal{L}(\theta, \phi, \beta ; \myX) &=  \mathbb{E}_{q_{\phi}(\myZ | \myX)}\big[ \log p_{\theta_{\myx}}(\myX | \myZ) \big] \! - \! \beta D_{\textit{KL}}\big(q_{\phi}(\myZ | \myX) \!\parallel p_{\theta_{\myz}}(\myZ)\big).\!\! \label{eq:VLB-beta} 
\end{align} 
A value of $\beta$ larger than $1$ favors the KL term; hence, it encourages  independence/disentanglement of the latent vector entries, although at the price of lower reconstruction/generation quality. For example, the images reconstructed with a $\beta$-VAE can be slightly blurred compared to those reconstructed with a standard VAE. However, the control of the properties of the objects represented in the image from the latent vectors is improved \citep{Higgins2017}. 

\citet{chen2018isolating} went a step ahead. Starting from the VLB (\ref{eq:VLB-b}), they introduced the \textit{aggregated posterior} $q_{\phi}(\myz) = \frac{1}{N_{tr}}\sum_{n=1}^{N_{tr}}q_{\phi}(\myz | \myx[n])$ and then decomposed the KL term of the VLB (summed over the training data) into a sum of three terms. The first one, referred to as the index-code mutual information, quantifies the mutual information between data and latent variables. The second one, referred to as the total correlation, is the KL divergence between the aggregated posterior and the product of its marginals (i.e., $D_{KL}\big(q_{\phi}(\myz) \parallel \prod_{l=1}^{L}q_{\phi}({\color{dred}z_l})\big)$). It quantifies the independence of the latent vector entries independently of data inputs (i.e., marginal independence as opposed to conditional independence). The third term is the sum over entries of the entry-wise KL divergence between the marginal aggregated posterior $q_{\phi}({\color{dred}z_l})$ and prior $p({\color{dred}z_l})$. The authors noted that minimizing the KL term of the VAE encourages the independence of the latent vector entries through minimization of the total correlation. However, it also penalizes the mutual information between the data and latent vectors, hence decreasing the power of latent components to explain the data. Therefore, they proposed applying a weighting factor (larger than $1$) to the total correlation only, leaving the mutual information term unchanged. They experimentally demonstrated the advantage of this strategy over the $\beta$-VAE. A similar idea was proposed independently by \citet{kim2018disentangling}, with a slightly different decomposition of the VLB KL term and a different implementation (based on an adversarial training of the model). For an extensive discussion and benchmark on disentangled representation learning with VAEs, see \citet{locatello2020sober}. 

 Such a general principle of enforcing latent factor disentanglement by modifying the loss function is independent of the issue of ``static'' against temporal modeling. It can thus, in principle, be applied to the DVAE framework. Finding a relevant decomposition of the loss function in the DVAE framework is still an open topic. Due to the more complex (temporal) intrications of the observed and latent variables, it is difficult to say if terms equivalent to total correlation or mutual information can be evidenced easily. Future studies should consider this aspect to make DVAE models more controllable and interpretable.

\subsection{Hierarchical VAEs and DVAEs}
\label{subsec:structured}

A structured VAE, or hierarchical VAE, is a general subclass of VAEs where the latent space is structured by setting a hierarchical prior distribution on a set of latent variables $\myz = \{\myz[0], \myz[1], ..., \myz[K]\}$ \citep[Chapter 4]{kingma2019introduction} \citep{salimans2016structured, sonderby2016ladder, sonderby2016train}. Here, the index denotes different latent variables, not a sample in a training set or a time index in a sequence as before. For example, a hierarchical ``multilevel'' VAE was proposed by \citet{bouchacourt2018multi}, with two latent vectors defined at different data scales: One latent vector encodes a common content for a group of data and the other latent vector encodes the style of subgroups of data within a group. In \citeauthor{bouchacourt2018multi}'s \citeyearpar{bouchacourt2018multi} paper, data grouping involves a certain amount of supervision during training. 

Another notable example of a hierarchical VAE was presented by \citep{salimans2016structured}, who proposed to use a (deep) first-order autoregressive prior model:
\begin{align}
p_{\theta_{\myz}}(\myz) &= p_{\theta_{\myz}}(\myz[0])\prod_{k=1}^K p_{\theta_{\myz}}(\myz[k] | \myz[k-1]). \label{eq:structured-prior} 
\end{align} 
In such an approach, the latent space is structured but not the data space (we still have a unique observed vector $\myx$). In this hierarchical latent variable model, there is no notion of time, but if we think of timely ordered variables, then we move toward Markov models (in the above example, a first-order one). In addition, if we consider both timely ordered latent variables and timely ordered observed variables, we step into the world of DVAEs. In other words, a DVAE can be considered a particular case of structured/hierarchical VAEs, with timely ordered latent and observed variables. 

At this point, we can make an interesting parallel among the DVAE papers that we have reviewed and the hierarchical VAE papers \citep{salimans2016structured, sonderby2016ladder, sonderby2016train, kingma2019introduction} concerning the design of the inference model. We have seen in Section~\ref{subsec:D-separation} that the D-separation methodology was not systematically used in the design of DVAE inference models. Interestingly and quite surprisingly, this methodology is not mentioned either in the above hierarchical VAE papers. Yet, we recall that D-separation is a major principled way to guide the design of inference models, including in this more general case. In these papers, the authors rather chose one model among different somewhat intuitive structures (e.g., top-down against bottom-up inference \citep{kingma2019introduction}). One strategy to make the choice is to favor an inference model with variable dependencies that ``mirror'' those of the generative model, so that some ``module'' and parameters can be shared between them. This is a suitable feature that we have discussed for DVAEs in Section~\ref{sec:sharing} and was applied in the VRNN model for example. 
In the hierarchical VAE papers, this ``module sharing'' strategy reportedly led to faster training and better fitting of model and data. 

That being said, module sharing is not incompatible with respecting the exact posterior distribution structure. For example, in \citeauthor{salimans2016structured}'s \citeyearpar{salimans2016structured} paper, the inference model is defined by
\begin{align}
q_{\phi}(\myz | \myx) &= q_{\phi}(\myz[0] | \myx)\prod_{k=1}^K q_{\phi}(\myz[k] | \myz[k-1], \myx). \label{eq:structured-posterior} 
\end{align} 
Again, it was chosen by the author because it ``mirrors'' the generative model. This inference model follows the structure of the exact posterior distribution, even though this fundamental latter point was not mentioned by \citet{salimans2016structured}. In the design of a DVAE, we can apply this principle: We can look for an inference model that both respects the structure of the exact posterior and shares some module(s) with the generative model. 

In DVAEs, the problem of disentangling the factors of data variation takes a new flavor, as different factors of variation can have different dynamics. In this context, one way to address the disentanglement challenge is therefore to apply different levels of hierarchical modeling of the latent factors on the time dimension; that is, we can design models with a different time resolution for different latent variables, which is of course not incompatible with other types of hierarchical models.
In particular, one general challenge is to separate the data dynamics (i.e., their temporal trajectories) and other factors of variations that are more constant over time (e.g., speaker identity for speech data, or objects present in the scene for videos).\footnote{Several papers dealing with disentanglement and separate control of content and dynamics in videos have reported impressive results in an adversarial training framework \citep{denton2017unsupervised,  villegas2017decomposing, tulyakov2018mocogan}.} 
For example, we have seen in this review the DSAE model \citep{yingzhen2018disentangled} and the FHVAE model \citep{hsu2017unsupervised}, which include latent variables defined at the sequence level, segment level (subsequence of consecutive frames), or frame level. For speech signal modeling, this appears as a promising way to separate the modeling and control of phonetic information, which is defined at the segment or frame level, and speaker/session information, which is defined at the sequence level. A generalization of this approach would be to impose a prior distribution of $\myz[1:T]$ that fits the dynamics of the latent factors to extract, which can be significantly different from the data dynamics.

In general, the issue of separating data dynamics and other factors of variation is still largely open in the literature on DVAE models with a sequence of latent vectors. For example, we were surprised to notice that there are very few experiments and information available on the explainability of the extracted sequence of latent factors. Experiments involving swapping of the extracted latent factors across two data sequences before resynthesizing them were reported by, for example, \citet{hsu2017unsupervised}. These experiments show that for speech signals, speaker identity can be exchanged between two sentences while preserving the same phonetic content, which is a very nice result. Yet, the issues of disentangling and controlling speech production factors separately remain largely open. Moreover, basic questions such as the impact of the size of $\myz[t]$ and $\myh[t]$ on modeling quality and the relevance of extracted latent factors have been poorly considered so far. For example, for speech processing, what happens if the size of $\myz[t]$ is reduced to a few entries, while the size of $\myh[t]$ is kept comparable to that of data $\myx[t]$? 

\citet{hsu2017unsupervised} indicated that \begin{quote}``to the best of our knowledge, there has not been any attempt to learn disentangled and interpretable representations without supervision from sequential data.''\end{quote} Regarding SRNN \citep{fraccaro2016sequential}, VRNN \citep{chung2015recurrent}, and SVAE \citep{johnson2016composing}, \citet{hsu2017unsupervised} said \begin{quote}``[...] it remains unclear whether independent attributes are disentangled in the latent space. Moreover, the learned latent variables in these models are not interpretable without manually inspecting or using labeled data.''\end{quote} Hence, the models such as STORN, VRNN, and SRNN provide an elegant and powerful mathematical and methodological framework for sequential data representation learning; however, there is still a lot of work to be done on the disentanglement challenge. Solutions for the disentanglement of $\myz[t]$ in DVAEs, inspired by or combined with existing structured or hierarchical models such as the ones presented by \citet{salimans2016structured} and \citet{sonderby2016ladder, sonderby2016train}, still have to be developed.

\section{Perspectives on source coding}
\label{sec:discussion-coding}

Although VAE and DVAE are excellent frameworks for extracting efficient and compact data representations, there are relatively few studies on their practical application to source coding (i.e., data compression including quantization and bitrate issues for data transmission or storage). We have mentioned above the problem of $\myz$ ``vanishing'' or ``being ignored'' when a powerful deterministic temporal encoder-decoder is used, and a few papers have related this problem to the need to better encode $\myz$, in the source coding sense, with an information-theoretic interpretation of a VAE as a lossy coder \citep{kingma2016improved, chen2017variational}. Among the few papers on the practical application of (D)VAE to data coding, we can mention the ConvDRAW model proposed by \citet{gregor2016towards}, which learns and encodes a hierarchy of latent variables, resulting in an image lossy compression that performs similarly to JPEG. Other examples include VQ-VAE, which is a mix of VAE and vector quantization of $\myz$, applied to speech coding \citep{van2017neural, garbacea2019low} and video compression with rate-distortion autoencoders \citep{habibian2019video}. 

As for a general approach to source coding based on DVAEs with a sequence of latent variables, we can mention two recent papers: \citet{lombardo2019deep} and
\citet{yang2020feedback}. \citet{lombardo2019deep} presented a video codec based on the DSAE model \citep{yingzhen2018disentangled}, which we reviewed in Chapter~\ref{sec:DSAE}. The sequence of latent vectors $\myz[1:T]$ extracted by the DSAE encoder is quantized and transformed into a binary minimum-length sequence by an arithmetic coder, which exploits the DSAE dynamical model $p_{\theta_{\myz}}(\myz[t] | \myz[1:t-1])$ for entropy coding. The chaining of inverse operations (i.e., arithmetic decoding, inverse quantization and DSAE decoder) enables to obtain the decoded data sequence $\myxh[1:T]$. The global variable $\myv$ (see Section~\ref{sec:DSAE-generative}) is encoded separately with a similar scheme. The resulting video codec is shown to exhibit rate-distortion performance that is comparable to the state-of-the art video codecs (such as VP9) on generic video sequences while drastically improving the performance on video sequences with specialized content (similar to the content of videos used to train the model). \citet{lombardo2019deep} provide no information on the control of the coded data sequence quality or that of the bitrate. They only mention that the arithmetic encoding of $\myz[1:T]$ requires a number of iterations. Moreover, the DSAE model is an SSM-like model; that is, $\myx[t]$ is generated from $\myz[t]$ ``alone,'' not considering the potential of using $\myx[t-1]$ (or its quantized version) for predicting $\myx[t]$ and encoding it more efficiently.  

In contrast, \citet{yang2020feedback} proposed different schemes for encoding a data sequence $\myx[1:T]$ through the inference and quantization of the corresponding sequence of latent vectors $\myz[1:T]$, with different options for recurrent connections. One of them, called feedback recurrent autoencoder (FRAE), has recurrent connections at both the encoder and decoder, and a feedback connection from the decoder to encoder. The recurrent connections are reminiscent of the \emph{predictive coding} principle that is classical in source coding theory \citep{gersho2012vector}. In fact, FRAE can be considered a \emph{nonlinear predictive coding} scheme, in which the encoder forms a latent code that encodes only the residual information that is missing when reconstructing a data vector from the deterministic internal state, which depends on past data vectors. This concept of predictive coding is strongly related to that of the predictive mode for the DVAE models that we discussed in general terms in Section~\ref{sec:DVAE-structure} and that we have seen implemented in different (autoregressive) DVAE models. Therefore, from this viewpoint, FRAE is strongly related to STORN, VRNN, and SRNN. The feedback connection from the decoder to encoder is reminiscent of another classical principle of source coding --\emph{closed-loop coding} \citep{gersho2012vector}-- even though \citet{yang2020feedback} did not refer to it explicitly. In short, closed-loop coding enables the decoder to use the quantized previous data vectors in place of the unquantized ones (not available at the decoder) for predicting the current data vector.

This line of research on nonlinear predictive coders based on DVAEs is quite promising and is only at its infancy. As \citet{yang2020feedback} wrote \begin{quote}
``There is no standard autoencoder architecture for temporally correlated data that has variable-length and long range dependencies such as video, speech, and text. The main challenge lies in the difficulty in capturing correlation information at different time-scales in an online/sequential fashion.''\end{quote}
This agrees with the concluding remark of \citet{chen2017variational}: \begin{quote}``We believe it’s exciting to extend this principle of learning lossy codes [of the latent variable $\myz$] to other forms of data, in particular those that have a temporal aspect like audio and video.''\end{quote}

\begin{acknowledgements}
This work was partially funded by the Multidisciplinary Institute in Artificial Intelligence MIAI@Grenoble-Alpes (ANR-19-P3IA-0003), the ANR ML3RI project (ANR-19-CE33-0008-01), and the H2020 SPRING
project (under GA \#871245).
\end{acknowledgements}

\appendix
\chapter{Marginalization of $\myh[1:T]$ in STORN}
\label{Appendix-A}

In this appendix, we present how the internal state vector sequence $\myh[1:T]$ can be ``marginalized'' in a DVAE model formulation; that is, how we can express $\myh[t]$ as a deterministic function of the other random variables (and thus move from the developed form of the model to the compact form). This is presented for the STORN model, but a similar derivation can be obtained for the other models as well. 

For conciseness, we replace here  \eqref{eq:STORN-1} with the generic notation $\myh[t] = f_{h}(\myx[t-1], \myz[t],\myh[t-1])$. From the dependencies between the different variables, represented by the graphical model in Figure~\ref{fig:STORN}, the joint distribution between all variables is given by
\begin{align}
p_{\theta}(\myx[1:T], \myh[1:T], \myz[1:T]) &=  \prod_{t=1}^{T} p_{\theta}(\myx[t] | \myh[t])p_{\theta}(\myh[t] | \myx[t-1], \myz[t],\myh[t-1])p(\myz[t]). \label{eq:STORN-joint}
\end{align}
As $\myh[t]$ is a deterministic function of $\myx[t-1]$, $\myz[t]$, and $\myh[t-1]$, its conditional density is a Dirac distribution with a mode given by $f_{h}(\myx[t-1], \myz[t],\myh[t-1])$:
\begin{align}
p_{\theta}(\myh[t] | \myx[t-1], \myz[t],\myh[t-1]) &= \delta(\myh[t]; f_{h}(\myx[t-1], \myz[t],\myh[t-1])).
\end{align}
Let us denote with $\mathbf{d}_{1:T}$ the sequence $\myh[1:T]$ considered a (deterministic) function of $\myx[1:T]$ and $\myz[1:T]$ only; that is, at each time step, we have $\mathbf{d}_{t} = \myh[t] = \myh[t](\myx[1:t-1], \myz[1:t]) =  f_{h}(\myx[t-1], \myz[t],f_{h}(\myx[t-2], \myz[t-1], ...)),$ with a recursive injection of the recurrent terms into this latter expression up to the first term $f_{h}(\myx[0], \myz[1],\myh[0])$. Marginalizing \eqref{eq:STORN-joint} with respect to $\myh[1:T]$ leads to
\begin{align}
p_{\theta}(\myx[1:T], \myz[1:T]) &=  \int_{\mathcal{R}^{H \times T}} \prod_{t=1}^{T} p_{\theta}(\myx[t] | \myh[t])\delta(\myh[t]; f_{h}(\myx[t-1], \myz[t],\myh[t-1]))p(\myz[t])d\myh[1:T] \label{eq:STORN-app1} \\
&= \prod_{t=1}^{T} p_{\theta}(\myx[t] | \mathbf{d}_{t})p(\myz[t]). \label{eq:STORN-app2}
\end{align}
To move from \eqref{eq:STORN-app1} to \eqref{eq:STORN-app2}, one can start by marginalizing over $\myh[T]$, so that $\myh[T]$ is replaced with $f_{h}(\myx[T-1], \myz[T],\myh[T-1])$, and then marginalizing over $\myh[T-1]$, and so on. Hereinafter, for simplification of notations, we identify $\mathbf{d}_{t}$ with $\myh[t]$, but we must keep in mind that when doing so, we see $\myh[t]$ as a deterministic function of $\myx[1:t-1]$ and $\myz[1:t]$ with the recurrence being ``unfolded,'' and not as a free random variable. Thus, we have 
\begin{align}
p_{\theta}(\myx[1:T], \myz[1:T]) &= \prod_{t=1}^{T} p_{\theta}(\myx[t] | \myh[t])p(\myz[t]) = \prod_{t=1}^{T} p_{\theta}\big(\myx[t] | \myh[t](\myx[1:t-1],\myz[1:t])\big)p(\myz[t]). \label{eq:STORN-app3}
\end{align}
From the above equation and \eqref{eq:STORN-prior-z}, we deduce the following conditional distribution:
\begin{align}
p_{\theta}(\myx[1:T] | \myz[1:T]) &= \prod_{t=1}^{T} p_{\theta}(\myx[t] | \myh[t]) = \prod_{t=1}^{T} p_{\theta}\big(\myx[t] | \myh[t](\myx[1:t-1],\myz[1:t])\big). \label{eq:STORN-app4}
\end{align}
We insist that, in the above equations, $\myh[1:T]$ is to be considered as the set of vectors $\{\myh[t](\myx[1:t-1],\myz[1:t])\}_{t=1}^T$ and not as a free random variable.

\chapter{DVAE implementation with speech data}
\label{Appendix-B}

 In this section, we provide the complete specifications of the DVAE models used in our experiments with the speech data. For each DVAE model, we will first present the generation network (decoder) and then the inference network (encoder). As many of the networks are MLPs, we define a notation to refer to these architectures concisely: MLP($\mathbf{y}$,~$n_1$,~$f_1$, \ldots,~$n_L$,~$f_L$) refers to an $L$-layer MLP with input $\mathbf{y}$, and $n_\ell$ and $f_\ell$ denote the output dimension and the (element-wise) activation function of the $\ell$-th layer, respectively. The possible activation functions are ReLU, Sigmoid, hyperbolic tangent (Tanh), and linear $\mathbb{I}$. This latter activation function is always used for the last layer of the networks computing the parameters of the random variables $\myx[t]$ or $\myz[t]$, whether they are the output of an MLP or of an RNN. Therefore, it will not be made explicit for conciseness. For example, if we define the generative network for $\myz[t]$ as MLP($\myh[t]$,~$64$,~Sigmoid,~$32$,~Sigmoid), this means that we use an MLP with two hidden layers of dimension $64$ and $32$, respectively, both with Sigmoid activation function, and an output layer of dimension $2\times 16$ (for mean and log-variance vectors, which are both the same size as $\myz[t]$) with linear activation.

\section{DKF}
\label{sec:DKF-implementation-speech}

\paragraph{Generation:} Following \citet{krishnan2017structured}, we use a gated transition function to implement $d_{\myz}$ in \eqref{eq:SSM-a}:
\begin{align}
    \boldsymbol{\nu}_t &= {\rm MLP}(\myz[t-1], 16, {\rm ReLU}, 16, \rm{Sigmoid}) \label{eq:dkf-gtf-a}\\
    \boldsymbol{\mu}_{t}^{\rm nonlin} &= {\rm MLP}(\myz[t-1], 16, {\rm ReLU}, 16, \mathbb{I}) \label{eq:dkf-gtf-b}\\
    \boldsymbol{\mu}_{t}^{\rm lin} &= {\rm MLP}(\myz[t-1], 16, \mathbb{I}) \label{eq:dkf-gtf-c}\\
    \boldsymbol{\mu}_{\theta_{\myz}} (\myz[t-1]) &= (1 - \boldsymbol{\nu}_t) \odot \boldsymbol{\mu}_{t}^{\rm lin} + \boldsymbol{\nu}_t \odot \boldsymbol{\mu}_{t}^{\rm nonlin} \label{eq:dkf-gtf-d}\\
    \boldsymbol{\sigma}_{\theta_{\myz}}^2 (\myz[t-1]) &= {\rm MLP}(\rm{ReLU}(\boldsymbol{\mu}_{t}^{\rm nonlin}), 16, \rm{Softplus}) \label{eq:dkf-gtf-e}, 
\end{align}
where $\odot$ denotes element-wise multiplication. $\boldsymbol{\mu}_{\theta_{\myz}} (\myz[t-1])$ is a gated combination of a linear and a nonlinear estimate of the mean vector. The nonlinear estimate is also used to compute the variance $\boldsymbol{\sigma}_{\theta_{\myz}}^2 (\myz[t-1])$. 

As for the generation of $\myx[t]$, the function $d_{\myx}(\myz[t])$ in \eqref{eq:SSM-c} is implemented with an MLP($\myz[t]$, $32$, Tanh, $64$, Tanh, $128$, Tanh, $256$, Tanh).

\vspace{-0.25cm}
\paragraph{Inference:} We implement the DKS inference model proposed by \citet{krishnan2017structured}, which follows equations \eqref{eq:DKF-inference-a}--\eqref{eq:DKF-inference-d}, where \eqref{eq:DKF-inference-b} is the combiner function shown in Figure~\ref{fig:arc} (a). The function $e_{\mygbw}$ is implemented using a backward LSTM fed with an MLP($\myx[t]$, $256$, Tanh). The affine function of $\myz$ in \eqref{eq:DKF-inference-b} is implemented with a one-layer MLP($\myz[t-1]$, $32$, Tanh), and the function $e_{\myz}(\myg[t])$ is implemented with an MLP($\myg[t]$, $32$, Tanh).

\section{STORN}

\paragraph{Generation:} We first recall that in STORN, the prior distribution of $\myz[1:T]$ is an i.i.d.~standard Gaussian distribution. Therefore, $\myz[1:T]$ can be sampled without requiring a dedicated network.
As for $\myh[t]$, the function $d_{\myh}(\myx[t-1], \myz[t], \myh[t-1])$ in \eqref{eq:deep-STORN-a} is implemented with the concatenation of an MLP($\myx[t-1]$, $256$, Tanh) and an MLP($\myz[t]$, $32$, Tanh, $64$, Tanh) (which can both be considered feature extractors), followed by a forward LSTM network.
As for $\myx[t]$, the function  $d_{\myx}(\myh[t])$ in \eqref{eq:deep-STORN-b} is implemented with an MLP($\myh[t]$, $256$, Tanh).

\vspace{-0.25cm}
\paragraph{Inference:} The function $e_{\myg}(\myx[t], \myg[t-1])$ in \eqref{eq:STORN-inference-1} is implemented with an MLP($\myx[t]$, $256$, Tanh), followed by a forward LSTM network. The function $e_{\myz}(\myg[t])$ in \eqref{eq:STORN-inference-2} is implemented with an MLP($\myg[t]$, $64$, Tanh, $32$, Tanh).

\section{VRNN}

We recall that, unlike STORN, VRNN employs a shared RNN for inference and generation with an internal state vector $\myh[t]$. Furthermore, VRNN explicitly introduces feature extractors for $\myx[t]$ and $\myz[t]$ and uses the extracted features to feed the different encoder and decoder modules. 

\vspace{-0.25cm}
\paragraph{Feature extraction:} $\varphi_{\myx}(\myx[t])$ is an MLP($\myx[t]$, $256$, Tanh) and  $\varphi_{\myz}(\myz[t])$ is an MLP($\myz[t]$, $32$, Tanh, $64$, Tanh).

\vspace{-0.25cm}
\paragraph{Generation:} The function $d_{\myh}(\varphi_{\myx}(\myx[t-1]), \varphi_{\myz}(\myz[t-1]), \myh[t-1])$ in \eqref{eq:VRNN-a} is implemented with an LSTM network with input $[\varphi_{\myx}(\myx[t-1]), \varphi_{\myz}(\myz[t-1])]$. The function $d_{\myz}(\myh[t])$ in \eqref{eq:VRNN-d} is implemented by directly mapping $\myh[t]$ to the dimension of latent space (i.e. the output layer mentioned in the beginning of this subsection). The function $d_{\myx}(\varphi_{\myz}(\myz[t]), \myh[t])$ in \eqref{eq:VRNN-b} is also implemented with an simple output layer.

\vspace{-0.25cm}
\paragraph{Inference:} The function $e_{\myz}(\varphi_{\myx}(\myx[t]), \myh[t])$ in \eqref{eq:VRNN-inference-a} is still implemented with an output layer where the input is the concatenation of $\varphi_{\myx}(\myx[t])$ and $\myh[t]$.

\section{SRNN}

We recall that, as with VRNN, SRNN shares an internal recurrent state vector $\myh[t]$ between the generation and inference models.

\vspace{-0.25cm}
\paragraph{Generation:} The function $d_{\myh}(\myx[t-1], \myh[t-1])$ in \eqref{eq:SRNN-a} is implemented with an LSTM network with input MLP($\myx[t-1]$, $256$, Tanh). The function $d_{\myz}(\myz[t-1], \myh[t])$ in \eqref{eq:SRNN-d} is implemented with an MLP([$\myz[t-1], \myh[t]$], $64$, Tanh, $32$, Tanh). The function  $d_{\myx}(\myz[t], \myh[t])$ in \eqref{eq:SRNN-b} is implemented with an MLP([$\myz[t], \myh[t]$], $256$, Tanh).

\vspace{-0.25cm}
\paragraph{Inference:} The function $e_{\mygbw}([\myh[t], \myx[t]],\mygbw[t+1])$ in \eqref{eq:SRNN-inference-b} is a backward LSTM network with input MLP([$\myh[t], \myx[t]$], $256$, Tanh). The function  $e_{\myz}(\myz[t-1], \mygbw[t])$ in \eqref{eq:SRNN-inference-c} is an MLP([$\myz[t-1], \mygbw[t]$], $64$, Tanh, $32$, Tanh).

\section{RVAE}

As STORN, RVAE assumes an i.i.d.~standard Gaussian prior for $\myz[1:T]$, so no network is required to generate $\myz[t]$. We recall that RVAE has causal and noncausal versions depending on whether the generation of $\myx[t]$ uses $\myz[t+1:T]$ or not, respectively.

\vspace{-0.25cm}
\paragraph{Generation:} Regarding the causal case, the function $d_{\myh} (\myz[t], \myh[t-1])$ in \eqref{eq:Simon-detail-1} is a forward LSTM network with input $\myz[t]$. The function $d_{\myx}(\myh[t])$ in \eqref{eq:Simon-detail-2} is an single output layer. For the noncausal case, the generation of $\myh[t]$ in \eqref{eq:Simon-detail-4}--\eqref{eq:Simon-detail-5} is implemented with a bidirectional LSTM with the same input as in the causal case, and $d_{\myx}(\myh[t])$ is also the same.

\vspace{-0.25cm}
\paragraph{Inference:} In the causal case, the function $e_{\mygfw}(\myz[t-1], \mygfw[t-1])$ in \eqref{eq:causal-RVAE-inference-model-a} is a forward LSTM network with input $\myz[t-1]$. The function $e_{\mygbw}(\myx[t], \mygbw[t+1])$ in \eqref{eq:causal-RVAE-inference-model-b} is a backward LSTM network with input $\myx[t]$. The function $e_{\myz}(\myg[t])$ in \eqref{eq:causal-RVAE-inference-model-d} is single output layer. In the noncausal case, the function $e_{\mygfwz}$ in \eqref{eq:non-causal-RVAE-inference-model-a} follows the same architecture as $e_{\mygfw}$ in the causal inference model, whereas the backward LSTM $e_{\mygbwx}$  is replaced with a bidirectional LSTM. The outputs are indicated as $e_{\mygfwx}$ and $e_{\mygbwx}$ in function \eqref{eq:non-causal-RVAE-inference-model-b} and \eqref{eq:non-causal-RVAE-inference-model-c}.


\section{DSAE}

We recall that compared to the other models, DSAE has an extra sequence-level latent variable $\myv$. We assume that $\myv$ has the same dimension as $\myz[t]$. As the total dimension of $\myz[1:T]$ is $T$ times that of $\myz[t]$, introducing this extra latent variable $\myv$ will not change the total number of latent variables much. Therefore, we can still consider that it is fair to compare DSAE to the other models in such a configuration. The generation of $\myv$ is not detailed in the original paper, and we assume it follows a standard Gaussian distribution.

\vspace{-0.25cm}
\paragraph{Generation:} The function $d_{\myh}(\myz[t-1], \myh[t-1])$ in \eqref{eq:DSAE-a} is a forward LSTM network with a hidden layer of dimension $128$. The function $d_{\myz}(\myh[t])$ in \eqref{eq:DSAE-b} is a one-layer linear network to project the dimension of $\myh[t]$ onto the dimension of $\myz[t]$. The function $d_{\myx}(\myz[t], \myv)$ in \eqref{eq:DSAE-d} is an MLP([$\myz[t],\myv$], $32$, Tanh, $64$, Tanh, $128$, Tanh, $256$, Tanh).

\vspace{-0.25cm}
\paragraph{Inference:} As for the inference of $\myv$, Eqs.~\eqref{eq:DSAE-inference-a}--\eqref{eq:DSAE-inference-d} are implemented with a bidirectional many-to-one LSTM network with input $\myx[t]$ and output $\mygv$, followed by an single output layer. As for the inference of $\myz[t]$, Eqs.~\eqref{eq:DSAE-inference-f}--\eqref{eq:DSAE-inference-h} are implemented with a bidirectional LSTM with input [$\myv$, $\myx[t]$]. Finally, the function $e_{\myz}(\mygz[t])$ in \eqref{eq:DSAE-inference-i} is an RNN with a hidden layer of dimension $128$.\footnote{In Figure~\ref{fig:arc} (f), we plot an MLP block after $\myx[t]$ and another one before the BRNN to output $\mygfwz[t]$ and $\mygbwz[t]$, whereas in the implementation we do not use them, or they can be considered as an identity layer. We made this choice because it provides better performance than applying several dense layers in our experiments, but we keep the possibility to use dense layers in our open-source code.}

\chapter{DVAE implementation with 3D human motion data}
\label{Appendix-C}

In this section, we provide the complete specifications of the DVAE models used in our experiments with the 3D human motion data. The notations have been defined in the previous section. We retain the general same architectures as for the speech data, because the inputs are still sequences of 1D vectors. Considering that the dimension of human pose vectors is smaller than that of speech vectors (96 vs 513), we reduce the dimension of the latent variable $\myz[t]$ (and also $\myv$ in DSAE) to 10 and use more lightweight LSTM networks with hidden state of dimension 64.

\section{DKF}
\label{sec:DKF-experiments-implementation-motion}

\paragraph{Generation:} The gated transition function for motion data is the same as for speech data, except that the hidden dimension for all MLP in \eqref{eq:dkf-gtf-a} - \eqref{eq:dkf-gtf-e} is reduced to 10. As for the generation of $\myx[t]$, the function $d_{\myx}(\myz[t])$ in \eqref{eq:SSM-c} is implemented with an MLP($\myz[t]$, $32$, Tanh, $64$, Tanh).

\vspace{-0.25cm}
\paragraph{Inference:} Similar to the implementation for speech data, the  function $e_{\mygbw}$ is implemented using a backward LSTM fed with an MLP($\myx[t]$, $64$, Tanh), the affine function of $\myz$ in \eqref{eq:DKF-inference-b} is replaced with a one-layer MLP($\myz[t-1]$, $16$, Tanh) and the function $e_{\myz}(\myg[t])$ is implemented with an MLP($\myg[t]$, $32$, Tanh).

\section{STORN}

\paragraph{Generation:} For $\myh[t]$, the function $d_{\myh}(\myx[t-1], \myz[t], \myh[t-1])$ in \eqref{eq:deep-STORN-a} is implemented with the concatenation of an MLP($\myx[t-1]$, $64$, Tanh) and an MLP($\myz[t]$, $32$, Tanh), followed by a forward LSTM network.
As for $\myx[t]$, the function  $d_{\myx}(\myh[t])$ in \eqref{eq:deep-STORN-b} is the output layer.

\vspace{-0.25cm}
\paragraph{Inference:} The function $e_{\myg}(\myx[t], \myg[t-1])$ in \eqref{eq:STORN-inference-1} is implemented with an MLP($\myx[t]$, $64$, Tanh), followed by a forward LSTM network. The function $e_{\myz}(\myg[t])$ in \eqref{eq:STORN-inference-2} is implemented with an MLP($\myg[t]$, $32$, Tanh).

\section{VRNN}

\paragraph{Feature extraction:} $\varphi_{\myx}(\myx[t])$ is an MLP($\myx[t]$, $64$, Tanh) and  $\varphi_{\myz}(\myz[t])$ is an MLP($\myz[t]$, $16$, Tanh, $32$, Tanh).

\vspace{-0.25cm}
\paragraph{Generation:} Same to speech data, the function $d_{\myh}(\varphi_{\myx}(\myx[t-1]), \varphi_{\myz}(\myz[t-1]), \myh[t-1])$ in \eqref{eq:VRNN-a} is implemented with an LSTM network with input $[\varphi_{\myx}(\myx[t-1]), \varphi_{\myz}(\myz[t-1])]$. The function $d_{\myz}(\myh[t])$ in \eqref{eq:VRNN-d} is implemented by directly mapping $\myh[t]$ to the dimension of latent space (i.e. the output layer mentioned in the beginning of this subsection). The function $d_{\myx}(\varphi_{\myz}(\myz[t]), \myh[t])$ in \eqref{eq:VRNN-b} is also implemented with an simple output layer.

\vspace{-0.25cm}
\paragraph{Inference:} Same to speech data, the function $e_{\myz}(\varphi_{\myx}(\myx[t]), \myh[t])$ in \eqref{eq:VRNN-inference-a} is still implemented with an output layer where the input is the concatenation of $\varphi_{\myx}(\myx[t])$ and $\myh[t]$.

\section{SRNN}

\paragraph{Generation:} The function $d_{\myh}(\myx[t-1], \myh[t-1])$ in \eqref{eq:SRNN-a} is implemented with an LSTM network with input MLP($\myx[t-1]$, $64$, Tanh). The function $d_{\myz}(\myz[t-1], \myh[t])$ in \eqref{eq:SRNN-d} is implemented with an MLP([$\myz[t-1], \myh[t]$], $32$, Tanh). The function  $d_{\myx}(\myz[t], \myh[t])$ in \eqref{eq:SRNN-b} is implemented with an MLP([$\myz[t], \myh[t]$], $64$, Tanh).

\vspace{-0.25cm}
\paragraph{Inference:} The function $e_{\mygbw}([\myh[t], \myx[t]],\mygbw[t+1])$ in \eqref{eq:SRNN-inference-b} is a backward LSTM network with input MLP([$\myh[t], \myx[t]$], $64$, Tanh). The function  $e_{\myz}(\myz[t-1], \mygbw[t])$ in \eqref{eq:SRNN-inference-c} is an MLP([$\myz[t-1], \mygbw[t]$], $32$, Tanh).

\section{RVAE}

\paragraph{Generation:} The implementation is the same as for speech data for both the causal case and the noncausal case, except the different dimension of the LSTM hidden state vector.

\vspace{-0.25cm}
\paragraph{Inference:} The implementation is the same as for speech data, except the different dimension of the LSTM hidden state vector.

\section{DSAE}

\paragraph{Generation:} The function $d_{\myx}(\myz[t], \myv)$ in \eqref{eq:DSAE-d} is an MLP([$\myz[t],\myv$], $32$, Tanh, $64$, Tanh). The others are the same as for speech data, except different hidden dimension of LSTM.

\vspace{-0.25cm}
\paragraph{Inference:} Same as for speech data, except different hidden dimension of LSTM.


\backmatter  

\printbibliography

@article{tipping1999probabilistic,
  title={Probabilistic principal component analysis},
  author={Tipping, Michael E and Bishop, Christopher M},
  journal={Journal of the Royal Statistical Society: Series B (Statistical Methodology)},
  volume={61},
  number={3},
  pages={611--622},
  year={1999},
}

@article{dayan1995helmholtz,
	author = {Peter Dayan and Geoffrey E Hinton and Radford M Neal and Richard S Zemel},
	journal = {Neural Computation},
	number = {5},
	pages = {889--904},
	title = {The {Helmholtz} machine},
	volume = {7},
	year = {1995}
	}

@book{frey98graphical,
	address = {Cambridge, MA},
	author = {Brendan J. Frey},
	editor = {Thomas G. Dietterich},
	publisher = {MIT Press},
	title = {Graphical models for machine learning and digital communication},
	year = {1998}
	}

@article{williams1989learning,
  title={A learning algorithm for continually running fully recurrent neural networks},
  author={Williams, Ronald J and Zipser, David},
  journal={Neural Computation},
  volume={1},
  number={2},
  pages={270--280},
  year={1989}
}

@article{bengio2015scheduled,
  title={Scheduled sampling for sequence prediction with recurrent neural networks},
  author={Bengio, Samy and Vinyals, Oriol and Jaitly, Navdeep and Shazeer, Noam},
  journal={arXiv preprint arXiv:1506.03099},
  year={2015}
}

@article{chen2018isolating,
  title={Isolating sources of disentanglement in variational autoencoders},
  author={Chen, Ricky T.Q. and Li, Xuechen and Grosse, Roger and Duvenaud, David},
  journal={arXiv preprint arXiv:1802.04942},
  year={2018}
}

@inproceedings{kim2018disentangling,
  title={Disentangling by factorising},
  author={Kim, Hyunjik and Mnih, Andriy},
  booktitle={International Conference on Machine Learning (ICML)},
  address={Stockholm, Sweden},
  year={2018}
}

@inproceedings{liu2019transformer,
  title={A {T}ransformer-based variational autoencoder for sentence generation},
  author={Liu, Danyang and Liu, Gongshen},
  booktitle={IEEE International Joint Conference on Neural Networks (IJCNN)},
  address={Budapest, Hungary},
  year={2019}
}

@inproceedings{wang2019t,
  title={T-CVAE: Transformer-based conditioned variational autoencoder for story completion},
  author={Wang, Tianming and Wan, Xiaojun},
  booktitle={AAAI International Joint Conference on Artificial Intelligence (IJCAI)},
  address={Macao, China},
  year={2019}
}

@inproceedings{jiang2020transformer,
  title={Transformer VAE: A hierarchical model for structure-aware and interpretable music representation learning},
  author={Jiang, Junyan and Xia, Gus and Carlton, Dave and Anderson, Chris and Miyakawa, Ryan},
  booktitle={IEEE International Conference on Acoustics, Speech and Signal Processing (ICASSP)},
  address={Virtual Barcelona},
  year={2020}
}

@inproceedings{vaswani2017attention,
  title={Attention is all you need},
  author={Vaswani, Ashish and Shazeer, Noam and Parmar, Niki and Uszkoreit, Jakob and Jones, Llion and Gomez, Aidan N and Kaiser, {\L}ukasz and Polosukhin, Illia},
  booktitle={Advances in Neural Information Processing Systems (NeurIPS)},
  address={Long Beach, CA},
  year={2017}
}

@article{bai2018empirical,
  title={An empirical evaluation of generic convolutional and recurrent networks for sequence modeling},
  author={Bai, Shaojie and Kolter, Zico and Koltun, Vladlen},
  journal={arXiv preprint arXiv:1803.01271},
  year={2018}
}

@article{luo2019conv,
  title={Conv-{T}asnet: Surpassing ideal time-frequency magnitude masking for speech separation},
  author={Luo, Yi and Mesgarani, Nima},
  journal={IEEE/ACM Transactions on Audio, Speech, and Language Processing},
  volume={27},
  number={8},
  pages={1256--1266},
  year={2019}
}

@inproceedings{aksan2018stcn,
  title={{STCN}: Stochastic temporal convolutional networks},
  author={Aksan, Emre and Hilliges, Otmar},
  booktitle={International Conference on Learning Representations (ICLR)},
  address={New Orleans, LA},
  year={2019}
}

@inproceedings{lea2016temporal,
  title={Temporal convolutional networks: A unified approach to action segmentation},
  author={Lea, Colin and Vidal, Rene and Reiter, Austin and Hager, Gregory D},
  booktitle={European Conference on Computer Vision (ECCV)},
  address={Amsterdam, The Netherlands},
 year={2016}
}

@article{Oord_wavenet2016,
  title={Wavenet: A generative model for raw audio},
  author={{v}an den Oord, A. and Dieleman, S. and Zen, H. and Simonyan, K. and Vinyals, O. and Graves, A. and Kalchbrenner, N. and Senior, A. and Kavukcuoglu, K.},
  journal={arXiv preprint arXiv:1609.03499},
  year={2016}
}

@article{hsu2017learning,
  title={Learning latent representations for speech generation and transformation},
  author={Hsu, W.-N. and Zhang, Y. and Glass, J.},
  journal={arXiv preprint arXiv:1704.04222},
  year={2017}
}

@inproceedings{blaauw2016modeling,
  title={Modeling and transforming speech using variational autoencoders},
  author={Blaauw, M. and Bonada, J.},
  booktitle={Conference of the International Speech Communication Association (Interspeech)},
address={San Francisco, CA},
year={2016}
}

@InProceedings{Kingma2014, 
author={Kingma, Diederik P and Welling, Max},
Booktitle= {International Conference on Learning Representations (ICLR)},
address={Banff, Canada},
title={Auto-encoding variational {Bayes}},
year={2014}
}

@inproceedings{Higgins2017,
  title={$\beta$-VAE: learning basic visual concepts with a constrained variational framework},
  author={Higgins, I. and Matthey, L. and Pal, A. and Burgess, C. and Glorot, X. and Botvinick, M. and Mohamed, S. and Lerchner, A.},
  booktitle={International Conference on Learning Representations (ICLR)},
  address={Toulon, France},
  year={2017}
}

@Article{Hinton2006,
  Title = {Reducing the dimensionality of data with neural networks},
  Author = {Hinton, G. and Salakhutdinov, R.},
  Journal = {Science},
  Year = {2006},
  Number = {5786},
  Pages = {504--507},
  Volume = {313}
}

@inproceedings{Goodfellow2014,
title = {Generative adversarial nets},
author = {Goodfellow, I. and Pouget-Abadie, J. and Mirza, M. and Xu, B. and Warde-Farley, D. and Ozair, S. and Courville, A. and Bengio, Y.},
booktitle = {Advances in Neural Information Processing Systems (NeurIPS)},
address = {Montr\'eal, Canada},
year = {2014}
}

@article{Adam,
  title={Adam: A method for stochastic optimization},
  author={Kingma, Diederik P and Ba, J.},
  journal={arXiv preprint arXiv:1412.6980},
  year={2014}
}

@article{hochreiter1997long,
  title={Long short-term memory},
  author={Hochreiter, Sepp and Schmidhuber, J{\"u}rgen},
  journal={Neural Computation},
  volume={9},
  number={8},
  pages={1735--1780},
  year={1997}
  }

@article{vincent2010stacked,
  title={Stacked denoising autoencoders: Learning useful representations in a deep network with a local denoising criterion},
  author={Vincent, P. and Larochelle, H. and Lajoie, I. and Bengio, Y. and Manzagol, P.-A.},
  journal={Journal of Machine Learning Research},
  volume={11},
  pages={3371--3408},
  year={2010}
}

@inproceedings{roche2019,
  title={Autoencoders for music sound synthesis: A comparison of linear, shallow, deep and variational models},
  author={Roche, F. and Hueber, T. and Limier, S. and Girin, L.},
  booktitle={Sound and Music Conference (SMC)},
  address={Malaga, Spain},
  year={2019}
}

@inproceedings{girin2019notes,
  title={Notes on the use of variational autoencoders for speech and audio spectrogram modeling},
  author={Girin, L. and Roche, F. and Hueber, T. and Leglaive, S.},
  booktitle={Digital Audio Effects Conference (DAFx)},
  address={Birmingham, UK},
  year={2019}
}

@inproceedings{leglaive2018variance,
  title={A variance modeling framework based on variational autoencoders for speech enhancement},
  author={Leglaive, S. and Girin, L. and Horaud, R.},
  booktitle={IEEE International Workshop on Machine Learning for Signal Processing (MLSP)},
  address={Aalborg, Denmark},
  year={2018}
}

@inproceedings{leglaive2019semi,
  title={Semi-supervised multichannel speech enhancement with variational autoencoders and non-negative matrix factorization},
  author={Leglaive, S. and Girin, L. and Horaud, R.},
  booktitle={IEEE International Conference on Acoustics, Speech and Signal Processing (ICASSP)},
  year={2019},
  address={Brighton, UK}
}

@inproceedings{bando2017statistical, 
    author={Bando, Y. and Mimura, M. and Itoyama, K. and Yoshii, K. and Kawahara, T.},
    booktitle={IEEE International Conference on Acoustics, Speech and Signal Processing (ICASSP)}, 
    title={Statistical speech enhancement based on probabilistic integration of variational autoencoder and non-negative matrix factorization}, 
    year={2018}, 
    address={Calgary, Canada}
}

@inproceedings{esling2018bridging,
  title={Bridging audio analysis, perception and synthesis with perceptually-regularized variational timbre space},
  author={Esling, P. and Chemla-Romeu-Santos, A. and Bitton, A.},
booktitle={International Society for Music Information Retrieval Conference (ISMIR)},
  address={Paris, France},
  year={2018}
}

@article{kameoka2018semi,
  title={Semi-blind source separation with multichannel variational autoencoder},
  author={Kameoka, H. and Li, L. and Inoue, S. and Makino, S.},
  journal={arXiv preprint arXiv:1808.00892},
  year={2018}
}

@inproceedings{li2017an,
  title={An {EM} algorithm for audio source separation based on the convolutive transfer function},
  author={Li, X. and Girin, L. and Horaud, R.},
  booktitle={IEEE Workshop on Applications of Signal Processing to Audio and Acoustics (WASPAA)},
  year={2017},
  address={New Paltz, NJ}
}

@inproceedings{rix2001perceptual,
  title={{Perceptual evaluation of speech quality (PESQ): A new method for speech quality assessment of telephone networks and codecs}},
  author={Rix, A.W. and Beerends, J.G. and Hollier, M.P. and Hekstra, A.P.},
  Booktitle = {IEEE International Conference on Acoustics, Speech and Signal Processing (ICASSP)},
  address={Salt Lake City, Utah},
  year={2001}
}

@article{neeser1993,
  Title = {Proper complex random processes with applications to information theory},
  Author = {F. D. Neeser and J. L. Massey},
  Journal = {IEEE Transactions on Information Theory},
  Year = {1993},
  Number = {4},
  Pages = {1293-1302},
  Volume = {39}
}

@article{daum2005nonlinear,
  title={Nonlinear filters: {B}eyond the {K}alman filter},
  author={Daum, Fred},
  journal={IEEE Aerospace and Electronic Systems Magazine},
  volume={20},
  number={8},
  pages={57--69},
  year={2005}
}

@book{moreno2009kalman,
  title={Kalman filter: {R}ecent advances and applications},
  author={Moreno, Victor M and Pigazo, Alberto},
  year={2009},
  publisher={BoD--Books on Demand}
}

@article{huang2016flexible,
  title={A flexible and efficient algorithmic framework for constrained matrix and tensor factorization},
  author={Huang, Kejun and Sidiropoulos, Nicholas D and Liavas, Athanasios P},
  journal={IEEE Transactions on Signal Processing},
  volume={64},
  number={19},
  pages={5052--5065},
  year={2016},
  publisher={IEEE}
}

@inproceedings{kuleshov2015tensor,
  title={Tensor factorization via matrix factorization},
  author={Kuleshov, Volodymyr and Chaganty, Arun and Liang, Percy},
  booktitle={Artificial Intelligence and Statistics},
  pages={507--516},
  year={2015}
}

@book{BishopBook,
  Title = {Pattern Recognition and Machine Learning},
  Author = {Bishop, C.},
  Publisher = {Springer},
  Year = {2006}
  }

@book{DLBook,
  Title = {Deep Learning},
  Author  = {I. Goodfellow and Y. Bengio and A. Courville},
  Publisher = {MIT Press},
  Year = {2016},
  Note  = {\url{http://www.deeplearningbook.org}}
}

@book{murphy2012machine,
  title={Machine learning: {A} probabilistic perspective},
  author={Murphy, Kevin P},
  year={2012},
  publisher={MIT Press}
}

@book{vsmidl2006variational,
  title={The variational {B}ayes method in signal processing},
  author={{\v{S}}m{\'i}dl, V{\'a}clav and Quinn, Anthony},
  year={2006},
  publisher={Springer Science \& Business Media}
}

@book{haykin2004kalman,
  title={Kalman filtering and neural networks},
  author={Haykin, Simon},
  year={2004},
  publisher={John Wiley \& Sons}
}

@book{gersho2012vector,
  title={Vector quantization and signal compression},
  author={Gersho, Allen and Gray, Robert M},
  year={2012},
  publisher={Springer Science \& Business Media}
}

@book{koller2009probabilistic,
  title={Probabilistic graphical models: {P}rinciples and techniques},
  author={Koller, Daphne and Friedman, Nir},
  year={2009},
  publisher={MIT Press}
}

@book{williams2006gaussian,
  title={Gaussian processes for machine learning},
  author={Williams, Christopher and Rasmussen, Carl Edward},
  year={2006},
  publisher={MIT Press}
}

@book{hamilton2020time,
  title={Time series analysis},
  author={Hamilton, James Douglas},
  year={2020},
  publisher={Princeton University Press}
}

@book{papoulis1977signal,
  title={Signal analysis},
  author={Papoulis, Athanasios},
  year={1977},
  publisher={McGraw-Hill New York}
}

@book{durbin2012time,
  title={Time series analysis by state space methods},
  author={Durbin, James and Koopman, Siem Jan},
  year={2012},
  publisher={Oxford University Press}
}

@article{archer2015black,
  title={Black box variational inference for state space models},
  author={Archer, Evan and Park, Il Memming and Buesing, Lars and Cunningham, John and Paninski, Liam},
  journal={arXiv preprint arXiv:1511.07367},
  year={2015}
}

@inproceedings{babaeizadeh2017stochastic,
  title={Stochastic variational video prediction},
  author={Babaeizadeh, Mohammad and Finn, Chelsea and Erhan, Dumitru and Campbell, Roy H and Levine, Sergey},
  booktitle={International Conference on Learning Representations (ICLR)},
  address={Vancouver, Canada},
  year={2018}
}

@inproceedings{denton2018stochastic,
  title={Stochastic video generation with a learned prior},
  author={Denton, Emily and Fergus, Rob},
  booktitle={International Conference on Machine Learning (ICML)},
  address={Stockholm, Sweden},
  year={2018}
}

@article{bayer2014learning,
  title={Learning stochastic recurrent networks},
  author={Bayer, J. and Osendorfer, C.},
  journal={arXiv preprint arXiv:1411.7610},
  year={2014}
}

@article{bitton2020neural,
  title={Neural granular sound synthesis},
  author={Bitton, Adrien and Esling, Philippe and Harada, Tatsuya},
  journal={arXiv preprint arXiv:2008.01393},
  year={2020}
}

@inproceedings{bouchacourt2018multi,
  title={Multi-level variational autoencoder: Learning disentangled representations from grouped observations},
  author={Bouchacourt, Diane and Tomioka, Ryota and Nowozin, Sebastian},
  booktitle={AAAI Conference on Artificial Intelligence},
  address={New Orleans, LA},
  year={2018}
}

@inproceedings{boulanger2012modeling,
  title={Modeling temporal dependencies in high-dimensional sequences: Application to polyphonic music generation and transcription},
  author={Boulanger-Lewandowski, Nicolas and Bengio, Yoshua and Vincent, Pascal and Gray, Patrick and Naguri, Chinmaya},
  booktitle={International Conference on Machine Learning (ICML)},
  address={Edinburgh, Scotland},
  year={2012}
}

@article{bowman2015generating,
  title={Generating sentences from a continuous space},
  author={Bowman, Samuel R and Vilnis, Luke and Vinyals, Oriol and Dai, Andrew M and Jozefowicz, Rafal and Bengio, Samy},
  journal={International Conference on Computational Natural Language Learning (CoNLL)},
  address={Berlin, Germany},
  year={2016}
}

@inproceedings{chen2017variational,
  title={Variational lossy autoencoder},
  author={Chen, Xi and Kingma, Diederik P and Salimans, Tim and Duan, Yan and Dhariwal, Prafulla and Schulman, John and Sutskever, Ilya and Abbeel, Pieter},
  booktitle={International Conference on Learning Representations (ICLR)},
  address={Toulon, France},
  year={2017}
}

@article{cho2014learning,
  title={Learning phrase representations using RNN encoder-decoder for statistical machine translation},
  author={Cho, Kyunghyun and Van Merri{\"e}nboer, Bart and Gulcehre, Caglar and Bahdanau, Dzmitry and Bougares, Fethi and Schwenk, Holger and Bengio, Yoshua},
  journal={arXiv preprint arXiv:1406.1078},
  year={2014}
}

@inproceedings{chung2015recurrent,
  title={A recurrent latent variable model for sequential data},
  author={Chung, J. and Kastner, K. and Dinh, L. and Goel, K. and Courville, A. and Bengio, Y.},
  booktitle={Advances in Neural Information Processing Systems (NeurIPS)},
  address = {Montr\'eal, Canada},
  year={2015}
}

@inproceedings{deng2017factorized,
  title={Factorized variational autoencoders for modeling audience reactions to movies},
  author={Deng, Zhiwei and Navarathna, Rajitha and Carr, Peter and Mandt, Stephan and Yue, Yisong and Matthews, Iain and Mori, Greg},
  booktitle={IEEE Conference on Computer Vision and Pattern Recognition (CVPR)}, 
  address={Honolulu, HW},
  year={2017}
}

@inproceedings{denton2017unsupervised,
  title={Unsupervised learning of disentangled representations from video},
  author={Denton, Emily and Birodkar, Vighnesh},
  booktitle={Advances in Neural Information Processing Systems (NeurIPS)},
  address={Long Beach, CA},
  year={2017}
}

@article{fabius2014variational,
  title={Variational recurrent auto-encoders},
  author={Fabius, Otto and van Amersfoort, Joost R},
  journal={arXiv preprint arXiv:1412.6581},
  year={2014}
}

@inproceedings{fortuin2020gp,
  title={{GP-VAE}: Deep probabilistic time series imputation},
  author={Fortuin, Vincent and Baranchuk, Dmitry and R{\"a}tsch, Gunnar and Mandt, Stephan},
  booktitle={International Conference on Artificial Intelligence and Statistics (AISTATS)},
  address={Palermo, Italy},
  year={2020}
}

@article{fox2011bayesian,
  title={Bayesian nonparametric inference of switching dynamic linear models},
  author={Fox, Emily and Sudderth, Erik B and Jordan, Michael I and Willsky, Alan S},
  journal={IEEE Transactions on Signal Processing},
  volume={59},
  number={4},
  pages={1569--1585},
  year={2011}
}

@inproceedings{fraccaro2016sequential,
  title={Sequential neural models with stochastic layers},
  author={Fraccaro, M. and S{\o}nderby, S. K. and Paquet, U. and Winther, O.},
  booktitle={Advances in Neural Information Processing Systems (NeurIPS)},
  address={Barcelona, Spain},
  year={2016}
}

@inproceedings{fraccaro2017disentangled,
  title={A disentangled recognition and nonlinear dynamics model for unsupervised learning},
  author={Fraccaro, Marco and Kamronn, Simon and Paquet, Ulrich and Winther, Ole},
  booktitle={Advances in Neural Information Processing Systems (NeurIPS)},
  address={Long Beach, CA},
  year={2017}
}

@inproceedings{gan2015deep,
  title={Deep temporal sigmoid belief networks for sequence modeling},
  author={Gan, Zhe and Li, Chunyuan and Henao, Ricardo and Carlson, David E and Carin, Lawrence},
  booktitle={Advances in Neural Information Processing Systems (NeurIPS)},
  address = {Montr\'eal, Canada},
  year={2015}
}

@inproceedings{garbacea2019low,
  title={Low bit-rate speech coding with VQ-VAE and a WaveNet decoder},
  author={G{\^a}rbacea, Cristina and van den Oord, A{\"a}ron and Li, Yazhe and Lim, Felicia SC and Luebs, Alejandro and Vinyals, Oriol and Walters, Thomas C},
  booktitle={IEEE International Conference on Acoustics, Speech and Signal Processing (ICASSP)},
  address={Brighton, UK},
  year={2019}
}

@article{geiger1990identifying,
  title={Identifying independence in Bayesian networks},
  author={Geiger, Dan and Verma, Thomas and Pearl, Judea},
  journal={Networks},
  volume={20},
  number={5},
  pages={507--534},
  year={1990}
}

@article{goodfellow2016nips,
  title={NIPS 2016 tutorial: Generative adversarial networks},
  author={Goodfellow, Ian},
  journal={arXiv preprint arXiv:1701.00160},
  year={2016}
}

@inproceedings{goyal2017z,
  title={Z-forcing: Training stochastic recurrent networks},
  author={Goyal, Anirudh and Sordoni, Alessandro and C{\^o}t{\'e}, Marc-Alexandre and Ke, Nan Rosemary and Bengio, Yoshua},
  booktitle={Advances in Neural Information Processing Systems (NeurIPS)},
  address={Long Beach, CA},
  year={2017}
}

@article{graves2013generating,
  title={Generating sequences with recurrent neural networks},
  author={Graves, Alex},
  journal={arXiv preprint arXiv:1308.0850},
  year={2013}
}

@inproceedings{graves2013speech,
  title={Speech recognition with deep recurrent neural networks},
  author={Graves, Alex and Mohamed, Abdel-rahman and Hinton, Geoffrey},
  booktitle={IEEE International Conference on Acoustics, Speech and Signal Processing (ICASSP)},
  address={Vancouver, Canada},
  year={2013}
}

@inproceedings{gregor2015draw,
  title={{DRAW}: A recurrent neural network for image generation},
  author={Gregor, Karol and Danihelka, Ivo and Graves, Alex and Rezende, Danilo Jimenez and Wierstra, Daan},
  booktitle={International Conference on Machine Learning (ICML)},
  address={Lille, France},
  year={2015}
}

@inproceedings{gregor2016towards,
  title={Towards conceptual compression},
  author={Gregor, Karol and Besse, Frederic and Rezende, Danilo Jimenez and Danihelka, Ivo and Wierstra, Daan},
  booktitle={Advances in Neural Information Processing Systems (NeurIPS)},
  address={Barcelona, Spain},
  year={2016}
}

@inproceedings{gu2015neural,
  title={Neural adaptive sequential {M}onte {C}arlo},
  author={Gu, Shixiang and Ghahramani, Zoubin and Turner, Richard E},
  booktitle={Advances in Neural Information Processing Systems (NeurIPS)},
  address = {Montr\'eal, Canada},
  year={2015}
}

@article{gulrajani2016pixelvae,
  title={Pixel{VAE}: A latent variable model for natural images},
  author={Gulrajani, Ishaan and Kumar, Kundan and Ahmed, Faruk and Taiga, Adrien Ali and Visin, Francesco and Vazquez, David and Courville, Aaron},
  journal={arXiv preprint arXiv:1611.05013},
  year={2016}
}

@inproceedings{habibian2019video,
  title={Video compression with rate-distortion autoencoders},
  author={Habibian, Amirhossein and Rozendaal, Ties van and Tomczak, Jakub M and Cohen, Taco S},
  booktitle={IEEE International Conference on Computer Vision (ICCV)},
  address={Seoul, Korea},
  year={2019}
}

@article{hafner2018learning,
  title={Learning latent dynamics for planning from pixels},
  author={Hafner, Danijar and Lillicrap, Timothy and Fischer, Ian and Villegas, Ruben and Ha, David and Lee, Honglak and Davidson, James},
  journal={arXiv preprint arXiv:1811.04551},
  year={2018}
}

@article{hinton1995wake,
  title={The ``{W}ake-{S}leep" algorithm for unsupervised neural networks},
  author={Hinton, Geoffrey and Dayan, Peter and Frey, Brendan and Neal, Radford},
  journal={Science},
  volume={268},
  number={5214},
  pages={1158--1161},
  year={1995}
  }

@inproceedings{hsu2017unsupervised,
  title={Unsupervised learning of disentangled and interpretable representations from sequential data},
  author={Hsu, Wei-Ning and Zhang, Yu and Glass, James},
  booktitle={Advances in Neural Information Processing Systems (NeurIPS)},
  address={Long Beach, CA},
  year={2017}
}

@inproceedings{hu2017toward,
  title={Toward controlled generation of text},
  author={Hu, Zhiting and Yang, Zichao and Liang, Xiaodan and Salakhutdinov, Ruslan and Xing, Eric P},
  booktitle={International Conference on Machine Learning (ICML)},
  address={Sydney, Australia},
  year={2017}
}

@article{jang2019recurrent,
  title={Recurrent neural network-based semantic variational autoencoder for sequence-to-sequence learning},
  author={Jang, Myeongjun and Seo, Seungwan and Kang, Pilsung},
  journal={Information Sciences},
  volume={490},
  pages={59--73},
  year={2019}
}

@inproceedings{johnson2016composing,
  title={Composing graphical models with neural networks for structured representations and fast inference},
  author={Johnson, Matthew J and Duvenaud, David K and Wiltschko, Alex and Adams, Ryan P and Datta, Sandeep R},
  booktitle={Advances in Neural Information Processing Systems (NeurIPS)},
  address={Barcelona, Spain},
  year={2016}
}

@article{jordan1999introduction,
  title={An introduction to variational methods for graphical models},
  author={Jordan, Michael I and Ghahramani, Zoubin and Jaakkola, Tommi S and Saul, Lawrence K},
  journal={Machine Learning},
  volume={37},
  number={},
  pages={183--233},
  year={1999}
}

@inproceedings{karl2016deep,
  title={Deep variational {B}ayes filters: Unsupervised learning of state space models from raw data},
  author={Karl, M. and Soelch, M. and Bayer, J. and van der Smagt, P.},
  booktitle={International Conference on Learning Representations (ICLR)},
  address={Toulon, France},
  year={2017}
}

@inproceedings{kingma2016improved,
  title={Improved variational inference with inverse autoregressive flow},
  author={Kingma, Durk P and Salimans, Tim and Jozefowicz, Rafal and Chen, Xi and Sutskever, Ilya and Welling, Max},
  booktitle={Advances in Neural Information Processing Systems (NeurIPS)},
  address={Barcelona, Spain},
  year={2016}
}

@article{kingma2019introduction,
  title={An introduction to variational autoencoders},
  author={Kingma, Diederik P and Welling, Max},
  journal={Foundations and Trends in Machine Learning},
 volume = {12},
 issue = {4},
 pages={307--392},
  year={2019}
}

@inproceedings{krishnan2015deep,
  title={Deep {K}alman filters},
  author={Krishnan, R.G. and Shalit, U. and Sontag, D.},
  booktitle={arXiv preprint arXiv:1511.05121},
  year={2015}
}

@inproceedings{krishnan2017structured,
  title={Structured inference networks for nonlinear state space models},
  author={Krishnan, R.G. and Shalit, U. and Sontag, D.},
  booktitle={AAAI Conference on Artificial Intelligence},
  address={San Francisco, CA},
  year={2017}
}

@inproceedings{lee2018acoustic,
  title={Acoustic modeling using adversarially trained variational recurrent neural network for speech synthesis},
  author={Lee, Joun Yeop and Cheon, Sung Jun and Choi, Byoung Jin and Kim, Nam Soo and Song, Eunwoo},
  booktitle={Conference of the International Speech Communication Association (Interspeech)},
  address={Hyderabad, India},
  year={2018}
}

@inproceedings{leglaive2020recurrent,
  title={A recurrent variational autoencoder for speech enhancement},
  author={Leglaive, Simon and Alameda-Pineda, Xavier and Girin, Laurent and Horaud, Radu},
  booktitle={IEEE International Conference on Acoustics, Speech and Signal Processing (ICASSP)},
  year={2020},
  address={Virtual Barcelona}
}

@article{linderman2016recurrent,
  title={Recurrent switching linear dynamical systems},
  author={Linderman, Scott W and Miller, Andrew C and Adams, Ryan P and Blei, David M and Paninski, Liam and Johnson, Matthew J},
  journal={arXiv preprint arXiv:1610.08466},
  year={2016}
}

@inproceedings{lucas2018auxiliary,
  title={Auxiliary guided autoregressive variational autoencoders},
  author={Lucas, Thomas and Verbeek, Jakob},
  booktitle={Joint European Conference on Machine Learning and Knowledge Discovery in Databases},
  address={Dublin, Ireland},
  year={2018}
}

@inproceedings{lombardo2019deep,
  title={Deep generative video compression},
  author={Lombardo, Salvator and Han, Jun and Schroers, Christopher and Mandt, Stephan},
  booktitle={Advances in Neural Information Processing Systems (NeurIPS)},
  address  = {Vancouver, Canada},
  year={2019}
}

@inproceedings{mathieu2016disentangling,
  title={Disentangling factors of variation in deep representation using adversarial training},
  author={Mathieu, Michael F and Zhao, Junbo Jake and Zhao, Junbo and Ramesh, Aditya and Sprechmann, Pablo and LeCun, Yann},
  booktitle={Advances in Neural Information Processing Systems (NeurIPS)},
  address={Barcelona, Spain},
  year={2016}
}

@inproceedings{miao2016neural,
  title={Neural variational inference for text processing},
  author={Miao, Yishu and Yu, Lei and Blunsom, Phil},
  booktitle={International Conference on Machine Learning (ICML)},
  address={New York, NY},
  year={2016}
}

@article{miladinovic2019disentangled,
  title={Disentangled state space representations},
  author={Miladinovi{\'c}, Dorde and Gondal, Muhammad Waleed and Sch{\"o}lkopf, Bernhard and Buhmann, Joachim M and Bauer, Stefan},
  journal={arXiv preprint arXiv:1906.03255},
  year={2019}
}

@article{mnih2014neural,
  title={Neural variational inference and learning in belief networks},
  author={Mnih, Andriy and Gregor, Karol},
  journal={arXiv preprint arXiv:1402.0030},
  year={2014}
}

@article{murphy1998switching,
  title={Switching {K}alman filters},
  author={Murphy, Kevin P},
  year={1998},
  journal={Unpublished technical report -- Available online}
}

@article{pariente2019statistically,
  title={A statistically principled and computationally efficient approach to speech enhancement using variational autoencoders},
  author={Pariente, Manuel and Deleforge, Antoine and Vincent, Emmanuel},
  journal={arXiv preprint arXiv:1905.01209},
  year={2019}
}

@inproceedings{pereira2018unsupervised,
  title={Unsupervised anomaly detection in energy time series data using variational recurrent autoencoders with attention},
  author={Pereira, Joao and Silveira, Margarida},
  booktitle={IEEE International Conference on Machine Learning and Applications (ICMLA)},
  address={Orlando, FL},
  year={2018}
}

@article{raiko2009variational,
  title={Variational {B}ayesian learning of nonlinear hidden state-space models for model predictive control},
  author={Raiko, Tapani and Tornio, Matti},
  journal={Neurocomputing},
  volume={72},
  number={16-18},
  pages={3704--3712},
  year={2009}
}

@inproceedings{rezende2014stochastic,
  title={Stochastic backpropagation and approximate inference in deep generative models},
  author={Rezende, D. J. and Mohamed, S. and Wierstra, D.},
 booktitle={International Conference on Machine Learning (ICML)},
  address={Beijing, China},
  year={2014}
}

@article{rezende2015variational,
  title={Variational inference with normalizing flows},
  author={Rezende, Danilo Jimenez and Mohamed, Shakir},
  journal={arXiv preprint arXiv:1505.05770},
  year={2015}
}

@article{roberts2018hierarchical,
  title={A hierarchical latent vector model for learning long-term structure in music},
  author={Roberts, Adam and Engel, Jesse and Raffel, Colin and Hawthorne, Curtis and Eck, Douglas},
  journal={arXiv preprint arXiv:1803.05428},
  year={2018}
}

@article{salimans2016structured,
  title={A structured variational auto-encoder for learning deep hierarchies of sparse features},
  author={Salimans, Tim},
  journal={arXiv preprint arXiv:1602.08734},
  year={2016}
}

@article{semeniuta2017hybrid,
  title={A hybrid convolutional variational autoencoder for text generation},
  author={Semeniuta, Stanislau and Severyn, Aliaksei and Barth, Erhardt},
  journal={arXiv preprint arXiv:1702.02390},
  year={2017}
}

@inproceedings{serban2017hierarchical,
  title={A hierarchical latent variable encoder-decoder model for generating dialogues},
  author={Serban, Iulian Vlad and Sordoni, Alessandro and Lowe, Ryan and Charlin, Laurent and Pineau, Joelle and Courville, Aaron and Bengio, Yoshua},
  booktitle={AAAI Conference on Artificial Intelligence},
  address={San Francisco, CA},
  year={2017}
}

@article{serban2016piecewise,
  title={Piecewise latent variables for neural variational text processing},
  author={Serban, Iulian V and Ororbia, Alexander G and Pineau, Joelle and Courville, Aaron},
  journal={arXiv preprint arXiv:1612.00377},
  year={2016}
}

@inproceedings{shang2018channel,
  title={Channel-recurrent autoencoding for image modeling},
  author={Shang, Wenling and Sohn, Kihyuk and Tian, Yuandong},
  booktitle={IEEE Winter Conference on Applications of Computer Vision (WACV)},
  address={Lake Tahoe, NV},
  year={2018}
}

@inproceedings{siddharth2017learning,
  title={Learning disentangled representations with semi-supervised deep generative models},
  author={Siddharth, Narayanaswamy and Paige, Brooks and Van de Meent, Jan-Willem and Desmaison, Alban and Goodman, Noah and Kohli, Pushmeet and Wood, Frank and Torr, Philip},
  booktitle={Advances in Neural Information Processing Systems (NeurIPS)},
  address={Long Beach, CA},
  year={2017}
}

@inproceedings{sohn2015learning,
  Title = {Learning structured output representation using deep conditional generative models},
  Author = {Sohn, Kihyuk and Lee, Honglak and Yan, Xinchen},
  Booktitle = {Advances in Neural Information Processing Systems (NeurIPS)},
  address = {Montr\'eal, Canada},
  Year = {2015}
}

@inproceedings{sonderby2016ladder,
  title={Ladder variational autoencoders},
  author={S{\o}nderby, Casper Kaae and Raiko, Tapani and Maal{\o}e, Lars and S{\o}nderby, S{\o}ren Kaae and Winther, Ole},
  booktitle={Advances in Neural Information Processing Systems (NeurIPS)},
  address={Barcelona, Spain},
  year={2016}
}

@inproceedings{sonderby2016train,
  title={How to train deep variational autoencoders and probabilistic ladder networks},
  author={S{\o}nderby, Casper Kaae and Raiko, Tapani and Maal{\o}e, Lars and S{\o}nderby, S{\o}ren Kaae and Winther, Ole},
  booktitle={International Conference on Machine Learning (ICML)},
    address={New York, NY},
  year={2016}
}

@inproceedings{su2018variational,
  title={Variational recurrent neural machine translation},
  author={Su, Jinsong and Wu, Shan and Xiong, Deyi and Lu, Yaojie and Han, Xianpei and Zhang, Biao},
  booktitle={AAAI International Conference on Artificial Intelligence},
  address={New Orleans, NV},
  year={2018}
}

@book{sutskever2013training,
  title={Training recurrent neural networks},
  author={Sutskever, Ilya},
  year={2013},
  publisher={University of Toronto}
}

@inproceedings{tulyakov2018mocogan,
  title={{M}o{C}o{G}an: Decomposing motion and content for video generation},
  author={Tulyakov, Sergey and Liu, Ming-Yu and Yang, Xiaodong and Kautz, Jan},
  booktitle={IEEE International Conference on Computer Vision and Pattern Recognition (CVPR)},
  address={Salt Lake City, Utah},
  year={2018}
}

@article{oord2016pixel,
  title={Pixel recurrent neural networks},
  author={{v}an den Oord, Aaron and Kalchbrenner, Nal and Kavukcuoglu, Koray},
  journal={International Conference on Machine Learning (ICML)},
  address={New York, NY},
  year={2016}
}

@inproceedings{van2016conditional,
  title={Conditional image generation with {P}ixel{CNN} decoders},
  author={{v}an den Oord, Aaron and Kalchbrenner, Nal and Espeholt, Lasse and Vinyals, Oriol and Graves, Alex and others},
  booktitle={Advances in Neural Information Processing Systems (NeurIPS)},
  address={Barcelona, Spain},
  year={2016}
}

@inproceedings{van2017neural,
  title={Neural discrete representation learning},
  author={{v}an den Oord, Aaron and Vinyals, Oriol and Kavukcuoglu, Koray},
  booktitle={Advances in Neural Information Processing Systems (NeurIPS)},
  address={Long Beach, CA},
  year={2017}
}

@inproceedings{villegas2017decomposing,
  title={Decomposing motion and content for natural video sequence prediction},
  author={Villegas, Ruben and Yang, Jimei and Hong, Seunghoon and Lin, Xunyu and Lee, Honglak},
  booktitle={International Conference on Learning Representations (ICLR)},
  address={Toulon, France},
  year={2017}
}

@inproceedings{wan2000unscented,
  title={The unscented {K}alman filter for nonlinear estimation},
  author={Wan, Eric and Van Der Merwe, Rudolph},
  booktitle={IEEE Adaptive Systems for Signal Processing, Communications, and Control Symposium},
  address={Lake Louise, Canada},
  year={2000}
}

@inproceedings{watter2015embed,
  title={Embed to control: A locally linear latent dynamics model for control from raw images},
  author={Watter, Manuel and Springenberg, Jost and Boedecker, Joschka and Riedmiller, Martin},
  booktitle={Advances in Neural Information Processing Systems (NeurIPS)},
  address = {Montr\'eal, Canada},
  year={2015}
}

@inproceedings{yang2017improved,
  title={Improved variational autoencoders for text modeling using dilated convolutions},
  author={Yang, Zichao and Hu, Zhiting and Salakhutdinov, Ruslan and Berg-Kirkpatrick, Taylor},
  booktitle={International Conference on Machine Learning (ICML)},
  address={Sydney, Australia},
  year={2017}
}

@inproceedings{yang2020feedback,
  title={Feedback recurrent autoencoder},
  author={Yang, Yang and Sauti{\`e}re, Guillaume and Ryu, J Jon and Cohen, Taco S},
  booktitle={IEEE International Conference on Acoustics, Speech and Signal Processing (ICASSP)},
  address={Virtual Barcelona},
  year={2020}
}

@inproceedings{yingzhen2018disentangled,
  title={Disentangled sequential autoencoder},
  author={Li, Yingzhen and Mandt, Stephan},
  booktitle={International Conference on Machine Learning (ICML)},
  address={Stockholm, Sweden},
  year={2018}
}

@article{zhao2017learning,
  title={Learning discourse-level diversity for neural dialog models using conditional variational autoencoders},
  author={Zhao, Tiancheng and Zhao, Ran and Eskenazi, Maxine},
  journal={arXiv preprint arXiv:1703.10960},
  year={2017}
}

@article{zhao2018unsupervised,
  title={Unsupervised discrete sentence representation learning for interpretable neural dialog generation},
  author={Zhao, Tiancheng and Lee, Kyusong and Eskenazi, Maxine},
  journal={arXiv preprint arXiv:1804.08069},
  year={2018}
}

@article{zhao2020variational,
  title={Variational online learning of neural dynamics},
  author={Zhao, Yuan and Park, Il Memming},
  journal={arXiv preprint arXiv:1707.09049},
  year={2017}
}

@article{zhao2019streaming,
  title={Streaming variational {M}onte {C}arlo},
  author={Zhao, Yuan and Nassar, Josue and Jordan, Ian and Bugallo, M{\'o}nica and Park, Il Memming},
  journal={arXiv preprint arXiv:1906.01549},
  year={2019}
}

@incollection{paszke2019pytorch,
title = {PyTorch: An imperative style, high-performance deep learning library},
author = {Paszke, Adam and Gross, Sam and Massa, Francisco and Lerer, Adam and Bradbury, James and Chanan, Gregory and Killeen, Trevor and Lin, Zeming and Gimelshein, Natalia and Antiga, Luca and Desmaison, Alban and Kopf, Andreas and Yang, Edward and DeVito, Zachary and Raison, Martin and Tejani, Alykhan and Chilamkurthy, Sasank and Steiner, Benoit and Fang, Lu and Bai, Junjie and Chintala, Soumith},
booktitle = {Advances in Neural Information Processing Systems (NeurIPS)},
address  = {Vancouver, Canada},
year = {2019}
}

@article{abadi2016tensorflow,
  title={Tensorflow: Large-scale machine learning on heterogeneous distributed systems},
  author={Abadi, Mart{\'\i}n and Agarwal, Ashish and Barham, Paul and Brevdo, Eugene and Chen, Zhifeng and Citro, Craig and Corrado, Greg S and Davis, Andy and Dean, Jeffrey and Devin, Matthieu and others},
  journal={arXiv preprint arXiv:1603.04467},
  year={2016}
}

@article{WSJ0,
  title={CSR-I (WSJ0) Sennheiser LDC93S6B. https://catalog.ldc.upenn.edu/LDC93S6B},
  author={Garofolo, John and Graff, David and Paul, Doug and Pallett, David},
  journal={Philadelphia: Linguistic Data Consortium},
  year={1993}
}

@incollection{neal1998view,
  title={A view of the EM algorithm that justifies incremental, sparse, and other variants},
  author={Neal, Radford M and Hinton, Geoffrey E},
  booktitle={Learning in graphical models},
  pages={355--368},
  year={1998},
  publisher={Springer}
}

@article{wei1990monte,
  title={A {Monte Carlo} implementation of the {EM} algorithm and the poor man's data augmentation algorithms},
  author={Wei, Greg C.G. and Tanner, Martin A.},
  journal={Journal of the American Statistical Association},
  volume={85},
  number={411},
  pages={699--704},
  year={1990}
}

@Article{dempster1977maximum,
  Title                    = {{Maximum likelihood from incomplete data via the EM algorithm}},
  Author                   = {Dempster, Arthur P. and Laird, Nan M. and Rubin, Donald B.},
  Journal                  = {Journal of the Royal Statistical Society. Series B (Methodological)},
  Year                     = {1977},
  Number                   = {1},
  Pages                    = {1--38},
  Volume                   = {39}
}

@article{blei2017variational,
  title={Variational inference: A review for statisticians},
  author={Blei, David M and Kucukelbir, Alp and McAuliffe, Jon D},
  journal={Journal of the American Statistical Association},
  volume={112},
  number={518},
  pages={859--877},
  year={2017}
}

@article{winn2005variational,
  title={Variational message passing},
  author={Winn, John and Bishop, Christopher M},
  journal={Journal of Machine Learning Research},
  volume={6},
  number={Apr},
  pages={661--694},
  year={2005}
}

@article{honkela2010approximate,
  title={Approximate Riemannian conjugate gradient learning for fixed-form variational Bayes},
  author={Honkela, Antti and Raiko, Tapani and Kuusela, Mikael and Tornio, Matti and Karhunen, Juha},
  journal={Journal of Machine Learning Research},
  volume={11},
  number={Nov},
  pages={3235--3268},
  year={2010}
}

@article{salimans2013fixed,
  title={Fixed-form variational posterior approximation through stochastic linear regression},
  author={Salimans, Tim and Knowles, David A},
  journal={Bayesian Analysis},
  volume={8},
  number={4},
  pages={837--882},
  year={2013}
}

@article{hoffman2013stochastic,
  title={Stochastic variational inference},
  author={Hoffman, Matthew D and Blei, David M and Wang, Chong and Paisley, John},
  journal={Journal of Machine Learning Research},
  volume={14},
  number={1},
  pages={1303--1347},
  year={2013}
}

@inproceedings{he2018lagging,
  title={Lagging inference networks and posterior collapse in variational autoencoders},
  author={He, Junxian and Spokoyny, Daniel and Neubig, Graham and Berg-Kirkpatrick, Taylor},
  booktitle={International Conference on Learning Representations (ICLR)},
  address={Vancouver, Canada},
  year={2018}
}

@inproceedings{yeung2017tackling,
  title={Tackling over-pruning in variational autoencoders},
  author={Yeung, Serena and Kannan, Anitha and Dauphin, Yann and Fei-Fei, Li},
  booktitle={International Conference on Machine Learning -- Workshop on Principled Approaches to Deep Learning},
  address={Sydney, Australia},
  year={2017}
}

@inproceedings{saul1996exploiting,
  title={Exploiting tractable substructures in intractable networks},
  author={Saul, Lawrence K and Jordan, Michael I},
 booktitle={Advances in Neural Information Processing Systems (NeurIPS)},
 address={Denver, CO},
  year={1996}
}

@article{robbins1951stochastic,
  title={A stochastic approximation method},
  author={Robbins, Herbert and Monro, Sutton},
  journal={The Annals of Mathematical Statistics},
  pages={400--407},
  year={1951}
}

@Inbook{Bottou2004,
author="Bottou, L{\'e}on",
title="Stochastic Learning",
bookTitle="Advanced Lectures on Machine Learning: ML Summer Schools 2003, Canberra, Australia",
year="2004",
pages="146--168"
}

@article{diggle1984monte,
  title={Monte Carlo methods of inference for implicit statistical models},
  author={Diggle, Peter J and Gratton, Richard J},
  journal={Journal of the Royal Statistical Society: Series B (Methodological)},
  volume={46},
  number={2},
  pages={193--212},
  year={1984}
}

@article{bie2021unsupervised,
  title={Unsupervised speech enhancement using dynamical variational autoencoders},
  author={Bie, Xiaoyu and Leglaive, Simon and Alameda-Pineda, Xavier and Girin, Laurent},
  journal={arXiv preprint arXiv:2106.12271},
  year={2021}
}

@inproceedings{cremer2018inference,
  title={Inference suboptimality in variational autoencoders},
  author={Cremer, Chris and Li, Xuechen and Duvenaud, David},
  booktitle={International Conference on Machine Learning (ICML)},
  address={Stockholm, Sweden},
  year={2018}
}

@inproceedings{krishnan2018challenges,
  title={On the challenges of learning with inference networks on sparse, high-dimensional data},
  author={Krishnan, Rahul and Liang, Dawen and Hoffman, Matthew},
  booktitle={International Conference on Artificial Intelligence and Statistics},
  address={Lanzarote, Spain},
  year={2018}
}

@inproceedings{maaloe2016auxiliary,
  title={Auxiliary deep generative models},
  author={Maal{\o}e, Lars and S{\o}nderby, Casper Kaae and S{\o}nderby, S{\o}ren Kaae and Winther, Ole},
  booktitle={International Conference on Machine Learning (ICML)},
  address={New York, NY},
  year={2016}
}

@inproceedings{ranganath2016hierarchical,
  title={Hierarchical variational models},
  author={Ranganath, Rajesh and Tran, Dustin and Blei, David},
  booktitle={International Conference on Machine Learning (ICML)},
  address={New York, NY},
  year={2016}
}

@inproceedings{salimans2015markov,
  title={Markov chain Monte Carlo and variational inference: Bridging the gap},
  author={Salimans, Tim and Kingma, Diederik and Welling, Max},
  booktitle={International Conference on Machine Learning (ICML)},
  address={Lille, France},
  year={2015}
}

@inproceedings{marino2018iterative,
  title={Iterative amortized inference},
  author={Marino, Joe and Yue, Yisong and Mandt, Stephan},
  booktitle={International Conference on Machine Learning (ICML)},
  address={Stockholm, Sweden},
  year={2018}
}

@inproceedings{kim2018semi,
  title={Semi-amortized variational autoencoders},
  author={Kim, Yoon and Wiseman, Sam and Miller, Andrew and Sontag, David and Rush, Alexander},
  booktitle={International Conference on Machine Learning (ICML)},
  address={Stockholm, Sweden},
  year={2018}
}

@inproceedings{marino2018general,
  title={A general method for amortizing variational filtering},
  author={Marino, Joseph and Cvitkovic, Milan and Yue, Yisong},
  booktitle={Advances in Neural Information Processing Systems (NeurIPS)},
  address={Montr\'eal, Canada},
  year={2018}
}

@inproceedings{andrychowicz2016learning,
  title={Learning to learn by gradient descent by gradient descent},
  author={Andrychowicz, Marcin and Denil, Misha and Gomez, Sergio and Hoffman, Matthew W and Pfau, David and Schaul, Tom and Shillingford, Brendan and De Freitas, Nando},
  booktitle={Advances in Neural Information Processing Systems (NeurIPS)},
  address={Barcelona, Spain},
  year={2016}
}

@inproceedings{dai2020usual,
  title={The usual suspects? Reassessing blame for {VAE} posterior collapse},
  author={Dai, Bin and Wang, Ziyu and Wipf, David},
  booktitle={International Conference on Machine Learning (ICML)},
  address={Virtual online},
  year={2020}
}

@inproceedings{lucas2019don,
  title={Don't blame the {ELBO}! a linear {VAE} perspective on posterior collapse},
  author={Lucas, James and Tucker, George and Grosse, Roger B and Norouzi, Mohammad},
  booktitle={Advances in Neural Information Processing Systems (NeurIPS)},
  address  = {Vancouver, Canada},
  year={2019}
}

@article{razavi2019preventing,
  title={Preventing posterior collapse with delta-{VAE}s},
  author={Razavi, Ali and {van den Oord}, A{\"a}ron and Poole, Ben and Vinyals, Oriol},
  journal={arXiv preprint arXiv:1901.03416},
  year={2019}
}

@inproceedings{le2018auto,
  title={Auto-encoding sequential Monte Carlo},
  author={Le, Tuan Anh and Igl, Maximilian and Rainforth, Tom and Jin, Tom and Wood, Frank},
  booktitle={International Conference on Learning Representations (ICLR)},
  address={Vancouver, Canada},
  year={2018}
}

@inproceedings{naesseth2018variational,
  title={Variational sequential Monte Carlo},
  author={Naesseth, Christian and Linderman, Scott and Ranganath, Rajesh and Blei, David},
  booktitle={International Conference on Artificial Intelligence and Statistics},
  address={Lanzarote, Spain},
  year={2018}
}

@inproceedings{maddison2017filtering,
  title={Filtering variational objectives},
  author={Maddison, Chris J and Lawson, John and Tucker, George and Heess, Nicolas and Norouzi, Mohammad and Mnih, Andriy and Doucet, Arnaud and Teh, Yee},
  booktitle={Advances in Neural Information Processing Systems (NeurIPS)},
  address={Long Beach, CA},
  year={2017}
}

@article{locatello2020sober,
  title={A sober look at the unsupervised learning of disentangled representations and their evaluation},
  author={Locatello, Francesco and Bauer, Stefan and Lucic, Mario and R{\"a}tsch, Gunnar and Gelly, Sylvain and Sch{\"o}lkopf, Bernhard and Bachem, Olivier},
  journal={Journal of Machine Learning Research},
  volume={21},
  pages={1--62},
  year={2020}
}

@inproceedings{le2019sdr,
  title={{SDR}: Half-baked or well done?},
  author={Le Roux, Jonathan and Wisdom, Scott and Erdogan, Hakan and Hershey, John R},
  booktitle={IEEE International Conference on Acoustics, Speech and Signal Processing (ICASSP)},
  year={2019},
  address={Brighton, UK}
}

@article{taal2011algorithm,
  title={An algorithm for intelligibility prediction of time-frequency weighted noisy speech},
  author={Taal, Cees H and Hendriks, Richard C and Heusdens, Richard and Jensen, Jesper},
  journal={IEEE Transactions on Audio, Speech, and Language Processing},
  volume={19},
  number={7},
  pages={2125--2136},
  year={2011}
}

@ARTICLE{h36m,
  author={Ionescu, Catalin and Papava, Dragos and Olaru, Vlad and Sminchisescu, Cristian},
  journal={IEEE Transactions on Pattern Analysis and Machine Intelligence}, 
  title={Human3.6M: Large scale datasets and predictive methods for 3D human sensing in natural environments}, 
  year={2014},
  volume={36},
  number={7},
  pages={1325-1339}
 }

@article{finn2016unsupervised,
  title={Unsupervised learning for physical interaction through video prediction},
  author={Finn, Chelsea and Goodfellow, Ian and Levine, Sergey},
  journal={Advances in Neural Information Processing Systems (NeurIPS)},
  address={Barcelona, Spain},
  year={2016}
}

@inproceedings{bogo2016keep,
  title={Keep it SMPL: Automatic estimation of 3D human pose and shape from a single image},
  author={Bogo, Federica and Kanazawa, Angjoo and Lassner, Christoph and Gehler, Peter and Romero, Javier and Black, Michael J},
  booktitle={European Conference on Computer Vision (ECCV)},
  address={Amsterdam, The Netherlands},
  year={2016}
}

@inproceedings{martinez2017human,
  title={On human motion prediction using recurrent neural networks},
  author={Martinez, Julieta and Black, Michael J and Romero, Javier},
  booktitle={IEEE Conference on Computer Vision and Pattern Recognition (CVPR)},
  address={Honolulu, HW},
  year={2017}
}

@inproceedings{mao2020history,
  title={History repeats itself: Human motion prediction via motion attention},
  author={Mao, Wei and Liu, Miaomiao and Salzmann, Mathieu},
  booktitle={European Conference on Computer Vision (ECCV)},
  address={Virtual online},
  year={2020}
}

@inproceedings{vahdat2020nvae,
 author = {Vahdat, Arash and Kautz, Jan},
 booktitle = {Advances in Neural Information Processing Systems (NeurIPS)},
 address={Virtual online},
 title = {NVAE: A deep hierarchical variational autoencoder},
 year = {2020}
}

@INPROCEEDINGS{petrovich21actor,
  title     = {Action-conditioned 3{D} human motion synthesis with Transformer {VAE}},
  author    = {Petrovich, Mathis and Black, Michael J. and Varol, G{\"u}l},
  booktitle = {International Conference on Computer Vision (ICCV)},
   address={Virtual online},
  year      = {2021}
}

\end{document}